\documentclass[letterpaper,preprint]{elsarticle}
\pdfoutput=1

\renewcommand{\subparagraph}{\paragraph}
\usepackage{times}
\usepackage{amsmath, amsthm, amssymb}
\usepackage{pstricks}
\usepackage{pst-tree}
\usepackage{makeidx}  
\usepackage{bm}
\usepackage{graphicx}
\usepackage{epsfig}
\usepackage{subfigure}
\usepackage{amsmath}
\usepackage{amsfonts}
\usepackage{color}
\usepackage{array}
\usepackage{colortbl}
\usepackage{framed}
\usepackage{url}
\usepackage{booktabs}
\usepackage{multirow}
\usepackage{sgame}
\usepackage{dsfont}

\usepackage{multicol}
\usepackage{lscape}
\usepackage{relsize}
\newcommand{\cut}[1]{}
\renewcommand{\blue}[1]{{#1}}
\renewcommand{\red}[1]{{#1}}
\newcommand{\revision}[1]{{\blue{#1}}}
\usepackage{rotating}

\DeclareFontFamily{OT1}{pzc}{}
\DeclareFontShape{OT1}{pzc}{m}{it}{<-> s * [1.200] pzcmi7t}{}
\DeclareMathAlphabet{\mathscr}{OT1}{pzc}{m}{it}

\newcommand{\note}[1]{}
\renewcommand{\note}[1]{~\\\frame{\begin{minipage}[c]{\textwidth}\vspace{2pt}\center{#1}\vspace{2pt}\end{minipage}}\vspace{3pt}\\}

\newcommand{\hlinespace}{~\vspace*{-0.15cm}~\\\hline\\\vspace*{0.15cm}}

\newcommand{\emcite}[1]{\citet{#1}}

\newcommand{\lightrule}{\specialrule{.03em}{.4em}{.4em}}

\usepackage{subfigure}
\usepackage{url}
\usepackage{bm}
\usepackage{graphicx}
\usepackage{times}
\usepackage{dsfont}
\usepackage{microtype}
\usepackage{amsmath, amssymb, amsfonts}

\newcommand{\RCW}{{\smaller{\texttt{\textsc{RCW}}}}}
\newcommand{\REG}{{\smaller{\texttt{\textsc{REG}}}}}
\newcommand{\CORLAT}{{\smaller{\texttt{\textsc{CORLAT}}}}}
\newcommand{\BIGMIX}{{\smaller{\texttt{\textsc{BIGMIX}}}}}
\newcommand{\TSPLIB}{{\smaller{\texttt{\textsc{TSPLIB}}}}}
\newcommand{\CORLATREG}{{\smaller{\texttt{\textsc{CR}}}}}
\newcommand{\CORLATREGRCW}{{\smaller{\texttt{\textsc{CRR}}}}}
\newcommand{\SWV}{{\smaller{\texttt{\textsc{SWV}}}}}
\newcommand{\IBM}{{\smaller{\texttt{\textsc{IBM}}}}}
\newcommand{\SWVIBM}{{\smaller{\texttt{\textsc{SWV-IBM}}}}}
\newcommand{\INDU}{{\smaller{\texttt{\textsc{INDU}}}}}
\newcommand{\HAND}{{\smaller{\texttt{\textsc{HAND}}}}}
\newcommand{\RAND}{{\smaller{\texttt{\textsc{RAND}}}}}
\newcommand{\RANDSAT}{{\smaller{\texttt{\textsc{RANDSAT}}}}}
\newcommand{\INDUHANDRAND}{{\smaller{\texttt{\textsc{COMPETITION}}}}}
\newcommand{\PORTGEN}{{\smaller{\texttt{\textsc{RUE}}}}}
\newcommand{\PORTCGEN}{{\smaller{\texttt{\textsc{RCE}}}}}

\newcommand{\hide}[1]{}

\newcommand{\hh}[1]{\note{HH says: #1}}
\renewcommand{\hh}[1]{}

\newcommand{\algofont}[1]{{\smaller{\texttt{#1}}}}

\newcommand{\satzilla}{\algofont{SATzilla}}

\newcommand{\etal}[0]{et al.{}}
\newcommand{\eg}[0]{\emph{e.{}g.{}}}
\newcommand{\ie}[0]{\emph{i.{}e.{}}}

\newcommand{\wrt}[0]{w.{}r.{}t.{}}

\newcommand{\vTheta}{{\bm{\Theta}}}
\newcommand{\vtheta}{{\bm{\theta}}}

\newcommand{\vy}{{\bm{y}}}

\newcommand{\calD}{\mbox{${\cal D}$}}
\newcommand{\gauss}{\mbox{${\cal N}$}}
\newcommand\transpose{{\textrm{\tiny{\sf{T}}}}}

\usepackage[outerbars,color]{changebar}

\cbcolor{red}

\newcommand{\denselist}{\itemsep 2pt plus 2pt minus 2pt\partopsep -20pt}

\usepackage{amsmath, amsthm, amssymb}
\newtheorem{thm}{Theorem}

\newtheorem{prop}[thm]{Proposition}

\newtheorem{define}[thm]{Definition}
\makeatletter
\renewcommand\bibsection%
{
  \section*{\refname
    \@mkboth{\MakeUppercase{\refname}}{\MakeUppercase{\refname}}}
}
\makeatother

\newcommand{\spear}{\algofont{SPEAR}}
\newcommand{\saps}{\algofont{SAPS}}

\newcommand{\tnm}{\algofont{tnm}}
\newcommand{\minisat}{\algofont{Minisat~2.0}}
\newcommand{\lkh}{\algofont{LK-H}}
\newcommand{\concorde}{\algofont{Concorde}}
\newcommand{\cryptominisat}{\algofont{CryptoMinisat}}
\newcommand{\satelite}{\algofont{SATElite}}

\newcommand{\lpsolve}{\algofont{lp\_solve}}
\newcommand{\scip}{\algofont{SCIP}}
\newcommand{\gurobi}{\algofont{Gu\-ro\-bi}}
\newcommand{\cplex}{\algofont{CPLEX}}

\newcommand{\SATzilla}{\algofont{SATzilla}}

\newcommand{\Var}{\ensuremath\text{Var}}
\newcommand{\indicator}{\ensuremath\mathds{I}}

\newcommand{\interrowspace}{.6em}

\journal{Artificial Intelligence}

\begin{document}

\begin{frontmatter}

\title{Algorithm Runtime Prediction: Methods \& Evaluation}

\author{Frank Hutter}
\author{Lin Xu}
\author{Holger H. Hoos}
\author{Kevin Leyton-Brown}

\address{
     Department of Computer Science\\
     University of British Columbia \\
     201-2366 Main Mall, BC V6T 1Z4, CANADA\\
     {\texttt{\{hutter, xulin730, hoos, kevinlb\}@cs.ubc.ca}}
}

\begin{abstract}

Perhaps surprisingly, it is possible to predict how long an algorithm will take to run on a
previously unseen input, using machine learning techniques to build a model of the algorithm's runtime as a function of \revision{problem-specific instance features}. Such models have important applications to
algorithm analysis, portfolio-based algorithm selection, and the automatic configuration of parameterized algorithms.
Over the past decade, a wide variety of techniques have been studied for building such models.
Here, we describe extensions and improvements of existing models, new families of models, and---perhaps most importantly---a much more thorough treatment of algorithm parameters as model inputs.
We also \revision{comprehensively describe new and existing features for predicting algorithm runtime for propositional satisfiability (SAT), travelling salesperson (TSP) and mixed integer programming (MIP) problems}.
We evaluate these innovations through the largest empirical analysis of its kind, \revision{comparing to a wide range of runtime modelling techniques from the literature}.
Our experiments consider 11 algorithms and 35 instance distributions; they also span a very wide range of SAT, MIP, and TSP instances, with the least structured having been generated uniformly at random and the most structured having emerged from real industrial applications. Overall, we demonstrate that our new models yield substantially better runtime predictions than previous approaches in terms of their generalization to new problem instances, to new algorithms from a parameterized space, and to both simultaneously.
\end{abstract}

\begin{keyword}
Supervised machine learning \sep Performance prediction \sep Empirical performance models
\sep Response surface models \sep Highly parameterized algorithms
\sep Propositional satisfiability \sep Mixed integer programming \sep Travelling salesperson problem

\MSC[2010] 68T20

\end{keyword}

\end{frontmatter}

\section{Introduction} \label{sec:introduction}

NP-complete problems are ubiquitous in AI. Luckily, while these problems may be hard to solve on worst-case inputs, it is often feasible to solve even large problem instances that arise in practice.
Less luckily, state-of-the-art algorithms often exhibit \revision{extreme}
runtime variation across instances from realistic distributions, even when problem size is held constant, and conversely the same instance can take
\revision{dramatically}
different amounts of time to solve depending on the algorithm used
\cite{GomSelCraKau00}. There is little theoretical understanding of what causes this variation.
\revision{Over the past decade, a considerable body of work has shown how to use supervised machine learning methods to build regression models that provide approximate answers to this question based on given  algorithm performance data; we survey this work in Section \ref{sec:eh_models_old_overview}.
In this article, we refer to such models as \emph{empirical performance models (EPMs)}.\footnote{In work aiming to gain insights into instance hardness beyond the worst case, we have used the term \emph{empirical hardness model}~\cite{LeyNudSho02,LeyNudSho09:jacm,EHM-CACM}.
Similar regression models can also be used to predict objectives other than runtime;
examples include an algorithm's success probability~\cite{HowDahHanSchMay00:exploiting,Roberts07learnedmodels}, the solution quality an optimization algorithm achieves in a fixed time~\cite{RidKud07,ChiGoe10,HutHooLey12-ParallelAC}, approximation ratio of greedy local search~\cite{MerEtAl13}, or the SAT competition scoring function~\cite{SATzilla-Full}. We reflect this broadened scope by using the term
EPMs, which we understand as an umbrella that includes EHMs.}
These models are useful in a variety of practical contexts:}
\revision{
\begin{itemize}
\denselist
        \item \textbf{Algorithm selection.} This classic problem of selecting the best from a given set of 
				\red{algorithms}
				on a per-instance basis~\cite{Ric76,SmithMilesAlgoSelectionSurvey}
        has been successfully addressed by using EPMs to predict the performance of all candidate algorithms and selecting the one
        predicted to perform best~\cite{Bre95,LobLem98,Fink98howto,HowDahHanSchMay00:exploiting,Roberts07learnedmodels,SATzilla-Full,KotGenMig12:eval}.
        \item \textbf{Parameter tuning and algorithm configuration.} EPMs are useful for these problems in at least two ways.
        First, they can model the performance of a parameterized algorithm dependent on the settings of its parameters; in a sequential model-based optimization process,
        one alternates between learning an EPM and using it to identify promising settings to evaluate next~\cite{JonSchWel98,spo_conf,HutHooLeyMur10-tbspo,HutHooLey11-SMAC,HutHooLey12-ParallelAC}.
        Second, EPMs can model algorithm performance dependent on both problem instance features and algorithm parameter settings; such models can then be used to select
        parameter settings with good predicted performance on a per-instance basis~\cite{HutHamHooLey06b}.
        \item \textbf{Generating hard benchmarks.} An EPM for one or more algorithms can be used to set the parameters
        of existing benchmark generators in order to create instances that are hard for the algorithms in question~\cite{boosting-CP,LeyNudSho09:jacm}.
        \item \textbf{Gaining insights into instance hardness and algorithm performance.} EPMs can be used to assess which instance features and algorithm parameter values most
        impact empirical performance. Some models support such assessments directly~\cite{RidKud07,MerEtAl13}.
                                For other models, generic feature selection methods, such as forward selection, can be used to
        identify a small number of key model inputs (often 
				\red{fewer}
				than five) that explain algorithm performance almost as well as the whole set of inputs~\cite{LeyNudSho09:jacm,HutHooLey13ImportanceFwdSel}.
\end{itemize}
}

\noindent{}\revision{While these applications motivate our work, in the following, we will not discuss them in detail; instead, we focus on the models themselves.
The idea of modelling algorithm runtime is no longer new; however, we have made substantial recent progress in making runtime prediction methods more general, scalable and accurate.} After a review of past work (Section \ref{sec:eh_models_old_overview}) and of the runtime prediction methods used by this work (Section \ref{sec:eh_models_old_methods}), we describe four new contributions.
\begin{enumerate}
\item We describe new, more sophisticated modeling techniques (based on random forests and approximate Gaussian processes) and methods for modeling runtime variation arising from the settings of a large number of (both categorical and continuous) algorithm parameters (Section \ref{sec:eh_models_new}).
\item We introduce new instance features for
\revision{propositional satisfiability (SAT),
travelling salesperson (TSP) and mixed integer programming (MIP) problems}---in particular, novel probing features and timing features---yielding comprehensive sets of 138, 121, and 64 features for SAT, MIP, and TSP, respectively (Section \ref{sec:eh_features}).
\item To assess the impact of these advances and to determine the current state of the art, we performed what we believe is the most comprehensive evaluation of runtime prediction methods to date. Specifically, we evaluated all methods of which we are aware on performance data for 11 algorithms and 35 instance distributions spanning SAT, TSP and MIP and considering three different problems: predicting runtime on novel instances (Section \ref{sec:ehms_better}), novel parameter configurations (Section \ref{sec:rms_better}), and both novel instances \emph{and} configurations (Section \ref{sec:combination_ehm_rsm}). 
\item Techniques from the statistical literature on survival analysis offer ways to better handle data from runs that were terminated prematurely. While these techniques were not used in most previous work---leading us to omit them from the comparison above---
    we show how to leverage them to achieve further improvements to our best-performing model, random forests (Section \ref{sec:censoring}).\footnote{We used early versions of the new modeling techniques described in Section \ref{sec:eh_models_new}, as well as the extensions to censored data described in Section \ref{sec:censoring} in recent conference and workshop publications on algorithm configuration \cite{HutHooLeyMur10-tbspo,HutHooLey11-SMAC,HutHooLey12-ParallelAC,HutHooLey11-censoredBO}. This article is the first to comprehensively evaluate the quality of these models.}
\end{enumerate}

\section{An Overview of Related Work}\label{sec:eh_models_old_overview}

Because the problems have been considered by substantially different communities, we separately consider related work on predicting the runtime of parameterless and parameterized algorithms, and applications of these predictions to gain insights into instance hardness and algorithm parameters.

\subsection{Related Work on Predicting Runtime of Parameterless Algorithms}

\revision{The use of statistical regression methods for runtime prediction has its roots in a range of different communities and dates back at least to the mid-1990s.
In the parallel computing literature, Brewer used linear regression models to predict the runtime of different implementations of portable, high-level libraries for multiprocessors, aiming to automatically select the best implementation on a novel architecture~\cite{Bre94,Bre95}.
In the AI planning literature, \emcite{Fink98howto} used linear regression to predict how the performance of three planning algorithms depends on problem size and used these predictions for deciding which algorithm to run for how long.
In the same community, Howe and co-authors \cite{HowDahHanSchMay00:exploiting,Roberts07learnedmodels}
used linear regression to 
predict how both a planner's runtime and its probability of success depend on various features of the planning problem; they also applied these predictions to
decide, on a per-instance basis, which of a finite set of algorithms should be run in order to optimize a performance objective such as expected runtime. Specifically, they constructed a \emph{portfolio} of planners
that ordered algorithms by their expected success probability divided by their expected runtime.
In the constraint programming literature, Leyton-Brown \etal{} \cite{LeyNudSho02,LeyNudSho09:jacm} studied the winner determination problem in combinatorial auctions and showed that accurate runtime predictions could be made for several different solvers and a wide variety of instance distributions. That work considered a variety of different regression methods (including lasso regression, multivariate adaptive regression splines, and support vector machine regression) but in the end settled on a relatively simpler method: ridge regression with preprocessing to select an appropriate feature subset, a quadratic basis function expansion, and a log-transformation of the response variable. (We formally define this and other regression methods in Section~\ref{sec:eh_models_old_methods}.) The problem-independent runtime modelling techniques from that work were
subsequently applied to the SAT problem \cite{NudLeyDevShoHoo04}, leading to the successful portfolio-based algorithm selection method \SATzilla{}~\cite{Satzilla03,NudLeyDevShoHoo04,cp-satzilla07,SATzilla-Full}.
Most recently, in the machine learning community, \emcite{HuaJiaYuChuManNai10} applied linear regression techniques to the modeling of algorithms with low-order polynomial runtimes.}

\revision{Due to the extreme runtime variation often exhibited by algorithms for solving combinatorial problems, it is common practice to terminate unsuccessful runs after they exceed a so-called \emph{captime}. Capped runs only yield a \emph{lower bound} on algorithm runtime, but are typically treated as having succeeded at the captime.
\emcite{Fink98howto} was the first to handle such \emph{right-censored} data points more soundly for runtime predictions of AI planning methods
and used the resulting predictions to compute captimes that maximize a given utility function.
Gagliolo \etal{} \cite{GagSch06AIMATH,Gagliolo2010Survival} made the connection to the statistical literature on survival analysis to handle right-censored data in their work on dynamic algorithm portfolios. Subsequently, similar techniques were used for \satzilla{}'s runtime predictions~\cite{cp-satzilla07} and in
model-based algorithm configuration~\cite{HutHooLey11-censoredBO}.}

\revision{Recently, Smith-Miles \etal{}  published a series of papers on learning-based approaches for characterizing instance hardness for a wide variety of hard combinatorial
problems~\cite{SmithMilesAlgoSelectionSurvey,tspfeat10,SmiTan12,SmiLop12}.
Their work considered a range of tasks, including not only performance prediction, but also clustering, classification into easy and hard instances, as well as visualization. In the context of performance prediction, on which we focus in this article, theirs is the only work known to us to use neural network models.
Also recently, \emcite{KotGenMig12:eval} compared regression, classification, and ranking algorithms for algorithm selection and showed that this choice matters: poor regression and classification methods yielded worse performance than the single best solver, while good methods yielded better performance.}

\revision{Several other veins of performance prediction research deserve mention.
\emcite{Haim:2008} extended linear methods to the problem of making online estimates of SAT solver runtimes.
Several researchers have applied supervised classification to select the fastest algorithm for a problem instance~\cite{GueMil04,milano04,Guo04,GebHniBriFre05,XuHutHooLey12} or to judge whether a particular run of a randomized algorithm would be good or bad~\cite{Horvitz01}  (in contrast to our topic of predicting performance directly using a regression model).
In the machine learning community, \emph{meta-learning} aims to predict the accuracy of learning algorithms~\cite{Vilalta02:aij_metalearning}.
Meta-level control for anytime algorithms computes estimates of an algorithm's performance in order to decide when to stop it and act on the solution found~\cite{HansenZilberstein:anytime}.
Algorithm scheduling in parallel and distributed systems has long relied on low-level performance predictions, for example based on source code analysis~\cite{NuddEtAl2000:PACE}.
In principle, the methods discussed in this article could also be applied to meta-level control and algorithm scheduling.}

\revision{Other research has aimed to identify single quantities that correlate with an algorithm's runtime. A famous early example is the clauses-to-variables ratio for uniform-random 3-SAT \cite{peter91,mitchell92}. Earlier still, Knuth showed how to use random probes of a search tree to estimate its size \cite{Knu75}; subsequent work refined this approach \cite{LobLem98,KilSlaThiWal06}. We incorporated such predictors as features in our own work and therefore do not evaluate them separately. (We note, however, that we have found Knuth's tree-size estimate to be very useful for predicting runtime in some \revision{cases, \eg{}, for}  complete SAT solvers on unsatisfiable 3-SAT instances~\cite{NudLeyDevShoHoo04}.)
The literature on search space analysis has proposed a variety of quantities correlated with the runtimes of (mostly) local search algorithms. Prominent examples include fitness distance correlation~\cite{Jones95fitnessdistance}
and autocorrelation length (ACL)~\cite{Wei90-landscape}.
With one exception (ACL for TSP) we have not included such measures in our feature sets, as computing them can be quite expensive.}

\subsection{Related Work on Predicting Runtime of Parameterized Algorithms}

In principle, it is not particularly harder to predict the runtimes of parameterized algorithms than the runtimes of their parameterless cousins: parameters can be treated as additional inputs to the model (notwithstanding the fact that they describe the algorithm rather than the problem instance, and hence are directly controllable by the experimenter), and a model can be learned in the standard way.
\revision{In past work, we
pursued precisely this approach, using both linear regression models and exact Gaussian processes to model the
dependency of runtime on both instance features and algorithm parameter values \cite{HutHamHooLey06b}. However, this direct application of methods designed for parameterless algorithms is effective only for small numbers of continuous-valued parameters (\eg{}, the experiments in \cite{HutHamHooLey06b} considered only two parameters).} Different methods are more appropriate when an algorithm's parameter space becomes very large. In particular, a careful sampling strategy must be used, making it necessary to consider issues raised in the statistics literature on experimental design. Separately, models must be adjusted to deal with \emph{categorical} parameters: parameters with finite, unordered domains (\eg{}, selecting which of various possible heuristics to use, \revision{or activating an optional preprocessing routine}).

\revision{The experimental design literature uses the term \emph{response surface model (RSM)} to refer to a predictor for the output of a process with controllable input parameters that can generalize from observed data to new, unobserved parameter settings \citep[see, \eg{}, ][]{BoxWil51,BoxDra07}.
Such RSMs are at the core of sequential model-based optimization methods for blackbox functions~\cite{JonSchWel98},
which have recently been adapted to applications in automated parameter tuning and algorithm configuration~\citep[see, \eg{},][]{spo_conf,Bartz06,HutHooLeyMur09-spo+,HutHooLeyMur10-tbspo,HutHooLey11-SMAC}.}

Most of the literature on RSMs of algorithm performance has limited its consideration to algorithms running on single problem instances and algorithms
only with continuous input parameters.
We are aware of a few papers beyond our own that relax these assumptions. \emcite{BarMar04} support categorical algorithm parameters (using regression tree models), and two
existing methods consider predictions across both different instances and parameter settings.
First, \citet{RidKud07} applied an analysis of variance (ANOVA) approach to
detect important parameters, using linear and quadratic models.
\revision{Second, \citet{ChiGoe10} noted that in constrast to algorithm parameters, instance characteristics cannot be controlled and should be treated as so-called \emph{random effects}.}
Their resulting \emph{mixed-effects} models are linear and, like Ridge \& Kudenko's ANOVA model, assume Gaussian performance distributions.
We note that this normality assumption is much more realistic in the context of predicting solution quality of local search algorithms (the problem addressed in \cite{ChiGoe10}) than in the context of the algorithm runtime prediction problem we tackle here.

\subsection{\revision{Related Work on Applications of Runtime Prediction to Gain Insights into Instance Hardness and Algorithm Parameters}}\label{sec:analyzing_models}

\revision{Leyton-Brown and co-authors \cite{LeyNudSho02,NudLeyDevShoHoo04,LeyNudSho09:jacm} employed forward selection with linear regression models to determine small sets of instance features that suffice to yield high-quality predictions, finding that often as little as five to ten features yielded predictions as good as the full feature set.
\citet{HutHooLey13ImportanceFwdSel} extended that work to predictions in the joint space of instance features and algorithm parameters, using arbitrary models. Two model-specific approaches for this joint identification of instance features and algorithm parameters are the ANOVA approach of \citet{RidKud07} and the mixed-effects model of \citet{ChiGoe10} mentioned previously.
Other approaches for quantifying parameter importance include an entropy-based measure~\cite{NanEib07:revac},
and visualization methods for interactive parameter exploration 
\cite{Bartz06}.}

\section{Methods Used in Related Work}\label{sec:eh_models_old_methods}

We now define the different machine learning methods that have been used to predict algorithm runtimes: ridge regression (used by \cite{Bre94,Bre95,LeyNudSho02,LeyNudSho09:jacm,Satzilla03,NudLeyDevShoHoo04,HutHamHooLey06b,cp-satzilla07,SATzilla-Full,HuaJiaYuChuManNai10}), neural networks (see \cite{MilvHem11}), Gaussian process regression (see \cite{HutHamHooLey06b}), and regression trees (see \cite{BarMar04}). This section provides the basis for the experimental evaluation of different methods in Sections~\ref{sec:ehms_better}, \ref{sec:rms_better}, and \ref{sec:combination_ehm_rsm}; thus, we also discuss implementation details.

\subsection{Preliminaries}

We describe a problem instance by a list of $m$ features $\bm{z} = [z_{1},\dots,z_{m}]^\transpose$, drawn from a given \emph{feature space} $\mathcal{F}$.
These features must be computable by a piece of \revision{problem-specific} code (usually provided by a domain expert) that
efficiently extracts characteristics for any given problem instance
(typically, in low-order polynomial time \wrt{} to the size of the given problem instance).
We define the \emph{configuration space} of a parameterized algorithm with $k$ parameters $\theta_1, \dots, \theta_k$ with respective domains $\Theta_1, \dots, \Theta_k$ as a subset of the cross-product of parameter domains: $\vTheta \subseteq \Theta_1 \times \cdots \times \Theta_k$.
The elements of $\vTheta$ are complete instantiations of the algorithm's $k$ parameters, and we refer to them as \emph{configurations}.
Taken together, the configuration and the feature spaces define the \emph{input space}: $\mathcal{I} = \vTheta \times \mathcal{F}$.

Let $\Delta(\mathds{R})$ denote the space of probability distributions over the real numbers; we will use these real numbers to represent an algorithm performance measure, such as runtime in seconds on some reference machine.
(In principle, EPMs can predict any type of performance measure that can be evaluated in single algorithm runs, such as runtime, solution quality, memory usage, energy consumption,
or communication overhead.)
Given an algorithm $\mathcal{A}$ with configuration space $\vTheta$ and a distribution of instances with feature space $\mathcal{F}$, an EPM is a stochastic process $f: \mathcal{I} \mapsto \Delta(\mathds{R})$ that defines a probability distribution over performance measures for each combination of a parameter configuration $\vtheta\in \vTheta$ of $\mathcal{A}$ and a problem instance with features $\mathbf{z}\in \mathcal{F}$.
The prediction of an entire distribution allows us to assess the model's \emph{confidence} at a particular input, which is essential, \eg{}, in model-based algorithm configuration~\cite{spo_conf,Bartz06,HutHooLeyMur09-spo+,HutHooLey11-SMAC}.
Nevertheless, since many of the methods we review yield only point-valued runtime predictions, our experimental analysis focuses on the accuracy of mean predicted runtimes. 
\red{For the models that define a predictive distribution (Gaussian processes and our variant of random forests), we study the accuracy of confidence values separately in the online appendix, with qualitatively similar results as for mean predictions.}

To construct an EPM for an algorithm $\mathcal{A}$ with configuration space $\vTheta$ on an instance set $\Pi$,
we run $\mathcal{A}$ on various combinations of configurations $\vtheta_i \in \vTheta$ and instances $\pi_i \in \Pi$, and record the resulting performance values $y_i$.
We record the $k$-dimensional parameter configuration $\vtheta_i$ and the $m$-dimensional feature vector $\bm{z}_i$ of the instance used in the $i$th run, and combine them to form a $p=k+m$-dimensional vector of \emph{predictor variables} $\bm{x}_i=[\vtheta_i^\transpose, \bm{z}_i^\transpose]^\transpose$. The training data for our regression models is then simply
$(\bm{x}_1, y_1), \dots, (\bm{x}_n, y_n)\}$.
We use $\bm{X}$ to denote the $n \times p$ matrix containing $[\bm{x}_1, \ldots, \bm{x}_n]^\transpose$ (the so-called \emph{design matrix}) and $\bm{y}$ for the vector of performance values $[y_1, \ldots, y_n]^\transpose$.

Various transformations can make this data easier to model. In this article, we focus on runtime as a performance measure and use a log-transformation, thus effectively predicting log runtime.\footnote{Due to the resolution of our CPU timer, runtimes below $0.01$ seconds are measured as $0$ seconds. To make $y_i=\log(r_i)$ well defined in these cases, we count them as $0.005$ (which, in log space, has the same distance from 0.01 as the next bigger value measurable with our CPU timer, 0.02).}
In our experience, we have found this transformation to be very important due to the large variation in runtimes for hard combinatorial problems.
We also transformed the predictor variables, discarding those input dimensions constant across all training data points and normalizing the remaining ones to have mean 0 and standard deviation 1 (\ie{}, for each input dimension we subtracted the mean and then divided by the standard deviation).

For some instances, certain feature values can be missing because of timeouts, crashes, or because they are undefined (when preprocessing has already solved an instance).
These \emph{missing values} occur relatively rarely, so we use a simple mechanism for handling them.
We disregard missing values for the purposes of normalization, and then set them to zero for training our models. This means that missing
feature values are effectively assumed to be equal to the mean for the respective distribution
and thus to be minimally informative.
In some models (ridge regression and neural networks), this mechanism leads us to ignore missing features, since their weight is multiplied by zero.

\revision{Most modeling methods discussed in this paper have free hyperparameters that can be set by minimizing some loss function, such as cross-validation error. We point out these hyper-parameters, as well as their default setting, when discussing each of the methods. While, to the best of our knowledge, all previous work on runtime prediction has used fixed default hyperparameters, we also experimented with optimizing them for every method in our experiments. For this purpose, we used the gradient-free optimizer DIRECT~\cite{JonesEtAl93:direct} to minimize 2-fold cross-validated root mean squared error (RMSE) on the training set with a budget of 30 function evaluations. This simple approach is a better alternative than the frequently-used grid search and random search~\cite{BerBen12}.}

\subsection{Ridge Regression}\label{sec:linear-reg}

Ridge regression \cite[see, \eg{},][]{Bis06} is a simple regression method
that fits a linear function $f_{\bm{w}}(\bm{x})$ of its inputs $\bm{x}$.
Due to its simplicity (both conceptual and computational) and its interpretability, combined with competitive predictive performance \revision{in most scenarios} we studied, this is the method that \revision{has been used most frequently in the past for building EPMs \cite{Fink98howto,HowDahHanSchMay00:exploiting,LeyNudSho02,LeyNudSho09:jacm,NudLeyDevShoHoo04,HutHamHooLey06b,lin07-hhm}}.

Ridge regression works as follows. Let $\bm{X}$ and $\bm{y}$ be as defined above,
let $\bm{I_p}$ be the $p \times p$ identity matrix, and let $\epsilon$ be a small constant.
Then, compute the weight vector
\[\label{eq:ridge}\bm{w} = (\bm{X^\transpose} \bm{X} + \epsilon \bm{I_p})^{-1} \bm{X^\top} \bm{y}.\]
Given a new feature vector, $\bm{x}_{n+1}$, ridge regression predicts $f_{\bm{w}}(\bm{x}_{n+1}) = \bm{w^\transpose} \bm{x}_{n+1}$.
Observe that with $\epsilon=0$, we recover standard linear regression. The effect of $\epsilon>0$ is to regularize the model by penalizing large coefficients $\bm{w}$; it is equivalent to a Gaussian prior favouring small coefficients under a Bayesian model (see, \eg{}, \cite{Bis06}). A beneficial side effect of this regularization is that numerical stability improves in the common case where $\bm{X}$ is rank deficient, or nearly so.
The computational bottleneck in ridge regression with $p$ input dimensions is the inversion of the $p\times p$ matrix $A=\bm{X^\transpose} \bm{X} + \epsilon \bm{I_p}$, which requires time cubic in $p$.

Algorithm runtime can often be better approximated by a polynomial function than by a linear one,
and the same holds for log runtimes.
For that reason, it can make sense to perform a basis function expansion to create new features that are products of two or more original features. In light of the resulting increase in the number of features,
a quadratic expansion is particularly appealing.
Formally, we augment each model input $\bm{x}_i = [x_{i,1},\dots,x_{i,p}]^\transpose{}$ with pairwise product inputs $x_{i,j}\cdot{} x_{i,l}$ for $j=1,\dots,p$ and $l=j,\dots,p$.

Even with ridge regularization, the generalization performance of linear regression (and, indeed, many other learning algorithms) can deteriorate when some inputs are uninformative or highly correlated with others;
in our experience, it is difficult to construct sets of instance features that do not suffer from these problems.
Instead, we reduce the set of input features by performing \emph{feature selection}.
Many different methods exist for feature expansion and selection; we review two different ridge regression variants from the recent literature that only differ in these design decisions.\footnote{We also considered a third ridge regression variant that was originally proposed by \citet{LeyNudSho09:jacm} (``ridge regression with elimination of redundant features'', or RR-el for short). Unfortunately, running this method was computationally infeasible, considering the large number of features we consider in this paper, (a) forcing us to approximate the method, and (b) nevertheless preventing us from performing 10-fold cross-validation. Because these hurdles made it impossible to fairly compare RR-el to other methods, we do not discuss RR-el here. However, for completeness, our online appendix includes both a definition of our approximation to RR-el and experimental results showing it to perform worse than ridge regression variant RR in 34/35 cases.}

\subsubsection{Ridge Regression Variant RR: Two-phase forward selection~\cite{cp-satzilla07,SATzilla-Full}}

For more than half a decade, we used a simple and scalable feature selection method based on forward selection~\cite[see \eg{},][]{Guyon06book} to build the regression models used by \satzilla{}~\cite{cp-satzilla07,SATzilla-Full}. This iterative method starts with an empty input set, greedily adds one linear input at a time to minimize cross-validation error at each step, and stops when $l$ linear inputs have been selected.
It then performs a full quadratic expansion of these $l$ linear features (using the original, unnormalized features, and then normalizing the resulting quadratic features again to have mean zero and standard deviation one). Finally, it carries out another forward selection with the expanded feature set, once more starting with an empty input set and stopping when $q$ features have been selected. The reason for the two-phase approach is scalability: this method prevents us from ever having to perform a full quadratic expansion of our features. (For example, we have often employed over $100$ features and a million runtime measurements; in this case, a full quadratic expansion would involve over $5$ billion feature values.)

Our implementation reduces the computational complexity of forward selection by exploiting the fact that the inverse matrix $(A')^{-1}$ resulting from including one additional feature can be computed incrementally by two rank-one updates of the previous inverse matrix $A^{-1}$, requiring quadratic time rather than cubic time~\cite{rank1update}.

In our experiments, we fixed the number of linear inputs to $l=30$ in order to keep the result of a full quadratic basis function expansion manageable in size
(with 1 million data points, the resulting matrix has $(\binom{30}{2}+30) \cdot 1\,000\,000$, or about $500$ million elements).
The maximum number of quadratic terms $q$ and the ridge penalizer $\epsilon$
are free parameters of this method; by default, we used $q=20$ and $\epsilon=10^{-3}$.

\subsubsection{Ridge Regression Variant SPORE-FoBa: Forward-backward selection~\cite{HuaJiaYuChuManNai10}}

Recently, \emcite{HuaJiaYuChuManNai10} described a method for predicting algorithm runtime that they called Sparse POlynomial REgression~(SPORE), which is based on ridge regression with forward-backward (FoBa) feature selection.\footnote{Although this is not obvious from their publication~\cite{HuaJiaYuChuManNai10}, the authors confirmed to us that FoBa uses ridge rather than LASSO regression, and also gave us their original code.}
\citeauthor{HuaJiaYuChuManNai10} concluded that SPORE-FoBa outperforms lasso regression, which is consistent with the comparison to lasso by \citet{LeyNudSho09:jacm}.
In contrast to the RR variants above, SPORE-FoBa employs a cubic feature expansion (based on its own normalizations of the original predictor variables). Essentially, it performs a single pass of forward selection, at each step adding a small \emph{set} of terms determined by a forward-backward phase on a feature's candidate set.
Specifically, having already selected a set of terms $T$ based on raw features $S$, SPORE-FoBa loops over all raw features $r \notin S$, constructing a candidate set $T_r$ that consists of all polynomial expansions of $S \cup \{r\}$ that include $r$ with non-zero degree and whose total degree is bounded by 3.
For each such candidate set $T_r$, the forward-backward phase iteratively adds the best term $t \in T\setminus{} T_r$, if its reduction
of root mean squared error (RMSE) exceeds a threshold $\gamma$ (forward step), and then removes the worst term $t \in T$, if its reduction of RMSE is below $0.5 \gamma$ (backward step). This phase terminates when no single term $t \in T\setminus{} T_r$ can be added to reduce RMSE by more than $\gamma$.
Finally, SPORE-FoBa's outer forward selection loop chooses the set of terms $T$ resulting from the best of its forward-backward phases, and iterates until the number of terms in $T$ reach a prespecified maximum of $t_{\text{max}}$ terms.
In our experiments, we used the original SPORE-FoBa code; its free parameters are the ridge penalizer $\epsilon$, $t_{\text{max}}$, and $\gamma$, with defaults $\epsilon=10^{-3}$, $t_{\text{max}}=10$, and $\gamma = 0.01$.

\subsection{Neural Networks}\label{sec:nns}
Neural networks are a well-known regression method inspired by information processing in the human brain.
The multilayer perceptron (MLP) is a particularly popular type of neural network that organizes single computational units (``neurons'') in layers (input, hidden, and output layers), using the outputs of all units in a layer as the inputs of all units in the next layer.
Each neuron $n_i$ in the hidden and output layers with $k$ inputs $\mathbf{a_i}=[a_{i,1}, \dots, a_{i,k}]$ has an associated weight term vector $\mathbf{w_i}=[w_{i,1}, \dots, w_{i,k}]$ and a bias term $b_i$, and computes a function $\mathbf{w_i}^{\transpose}\mathbf{a_i} + b_i$. For neurons in the hidden layer, the result of this function is further propagated through a nonlinear activation function $g:\mathds{R}\rightarrow\mathds{R}$ (which is often chosen to be $\tanh$).
Given an input $\mathbf{x} = [x_1, \dots, x_p]$, a network with a single hidden layer of $h$ neurons $n_1, \dots, n_h$ and a single output neuron $n_{h+1}$ then computes output
\[\hat{f}(\mathbf{x}) = \left(\sum_{j=1}^{h}g(\mathbf{w_j}^{\transpose} \mathbf{x} + b_j) \cdot w_{h+1,j}\right) + b_{h+1}.\]
The $p\cdot{} h + h$ weight terms and $h+1$ bias terms
can be combined into a single weight vector $\mathbf{w}$, which can be set to minimize the network's prediction error using any continuous optimization algorithm (\eg{}, the classic ``backpropagation'' algorithm performs gradient descent to minimize squared prediction error).

\citet{MilvHem11} used an MLP with one hidden layer of 28 neurons to predict the runtime of local search algorithms for solving timetabling instances.
They used the proprietary neural network software Neuroshell, but advised us to compare to an off-the-shelf Matlab implementation instead.
We thus employed the popular Matlab neural network package NETLAB~\cite{Nab02}.
NETLAB uses activation function $g=\tanh$ and supports a regularizing prior to keep weights small, minimizing the error metric $\sum_{i}^N (\hat{f}(\mathbf{x_i}) - y_i)^2 + \alpha \mathbf{w}^\transpose \mathbf{w}$, where $\alpha$ is a parameter determining the strength of the prior.
In our experiments, we used NETLAB's default optimizer (scaled conjugate gradients, SCG) to minimize this error metric,
stopping the optimization after the default of 100 SCG steps.
Free parameters are the regularization factor $\alpha$ and the number of hidden neurons $h$; we used NETLAB's default $\alpha=0.01$ and, like \citet{MilvHem11}, $h=28$.

\subsection{Gaussian Process Regression}\label{sec:GPR}

Stochastic Gaussian processes (GPs)~\cite{RasWil06} are a popular
class of regression models with roots in geostatistics, where they
are also called Kriging models~\cite{kriging51}.
GPs are the dominant modern approach for building response surface models~\cite{SacWelMitWyn89,JonSchWel98,SanWilNot03,Bartz06}.
\revision{They were first applied to runtime prediction by \citet{HutHamHooLey06b}, who found them to yield better results than ridge regression, albeit at greater computational expense.}

\revision{To construct a GP regression model, we first need to select a kernel
function $k: \mathcal{I} \times \mathcal{I} \mapsto \mathds{R}^+$, characterizing the degree of similarity between pairs of elements of the input space $\mathcal{I}$.
A variety of kernel functions are possible, but the most common choice for continuous inputs is the squared exponential kernel \begin{equation}
\label{eqn:kernel-cont}
k_{\text{cont}}(\bm{x}_i,\bm{x}_j) = \exp \left(\sum_{l=1}^p
\left(-\lambda_l \cdot (x_{i,l}-x_{j,l})^2\right) \right),
\end{equation}
where $\lambda_1, \dots, \lambda_p$ are kernel parameters.
It is based on the idea that correlations decrease with weighted Euclidean distance in the input space (weighing each dimension $l$ by a kernel parameter $\lambda_l$).
In general, such a kernel defines a prior distribution over the type of functions we expect. This distribution takes the form of a \emph{Gaussian stochastic process}: a collection of random variables such that any finite subset of them has a joint Gaussian distribution.
What remains to be specified is the tradeoff between the strength of this prior and fitting observed data, which is set by specifying the \emph{observation noise}. Standard GPs assume normally distributed observation noise with mean zero and variance $\sigma^2$, where $\sigma^2$, like the kernel parameters $\lambda_l$, can be optimized to improve the fit.}
\revision{Combining the prior specified above with the training data $\calD = \{(\bm{x}_1,y_1),\ldots,(\bm{x}_n,y_n)\}$ yields the \emph{posterior} distribution at a new input point $\bm{x}_{n+1}$(see the book by \citet{RasWil06} for a derivation):}
\begin{equation}
\label{eq:gp-standard}p(y_{n+1} \mid \bm{x}_{n+1}, \bm{x}_{1:n}, \vy{}_{1:n}) = \gauss(y_{n+1} \mid \mu_{n+1},\Var_{n+1})
\end{equation}
with mean and variance
\begin{eqnarray}
\nonumber{}\mu_{n+1} & = & \bm{k}{_*}^\transpose  [\bm{K} + \sigma^2 \cdot \bm{I_n}]^{-1}  \vy_{1:n}\\
\nonumber{}\Var_{n+1} & = & k_{**} - \bm{k}{_*}^{\transpose}  [\bm{K} + \sigma^2 \mathbf{I}]^{-1} \bm{k}{_*},
\end{eqnarray}
where
\begin{eqnarray}
\nonumber{}\bm{K} &=&
\begin{pmatrix}
k(\bm{x}_1,\bm{x}_1) & \ldots & k(\bm{x}_1,\bm{x}_n) \\
&\ddots & \\
k(\bm{x}_n,\bm{x}_1) & \ldots & k(\bm{x}_n,\bm{x}_n)
\end{pmatrix}\\
\nonumber{}\bm{k}{_*} &=& (k(\bm{x}_1,\bm{x}_{n+1}), \ldots, k(\bm{x}_n,\bm{x}_{n+1}))^\transpose \\
\nonumber{}k_{**} &=& k(\bm{x}_{n+1},\bm{x}_{n+1}) + \sigma^2.
\end{eqnarray}

\revision{The GP equations above assume fixed kernel parameters $\lambda_1, \dots, \lambda_p$ and fixed observation noise variance $\sigma^2$. These constitute the GP's \emph{hyperparameters}.
In contrast to hyperparameters in other models, the number of GP hyperparameters grows with the input dimensionality, and their optimization is an integral part of fitting a GP: they are typically set by maximizing the \emph{marginal likelihood} $p(\vy_{1:n})$ of the data with a gradient-based optimizer (again, see \citet{RasWil06} for details).}
The choice of
optimizer can make a big difference in practice; we used the {\smaller\texttt{minFunc}} \cite{url:minFunc} implementation of a limited-memory version of BFGS~\cite{NocWri06}.

Learning a GP model from data can be computationally expensive.
Inverting the $n \times n$ matrix $[\bm{K} + \sigma^2 \mathbf{I_n}]$ takes $O(n^3)$ time and has to be done in every of the $h$ hyperparameter
optimization steps, yielding a total complexity of $O(h \cdot n^3)$.
Subsequent predictions at a new input require only time
$O(n)$ and $O(n^2)$ for the mean and the variance, respectively.

\subsection{Regression Trees}\label{sec:regression-trees}

Regression trees~\cite{brei84a} are simple tree-based regression models.
They are known to handle discrete inputs well; their first application to the prediction of algorithm performance 
was by \emcite{BarMar04}.
The leaf nodes of regression trees partition the input space into disjoint regions $R_1, \dots, R_M$,
and use a simple model for prediction in each region $R_m$; the most common choice is to predict a constant $c_m$. This leads to the following prediction for an input point $\bm{x}$:
\begin{equation}
\nonumber{}\hat{\mu}(\bm{x}) = \sum_{m=1}^M c_m \cdot \indicator_{\bm{x}\in R_m},
\end{equation}
where the indicator function $\indicator_z$ takes value $1$ if the proposition $z$ is true and $0$ otherwise.
Note that since the regions $R_m$ partition the input space, this sum will always involve exactly one non-zero term.
We denote the subset of training data points in region $R_m$ as $\mathcal{D}_m$.
Under the standard squared error loss function $\sum_{i=1}^n \left(y_i - \hat{\mu}(\bm{x}_i)\right)^2$, the error-minimizing
choice of constant $c_m$ in region $R_m$ is then the sample mean of the data points in $\mathcal{D}_m$:
\begin{equation}
c_m = \frac{1}{|\mathcal{D}_m|} \sum_{\bm{x}_i \in R_m} y_i\label{eqn:constant_in_leaf}.
\end{equation}

To construct a regression tree, we use the following standard recursive procedure, which starts at the root of the tree with all available training data points $\mathcal{D} = \{(\bm{x}_{1}, y_{1}), \ldots (\bm{x}_{n}, y_{n})\}$.
We consider binary partitionings of a given node's data along \emph{split variables j} and \emph{split points s}. 
For a real-valued split variable $j$, $s$ is a scalar and data point $\bm{x}_i$ is assigned to region $R_1(j,s)$ if $x_{i,j} \le s$ and to region $R_2(j,s)$ otherwise. For a categorical split variable $j$, $s$ is a set, and
data point $\bm{x}_i$ is assigned to region $R_1(j,s)$ if $x_{i,j} \in s$ and to region $R_2(j,s)$ otherwise.
At each node, we select split variable $j$ and split point $s$ to minimize the sum of squared differences to the regions' means,
\begin{equation}
l(j,s) = \sum_{\bm{x}_i \in R_{1}(j,s)} (y_i - c_1)^2 + \sum_{\bm{x}_i \in R_{2}(j,s)} (y_i - c_2)^2 \label{eqn:loss_of_split},
\end{equation}
where $c_1$ and $c_2$ are chosen according to Equation (\ref{eqn:constant_in_leaf})
as the sample means in regions $R_{1}(j,s)$ and $R_{2}(j,s)$, respectively.
We continue this procedure recursively, finding the best split variable and split point, partitioning the data into two child nodes, and recursing into the child nodes.
The process terminates when all training data points in a node share the same $\bm{x}$ values, meaning that no more splits are possible.
This procedure tends to overfit data, which can be mitigated by recursively pruning away branches that contribute little to the model's predictive accuracy. We use cost-complexity pruning with 10-fold cross-validation to identify the best
tradeoff between complexity and predictive quality; see the book by \citet{HasTibFri09} for details.

In order to predict the response value at a new input, $\bm{x}_i$, we
\emph{propagate $\bm{x}$ down the tree}, that is, at each node with split
variable $j$ and split point $s$, we continue to the left child node if $\bm{x}_{i,j} \le s$ (for real-valued variable $j$) or $\bm{x}_{i,j} \in s$ (for categorical variable $j$), and to the right child node otherwise. The predictive mean for $\bm{x}_i$ is the constant $c_m$ in the leaf that this process selects; there is no variance predictor.

\hide{
The computational cost for finding the best split at a node is quite manageable.
Denote the total number of data points as $n$.
For each continuous split variable $j$, we can sort the $n$ values $\vtheta_{d1},\dots,\vtheta_{dn}$ and only
consider up to $n-1$ possible split points between different values.
For the squared error loss function above the computation of $l(j,s)$ can be done in amortized $O(1)$ for each of $j$'s split points $s$,
such that the total time required for determining the best split point of a single continuous variable is $O(n \log{n})$.
For categorical predictor variables with $k$ possible values $v_1, \dots, v_k$, we have to select the best out of $2^k$ subsets for each region.
For the squared error loss function, there is an efficient procedure for this: sort values $v_l$ by the average response of
data points whose $j$th predictor variable has the value $v_l$ and only consider $k$ different subsets.
This procedure takes time $O(n + k \log{k})$. Since the number of values present is lower-bounded by the number of data points,
this is also in $O(n \log{n})$.
Thus, at a node with $n$ data points and $d$ predictor variables, it is possible to find the optimal combination of split variable \& point in $O(d \cdot{} n \log{n})$.
}

\subsubsection{Complexity of Constructing Regression Trees} \label{sec:complexity-rt}

If implemented efficiently, the computational cost of fitting a regression tree is small.
At a single node with $n$ data points of dimensionality $p$, it takes $O(p \cdot n \log{n})$ time to
identify the best combination of split variable and point,
because for each continuous split variable $j$, we can sort the $n$ values
$\mathbf{x}_{1,j},\dots,\mathbf{x}_{n,j}$ and only consider up to $n-1$ possible split points between different values. The procedure for categorical split variables has the same complexity: we consider each of the variable's $k$ categorical values $u_l$, compute score $s_l = \textrm{mean}(\{ y_i \mid \bm{x}_{i,j}=u_l \})$ across the node's data points, sort $(u_1, \dots, u_k)$ by these scores, and only consider the $k$ binary partitions with consecutive scores in each set.
For the squared error loss function we use, the computation of $l(j,s)$ (see Equation \eqref{eqn:loss_of_split}) can be performed in  amortized $O(1)$ time for each of $j$'s split points $s$,
such that the total time required for determining the best split point of a single variable is $O(n \log{n})$.
The complexity of building a regression tree depends on how balanced it is. In the worst case, one data point is split off at a time, leading to a tree of depth $n-1$ and a complexity of
$O(p \sum_{i=1}^n (n-i) \log{(n-i)})$, which is
$O(p \cdot n^2 \log n)$.
In the best case---a balanced tree---we have the recurrence relation $T(n) = v \cdot{} n \log{n} + 2 T(n/2)$, leading to a
complexity of $O(p \cdot n \log^2 n)$.
In our experience, trees are not perfectly balanced, but are much closer to the best case than to the worst case.
For example, $10\,000$ data points typically led to tree depths between 25 and 30 (whereas $\log_2(10\,000) \approx 13$).

Prediction with regression trees is cheap; we merely need to propagate new query points $\mathbf{x}_{n+1}$
down the tree. At each node with continuous split variable $j$ and split point $s$, we only
need to compare $\mathbf{x}_{n+1,j}$ to $s$, an $O(1)$ operation.
For categorical split variables, we can store a bit mask of the values in $s$ to enable $O(1)$ member queries.
In the worst case (where the tree has depth $n-1$), prediction thus takes $O(n)$ time, and in the best (balanced) case it takes $O(\log n)$ time.

\section{New Modeling Techniques for EPMs} \label{sec:eh_models_new}

In this section we extend existing modeling techniques for EPMs, with the primary goal of improving runtime predictions for highly parameterized algorithms. The methods described here draw on advanced machine learning techniques, but, to the best of our knowledge, our work is the first to have applied them for algorithm performance prediction. More specifically, we show how to extend all models to handle categorical inputs (required for predictions in partially categorical configuration spaces) and describe two new model families well-suited to modeling the performance of highly parameterized algorithms based on potentially large amounts of data: the projected process approximation to Gaussian processes and random forests of regression trees.

\subsection{Handling Categorical Inputs} \label{sec:categorical}

Empirical performance models have historically been limited to continuous-valued inputs; the only approach that has so far been used for performance predictions based on discrete-valued inputs is regression trees~\cite{BarMar04}.
In this section, we first present a standard method for encoding categorical parameters as real-valued parameters,
and then present a kernel for handling categorical inputs more directly in Gaussian processes.

\subsubsection{Extension of Existing Methods Using 1-in-$\mathscr{K}$ Encoding}\label{sec:one-in-k}

A standard solution for extending arbitrary modeling techniques to handle categorical inputs is the so-called $1$-in-$\mathscr{K}$ encoding scheme~\citep[see, \eg{}, ][]{Bis06}, which encodes categorical inputs with finite domain size $\mathscr{K}$ as $\mathscr{K}$ binary inputs.
Specifically, if the $i$th column of the design matrix $\bm{X}$ is categorical with domain $D_i$, we replace it with $|D_i|$ binary indicator columns, where the new column corresponding to each $d \in D_i$ contains values $[\indicator_{x_{1,i}=d}, \dots, \indicator_{x_{n,i}=d}]^\transpose$; for each data point, exactly one of the new columns is 1, and the rest are all 0. After this transformation, the new columns are treated exactly like the original real-valued columns, and arbitrary modeling techniques for numerical inputs become applicable.

\subsubsection{A Weighted Hamming Distance Kernel for Categorical Inputs in GPs}\label{sec:GPR-kernel}

\revision{A problem with the 1-in-$\mathscr{K}$ encoding is that using it increases the size of the input space considerably, causing some
regression methods  to perform poorly.
We now define a kernel for handling categorical inputs in GPs
more directly.}
\hide{
\revision{A problem when applying the 1-in-$\mathscr{K}$ encoding out-of-the-box to GPs is that we would end up with one kernel parameter $\lambda_l$ for each of the \emph{encoded} dimensions $l$,
rather than the \emph{original} dimensions. For problems with many categorical parameters, especially those that can take many possible values, this would substantially increase the input dimensionality.
This would complicate the optimization of GP hyperparameters, thereby slowing down model learning and potentially decreasing predictive performance.
One way around this problem is to share hyperparameters across all encoded parameters that originate from a common original input dimension.
We now define a new kernel for categorical parameters that can be understood as doing precisely this: encode discrete dimensions using a 1-in-$\mathscr{K}$ encoding, but use a common hyperparameter for the resulting encoded dimensions. The benefits of defining this kernel directly in the categorical space is twofold: it facilitates mathematical rigor (which is often lost when using a 1-in-$\mathscr{K}$) and enables clean interfaces from a software engineering perspective (one only reasons in the original space, without having to keep track of the encoded feature space and its relation to the original space).}
}
Our kernel is similar to the standard squared exponential kernel of Equation \eqref{eqn:kernel-cont}, but instead of measuring the (weighted) squared distance, it computes a (weighted) Hamming distance:
\begin{equation}
\label{eqn:kernel-cat}K_{\text{cat}}(\bm{x}_i,\bm{x}_j) = \exp \left(\sum_{l=1}^p
(-\lambda_l \cdot \indicator_{x_{i,l} \neq x_{j,l}}) \right).
\end{equation}
For a combination of continuous and categorical input dimensions $\mathcal{P}_{\text{cont}}$
and $\mathcal{P}_{\text{cat}}$, we combine the two kernels:
\begin{equation}
\nonumber{}K_{\text{mixed}}(\bm{x}_i,\bm{x}_j) = \exp \left(
\sum_{l \in \mathcal{P}_{\text{cont}}}
\left(-\lambda_l \cdot (x_{i,l}-x_{j,l})^2\right)
+
\sum_{l \in \mathcal{P}_{\text{cat}}}
\left(-\lambda_l \cdot \indicator_{x_{i,l} \neq x_{j,l}}\right)
 \right).
\end{equation}
Although $K_{\text{mixed}}$ is a straightforward adaptation of the standard kernel in Equation \eqref{eqn:kernel-cont}, we are not aware of any prior use of it.
\revision{To use this kernel in GP regression, we have to show that it is \emph{positive definite}.
\begin{define}[Positive definite kernel]
\label{def:kernel}
A function $k: \mathcal{I} \times \mathcal{I} \mapsto \mathds{R}$ is a \emph{positive definite kernel} iff it is (1) \emph{symmetric}: for any pair of inputs $\bm{x}_i, \bm{x}_j \in \mathcal{I}$, $k$ satisfies $k(\bm{x}_i, \bm{x}_j) = k(\bm{x}_j, \bm{x}_i)$;  and (2) \emph{positive definite}: for any $n$ inputs $\bm{x}_1, \dots, \bm{x}_n \in \mathcal{I}$ and any $n$ constants $c_1,\dots,$$c_n \in \mathds{R}$, $k$ satisfies $\sum_{i=1}^n \sum_{j=1}^n \left(c_i \cdot c_j \cdot k(\bm{x}_i, \bm{x}_j)\right) \ge 0$.
\end{define}}%
\revision{\begin{prop}[$K_{\text{mixed}}$ is positive definite]
\label{prop:valid_mixed}\label{prop:valid_kernel}
For any combination of continuous and categorical input dimensions $\mathcal{P}_{\text{cont}}$
and $\mathcal{P}_{\text{cat}}$, $K_{\text{mixed}}$ is a positive definite kernel function.
\end{prop}
\noindent{}Appendix B in the online appendix provides the proof, which shows that
$K_{\text{mixed}}$ can be constructed from simpler positive definite functions, and uses the facts that the space of
positive definite kernel functions is closed under addition and multiplication.}

\revision{Our new kernel can be understood as implicitly performing a 1-in-$\mathscr{K}$ encoding.
Note that Kernel $K_{\text{mixed}}$ has one hyperparameter $\lambda_i$ for each input dimension.
By using a 1-in-$\mathscr{K}$ encoding and kernel $K_{\text{cont}}$ instead, we end up with one hyperparameter $\lambda_i$ for each \emph{encoded} dimension;
if we then reparameterize $K_{\text{cont}}$ to share a single hyperparameter $\lambda_l$ across the encoded dimensions resulting from a single original input dimension $l$,
we recover $K_{\text{mixed}}$.}

Since $K_{\text{mixed}}$ is rather expressive, one may worry about overfitting. Thus, we also experimented with two variations: (1)~sharing the same hyperparameter $\lambda$ across all input dimensions; and (2)~sharing $\lambda_1$ across algorithm parameters and $\lambda_2$ across instance features. We found that neither variation outperformed $K_{\text{mixed}}$.

\subsection{Scaling to Large Amounts of Data with Approximate Gaussian Processes}\label{sec:GPR-approx}

The time complexity of fitting Gaussian processes is cubic in the number of data points, which limits the amount of data that can be used in practice to fit these models.
To deal with this obstacle, the machine learning literature has proposed various approximations to Gaussian processes \citep[see, \eg{},][]{QuiRasWil07}. To the best of our knowledge, these approximate GPs have previously been applied to runtime prediction only in our work on parameter optimization \cite{HutHooLeyMur10-tbspo} (considering parameterized algorithms, but only single problem instances).
We experimented with the Bayesian committee machine~\cite{BCM}, the informative vector machine~\cite{IVM}, and the projected process (PP) approximation~\cite{RasWil06}. All of these methods performed similarly, with the PP approximation having a slight edge. Below, we give the equations for the PP's predictive mean and variance; for a derivation, see the \citet{RasWil06}.

The PP approximation to GPs uses a subset of $a$ of the $n$ training data points, the so-called \emph{active set}.
Let $v$ be a vector consisting of the indices of these $a$ data points.
We extend the notation for exact GPs (see Section \ref{sec:GPR}) as follows:
let $K_{\text{aa}}$ denote the $a$ by $a$ matrix with $K_{\text{aa}}(i,j) = k(\bm{x}_{v(i)},\bm{x}_{v(j)})$ and let $\bm{K_{\text{an}}}$ denote the $a$ by $n$ matrix with $\bm{K_{\text{an}}}(i,j) = k(\bm{x}_{v(i)},\bm{x}_j)$.
The predictive distribution of the PP approximation is then a normal distribution with mean and variance
\begin{eqnarray}
\nonumber{}\mu_{n+1} & = & \bm{k}{_*}^\transpose (\sigma^2 \bm{K_{\text{aa}}} + \bm{K_{\text{an}}} \bm{K_{\text{an}}}^\transpose)^{-1} \bm{K_{\text{an}}} y_{1:n}\\
\nonumber{}\Var_{n+1} & = & k_{**} - \bm{k}{_*}^{\transpose} \bm{K_{\text{aa}}}^{-1} \bm{k}{_*} + \sigma^2 \bm{k}{_*}^{\transpose} (\sigma^2 \bm{K_{\text{aa}}} + \bm{K_{\text{an}}} \bm{K_{\text{an}}}^\transpose)^{-1} \bm{k}{_*}.
\end{eqnarray}

We perform $h$ steps of hyperparameter optimization based on a standard GP, trained using a set of $a$ data points sampled uniformly at random without replacement from the $n$ input data points.
We then use the resulting hyperparameters and another independently sampled set of $a$ data points (sampled in the same way) for the subsequent PP approximation.
In both cases, if $a > n$, we only use $n$ data points.

The complexity of the PP approximation is superlinear only in $a$; therefore, the approach is much faster when we choose $a \ll n$.
The hyperparameter optimization based on $a$ data points takes time $O(h \cdot a^3)$.
In addition, there is a one-time cost of $O(a^2 \cdot n)$ for evaluating the PP equations.
Thus, the  time complexity for fitting the approximate GP model is $O( [h \cdot a + n] \cdot a^2)$,
as compared to $O(h \cdot n^3)$ for the exact GP model.
The time complexity for predictions with this PP approximation is $O(a)$ for the
mean and $O(a^2)$ for the variance of the predictive distribution~\cite{RasWil06},
as compared to $O(n)$ and $O(n^2)$, respectively, for the exact version. In our experiments, we set $a=300$ and $h=50$ to achieve a good compromise between speed and predictive accuracy.

\subsection{Random Forest Models}\label{sec:rf-def}

\revision{Regression trees, as discussed in Section \ref{sec:regression-trees}, are a flexible modeling technique that is particularly effective for discrete input data.
However, they are also well known to be sensitive to small changes in the data and are thus prone to overfitting.}
Random forests~\cite{Bre01} overcome this problem by combining multiple regression trees into an ensemble.
Known for their strong predictions for high-dimensional and discrete input data,
random forests are an obvious choice for runtime predictions of highly parameterized algorithms. Nevertheless, to the best of our knowledge, they have not been used for algorithm runtime prediction except in our own recent work on algorithm configuration~\cite{HutHooLeyMur10-tbspo,HutHooLey11-SMAC,HutHooLey11-censoredBO,HutHooLey12-ParallelAC}, which used a prototype implementation of the models we describe here.\footnote{\blue{Note that random forests have also been found to be effective in predicting the approximation ratio of 2-opt on Euclidean TSP instances~\cite{MerEtAl13}.}}
In the following, we describe the standard RF framework and some nonstandard implementation choices we made.

\subsubsection{The Standard Random Forest Framework}

A random forest (RF) consists of a set of regression trees. If grown to sufficient depths, regression trees are extraordinarily flexible predictors, able to capture very complex interactions and thus having low bias.
However, this means they can also have high variance: small changes in the
data can lead to a dramatically different tree.
Random forests~\cite{Bre01} reduce this variance by aggregating predictions across multiple different trees.  (This is an alternative to the pruning procedure described previously; thus, the trees in random forests are not pruned, but are rather grown until each node contains no more than $n_{\min}$ data points.)
These trees are made to be different by training them on different subsamples of the training data, and/or by permitting only a random subset of the variables as split variables at each node. We chose the latter option, using the full training set for each tree. (We did experiment with a combination of the two approaches, but found that it yielded slightly worse performance.)

Mean predictions for a new input $\bm{x}$ are trivial: predict the response for $\bm{x}$ with each tree and average the predictions.
The predictive quality improves as the number of trees, $B$, grows, but computational cost also grows linearly in $B$.
We used $B=10$ throughout our experiments to keep computational costs low.
Random forests have two additional hyperparameters: the percentage of variables to consider at each split point, \emph{perc},
and the minimal number of data points required in a node to make it eligible to be split further, $n_{\min}$.
We set $perc = 0.5$ and $n_{\min}=5$ by default.

\subsubsection{Modifications to Standard Random Forests}

We introduce a simple, yet effective, method for quantifying predictive uncertainty in random forests. (Our method is similar in spirit to that of \citet{Mei06}, who recently introduced quantile regression trees, which allow for predictions of quantiles of the predictive distribution; in contrast, we predict a mean and a variance.)
In each leaf of each regression tree, in addition to the empirical mean of the training data associated with that leaf, we store the empirical variance of that data. To avoid making deterministic predictions for leaves with few data points, we round the stored variance up to at least the constant $\sigma{^2}_{\min}$; we set $\sigma{^2}_{\min}=0.01$ throughout.
For any input, each regression tree $T_b$ thus yields a predictive mean $\mu_b$ and a predictive variance $\sigma^2_b$. To combine these estimates into a single estimate, we treat the forest as a mixture model of $B$ different models. We denote the random variable for the prediction of tree $T_b$ as $L_b$ and the overall prediction as $L$, and then have $L = L_b \;\; \textrm{if} \;\; Y=b,$
where $Y$ is a multinomial variable with $p(Y=i)=1/B$ for $i=1,\dots,B$.
The mean and variance for $L$ can then be expressed as:
\begin{eqnarray}
\nonumber{}\mu = \mathds{E}[L] & = & \frac{1}{B} \sum_{b=1}^B \mu_b;\\
\nonumber{}\sigma^2 = \Var(L) & = & \mathds{E}[\Var(L|Y)] + \Var(\mathds{E}[L|Y]) \\
\nonumber{}& = & \left(\frac{1}{B} \sum_{b=1}^B \sigma^2_b\right) + \Big(\mathds{E}[\mathds{E}(L|Y)^2] - \mathds{E}[\mathds{E}(L|Y)]^2\Big)\\
\nonumber{}& = & \left(\frac{1}{B} \sum_{b=1}^B \sigma^2_b\right) + \left(\frac{1}{B} \sum_{b=1}^B \mu^2_b\right) - \mathds{E}[L]^2\\
\nonumber{}& = & \left(\frac{1}{B} \sum_{b=1}^B \sigma^2_b + \mu^2_b \right) - \mu^2.
\end{eqnarray}

Thus, our predicted mean is simply the mean across the means predicted by the individual trees in the random forest.
To compute the variance prediction, we used the law of total variance~\citep[see, \eg{}, ][]{total_variance}, which allows us to write the total variance as the variance across the means predicted by the individual trees (predictions are uncertain if the trees disagree), plus the average variance of each tree (predictions are uncertain if the predictions made by individual trees tend to be uncertain).

A second non-standard ingredient in our models concerns the choice of split points.
Consider splits on a real-valued variable $j$.
Note that when the loss in Equation (\ref{eqn:loss_of_split}) is minimized by choosing
split point $s$ between the values of $\bm{x}_{k,j}$ and $\bm{x}_{l,j}$,
we are still free to choose the exact location of $s$ anywhere in the interval $(\bm{x}_{k,j}, \bm{x}_{l,j})$.
Traditionally, $s$ is chosen as the midpoint
between $\bm{x}_{k,j}$ and $\bm{x}_{l,j}$. Instead, here we draw it uniformly at random from $(\bm{x}_{k,j}, \bm{x}_{l,j})$.
In the limit of an infinite number of trees, this leads to a linear interpolation of the training data instead of
a partition into regions of constant prediction.
Furthermore, it causes variance estimates to vary smoothly and to grow with the distance from observed data points.

\hide{
Secondly, consider splits on a categorical variable $j$
at a node $N$ whose associated data does not contain all of $j$'s possible values; for example, say variable $j$ has domain $\{A, B, C\}$, but only values $A$ and $B$ are present in the data at $N$.
A standard solution is to only partition the values present at $N$ (here: $\{A,B\}$); new data points with a different value (here: $C$) would then be predicted using the mean and the variance of (the inner node) $N$.
In contrast, in our implementation, we assign each value that is not present in the training data at node $N$ to one of $N$'s children, chosen uniformly at random.
In expectation, this does not affect mean predictions, but it much better reflects our uncertainty.
This implementation has the convenient side effect that each tree partitions the input space in its leaves and we need not worry about special cases concerning regions associated with inner nodes.
}

\subsubsection{Complexity of Fitting Random Forests} \label{sec:complexity-rf}

The computational cost for fitting a random forest is relatively low. We need to fit $B$ regression trees, each of which is somewhat easier to fit than a normal regression tree, since at each node we only consider
$v = \max(1, \left\lfloor{} \textrm{\emph{perc}} \cdot p \right\rfloor)$ out of the $p$ possible split variables.
Building $B$ trees simply takes $B$ times as long as building a single tree. Thus---by the same argument as for regression trees---the
complexity of learning a random forest is $O(B \cdot v \cdot n^2 \cdot \log n)$ in the worst case (splitting off one data point at a time) and
$O(B \cdot v \cdot n \cdot \log^2 n)$ in the best case (perfectly balanced trees).
Our random forest implementation is based on a port of Matlab's regression tree code to C, which yielded
speedups of between one and two orders of magnitude.

Prediction with a random forest model entails predicting with $B$ regression trees (plus an $O(B)$ computation to compute the mean and variance across those predictions). The time complexity of a single prediction is thus $O(B \cdot n)$ in the worst case and $O(B \cdot \log n)$ for perfectly balanced trees.

\section{Problem-Specific Instance Features} \label{sec:eh_features}

While the methods we have discussed so far could be used to model the performance of any algorithm for solving any problem,
in our experiments, we investigated specific NP-complete problems. In particular, we considered the propositional satisfiability problem (SAT), mixed integer programming (MIP) problems, and the travelling salesperson problem (TSP). Our reasons for choosing these three problems are as follows.
SAT is the prototypical NP-hard decision problem and is thus interesting from a theory perspective; modern SAT solvers are also
one of the most prominent approaches in hardware and software verification~\cite{prasad05survey}.
MIP is a canonical representation for constrained optimization problems with integer-valued and continuous variables, which serves as a unifying framework for NP-complete problems and combines the expressive power of integrality constraints with the efficiency of continuous optimization. As a consequence, it is very widely used both in academia and industry~\cite{cplex-datasheet}.
Finally, TSP is one of the most widely studied NP-hard optimization problems, and also of considerable interest for industry~\cite{url:tsp_applications}.

We tailor EPMs to a particular \revision{problem} through the choice of instance features.\footnote{If features are unavailable for an NP-complete problem of interest, one alternative is to reduce the problem to SAT, MIP, or TSP---a polynomial-time operation---and then compute some of the features we describe here. We do not expect this approach to be computationally efficient, but do observe that it extends the reach of existing EPM construction techniques to any NP-complete problem.}
Here we describe comprehensive sets of features for SAT, MIP, and TSP. \revision{For each of these problems, we summarize \blue{sets of} features found in the literature
and introduce many novel features.}
While all these features are polynomial-time computable, we note that some of them can be computationally expensive for very large instances (\eg{}, taking cubic time).
\revision{For some applications such expensive features will be reasonable---in particular, we note that for applications that take features as a one-time input, but build models repeatedly, it can even make sense to use features whose cost exceeds that of solving the instance; examples of such applications include model-based algorithm configuration~\cite{HutHooLey11-SMAC} and complex empirical analyses based on performance predictions~\cite{HutHooLey10-amai,HutHooLey13ImportanceFwdSel}. In runtime-sensitive applications, on the other hand, it may make sense to use only a subset of the features described here. To facilitate this, we categorize all features into one of four ``cost classes'': trivial, cheap, moderate, and expensive. In our experimental evaluation, we report the empirical cost of these feature classes and the predictive performance that can be achieved using them (see Table \ref{tab:feature_space_featuresubsetexp} on page \pageref{tab:feature_space_featuresubsetexp}). We also identify features introduced
 in this work and quantify their contributions to model performance.}

\emph{Probing features} are a \revision{generic} family of features that deserves special mention.
They are computed by briefly running an existing algorithm for the given problem on the given instance and extracting characteristics from that algorithm's trajectory---an idea closely related to that of landmarking in meta-learning~\cite{Pfahringer00meta-learning}. Probing features can be defined with little effort for a wide variety of problems;
indeed, in earlier work, we introduced the first probing features for SAT \cite{NudLeyDevShoHoo04} \revision{and showed that probing features based on one type of algorithm (\eg{}, local search) are often useful for predicting the performance of another type of algorithm (\eg{}, tree search).}
Here we introduce the first probing features for MIP and TSP.
\revision{Another new, generic family of features are \emph{timing features}, which measure the time other groups of features take to compute.}
Code and binaries for computing all our features, along with documentation providing additional details, are available online at {\footnotesize\url{http://www.cs.ubc.ca/labs/beta/Projects/EPMs/}}.

\subsection{Features for Propositional Satisfiability (SAT)}\label{sec:sat-features}

\begin{figure}[tp]
{\fontfamily{cmss}\fontseries{c}\fontsize{8pt}{8pt}\selectfont
\setlength{\columnsep}{0.8cm}
\begin{multicols}{2}[][0]

\textbf{Problem Size Features:}
\begin{enumerate}
\denselist
\item[1--2.] \textbf{Number of variables and clauses in original formula (trivial)}: denoted \texttt{v} and \texttt{c}, respectively
\item[3--4.] \textbf{Number of variables and clauses after simplification with \satelite{} (cheap)}: denoted \texttt{v}' and \texttt{c}', respectively
\item[5--6.] \textbf{Reduction of variables and clauses by simplification (cheap)}: (\texttt{v}-\texttt{v}')/\texttt{v}' and (\texttt{c}-\texttt{c}')/\texttt{c}'
\item[7.] \textbf{Ratio of variables to clauses (cheap)}: \texttt{v}'/\texttt{c}'
\end{enumerate}

\textbf{Variable-Clause Graph Features:}
\begin{enumerate}
\denselist
\item[8--12.] \textbf{Variable node degree statistics (expensive):} mean, variation
coefficient, min, max, and entropy
\item[13--17.] \textbf{Clause node degree statistics (cheap):} mean, variation
coefficient, min, max, and entropy
\end{enumerate}

\textbf{Variable Graph Features (expensive):}
\begin{enumerate}
\denselist
\item[18--21.] \textbf{Node degree statistics:} mean, variation coefficient, min, and
 max
 \item[22--26.] \textbf{Diameter$^*$:} mean, variation coefficient, min, max, and entropy
\end{enumerate}

\textbf{Clause Graph Features (expensive):}
\begin{enumerate}
\denselist
\item[27--31.] \textbf{Node degree statistics:} mean, variation coefficient, min, max, and entropy
\item[32--36.] \textbf{Clustering Coefficient*:} mean, variation coefficient, min, max, and entropy
\end{enumerate}

\textbf{Balance Features:}
\begin{enumerate}
\denselist
\item[37--41.] \textbf{Ratio of positive to negative literals in each
clause (cheap):} mean, variation coefficient, min, max, and entropy
\item[42--46.] \textbf{Ratio of positive to negative occurrences of each
variable (expensive):} mean, variation coefficient, min, max, and entropy
\item[47--49.] \textbf{Fraction of unary, binary, and ternary clauses (cheap)}
\end{enumerate}

\textbf{Proximity to Horn Formula (expensive):}
\begin{enumerate}
\denselist
\item[50.] \textbf{Fraction of Horn clauses}
\item[51--55.] \textbf{Number of occurrences in a Horn clause for each
variable:} mean, variation coefficient, min, max, and entropy
\end{enumerate}

\textbf{DPLL Probing Features:}
\begin{enumerate}
\denselist
\item[56--60.] \textbf{Number of unit propagations (expensive):} computed at depths 1,
4, 16, 64 and 256
\item[61--62.] \textbf{Search space size estimate (cheap):} mean depth to
contradiction, estimate of the log of number of nodes
\end{enumerate}

\newpage

\textbf{LP-Based Features (moderate):}
\begin{enumerate}
\denselist
\item[63--66.] \textbf{Integer slack vector :} mean, variation coefficient, min, and max
\item[67.] \textbf{Ratio of integer vars in LP solution }
\item[68.] \textbf{Objective value of LP solution }
\end{enumerate}

\textbf{Local Search Probing Features, based on 2 seconds of running each of SAPS and GSAT (cheap):}
\begin{enumerate}
\denselist
\item[69--78.] \textbf{Number of steps to the best local minimum in a run:}
mean, median, variation coefficient, 10th and 90th percentiles
\item[79--82.] \textbf{Average improvement to best in a run:} mean and coefficient of variation of improvement per step to best solution
\item[83--86.] \textbf{Fraction of improvement due to first local minimum:}
mean and variation coefficient
\item[87--90.] \textbf{Best solution:} mean and variation coefficient
\end{enumerate}

\textbf{Clause Learning Features$^*$ (based on 2 seconds of running \textsc{Zchaff\_rand}; cheap):}
\begin{enumerate}
\denselist
\item[91--99.] \textbf{Number of learned clauses:} mean, variation coefficient, min, max, 10\%, 25\%, 50\%, 75\%, and 90\% quantiles
\item[100--108.] \textbf{Length of learned clause:} mean, variation coefficient, min, max, 10\%, 25\%, 50\%, 75\%, and 90\% quantiles
\end{enumerate}

\textbf{Survey Propagation Features$^*$ (moderate)}
\begin{enumerate}
\denselist
\item[109--117.] \textbf{Confidence of survey propagation:} For each variable, compute the higher of $P(true)/P(false)$ or $P(false)/P(true)$. Then compute statistics across variables: mean, variation coefficient, min, max, 10\%, 25\%, 50\%, 75\%, and 90\% quantiles
\item[118--126.] \textbf{Unconstrained variables:} For each variable, compute $P(unconstrained)$. Then compute statistics across variables: mean, variation coefficient, min, max, 10\%, 25\%, 50\%, 75\%, and 90\% quantiles
\end{enumerate}

\textbf{Timing Features*}
\begin{enumerate}
\denselist
\item[127--138.] \textbf{CPU time required for feature computation:} one feature for each of 12 subsets of features
(see text for details)
\end{enumerate}%
\end{multicols}%
}%
\vspace{-1.8em}\caption{\small SAT instance features. New features are marked with $^*$. \label{fig:sat-features}}%
\end{figure}

Figure \ref{fig:sat-features} summarizes 138 features for SAT.
Since various preprocessing techniques are routinely used before applying a general-purpose SAT solver
and typically lead to substantial reductions in instance size and difficulty
(especially for industrial-like instances), we apply the preprocessing procedure \satelite~\cite{satelite} on all
instances first, and then compute instance features on the
preprocessed instances.
The first 90 features, with the exception of features
22--26 and 32--36, were introduced in our previously published work on \satzilla{}~\cite{NudLeyDevShoHoo04,SATzilla-Full}.
They can be
categorized as \emph{problem size} features
(1--7), \emph{graph-based} features (8--36), \emph{balance} features
(37--49), \emph{proximity to Horn formula} features (50--55),
\emph{DPLL probing} features (56--62), \emph{LP-based} features (63--68), and \emph{local search probing} features (69--90).

\revision{Our new features (devised over the last five years in our ongoing work on \satzilla{} and so far only mentioned in short solver descriptions~\cite{SATzilla2009,Satzilla2012}) fall into four categories.
First, we added two additional subgroups of graph-based features.}
Our new \emph{diameter} features 22--26 are based on the variable graph~\cite{paul:ms}. For each node $i$ in that graph, we compute the longest shortest path between $i$ and any other node. As with most of the features that follow, we then compute various statistics over this vector (\eg{}, mean, max); we do not state the exact statistics for each vector below but list them in Figure \ref{fig:sat-features}.
Our new \emph{clustering coefficient} features 32--36 measure the local cliqueness of the clause graph.
For each node in the clause graph, let $p$ denote the number of edges present between the node and its neighbours, and let $m$ denote the
maximum possible number of such edges; we compute $p/m$ for each node.

Second, our new \emph{clause learning}
features (91--108) are based on statistics gathered in 2-second runs of {\smaller\texttt{Zchaff\_rand}}~\cite{zchaff_rand}.
We measure the number of learned clauses (features 91--99) and the length of the learned clauses
(features 100--108) after every 1000 search steps.
Third, our new \emph{survey propagation} features (109--126) are based on
estimates of variable bias in a SAT formula obtained using probabilistic inference~\cite{hsu-cp08}.
We used {\smaller\texttt{VARSAT}}'s implementation to estimate the probabilities that each variable is true in every satisfying assignment, false in every satisfying assignment, or unconstrained. Features
109--117 measure the confidence of survey propagation
(that is, $\max(P_\text{true}(i)/P_\text{false}(i), P_\text{false}(i)/P_\text{true}(i))$ for each variable $i$) and
features 118--126 are based on the $P_\text{unconstrained}$ vector.

Finally, our new \emph{timing features} (127--138) measure the time taken by 12 different blocks of feature computation code: instance preprocessing by \satelite, problem size (1--6), variable-clause graph (clause node) and balance features (7, 13--17, 37--41, 47--49); variable-clause graph (variable node), variable graph and proximity to Horn formula features (8--12, 18--21, 42--46, 50--55); diameter-based features (22--26); clause graph features (27--36); unit propagation features (56--60); search space size estimation (61--62); LP-based features (63--68); local search probing features (69--90) with SAPS and GSAT; clause learning features (91--108); and survey propagation features (109--126).

\subsection{Features for Mixed Integer Programs}\label{sec:mip-features}

\begin{figure}[tp]
{\fontfamily{cmss}\fontseries{c}\fontsize{8pt}{8pt}\selectfont
\setlength{\columnsep}{0.8cm}

\begin{multicols}{2}[][0]

\textbf{Problem Type (trivial):}
\begin{enumerate}
\denselist
\item[1.] {\bf Problem type}: LP, MILP, FIXEDMILP, QP, MIQP, FIXEDMIQP, MIQP, QCP, or MIQCP, as attributed by \cplex{}
\end{enumerate}

\textbf{Problem Size Features (trivial):}
\begin{enumerate} \setcounter{enumi}{2}
\denselist
\item[2--3.] \textbf{Number of variables and constraints}: denoted $n$ and $m$, respectively
\item[4.] \textbf{Number of non-zero entries in the linear constraint matrix, $\bm{A}$}
\item[5--6.] \textbf{Quadratic variables and constraints}: number of variables with quadratic constraints and number of quadratic constraints
\item[7.] \textbf{Number of non-zero entries in the quadratic constraint matrix, $\bm{Q}$}
\item[8--12.] \textbf{Number of variables of type}: Boolean, integer, continuous, semi-continuous, semi-integer
\item[13--17.] \textbf{Fraction of variables of type} (summing to 1): Boolean, integer, continuous, semi-continuous, semi-integer
\item[18-19.] \textbf{Number and fraction of non-continuous variables} (counting Boolean, integer, semi-continuous, and semi-integer variables)
\item[20-21.] \textbf{Number and fraction of unbounded non-continuous variables}: fraction of non-continuous variables that has infinite lower or upper bound
\item[22-25.] \textbf{Support size}: mean, median, vc, q90/10 for vector composed of the following values for bounded variables:
domain size for binary/integer, 2 for semi-continuous, 1+domain size for semi-integer variables.
\end{enumerate}

\textbf{Variable-Constraint Graph Features (cheap):} each feature is replicated three times, for $X\in \{C,NC,V\}$
\begin{enumerate}
\denselist
\item[26--37.] {\bf Variable node degree statistics}: characteristics of vector $(\sum_{c_j \in C} \indicator(A_{i,j} \neq 0))_{x_i \in X}$: mean, median, vc, q90/10
\item[38--49.] {\bf Constraint node degree statistics}: characteristics of vector $(\sum_{x_i \in X} \indicator(A_{i,j} \neq 0))_{c_j \in C}$: mean, median, vc, q90/10
\end{enumerate}

\textbf{Linear Constraint Matrix Features (cheap):} each feature is replicated three times, for $X\in \{C,NC,V\}$
\begin{enumerate}
\denselist
\item[50--55.] {\bf Variable coefficient statistics}: characteristics of vector $(\sum_{c_j \in C} A_{i,j})_{x_i \in X}$: mean, vc
\item[56--61.] {\bf Constraint coefficient statistics}: characteristics of vector $(\sum_{x_i \in X} A_{i,j})_{c_j \in C}$: mean, vc
\item[62--67.] {\bf Distribution of normalized constraint matrix entries, $\bm{A}_{i,j}/b_i$}: mean and vc (only of elements where $b_i \neq 0$)
\item[68--73.] {\bf Variation coefficient of normalized absolute non-zero entries per row} (the normalization is by dividing by sum of the row's absolute values):
mean, vc 
\end{enumerate}

\textbf{Objective Function Features (cheap):} each feature is replicated three times, for $X\in \{C,NC,V\}$
\begin{enumerate}
\denselist
\item[74-79.] {\bf Absolute objective function coefficients} $\{|c_i|\}_{i=1}^n$: mean and stddev
\item[80-85.] {\bf Normalized absolute objective function coefficients $\{|c_i|/n_i\}_{i=1}^n$}, where $n_i$ denotes the number of non-zero entries in column $i$ of $\bm{A}$: mean and stddev
\item[86-91.] {\bf squareroot-normalized absolute objective function coefficients $\{|c_i|/\sqrt{n_i}\}_{i=1}^n$}: mean and stddev
\end{enumerate}

\textbf{LP-Based Features (expensive):}
\begin{enumerate}
\denselist
\item[92--94.] \textbf{Integer slack vector:} mean, max, $L_2$ norm
\item[95.] \textbf{Objective function value of LP solution}
\end{enumerate}

\textbf{Right-hand Side Features (trivial):}
\begin{enumerate}
\denselist
\item[96-97.] {\bf Right-hand side for $\le$ constraints}: mean and stddev
\item[98-99.] {\bf Right-hand side for $=$ constraints}: mean and stddev
\item[100-101.] {\bf Right-hand side for $\ge$ constraints}: mean and stddev
\end{enumerate}

\textbf{Presolving Features$^*$ (moderate):}
\begin{enumerate}
\denselist
\item[102-103.] {\bf CPU times}: presolving and relaxation CPU time
\item[104-107.] {\bf Presolving result features:} \# of constraints, variables, non-zero entries in the constraint matrix, and clique table inequalities after presolving.
\end{enumerate}

\textbf{Probing Cut Usage Features$^*$ (moderate):}
\begin{enumerate}
\denselist
\item[108-112.] {\bf Number of specific cuts}: clique cuts, Gomory fractional cuts, mixed integer rounding cuts, implied bound cuts, flow cuts
\end{enumerate}

\textbf{Probing Result features$^*$ (moderate):}
\begin{enumerate}
\denselist
\item[113-116.] {\bf Performance progress}: MIP gap achieved, \# new incumbent found by primal heuristics,  \# of feasible solutions found, \# of solutions or incumbents found
\end{enumerate}

\textbf{Timing Features*}
\begin{enumerate}
\denselist
\item[117--121.] \textbf{CPU time required for feature computation:} one feature for each of 5 groups of features (see text for details)
\end{enumerate}
\end{multicols}
}\vspace{-1em}\caption{\small MIP instance features; for the variable-constraint graph, linear constraint matrix, and objective function features, each feature is computed with respect to three subsets of variables: continuous, $C$, non-continuous, $NC$, and all, $V$. Features introduced for the first time are marked with $^*$.\label{fig:mip-featurelist}}
\end{figure}

Figure \ref{fig:mip-featurelist} summarizes 121 features for mixed integer programs (\ie{}, MIP instances).
These include 101 features based on existing work~\cite{LeyNudSho09:jacm,Hut09:phd,isac}, 15 new \emph{probing} features, and 5 new \emph{timing} features.
Features 1--101 are primarily based on features for the combinatorial winner determination problem from our past work \cite{LeyNudSho09:jacm}, generalized to MIP and previously only described in a Ph.D.\ thesis~\cite{Hut09:phd}.
These features can be categorized as \emph{problem type \& size} features (1--25),
\emph{variable-constraint graph} features (26--49), \emph{linear constraint matrix} features (50--73),
\emph{objective function} features (74--91), and \emph{LP-based} features
(92--95).
We also integrated ideas from the feature set used by \emcite{isac} (\emph{right-hand side} features (96--101) and the computation of separate statistics for continuous variables, non-continuous variables, and their union).
We extended existing features by adding richer statistics where applicable: medians, variation coefficients (vc), and
percentile ratios (q90/q10) of vector-based features.

\revision{We introduce two new sets of features. Firstly, our new \emph{MIP probing features} 102--116 are based on 5-second runs of \cplex{} with default settings. They are obtained via the \cplex{} API and include 6 \emph{presolving} features based on the output of \cplex{}'s presolving phase (102--107);
5 \emph{probing cut usage} features describing the different cuts \cplex{} used during probing (108--112);
and 4 \emph{probing result} features summarizing probing runs (113--116).
\hide{
The remaining 22 \emph{progress over time} features are based on vectors tracking the progress made by \cplex{} by measuring the trajectory of
six different quantities: the number of integer-infeasible variables, the number of nodes left to be explored, the objective value of the LP relaxation of the current node, the best integer feasible solution found so far,
the best known dual bound, and the MIP gap.
For each of these quantities, we parsed the corresponding entry from each line of \cplex{}'s
log file output. For the number of integer-infeasible variables and the number of nodes left to be explored
(122--127), the vector contains one element for each line. The remaining four quantities
(128--143) measure improvement over time and their respective vectors thus contain the difference
between two successive entries.}
Secondly, our new \emph{timing features} 117--121 capture the CPU time required for computing five different groups of features: variable-constraint graph, linear constraint matrix, and objective features for three subsets of variables (``continuous'', ``non-continuous'', and ``all'', 26--91); LP-based features (92--95); and CPLEX probing features (102--116). The cost of computing the remaining features (1--25, 96--101) is small (linear in the number of variables or constraints).}

\subsection{Features for the Travelling Salesperson Problem (TSP)}\label{sec:tsp-features}

\begin{figure}[tp]
{\fontfamily{cmss}\fontseries{c}\fontsize{8pt}{8pt}\selectfont
\setlength{\columnsep}{0.8cm}
\begin{multicols}{2}[][0]

\textbf{Problem Size Features$^*$ (trivial):}
\begin{enumerate}
\denselist
\item[1.] \textbf{Number of nodes}: denoted $n$
\end{enumerate}

\textbf{Cost Matrix Features$^*$ (trivial):}
\begin{enumerate}
\denselist
\item[2--4.] \textbf{Cost statistics:} mean, variation coefficient,
skew
\end{enumerate}

\textbf{Minimum Spanning Tree Features$^*$ (trivial):}
\begin{enumerate}
\denselist
\item[5--8.] \textbf{Cost statistics:} sum, mean, variation coefficient,
skew
 \item[9--11.] \textbf{Node degree statistics:} mean, variation coefficient,
 skew
\end{enumerate}

\textbf{Cluster Distance Features$^*$ (moderate):}
\begin{enumerate}
\denselist
\item[12--14.] \textbf{Cluster distance:} mean, variation coefficient,
skew
\end{enumerate}

\textbf{Local Search Probing Features$^*$ (expensive):}
\begin{enumerate}
\denselist
\item[15--17.] \textbf{Tour cost from construction heuristic:} mean, variation coefficient,
skew
\item[18--20.] \textbf{Local minimum tour length:} mean, variation coefficient,
skew
\item[21--23.] \textbf{Improvement per step:} mean, variation coefficient,
skew
\item[24--26.] \textbf{Steps to local minimum}: mean, variation coefficient,
skew
\item[27--29.] \textbf{Distance between local minima}: mean, variation coefficient,
skew
\item[30--32.] \textbf{Probability of edges in local minima}: mean, variation coefficient,
skew
\end{enumerate}

\textbf{Branch and Cut Probing Features$^*$ (moderate):}
\begin{enumerate}
\denselist
\item[33--35.] \textbf{Improvement per cut:} mean, variation coefficient, skew
\item[36.] \textbf{Ratio of upper bound and lower bound}
\item[37--43.] \textbf{Solution after probing:} Percentage of integer values and non-integer values in the final solution after probing. For non-integer
values, we compute statics across nodes: min,max, 25\%,50\%, 75\%
quantiles
\end{enumerate}

\textbf{Ruggedness of Search Landscape$^*$ (cheap):}
\begin{enumerate}
\denselist
\item[44.] \textbf{Autocorrelation coefficient}
\end{enumerate}

\textbf{Timing Features*}
\begin{enumerate}
\denselist
\item[45--50.] \textbf{CPU time required for feature computation:} one feature for each of 6 groups (see text)
\end{enumerate}

\textbf{Node Distribution Features (after instance normalization, moderate)}
\begin{enumerate}
\denselist
\item[51.] \textbf{Cost matrix standard deviation:} standard deviation of cost matrix after instance has been normalized to the rectangle
                $[(0,0), (400, 400)]$.
\item[52--55.] \textbf{Fraction of distinct distances}: precision to 1, 2, 3, 4 decimal places
\item[56--57.] \textbf{Centroid:} the $(x, y)$ coordinates of the instance centroid
\item[58.] \textbf{Radius:} the mean distances from each node to the centroid
\item[59.] \textbf{Area:} the are of the rectangle in which nodes lie
\item[60--61.] \textbf{nNNd}: the standard deviation and coefficient variation of the normalized nearest neighbour distance
\item[62--64.] \textbf{Cluster:} \#clusters / $n$ , \#outliers / $n$, variation of \#nodes in clusters
\end{enumerate}

\end{multicols}
}\vspace{-1em}\caption{\small TSP instance features. Features introduced for the first time are marked with $^*$. \label{fig:tsp-features}}
\end{figure}

Figure \ref{fig:tsp-features} summarizes 64 features for the travelling salesperson problem (TSP). Features 1--50 are new, while Features 51--64 were introduced by \emcite{tspfeat10}.
Features 51--64 capture the spatial distribution of nodes (features 51--61) and clustering of nodes
(features 62--64);
we used the authors' code (available at {\smaller{\url{http://www.vanhemert.co.uk/files/}}}\-{\smaller{\url{TSP-feature-extract-20120212.tar.gz}}}) to compute these features.

\revision{Our 50 new TSP features are as follows.\footnote{\blue{In independent work, \citet{MerEtAl13} have introduced feature sets similar to some of those described here.}}
The \emph{problem size} feature (1) is the number of nodes in the
given TSP. The \emph{cost matrix} features (2--4) are statistics of
the cost between two nodes. Our \emph{minimum spanning tree}
features (5--11) are based on constructing a minimum spanning tree over all
nodes in the TSP: features 5--8 are the statistics of the edge costs
in the tree and features 9--11 are based on its node degrees.
Our \emph{cluster distance} features (12--14) are based on the cluster distance
between every pair of nodes, which is the minimum bottleneck
cost of any path between them; here, the bottleneck cost of a path is defined as the largest cost along the path.
Our \emph{local search probing} features (15--32) are based on
20 short runs (1000 steps each) of LK~\cite{LK}, using the implementation available from~\cite{url:LK-code}.
Specifically, features 15--17 are based on the tour length obtained by LK;
features 18--20, 21--23, and
24--26 are based on the tour length of local minima, the tour quality improvement per search step, and the number of search steps to reach a local minimum, respectively; features 27--29 measure the
Hamming distance between two local minima; and features 30--32 describe
the probability of edges appearing in any local minimum encountered
during probing.
Our \emph{branch and cut probing} features (33--43) are based on 2-second runs of {\smaller\texttt{Concorde}}.
Specifically, features 33--35 measure the improvement of lower bound per cut; feature 36 is the ratio of upper and lower bound at the end of the probing run; and features 37--43 analyze the final LP solution.
Feature 44 is the autocorrelation coefficient: a measure of the
ruggedness of the search landscape, based on an uninformed
random walk~(see, \eg{}, \cite{HooStu05}).
Finally, our \emph{timing features} 45--50 measure the CPU time required for computing feature groups 2--7 (the cost of computing the number of nodes can be ignored).}

\section{Performance Predictions for New Instances} \label{sec:ehms_better}

\revision{
We now
study the performance of the models described in Sections \ref{sec:eh_models_old_methods} and \ref{sec:eh_models_new}, using (various subsets of) the features described in Section \ref{sec:eh_features}.
In this section, we consider the (only) problem considered by most past work: predicting the performance achieved
by the default configuration of a given algorithm on new instances.
(We go on to consider making predictions for novel algorithm configurations in Sections \ref{sec:rms_better} and \ref{sec:combination_ehm_rsm}.)
For brevity, we only present representative empirical results. The full results of our experiments are available in an online appendix at {\footnotesize\url{http://www.cs.ubc.ca/labs/beta/Projects/EPMs}}. All of our data, features, and source code for replicating our experiments is available
from the same site.
}

\begin{table}[t]
\setlength{\tabcolsep}{3pt}
    {\footnotesize
\centering
    \begin{tabular}{ccl}
\toprule
\textbf{Abbreviation} & \textbf{Reference Section} & \textbf{Description}\\
            \midrule
RR & \ref{sec:linear-reg} & Ridge regression with 2-phase forward selection\\
SP & \ref{sec:linear-reg} & SPORE-FoBa (ridge regression with forward-backward selection)\\
NN & \ref{sec:nns} & Feed-forward neural network with one hidden layer\\
PP & \ref{sec:GPR-approx} & Projected process (approximate Gaussian process)\\
RT & \ref{sec:regression-trees} & Regression tree with cost-complexity pruning\\
RF & \ref{sec:rf-def} & Random forest\\
\bottomrule
        \end{tabular}
  \vspace{-1em}
        \small\caption{Overview of our models.\label{tab:all-models}}
}
\end{table}

\subsection{Instances and Solvers} \label{sec:exp_setup}

For SAT, we used a wide range of instance distributions: \INDU{}, \HAND{}, and \RAND{} are collections of industrial, handmade, and random instances from the international SAT competitions and races, and  \INDUHANDRAND{} is their union;
\SWV{} and \IBM{} are sets of software and hardware verification instances, and \SWVIBM{} is their union;
\RANDSAT{} is a subset of \RAND{} containing only satisfiable instances. We give more details about these distributions in \ref{app:benchmarks-sat}.
For all distributions except \RANDSAT{}, we ran the popular tree search solver, \minisat~\cite{minisat}.
For \INDU, \SWV{} and \IBM, we also ran two additional solvers: \cryptominisat{}~\cite{CryptoMiniSat2010} (which won SAT Race 2010 and received gold and silver medals in the 2011 SAT competition) and \spear{}~\cite{BabHut07} (which has shown state-of-the-art performance on \IBM{} and \SWV{} with optimized parameter settings~\cite{HutBabHooHu07}).
Finally, to evaluate predictions for local search algorithms, we used the \RANDSAT{} instances, and considered two solvers: \tnm{}~\cite{TNM2009} (which won the random satisfiable category of the 2009 SAT Competition) and the dynamic local search algorithm \saps{}~\cite{HutTomHoo02} (a baseline).

For MIP, we used two instance distributions from computational sustainability (\RCW{} and \CORLAT{}), one from winner determination in combinatorial auctions (\REG{}), two unions of these (\CORLATREG{} := \CORLAT{} $\cup$ \RCW{} and \CORLATREGRCW{} := \CORLAT{} $\cup$ \REG{} $\cup$ \RCW{}),
and a large and diverse set of publicly available MIP instances (\BIGMIX{}).
Details about these distributions are given in \ref{app:benchmarks-mip}.
We used the two state-of-the-art commercial solvers \cplex{}~\cite{url:CPLEX2} and \gurobi{}~\cite{url:gurobi} (versions 12.1 and 2.0, respectively) and the two strongest non-commercial solvers, \scip{}~\cite{url:scip2} and \lpsolve{}~\cite{url:lpsolve} (versions 1.2.1.4 and 5.5, respectively).

For TSP, we used three instance distributions (detailed in \ref{app:benchmarks-tsp}): random uniform Euclidean instances (\PORTGEN{}),
random clustered Euclidean instances (\PORTCGEN{}), and \TSPLIB{}, a heterogeneous set of prominent TSP instances.
On these instance sets, we ran the state-of-the-art systematic and local search algorithms, \concorde{}~\cite{Applegate06book} and \lkh{}~\cite{helsgaun00}.
For the latter, we computed runtimes as the time required to find an optimal solution.

\subsection{Experimental Setup} \label{sec:exp_setup_eh}

To collect algorithm runtime data, for each \mbox{algorithm--distribution} pair, we executed the algorithm using default parameters on all instances of the distribution, measured its runtimes, and collected the results in a database.
All algorithm runs were executed on a cluster of 55 dual 3.2GHz Intel Xeon PCs with 2MB cache and 2GB RAM, running OpenSuSE Linux 11.1; runtimes were measured as CPU time on these reference machines.
We terminated each algorithm run after one CPU hour; this gave rise to \emph{capped} runtime observations, because for each run that was
terminated in this fashion, we only observed a lower bound on the runtime. Like most past work on runtime modeling, we simply counted such capped runs as having taken one hour. (In Section \ref{sec:censoring} we investigate alternatives and conclude that a better treatment of capped runtime data improves predictive performance for our best-performing model.)
\revision{Basic statistics of the resulting runtime distributions are given in Table \ref{tab:feature_space_featuresubsetexp}; Table C.1 in the online appendix lists all the details.}

We evaluated different model families by building models on a subset of the data and assessing their performance
on data that had not been used to train the models. This can be done visually (as, \eg{}, in the scatterplots in Figure \ref{fig:feature_space_def_2fold_cv} on Page \pageref{fig:feature_space_def_2fold_cv}, which show cross-validated predictions for a random subset of up to 1\,000 data points), or quantitatively.
We considered three complementary quantitative metrics to evaluate mean predictions $\mu_1, \ldots, \mu_n$ and
predictive variances $\sigma^2_1, \ldots, \sigma^2_n$ given true performance values $y_1, \ldots, y_n$.
\emph{Root mean squared error (RMSE)} is defined as $\sqrt{1/n \sum_{i=1}^n (y_i - \mu_i)^2}$; \emph{Pearson's correlation coefficient (CC)} is defined as $(\sum_{i=1}^n (\mu_i y_i) - n\cdot\bar{\mu}\cdot\bar{y}) / ( (n-1) \cdot s_\mu \cdot s_y)$,
where $\bar{x}$ and $s_x$ denote sample mean and standard deviation of $x$; and
\emph{log likelihood (LL)} is defined as $\sum_{i=1}^n \log \varphi( \frac{y_i-\mu_i}{\sigma_i} )$, where $\varphi$ denotes the probability density function (PDF) of a standard normal distribution. Intui\-tively, LL is the log probability of observing the true values $y_i$ under the predicted distributions $\mathcal{N}(\mu_i, \sigma^2_i)$.
For CC and LL, higher values are better, while for RMSE lower values are better.
We used $10$-fold cross-validation and report means of these measures across the $10$ folds.
\revision{We assessed the statistical significance of our findings using a Wilcoxon signed-rank test (we use this paired test, since cross-validation folds are correlated).} 

\subsection{Predictive Quality} \label{sec:basic_pred_perf}

\begin{table}[tp]
\setlength{\tabcolsep}{1.8pt}
    {\scriptsize
\centering
\begin{tabular}{l@{\hskip 1.6em}cccccc@{\hskip 3.5em}cccccc}
    \toprule
 & \multicolumn{6}{c}{\textbf{RMSE}} & \multicolumn{6}{c}{\textbf{Time to learn model (s)}}\\
\cmidrule(r{2.25em}){2-7}\cmidrule{8-13}
\textbf{Domain} &\textbf{RR} &\textbf{SP} &\textbf{NN} &\textbf{PP} &\textbf{RT} &\textbf{RF} &\textbf{RR} &\textbf{SP} &\textbf{NN} &\textbf{PP} &\textbf{RT} &\textbf{RF}\\
\cmidrule(r{2.25em}){1-1} \cmidrule(r{2.25em}){2-7}\cmidrule{8-13}
\minisat{}-\INDUHANDRAND{} &1.01 &1.25 &0.62 &0.92 &0.68 &\textbf{0.47} & \textbf{6.8} &28.08 &21.84 &46.56 &20.96 &22.42\\
\minisat{}-\HAND{} &1.05 &1.34 &0.63 &0.85 &0.75 &\textbf{0.51} & \textbf{3.7} &7.92 &6.2 &44.14 &6.15 &5.98\\
\minisat{}-\RAND{} &0.64 &0.76 &\textbf{0.38} &0.55 &0.5 &\textbf{0.37} & \textbf{4.46} &7.98 &10.81 &46.09 &7.15 &8.36\\
\minisat{}-\INDU{} &0.94 &1.01 &0.78 &0.86 &0.71 &\textbf{0.52} & \textbf{3.68} &7.82 &5.57 &48.12 &6.36 &4.42\\
\minisat-\SWVIBM{} &0.53 &0.76 &0.32 &0.52 &0.25 &\textbf{0.17} & 3.51 &6.35 &4.68 &51.67 &4.8 &\textbf{2.78}\\
\minisat{}-\IBM{} &0.51 &0.71 &0.29 &0.34 &0.3 &\textbf{0.19} & 3.2 &5.17 &2.6 &46.16 &2.47 &\textbf{1.5}\\
\minisat{}-\SWV{} &0.35 &0.31 &0.16 &0.1 &0.1 &\textbf{0.08} & 3.06 &4.9 &2.05 &53.11 &2.37 &\textbf{1.07}\\
\addlinespace[\interrowspace]
\cryptominisat-\INDU{} &0.94 &0.99 &0.94 &0.9 &0.91 &\textbf{0.72} & \textbf{3.65} &7.9 &5.37 &45.82 &5.03 &4.14\\
\cryptominisat-\SWVIBM{} &0.77 &0.85 &0.66 &0.83 &0.62 &\textbf{0.48} & 3.5 &10.83 &4.49 &48.99 &4.75 &\textbf{2.78}\\
\cryptominisat-\IBM{} &0.65 &0.96 &0.55 &0.56 &0.53 &\textbf{0.41} & 3.19 &4.86 &2.59 &44.9 &2.41 &\textbf{1.49}\\
\cryptominisat-\SWV{} &0.76 &0.78 &0.71 &0.66 &0.63 &\textbf{0.51} & 3.09 &4.62 &2.09 &53.85 &2.32 &\textbf{1.03}\\
\addlinespace[\interrowspace]
\spear-\INDU{} &0.95 &0.97 &0.85 &0.87 &0.8 &\textbf{0.58} & \textbf{3.55} &9.53 &5.4 &45.47 &5.52 &4.25\\
\spear-\SWVIBM{} &0.67 &0.85 &0.53 &0.78 &0.49 &\textbf{0.38} & 3.49 &6.98 &4.32 &48.48 &4.9 &\textbf{2.82}\\
\spear-\IBM{} &0.6 &0.86 &0.48 &0.66 &0.5 &\textbf{0.38} & 3.18 &5.77 &2.58 &45.72 &2.5 &\textbf{1.56}\\
\spear-\SWV{} &0.49 &0.58 &0.48 &0.44 &\textbf{0.47} &\textbf{0.34} & 3.09 &6.24 &2.09 &56.09 &2.38 &\textbf{1.13}\\
\addlinespace[\interrowspace]
\tnm{}-\RANDSAT{} &1.01 &1.05 &\textbf{0.94} &\textbf{0.93} &1.22 &\textbf{0.88} & \textbf{3.79} &8.63 &6.57 &46.21 &7.64 &5.42\\
\saps{}-\RANDSAT{} &0.94 &1.09 &0.73 &0.78 &0.86 &\textbf{0.66} & \textbf{3.81} &8.54 &6.62 &49.33 &6.59 &5.04\\
\addlinespace[\interrowspace]
\addlinespace[\interrowspace]
\cplex{}-\BIGMIX{} &2.7E8 &0.93 &1.02 &1 &0.85 &\textbf{0.64} & \textbf{3.39} &8.27 &4.75 &41.25 &5.33 &\textbf{3.54}\\
\gurobi{}-\BIGMIX{} &1.51 &\textbf{1.23} &1.41 &1.26 &1.43 &\textbf{1.17} & \textbf{3.35} &5.12 &4.55 &40.72 &5.45 &3.69\\
\scip{}-\BIGMIX{} &4.5E6 &0.88 &0.86 &0.91 &0.72 &\textbf{0.57} & \textbf{3.43} &\textbf{5.35} &4.48 &39.51 &5.08 &\textbf{3.75}\\
\lpsolve{}-\BIGMIX{} &1.1 &0.9 &0.68 &1.07 &0.63 &\textbf{0.5} & 3.35 &4.68 &4.62 &43.27 &\textbf{2.76} &4.92\\
\addlinespace[\interrowspace]
\cplex{}-\CORLAT{} &0.49 &0.52 &0.53 &\textbf{0.46} &0.62 &\textbf{0.47} & \textbf{3.19} &7.64 &5.5 &27.54 &4.77 &\textbf{3.4}\\
\gurobi{}-\CORLAT{} &\textbf{0.38} &0.44 &0.41 &\textbf{0.37} &0.51 &\textbf{0.38} & \textbf{3.21} &5.23 &5.52 &28.58 &4.71 &\textbf{3.31}\\
\scip{}-\CORLAT{} &0.39 &0.41 &0.42 &\textbf{0.37} &0.5 &\textbf{0.38} & \textbf{3.2} &7.96 &5.52 &26.89 &5.12 &3.52\\
\lpsolve{}-\CORLAT{} &\textbf{0.44} &0.48 &\textbf{0.44} &\textbf{0.45} &0.54 &\textbf{0.41} & 3.25 &5.06 &5.49 &31.5 &\textbf{2.63} &4.42\\
\addlinespace[\interrowspace]
\cplex{}-\RCW{} &0.25 &0.29 &0.1 &0.03 &0.05 &\textbf{0.02} & 3.11 &7.53 &5.25 &25.84 &4.81 &\textbf{2.66}\\
\cplex{}-\REG{} &\textbf{0.38} &\textbf{0.39} &0.44 &\textbf{0.38} &0.54 &0.42 & \textbf{3.1} &6.48 &5.28 &24.95 &4.56 &3.65\\
\cplex{}-\CORLATREG{} &0.46 &0.58 &0.46 &\textbf{0.43} &0.58 &0.45 & \textbf{4.25} &11.86 &11.19 &29.92 &11.44 &8.35\\
\cplex{}-\CORLATREGRCW{} &0.44 &0.54 &0.42 &\textbf{0.37} &0.47 &\textbf{0.36} & \textbf{5.4} &18.43 &17.34 &35.3 &20.36 &13.19\\
\addlinespace[\interrowspace]
\addlinespace[\interrowspace]
\lkh{}-\PORTGEN{} &\textbf{0.61} &0.63 &0.64 &\textbf{0.61} &0.89 &0.67 & 4.14 &\textbf{1.14} &12.78 &22.95 &11.49 &11.14\\
\lkh{}-\PORTCGEN{} &\textbf{0.71} &0.72 &0.75 &\textbf{0.71} &1.02 &0.76 & 4.19 &\textbf{2.7} &12.93 &24.78 &11.54 &10.79\\
\lkh{}-\TSPLIB{} &9.55 &\textbf{1.11} &1.77 &\textbf{1.3} &\textbf{1.21} &\textbf{1.06} & 1.61 &3.02 &0.51 &4.3 &0.17 &\textbf{0.11}\\
\addlinespace[\interrowspace]
\concorde{}-\PORTGEN{} &\textbf{0.41} &0.43 &0.43 &\textbf{0.42} &0.59 &0.45 & 4.18 &\textbf{3.6} &12.7 &22.28 &10.79 &9.9\\
\concorde{}-\PORTCGEN{} &\textbf{0.33} &0.34 &0.34 &0.34 &0.46 &0.35 & 4.17 &\textbf{2.32} &12.68 &24.8 &11.16 &10.18\\
\concorde{}-\TSPLIB{} &120.6 &\textbf{0.69} &0.99 &\textbf{0.87} &0.64 &\textbf{0.52} & 1.54 &2.66 &0.47 &4.26 &0.22 &\textbf{0.12}\\
\bottomrule
 \end{tabular}
 \small
    \caption{Quantitative comparison of models for runtime predictions on previously unseen instances.
    We report 10-fold cross-validation performance. Lower RMSE values are better (0 is optimal).
    Note the very large RMSE values for ridge regression on some data sets (we use scientific notation, denoting ``$\times 10^x$'' as ``$Ex$''); these large errors are due to extremely small/large predictions for a few data points. \revision{Boldface indicates performance not statistically significantly different from the best method in each row.}
\label{tab:feature_space_def_10fold_cv}}
  }
\end{table}

\renewcommand{\hlinespace}{\\[.2em]}

\begin{figure}[tbp]
    {\scriptsize
 \setlength{\tabcolsep}{2pt}
\centering
    \begin{tabular}{ccccc}
      ~ & \minisat{}-\INDUHANDRAND{} & \cplex{}-\BIGMIX{} & \cplex{}-\RCW{} & \concorde-\PORTGEN{}\\
     \begin{sideways}Ridge regression (RR)\end{sideways} &
\includegraphics[scale=0.2]{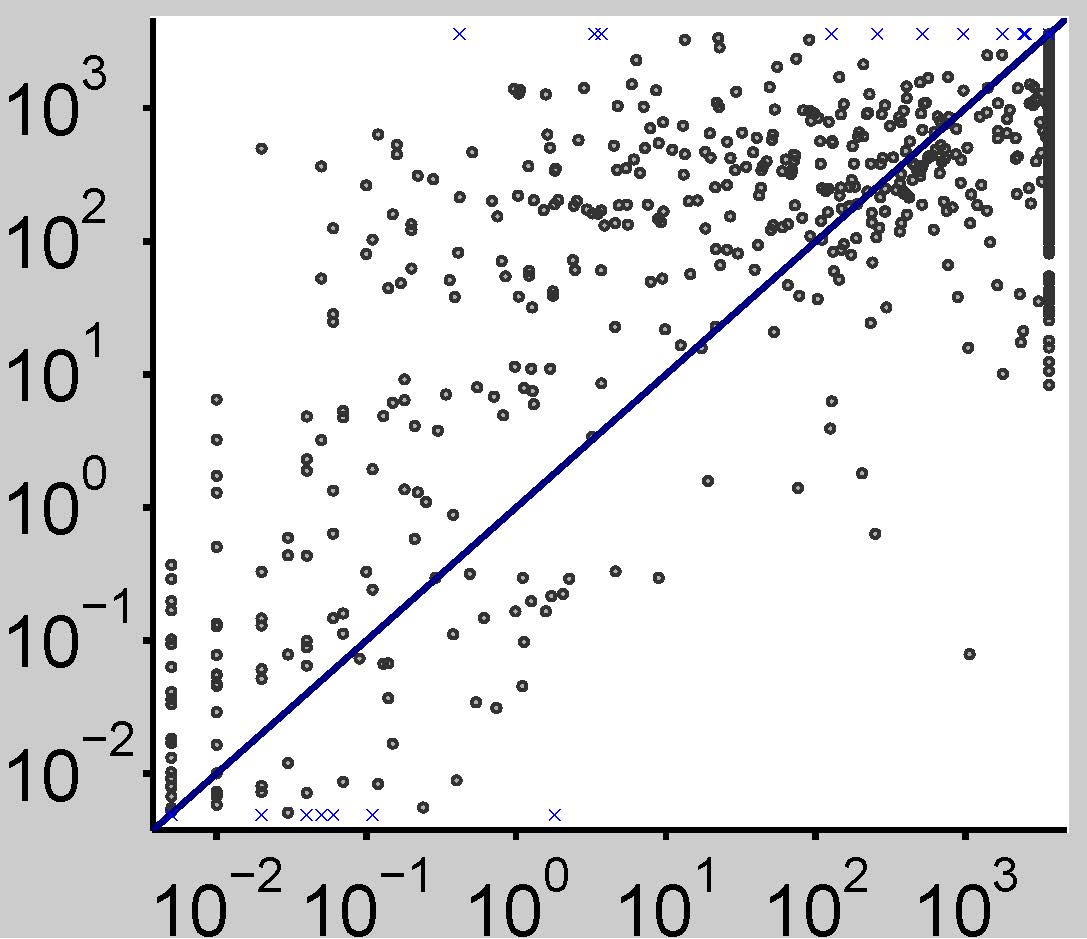} &
\includegraphics[scale=0.2]{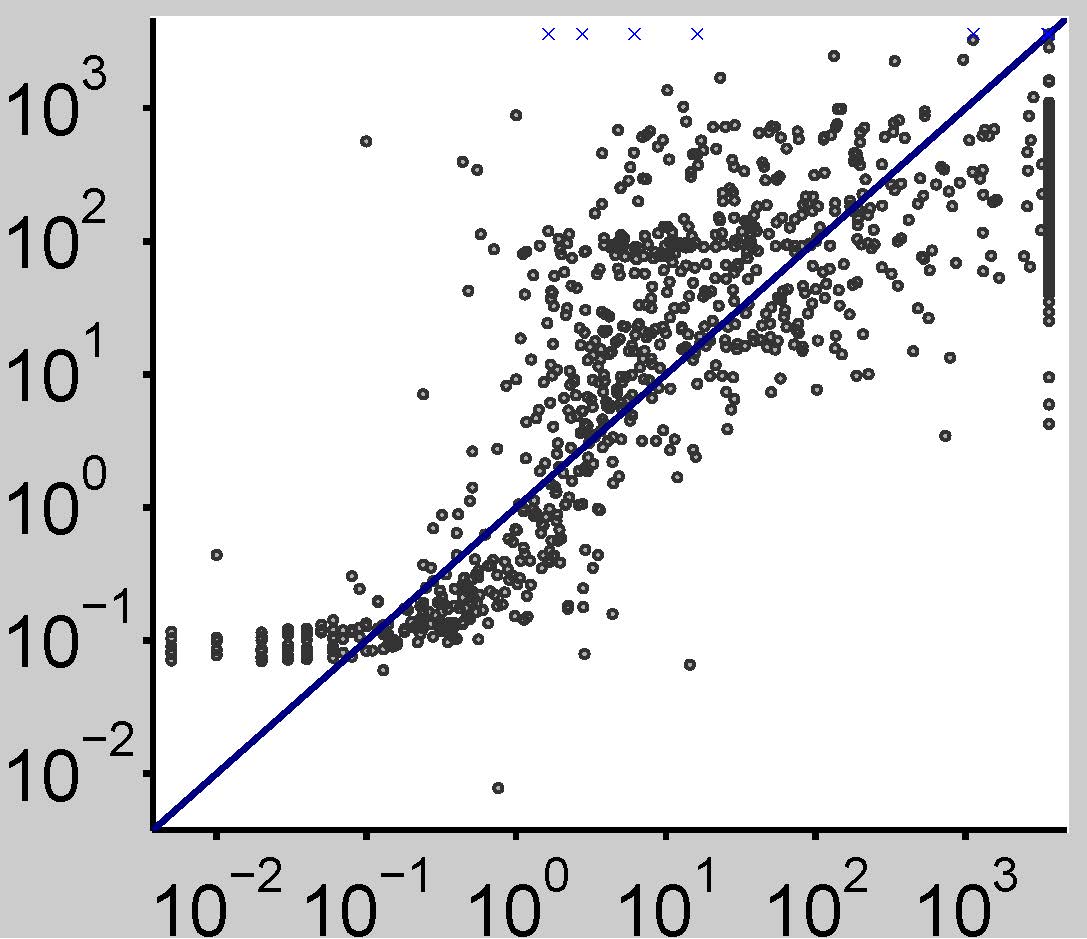} &
\includegraphics[scale=0.2]{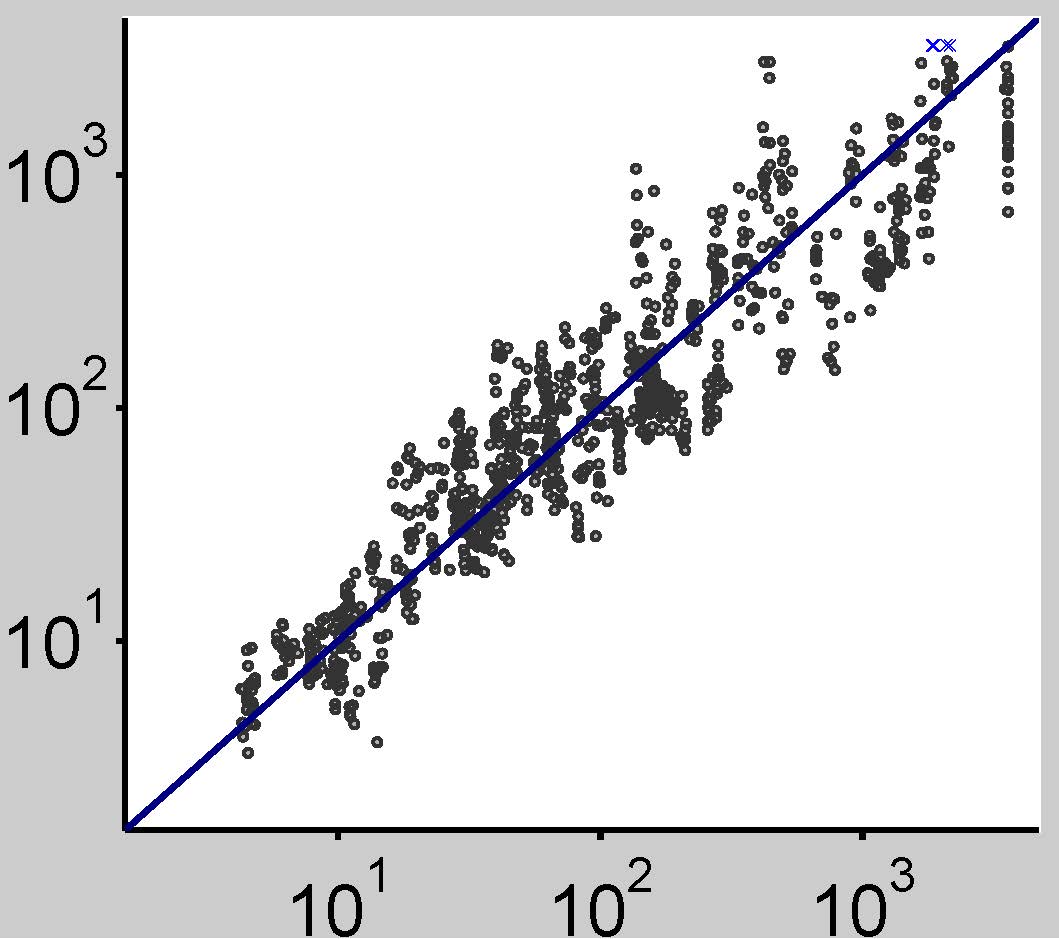} &
\includegraphics[scale=0.2]{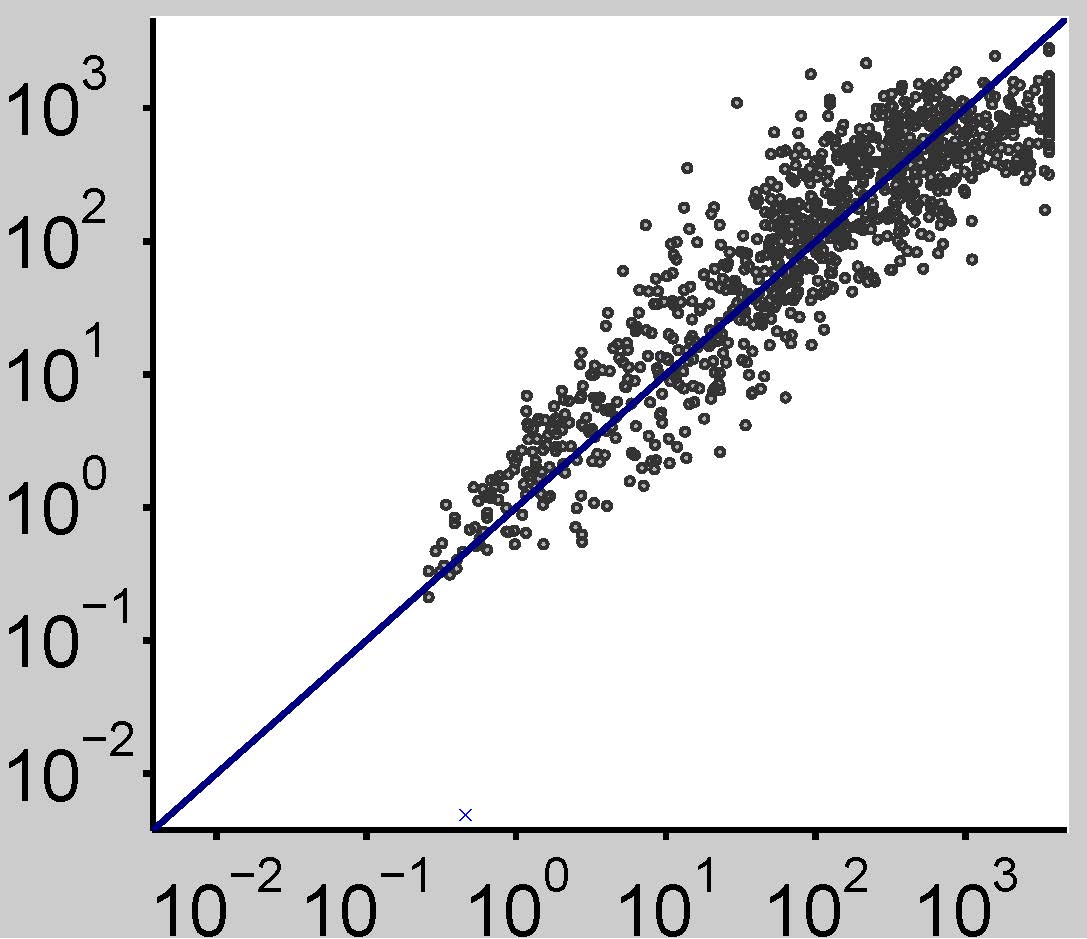}
            \hlinespace{}
     \begin{sideways}SPORE FoBa\end{sideways} &
\includegraphics[scale=0.2]{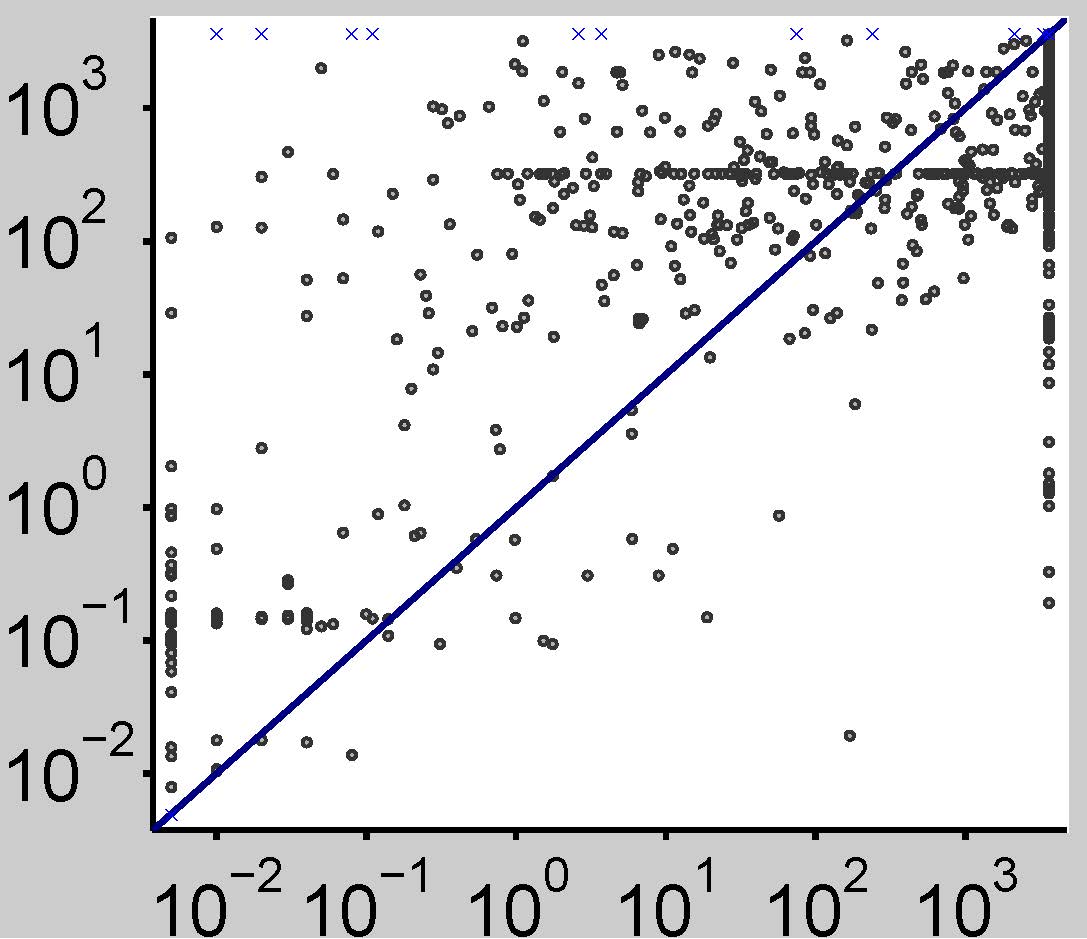} &
\includegraphics[scale=0.2]{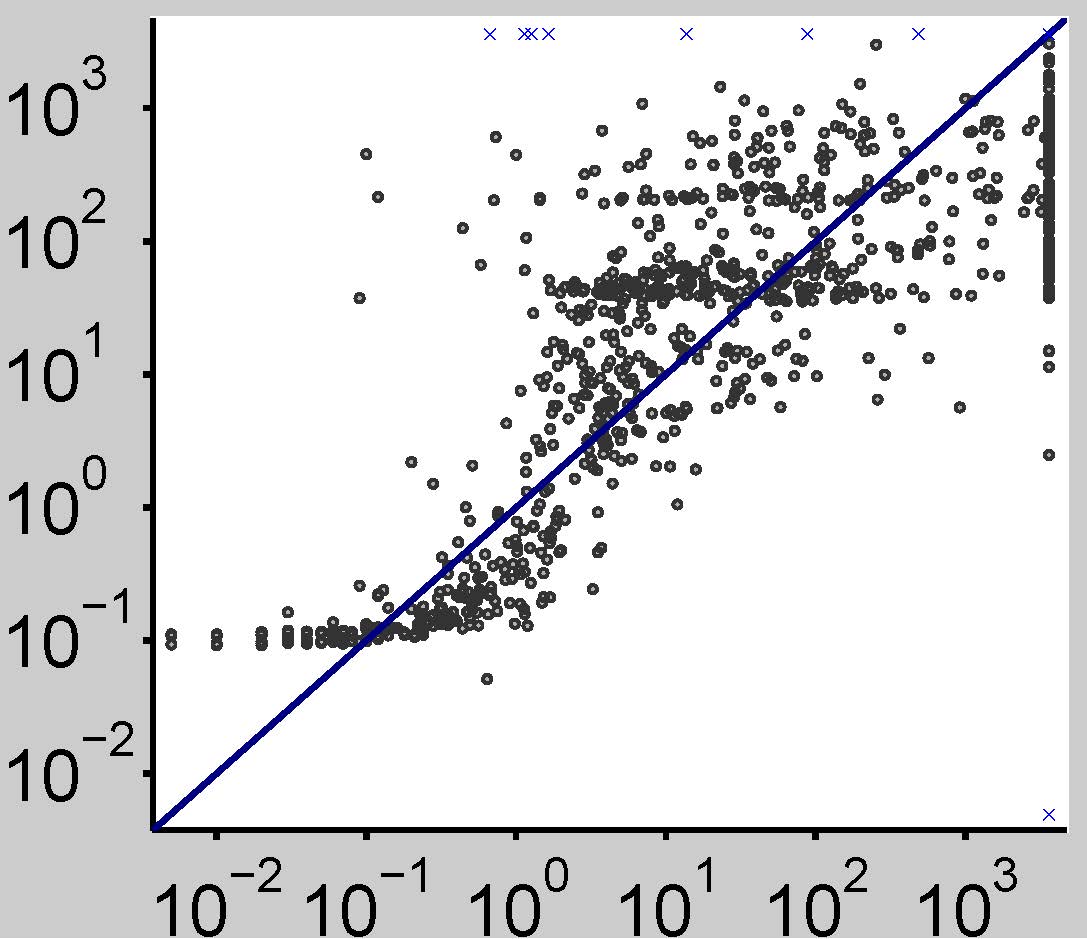} &
\includegraphics[scale=0.2]{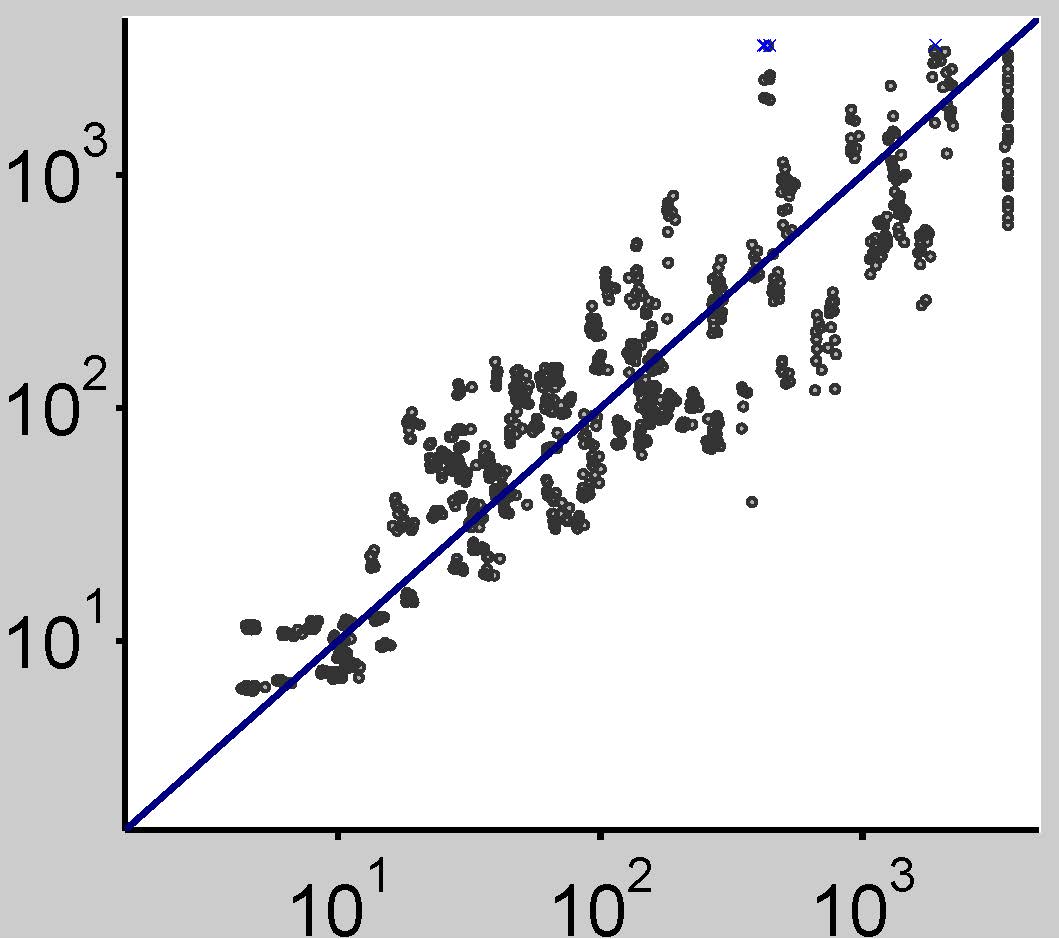} &
\includegraphics[scale=0.2]{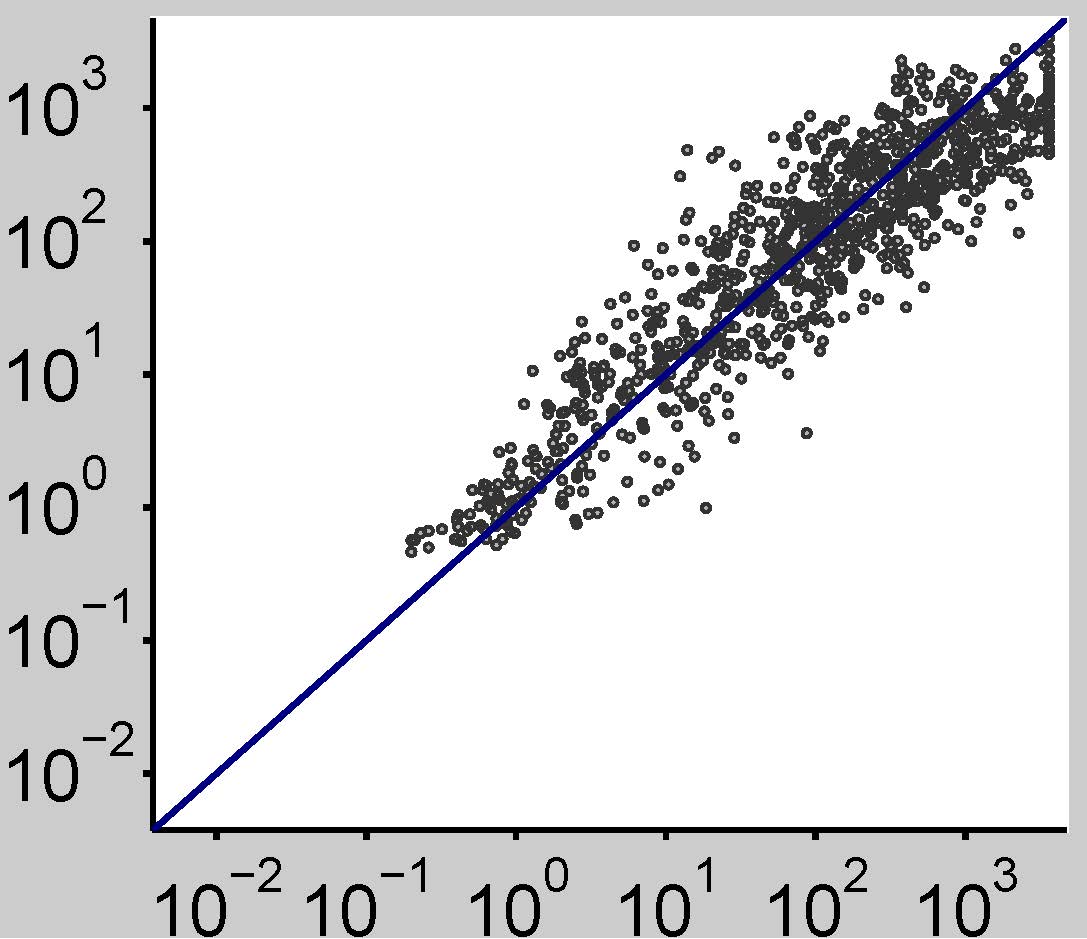}
            \hlinespace{}
    \begin{sideways}Projected process\end{sideways} &
\includegraphics[scale=0.2]{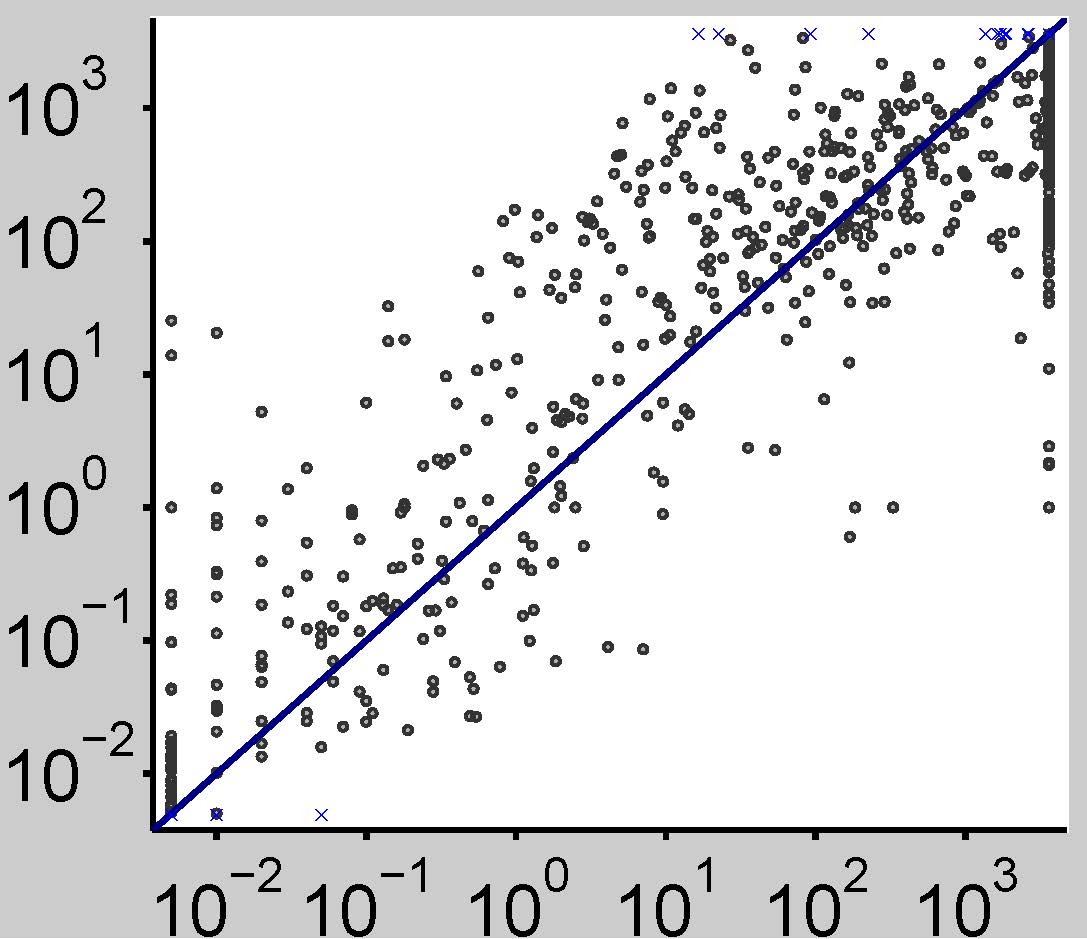} &
\includegraphics[scale=0.2]{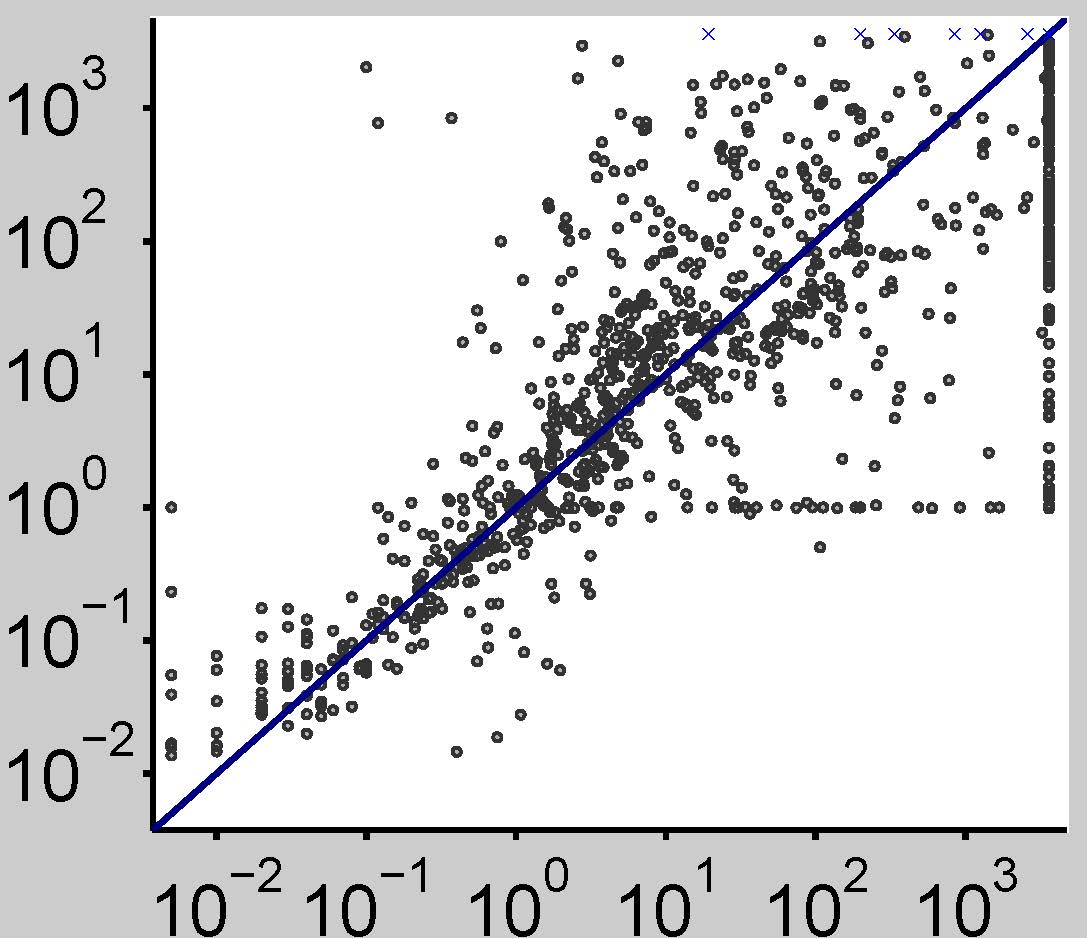} &
\includegraphics[scale=0.2]{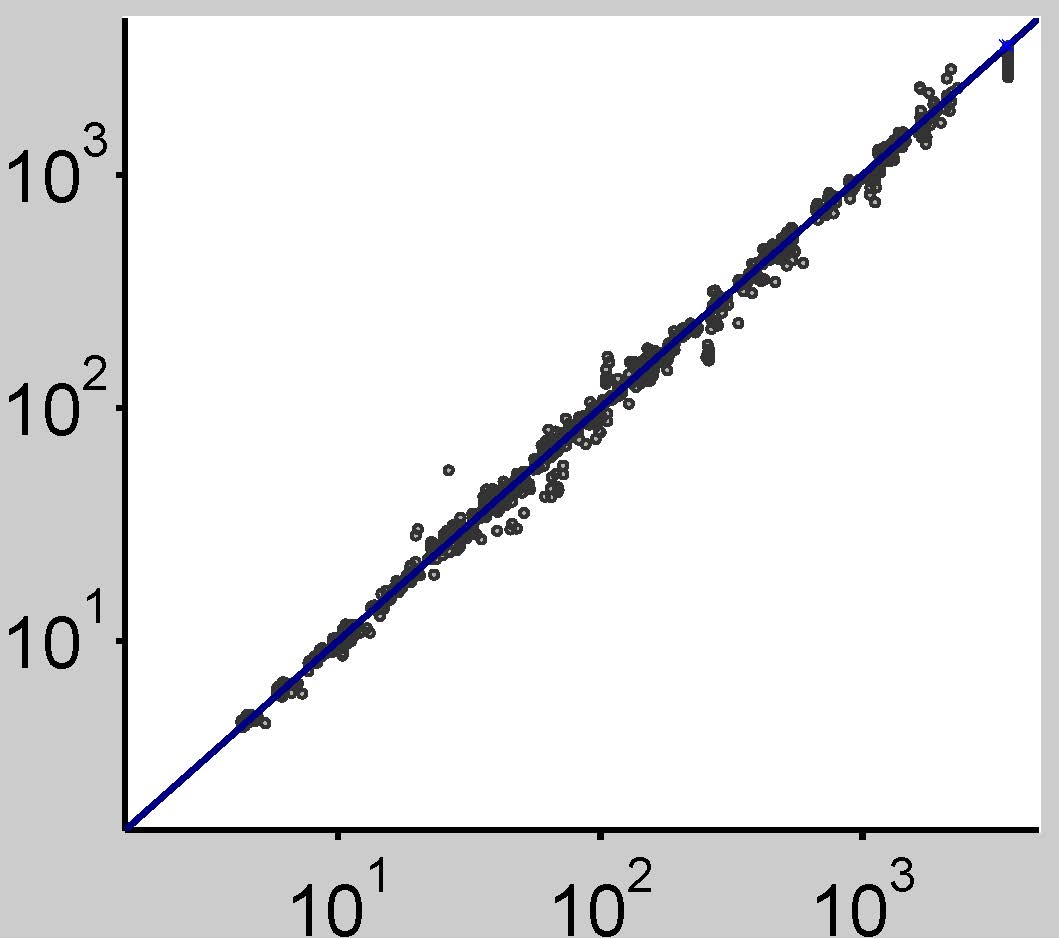} &
\includegraphics[scale=0.2]{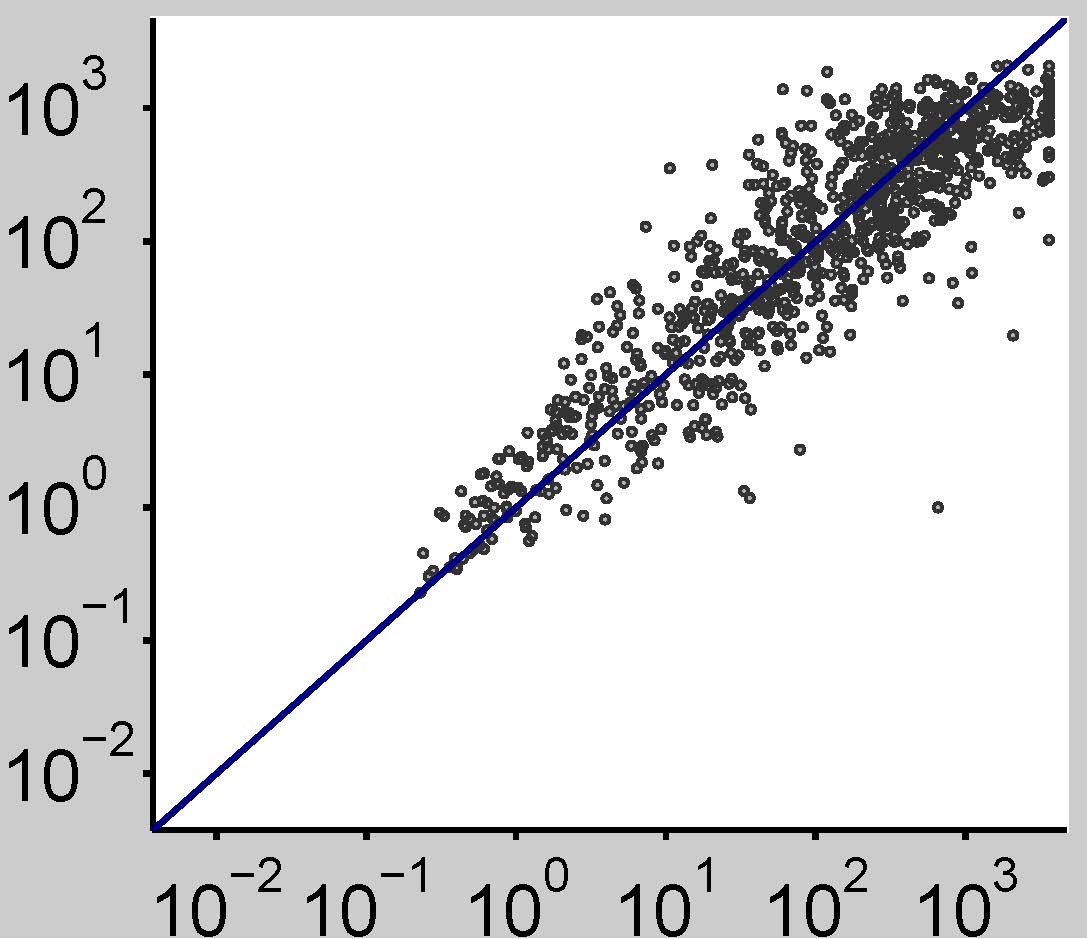}
            \hlinespace{}
     \begin{sideways}Neural network\end{sideways} &
\includegraphics[scale=0.2]{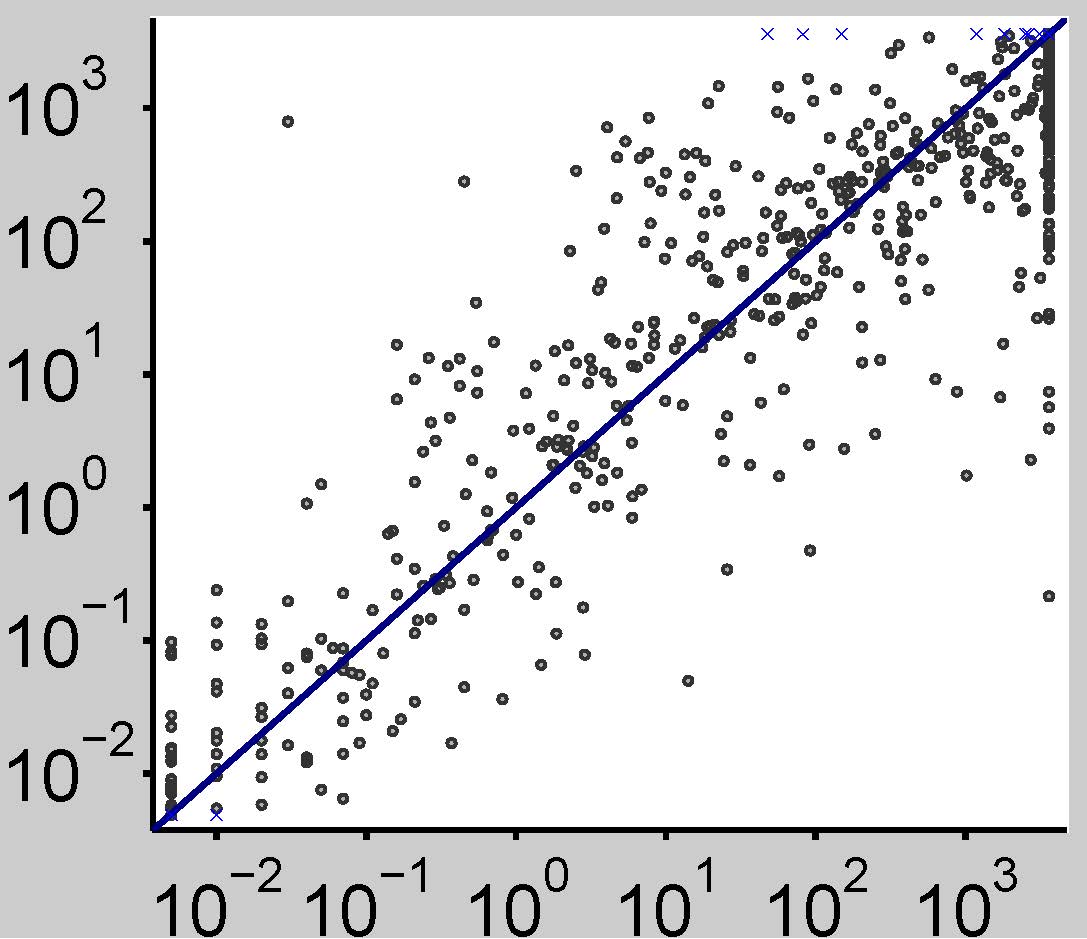} &
\includegraphics[scale=0.2]{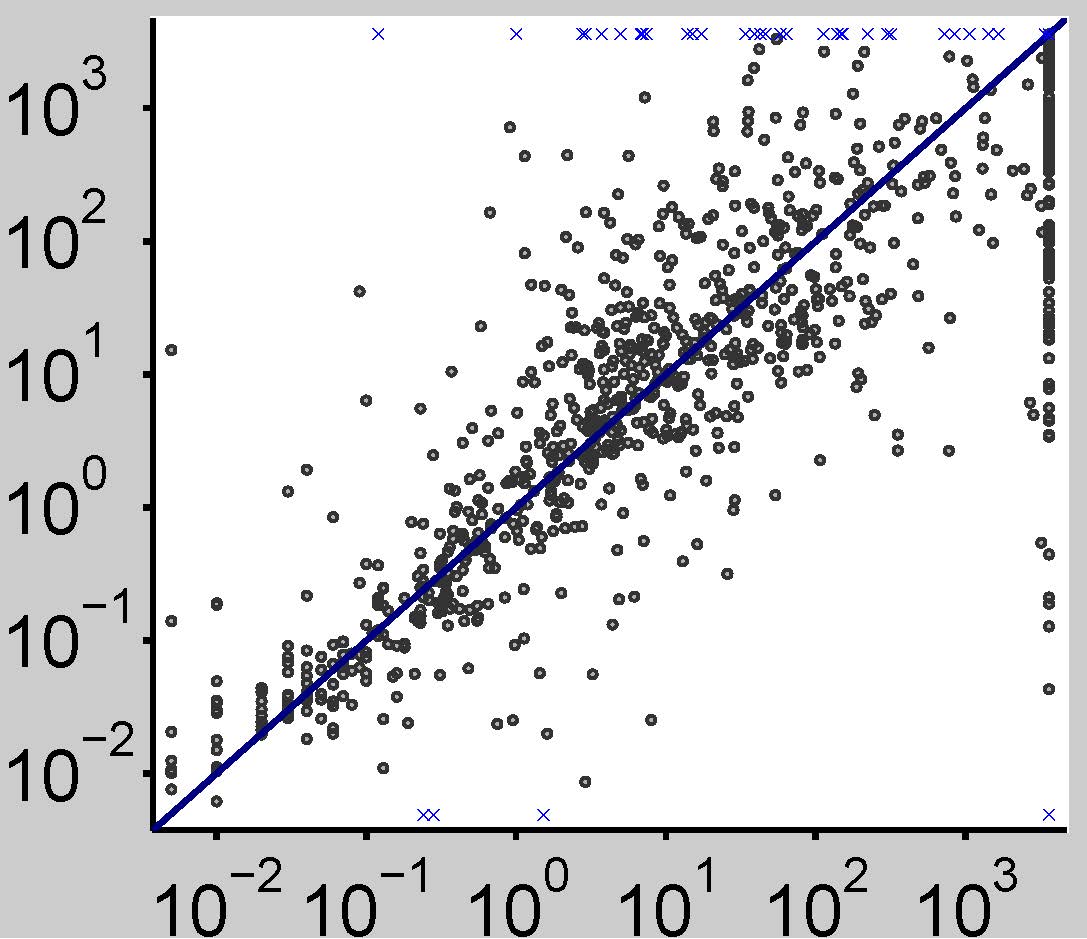} &
\includegraphics[scale=0.2]{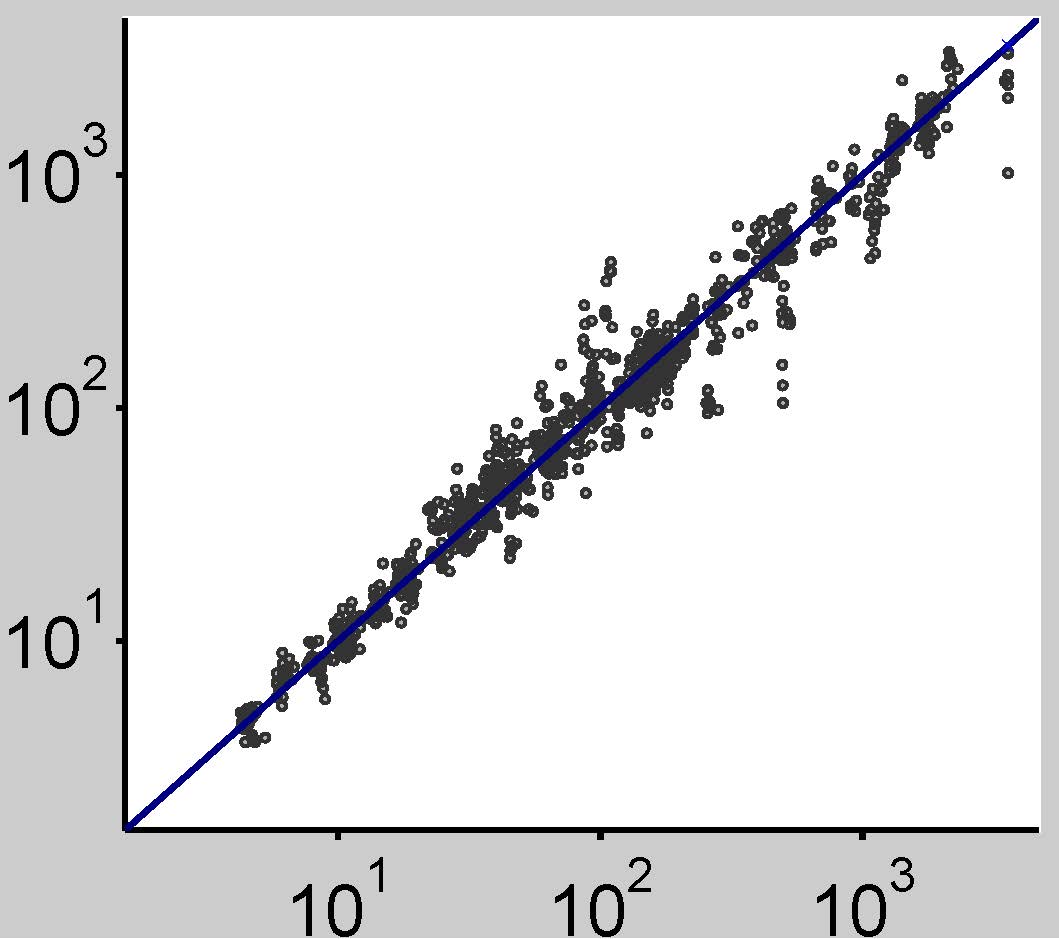} &
\includegraphics[scale=0.2]{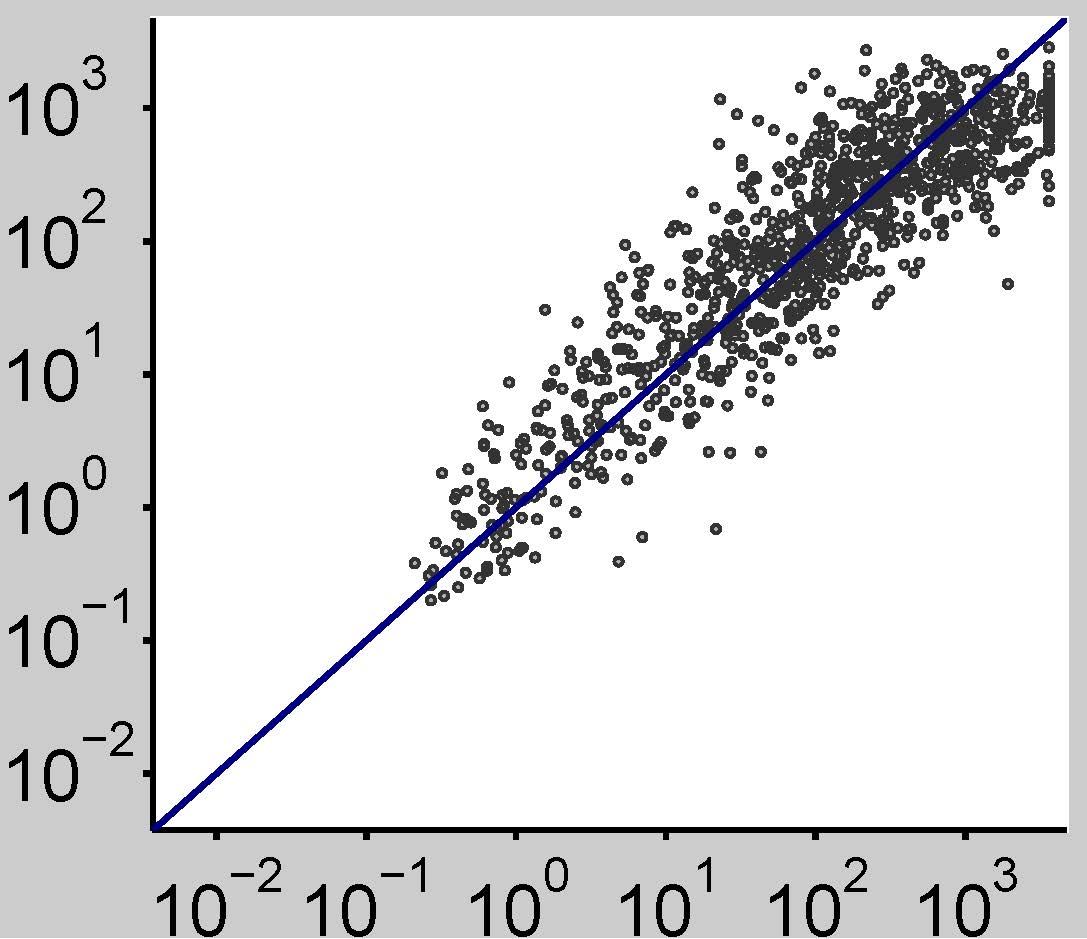}
            \hlinespace{}
    \begin{sideways}Random Forest \end{sideways} &
\includegraphics[scale=0.2]{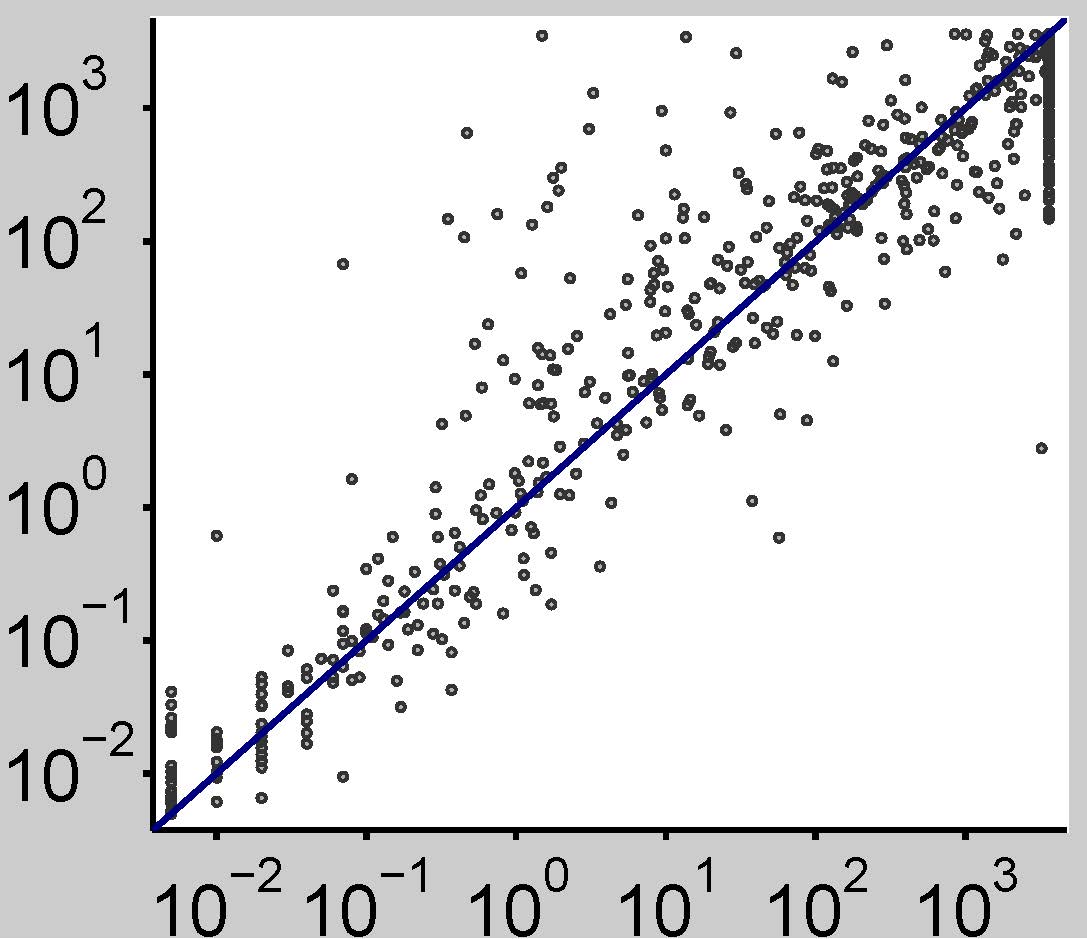} &
\includegraphics[scale=0.2]{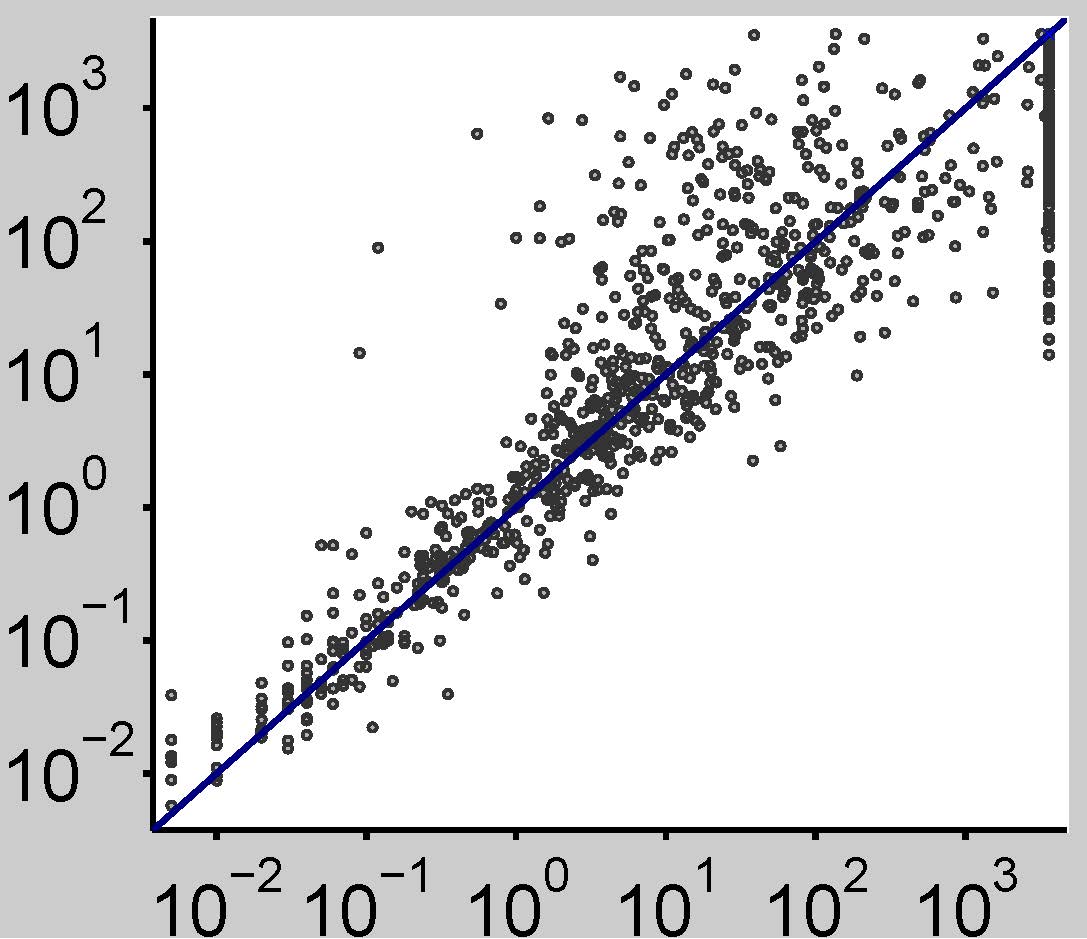} &
\includegraphics[scale=0.2]{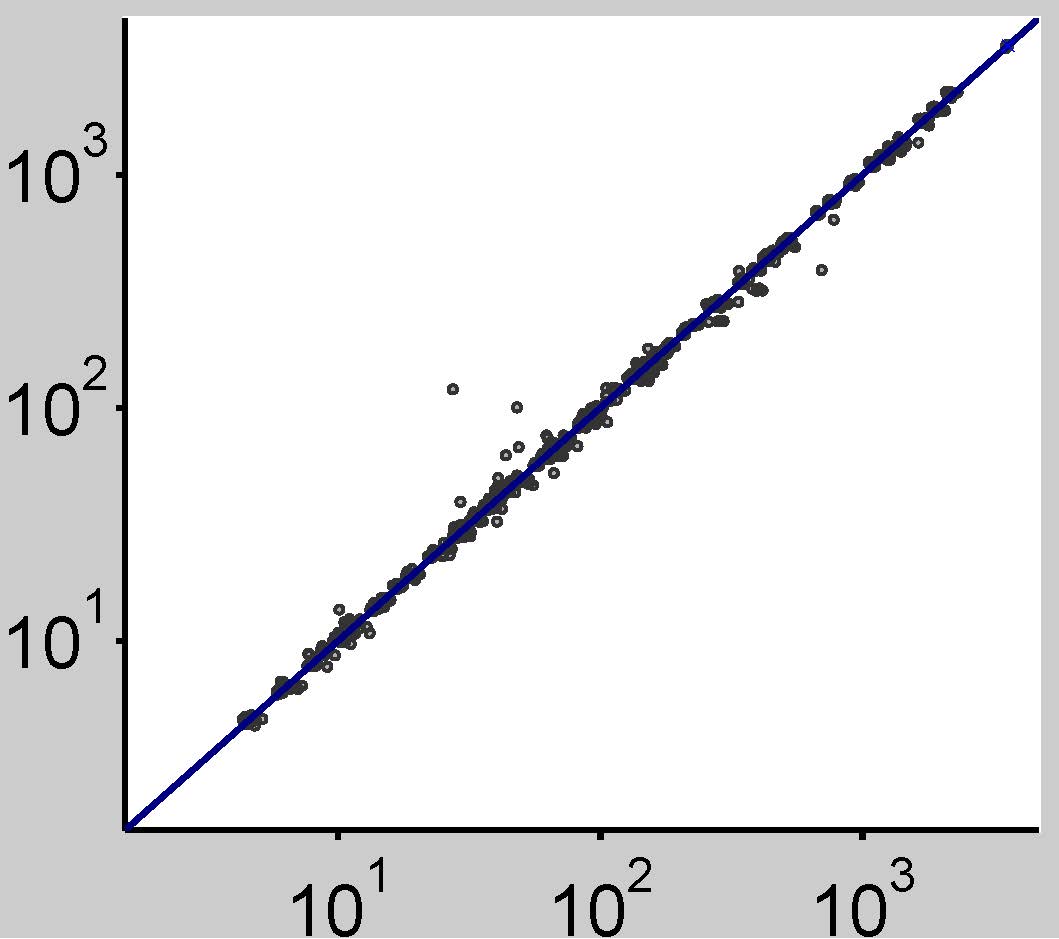} &
\includegraphics[scale=0.2]{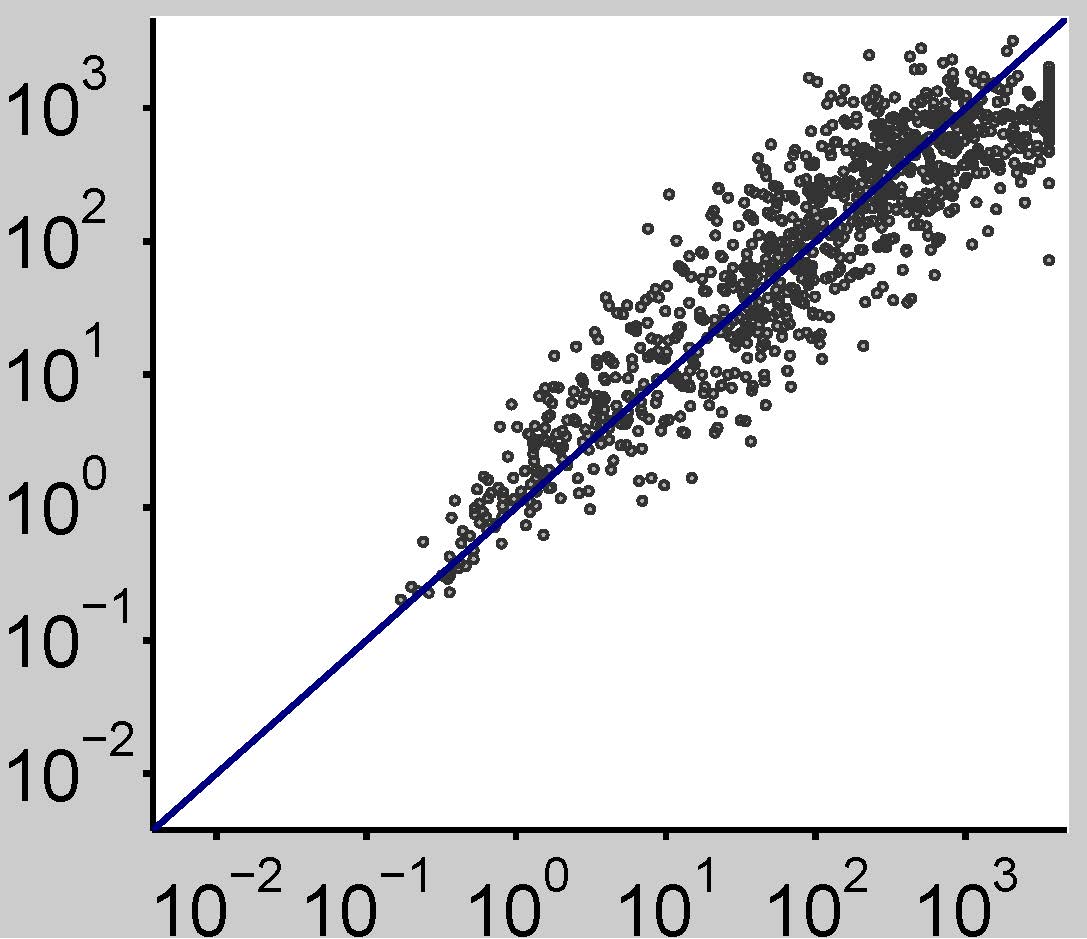}
        \small
    \end{tabular}
    \caption{Visual comparison of models for runtime predictions on previously unseen test instances. The data sets used in  each column
    are shown at the top. The $x$-axis of each scatter plot denotes true runtime and the $y$-axis 2-fold cross-validated runtime as predicted by the respective model; each dot represents one instance. Predictions above $3\,000$ or below $0.001$ are denoted by a blue cross rather than a black dot.
Figures D.1--D.10 (in the online appendix) show equivalent plots for the other benchmarks and also include regression trees (whose predictions were similar to those of random forests but had larger spread).
\label{fig:feature_space_def_2fold_cv}}
  }
\end{figure}

Table \ref{tab:feature_space_def_10fold_cv} provides quantitative results for all benchmarks, and Figure \ref{fig:feature_space_def_2fold_cv} visualizes results.
At the broadest level, we can conclude that most of the methods were able to capture enough about algorithm performance on training data to make meaningful predictions on test data, most of the time: easy instances tended to be predicted as being easy, and hard ones as being hard.
Take, for example the case of predicting the runtime of \minisat{} on a heterogeneous mix of SAT competition instances
(see the leftmost column in Figure \ref{tab:feature_space_def_10fold_cv} and the top row of Table \ref{tab:feature_space_def_10fold_cv}). \minisat{} runtimes varied by almost six orders of magnitude, while predictions with the better models rarely were off by more than one order of magnitude \revision{(outliers may draw the eye in the scatterplot, but quantitatively, the RMSE for predicting $\log_{10}$ runtime was low -- \eg{}, 0.47 for random forests, which means an average misprediction of a factor of $10^{0.47}<3$).}
While the models were certainly not perfect, note that even the relatively poor predictions of ridge regression variant RR tended to be accurate within about an order of magnitude, which was enough to enable the portfolio-based algorithm selector \satzilla{}~\cite{SATzilla-Full} to win five medals in each of the 2007 and 2009 SAT competitions. (Switching to random forest models after 2009 further improved \satzilla{}'s performance \cite{XuHutHooLey12}.)

In our experiments, random forests were the overall winner among the different methods, yielding the best predictions in terms of all our quantitative measures.\footnote{For brevity, we only report RMSE values in the tables here; comparative results for correlation coefficients and log likelihoods, given in Table D.3 in the online appendix, are qualitatively similar.} 
For SAT, random forests were always the best method, and for MIP they yielded the best performance for the most heterogeneous instance set, \BIGMIX{} (see Column 2 of Figure \ref{fig:feature_space_def_2fold_cv}).
We attribute the strong performance of random forests on highly heterogeneous data sets to the fact that, as a tree-based approach, they can model very different parts of the data separately; in contrast, the other methods allow the fit in a given part of the space to be influenced more by data in distant parts of the space.
Indeed, the ridge regression variants made extremely bad predictions for some outlying points on \BIGMIX{}.
For the more homogeneous MIP data sets, either random forests or projected processes performed best, often followed closely by ridge regression variant RR.
The performance of \cplex{} on set \RCW{} was a special case in that it could be predicted extremely well by all models (see Column 3 of Figure \ref{fig:feature_space_def_2fold_cv}).
Finally, for TSP, projected processes and ridge regression had a slight edge for the homogeneous \PORTGEN{} and \PORTCGEN{} benchmarks, whereas tree-based methods
(once again)
performed best on the most heterogeneous benchmark, \TSPLIB{}.
The last column of Figure \ref{fig:feature_space_def_2fold_cv}
shows that, in the case where random forests performed worst, the qualitative differences in predictions were small.
In terms of computational requirements, random forests were among the cheapest methods, taking between $0.1$ and $11$ seconds to learn a model.

\subsection{\revision{Results based on Different Classes of Instance Feature}}\label{sec:feature_qual_results}

\revision{While the previous experiments focussed on the performance of the various models based on our entire feature set,
we now study the performance of different subsets of features when using the overall best-performing model, random forests.
Table \ref{tab:feature_space_featuresubsetexp} presents the results and also lists the cost of the various feature subsets (which in most cases is much smaller than the runtime of the algorithm being modeled).
On the broadest level, we note that predictive performance improved as we used more computationally expensive features: \eg{}, while the trivial features
were basically free, they yielded rather poor performance, whereas using the entire feature set almost always led to the best performance.
Interestingly, however, for all SAT benchmarks, using at most moderately expensive features yielded results statistically insignificantly different from the best,
with substantial reductions in feature computation time. The same was even true for several SAT benchmarks when considering at most cheap features.
Our new features clearly showed value: for example, our cheap feature set yielded similar predictive performance as the set of previous features at a much lower cost; and our moderate feature set tended to yield better performance than the previous one at comparable cost.
Our new features led to especially clear improvements for MIP, yielding significantly better predictive performance than the previous features in 11/12 cases.
Similarly, for TSP, our new features improved performance significantly in 4/6 cases (TSPLIB was too small to
achieve reliable results in the two remaining cases, with even the trivial features performing insignificantly worse than the best).}

\hide{
\minisat{}-\INDUHANDRAND{} &188.8765 &3600 & 
\minisat{}-\HAND{} &138.0729 &3600 & 
\minisat{}-\RAND{} &405.1123 &3600 & 
\minisat{}-\INDU{} &58.3941 &3600 & 
\minisat-\SWVIBM{} &1.7842 &3600 & 
\minisat{}-\IBM{} &8.3686 &3600 & 
\minisat{}-\SWV{} &0.25194 &5.32 & 
\cryptominisat-\INDU{} &159.7472 &3600 & 
\cryptominisat-\SWVIBM{} &7.2257 &3600 & 
\cryptominisat-\IBM{} &20.7899 &3600 & 
\cryptominisat-\SWV{} &1.8948 &3600 & 
\spear-\INDU{} &134.4069 &3600 & 
\spear-\SWVIBM{} &3.1426 &3600 & 
\spear-\IBM{} &13.4376 &3600 & 
\spear-\SWV{} &0.49893 &3600 & 
\tnm{}-\RANDSAT{} &2.2086 &3600 & 
\saps{}-\RANDSAT{} &3.8821 &3600 & 
\cplex{}-\BIGMIX{} &16.6469 &3628.04 & 
\gurobi{}-\BIGMIX{} &33.9222 &3640.43 & 
\scip{}-\BIGMIX{} &99.2196 &3626.51 & 
\lpsolve{}-\BIGMIX{} &1456.2723 &3933.68 & 
\cplex{}-\CORLAT{} &5.0539 &3600.44 & 
\gurobi{}-\CORLAT{} &2.3812 &2159.76 & 
\scip{}-\CORLAT{} &5.0676 &3600 & 
\lpsolve{}-\CORLAT{} &233.5159 &3788.158 & 
\cplex{}-\RCW{} &89.9063 &3600.11 & 
\cplex{}-\REG{} &125.1273 &3600.03 & 
\cplex{}-\CORLATREG{} &25.1471 &3600.44 & 
\cplex{}-\CORLATREGRCW{} &38.3432 &3600.44 & 
\lkh{}-\PORTGEN{} &4.2644 &3600.22 & 
\lkh{}-\PORTCGEN{} &11.8625 &3600.22 & 
\lkh{}-\TSPLIB{} &0.68141 &3600 & 
\concorde{}-\PORTGEN{} &89.7058 &3600 & 
\concorde{}-\PORTCGEN{} &39.6196 &3600 & 
\concorde{}-\TSPLIB{} &25.5155 &3604 & 
}

\begin{table}[tp]
\setlength{\tabcolsep}{1.1pt}
    {\scriptsize
\centering
\revision{
\begin{tabular}{l@{\hskip .8em}c@{\hskip .5em}c@{\hskip .8em}c@{\hskip .5em}c@{\hskip .5em}c@{\hskip .5em}c@{\hskip .5em}c@{\hskip 2em}cccc@{\hskip 2em}cccc}
    \toprule
& \multicolumn{2}{c}{\textbf{Alg. runtime}} & \multicolumn{5}{c}{\textbf{RMSE}} & \multicolumn{4}{c}{\textbf{Avg.\ feature time [s]}} & \multicolumn{4}{c}{\textbf{Max.\ feature time [s]}}\\
\cmidrule(r{0.75em}){2-3}\cmidrule(r{1.75em}){4-8}\cmidrule{9-12}\cmidrule{13-16}
\textbf{Scenario} &avg & max &trivial & prev & cheap & mod & exp & prev & cheap & mod & exp & prev & cheap & mod & exp \\
\cmidrule(r{0.75em}){1-1} \cmidrule(r{0.75em}){2-3} \cmidrule(r{1.75em}){4-8}\cmidrule{9-12}\cmidrule{13-16}
\minisat{}-\INDUHANDRAND{} &2009 &3600 &1.01 &\textbf{0.5} &\textbf{0.49} &\textbf{0.47} &\textbf{0.47} & 102 &\textbf{21} &59 &109 & 6E3 &\textbf{1E3} &1E3 &6E3\\
\minisat{}-\HAND{} &1903 &3600 & 1.25 &\textbf{0.57} &\textbf{0.53} &\textbf{0.52} &\textbf{0.51} & 74 &\textbf{14} &48 &79 & 3E3 &\textbf{96} &179 &3E3\\
\minisat{}-\RAND{} &2497 &3600 & 0.82 &\textbf{0.39} &\textbf{0.38} &\textbf{0.37} &\textbf{0.37} & 59 &\textbf{23} &52 &65 & 221 &\textbf{35} &145 &230\\
\minisat{}-\INDU{} &1146 &3600 &0.94 &0.58 &0.57 &\textbf{0.55} &\textbf{0.52} & 222 &\textbf{24} &85 &231 & 6E3 &\textbf{1E3} &1E3 &6E3\\
\minisat-\SWVIBM{} &466 &3600 &0.85 &\textbf{0.16} &\textbf{0.17} &\textbf{0.17} &\textbf{0.17} & 98 &\textbf{8.4} &59 &102 & 1E3 &\textbf{74} &153 &1E3\\
\minisat{}-\IBM{} &834 &3600 &1.1 &\textbf{0.21} &0.25 &\textbf{0.21} &\textbf{0.19} & 130 &\textbf{11} &78 &136 & 1E3 &\textbf{74} &153 &1E3\\
\minisat{}-\SWV{} &0.89 &5.32 &0.25 &\textbf{0.08} &0.09 &\textbf{0.08} &\textbf{0.08} & 57 &\textbf{4.9} &34 &59 & 217 &\textbf{17} &123 &226\\
\addlinespace[\interrowspace]
\cryptominisat-\INDU{} &1921 &3600 &1.1 &0.81 &\textbf{0.73} &\textbf{0.74} &\textbf{0.72} & 222 &\textbf{24} &85 &231 & 6E3 &\textbf{1E3} &1E3 &6E3\\
\cryptominisat-\SWVIBM{} &873 &3600 & 1.07 &\textbf{0.47} &\textbf{0.5} &\textbf{0.49} &\textbf{0.48} & 98 &\textbf{8.4} &59 &102 & 1081 &\textbf{74} &153 &1103\\
\cryptominisat-\IBM{} &1178 &3600 &1.2 &\textbf{0.42} &\textbf{0.45} &\textbf{0.42} &\textbf{0.41} & 130 &\textbf{11} &78 &136 & 1081 &\textbf{74} &153 &1103\\
\cryptominisat-\SWV{} &486 &3600 &0.89 &\textbf{0.51} &\textbf{0.53} &\textbf{0.49} &\textbf{0.51} & 57 &\textbf{4.9} &34 &59 & 217 &\textbf{17} &123 &226\\
\addlinespace[\interrowspace]
\spear-\INDU{} &1685 &3600 &1.01 &0.67 &0.62 &\textbf{0.61} &\textbf{0.58} & 222 &\textbf{24} &85 &231 & 6E3 &\textbf{1E3} &1E3 &6E3\\
\spear-\SWVIBM{} &587 &3600 &0.97 &\textbf{0.38} &\textbf{0.39} &\textbf{0.39} &\textbf{0.38} & 98 &\textbf{8.4} &59 &102 & 1E3 &\textbf{74} &153 &1E3\\
\spear-\IBM{} &1004 &3600 &1.18 &\textbf{0.39} &\textbf{0.42} &\textbf{0.42} &\textbf{0.38} & 130 &\textbf{11} &78 &136 & 1E3 &\textbf{74} &153 &1E3\\
\spear-\SWV{} &60 &3600 & 0.54 &\textbf{0.36} &\textbf{0.34} &\textbf{0.34} &\textbf{0.34} & 57 &\textbf{4.9} &34 &59 & 217 &\textbf{17} &123 &226\\
\addlinespace[\interrowspace]
\tnm{}-\RANDSAT{} &568 &3600 & 1.05 &\textbf{0.88} &\textbf{0.97} &\textbf{0.9} &\textbf{0.88} & 63 &\textbf{26} &56 &70 & 221 &\textbf{35} &145 &230\\
\saps{}-\RANDSAT{} &1019 &3600 & 1 &\textbf{0.67} &0.71 &\textbf{0.65} &\textbf{0.66} & 63 &\textbf{26} &56 &70 & 221 &\textbf{35} &145 &230\\
\addlinespace[\interrowspace]
\addlinespace[\interrowspace]
\cplex{}-\BIGMIX{} &719 &3600 & 0.96 &0.84 &0.85 &\textbf{0.63} &\textbf{0.64} & 17 &\textbf{0.13} &6.7 &23 & 1E4 &\textbf{6.6} &54 &1E4\\
\gurobi{}-\BIGMIX{} &992 &3600 &1.31 &1.28 &1.31 &\textbf{1.19} &\textbf{1.17} & 17 &\textbf{0.13} &6.7 &23 & 1E4 &\textbf{6.6} &54 &1E4\\
\scip{}-\BIGMIX{} &1153 & 3600 &0.77 &0.67 &0.72 &\textbf{0.58} &\textbf{0.57} & 17 &\textbf{0.13} &6.7 &23 & 1E4 &\textbf{6.6} &54 &1E4\\
\lpsolve{}-\BIGMIX{} &3034 & 3600 &0.58 &\textbf{0.51} &\textbf{0.53} &\textbf{0.49} &\textbf{0.49} & 17 &\textbf{0.13} &6.7 &23 & 1E4 &\textbf{6.6} &54 &1E4\\
\addlinespace[\interrowspace]
\cplex{}-\CORLAT{} &430 &3600 &0.77 &0.62 &0.65 &\textbf{0.47} &\textbf{0.47} & 0.02 &\textbf{0.01} &5.0 &5.0 & 0.05 &\textbf{0.03} &8.5 &8.5\\
\gurobi{}-\CORLAT{} &52 &2159 &0.6 &0.48 &0.5 &\textbf{0.37} &\textbf{0.38} & 0.02 &\textbf{0.01} &5.0 &5.0 & 0.05 &\textbf{0.03} &8.5 &8.5\\
\scip{}-\CORLAT{} &99 &3600 & 0.59 &0.47 &0.48 &\textbf{0.38} &\textbf{0.38} & 0.02 &\textbf{0.01} &5.0 &5.0 & 0.05 &\textbf{0.03} &8.5 &8.5\\
\lpsolve{}-\CORLAT{} &2328 &3600 & 0.57 &0.52 &0.54 &0.43 &\textbf{0.41} & 0.02 &\textbf{0.01} &5.0 &5.0 & 0.05 &\textbf{0.03} &8.5 &8.5\\
\addlinespace[\interrowspace]
\cplex{}-\RCW{} &364 &3600 & \textbf{0.02} &0.02 &0.02 &0.03 &0.02 & 11 &\textbf{3.3} &13 &20 & 18 &\textbf{3.5} &14 &27\\
\cplex{}-\REG{} &402 &3600 & 0.77 &0.55 &0.6 &\textbf{0.42} &\textbf{0.42} & 0.47 &\textbf{0.02} &8.4 &8.9 & 0.8 &\textbf{0.05} &8.7 &9.2\\
\cplex{}-\CORLATREG{} &416 &3600 & 0.78 &0.59 &0.61 &\textbf{0.45} &\textbf{0.44} & 0.25 &\textbf{0.02} &6.7 &6.9 & 0.8 &\textbf{0.05} &8.7 &9.2\\
\cplex{}-\CORLATREGRCW{} &399 &3600 & 0.64 &0.48 &0.51 &\textbf{0.37} &\textbf{0.36} & 3.7 &\textbf{1.1} &8.7 &11 & 18 &\textbf{3.5} &14 &27\\
\addlinespace[\interrowspace]
\addlinespace[\interrowspace]
\lkh{}-\PORTGEN{} &109 &3600 &0.69 &0.71 &\textbf{0.69} &\textbf{0.69} &\textbf{0.67} & 6.0 &\textbf{1.3} &18 &49 & 9.9 &\textbf{2.6} &60 &97\\
\lkh{}-\PORTCGEN{} &203 &3600 & 0.82 &0.81 &0.81 &0.8 &\textbf{0.76} & 5.7 &\textbf{1.3} &12 &129 & 9.3 &\textbf{2.7} &27 &235\\
\lkh{}-\TSPLIB{} &429 &3600 &\textbf{1.01} &\textbf{1.1} &\textbf{0.84} &\textbf{0.94} &\textbf{1.1} & 8.2 &\textbf{1.8} &33 &68 & 72 &\textbf{16} &525 &559\\
\addlinespace[\interrowspace]
\concorde{}-\PORTGEN{} &490 &3600 & 0.53 &0.53 &0.53 &0.5 &\textbf{0.45} & 6.0 &\textbf{1.3} &18 &49 & 9.9 &\textbf{2.6} &60 &97\\
\concorde{}-\PORTCGEN{} &118 &3600 & 0.39 &0.39 &0.4 &0.37 &\textbf{0.35} & 5.7 &\textbf{1.3} &12 &129 & 9.3 &\textbf{2.7} &27 &235\\
\concorde{}-\TSPLIB{} &782 &3600 & \textbf{0.56} &\textbf{0.47} &\textbf{0.58} &\textbf{0.56} &\textbf{0.52} & 8.2 &\textbf{1.8} &33 &68 & 72 &\textbf{16} &525 &559\\
\bottomrule
\end{tabular}
} 
 \small
    \caption{\revision{Quantitative comparison of random forests based on different feature subsets: `prev' = previous features only; `mod' = moderate; `exp' = expensive.
		Feature sets `cheap', `mod', and `exp' include all cheaper features; e.g., `exp' uses the entire feature set.
    We report 10-fold cross-validation performance. Lower RMSE values are better (0 is optimal). Boldface denotes results not statistically significantly different from the best. (Note that depending on the variance, results can have the same rounded mean but still be statistically significantly different; this happens in the cases of \cplex{}-\RCW{} and \lkh{}-\PORTGEN{}.)}
\label{tab:feature_space_featuresubsetexp}}
  }
\end{table}

\subsection{Impact of Hyperparameter Optimization}\label{sec:opt_hyp_results}

\hide{Most modeling methods discussed in this paper have free hyperparameters that can be set by minimizing cross-validation loss.
While, to the best of our knowledge, all previous work on runtime prediction has used fixed default parameters, we also experimented with optimizing these hyperparameters for every method in our experiments (using the gradient-free optimizer DIRECT~\cite{JonesEtAl93:direct} to minimize 2-fold cross-validated RMSE loss on the training set with a budget of 30 function evaluations).}

Table \ref{tab:results-hyperparam-opt} shows representative results for the optimization of hyperparameters: it improved robustness somewhat for the ridge regression methods (decreasing the number of extreme outlier predictions) and improved most models slightly across the board. However, these improvements came at the expense of dramatically slower training.\footnote{Although we fixed the number of hyperparameter optimization steps, 
variation in model parameters affected learning time more for some model families than for others; for SP, slowdowns reached up to a factor of 3\,000 (dataset \minisat{}-\RAND{}).}
In practice, the small
improvements in predictive performance that can be obtained via hyperparameter optimization appear likely not to justify this drastic increase in computational cost (\eg{}, consider model-based algorithm configuration procedures, which iterate between model construction and data gathering, constructing thousands of models during typical algorithm configuration runs~\cite{HutHooLey11-SMAC}). Thus, we evaluate model performance based on fixed default hyperparameters in the rest of this article.
For completeness, our online appendix reports analogous results for models
with optimized hyperparameters.

\newcommand{\clipwidth}{.35em}

\begin{table}[t]
\setlength{\tabcolsep}{1.3pt}
    {\fontsize{6.4pt}{7.4pt}\selectfont
\centering
\begin{tabular}{l@{\hskip 1em}cc@{\hskip 1.3em}cc@{\hskip 1.3em}cc@{\hskip 1.3em}cc@{\hskip 2.5em}cc@{\hskip 1.3em}cc@{\hskip 1.3em}cc@{\hskip 1.3em}cc}
\toprule
& \multicolumn{8}{c}{\textbf{RMSE}} & \multicolumn{8}{c}{\textbf{Time to learn model (s)}}\\
\cmidrule(r{2.25em}){2-9}\cmidrule{10-17}
 & \multicolumn{2}{c}{RR} & \multicolumn{2}{c}{SP} & \multicolumn{2}{c}{NN} & \multicolumn{2}{c}{RF~~~~~~~} & \multicolumn{2}{c}{RR} & \multicolumn{2}{c}{SP} & \multicolumn{2}{c}{NN} & \multicolumn{2}{c}{RF}\\
\textbf{Domain}  & $\mathbf{\lambda}_{\text{def}}$ & $\mathbf{\lambda}_{\text{opt}}$ & $\mathbf{\lambda}_{\text{def}}$ & $\mathbf{\lambda}_{\text{opt}}$ & $\mathbf{\lambda}_{\text{def}}$ & $\mathbf{\lambda}_{\text{opt}}$ & $\mathbf{\lambda}_{\text{def}}$ & $\mathbf{\lambda}_{\text{opt}}$ & $\mathbf{\lambda}_{\text{def}}$ & $\mathbf{\lambda}_{\text{opt}}$ & $\mathbf{\lambda}_{\text{def}}$ & $\mathbf{\lambda}_{\text{opt}}$ & $\mathbf{\lambda}_{\text{def}}$ & $\mathbf{\lambda}_{\text{opt}}$ & $\mathbf{\lambda}_{\text{def}}$ & $\mathbf{\lambda}_{\text{opt}}$\\
\midrule
\minisat{}-\INDUHANDRAND{} &1.01 &\textbf{0.93} & \textbf{1.25} &\textbf{1.12} & \textbf{0.62} &\textbf{0.61} & \textbf{0.47} &\textbf{0.47} & \textbf{6.8} &478 & \textbf{28} &3.9E4 & \textbf{22} &6717 & \textbf{22} &631\\
\spear-\INDU{} &\textbf{0.95} &\textbf{0.96} & \textbf{0.97} &\textbf{29.5} & \textbf{0.85} &\textbf{0.89} & \textbf{0.58} &\textbf{0.6} & \textbf{3.6} &212 & \textbf{9.5} &6402 & \textbf{5.4} &1069 & \textbf{4.3} &139\\
\addlinespace[\interrowspace]
\cplex{}-\BIGMIX{} &\textbf{3E8} &\textbf{0.91} & \textbf{0.93} &\textbf{0.93} & 1.02 &\textbf{0.91} & \textbf{0.64} &\textbf{0.64} & \textbf{3.4} &140 & \textbf{8.3} &1257 & \textbf{4.8} &213 & \textbf{3.5} &111\\
\gurobi{}-\CORLAT{} &\textbf{0.38} &\textbf{0.38} & 0.44 &\textbf{0.37} & \textbf{0.41} &\textbf{0.4} & \textbf{0.38} &\textbf{0.37} & \textbf{3.2} &254 & \textbf{5.2} &1.0E4 & \textbf{5.5} &408 & \textbf{3.3} &101\\
\addlinespace[\interrowspace]
\lkh{}-\TSPLIB{} &9.55 &\textbf{1.09} & \textbf{1.11} &\textbf{0.93} & \textbf{1.77} &\textbf{1.67} & 1.06 &\textbf{0.88} & \textbf{1.6} &50 & \textbf{3.0} &406 & \textbf{0.5} &57 & \textbf{0.1} &5.0\\
\concorde{}-\PORTGEN{} &\textbf{0.41} &\textbf{0.41} & \textbf{0.43} &\textbf{0.42} & 0.43 &\textbf{0.41} & \textbf{0.45} &\textbf{0.44} & \textbf{4.2} &243 & \textbf{3.6} &7362 & \textbf{13} &574 & \textbf{9.9} &283\\
\bottomrule
        \end{tabular}
        \small
    \caption{Quantitative evaluation of the impact of hyperparameter optimization on predictive accuracy.
		    For each model family with hyperparameters, we report performance achieved with and without hyperparameter optimization ($\mathbf{\lambda}_{\text{def}}$ and $\mathbf{\lambda}_{\text{opt}}$, respectively). We show 10-fold cross-validation performance for the default and for hyperparameters optimized using DIRECT with 2-fold cross-validation. \revision{For each dataset and model class, boldface denotes which of $\mathbf{\lambda}_{\text{def}}$ and $\mathbf{\lambda}_{\text{opt}}$ were not statistically significant from the better of the two (boldfacing 3E8 for RR and \cplex{}-\BIGMIX{} is not an error: its poor mean performance stems from a single outlier).}
    Tables D.4 and D.5 (in the online appendix) provide results for all benchmarks.
    \label{tab:results-hyperparam-opt}}
  }
\end{table}

\subsection{Predictive Quality with Sparse Training Data} \label{sec:scaling_N}

\begin{figure}[tpb]
\centering
        \mbox{
\subfigure[\minisat-\INDUHANDRAND{}]{\includegraphics[scale=0.25]{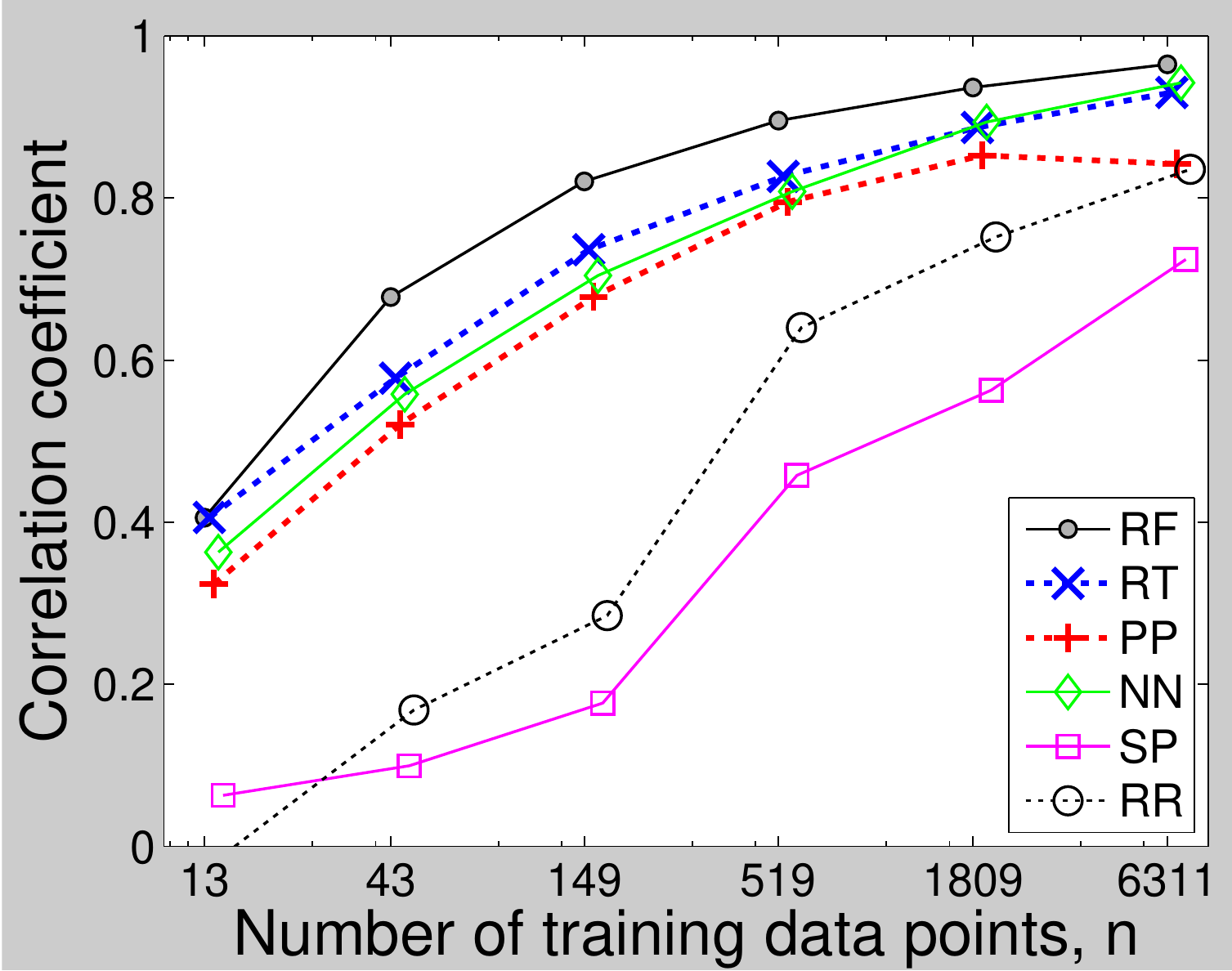}}
            ~~
\subfigure[\cplex{}-\BIGMIX{}]{\includegraphics[scale=0.25]{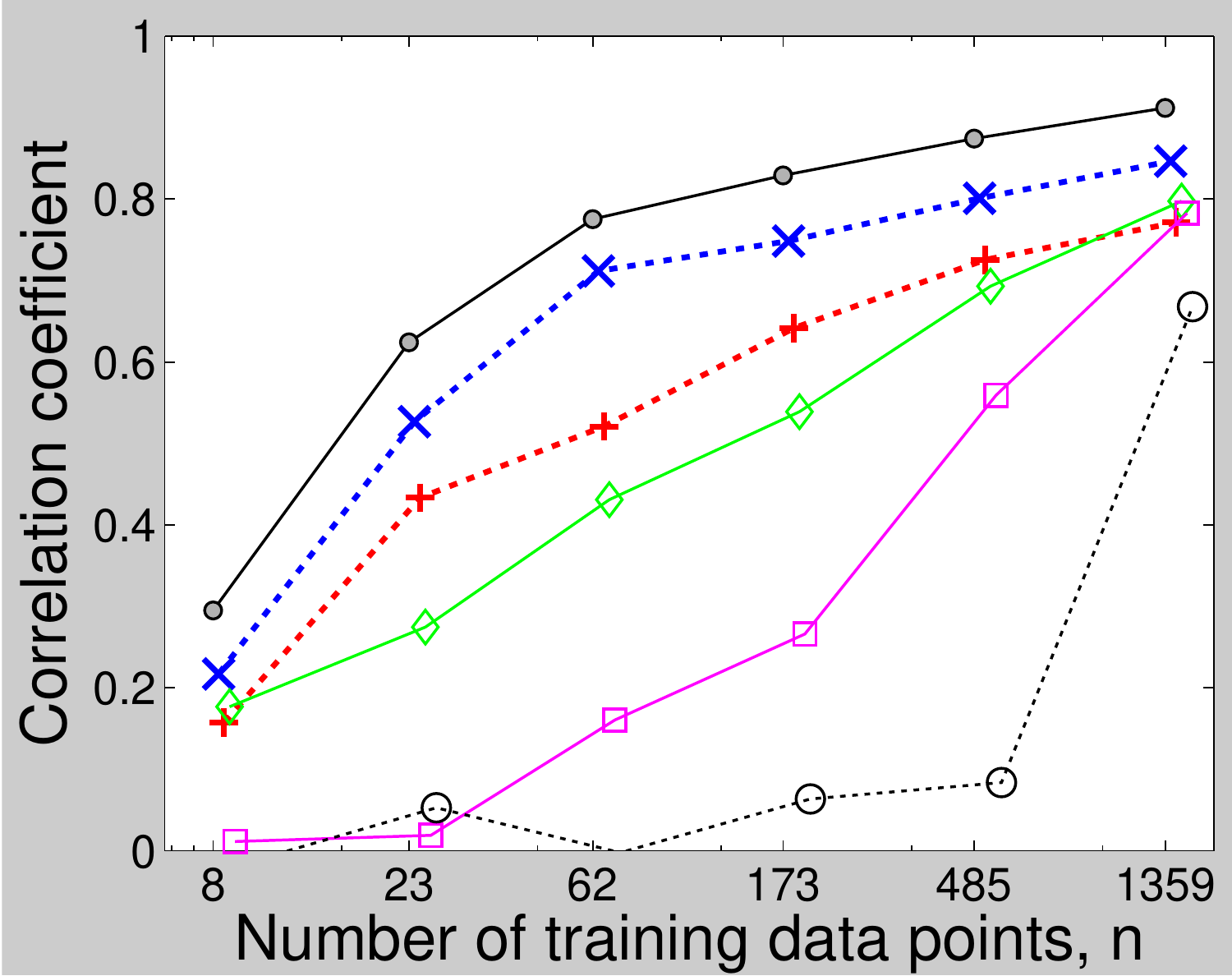}}
            ~~
\subfigure[\cplex{}-\CORLAT{}]{\includegraphics[scale=0.25]{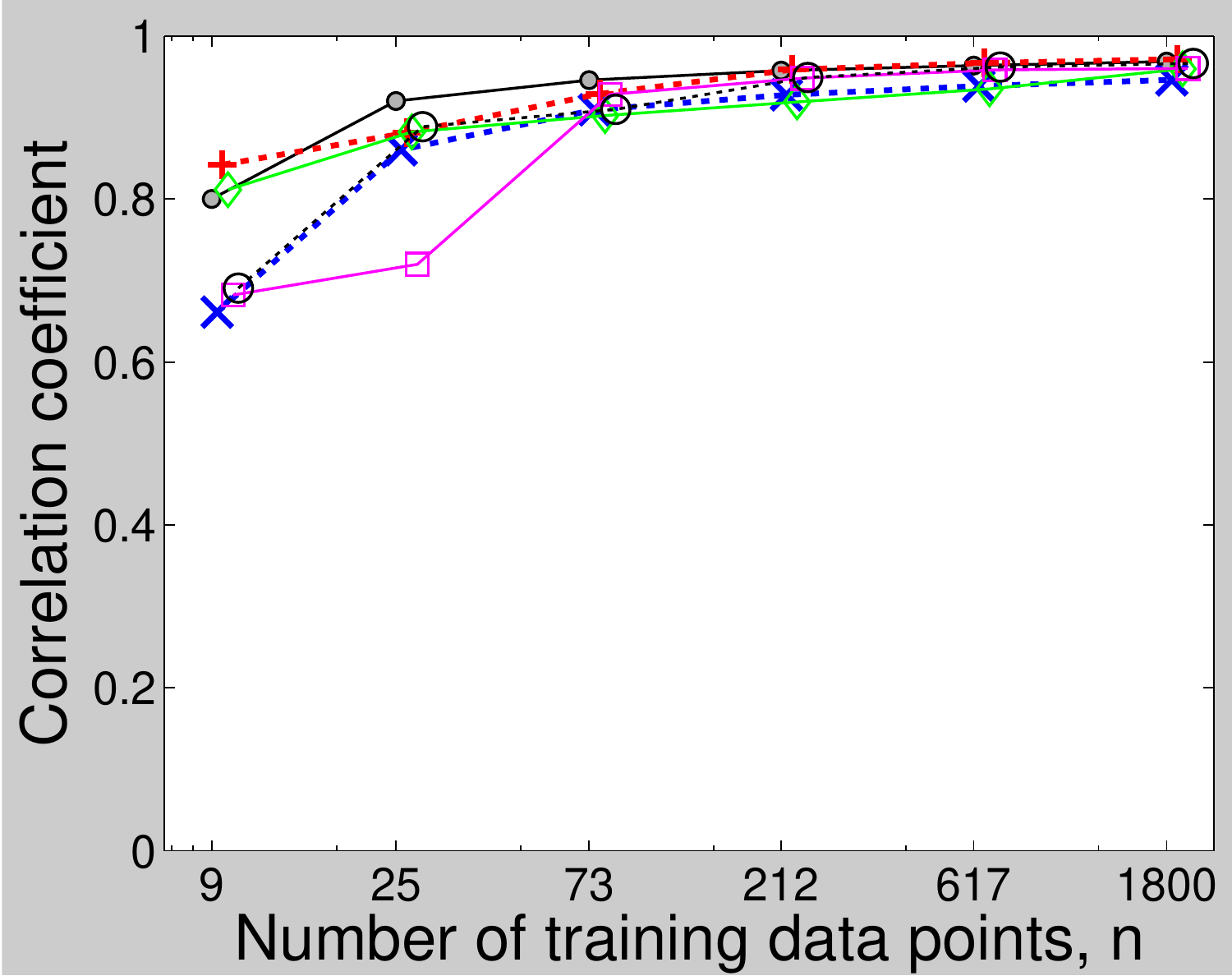}}
        }
        \mbox{
\subfigure[\minisat{}-\SWVIBM{}]{\includegraphics[scale=0.25]{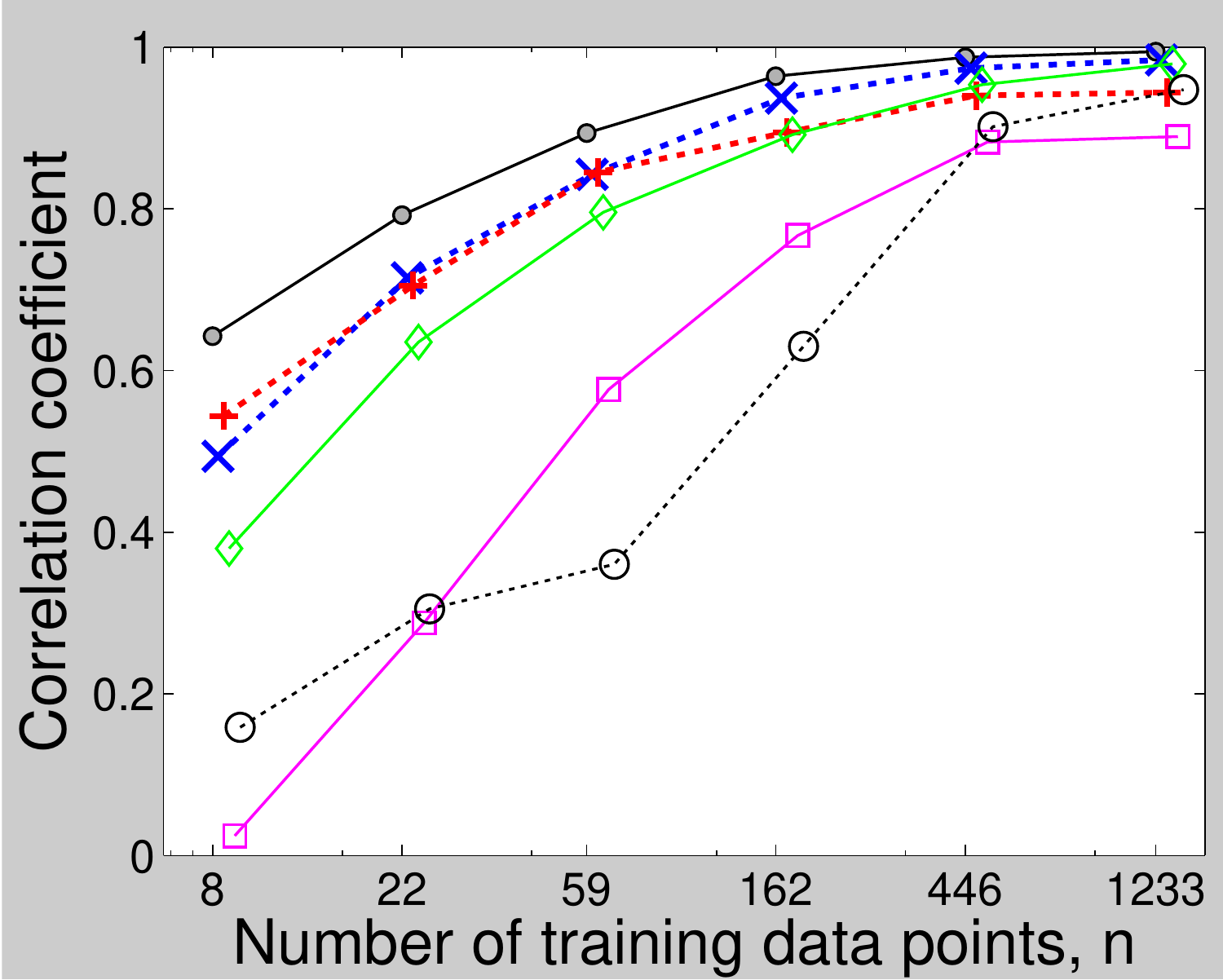}}
            ~~
\subfigure[\concorde{}-\PORTGEN{}\label{fig:scalingN-just-instances-concorde}]{\includegraphics[scale=0.25]{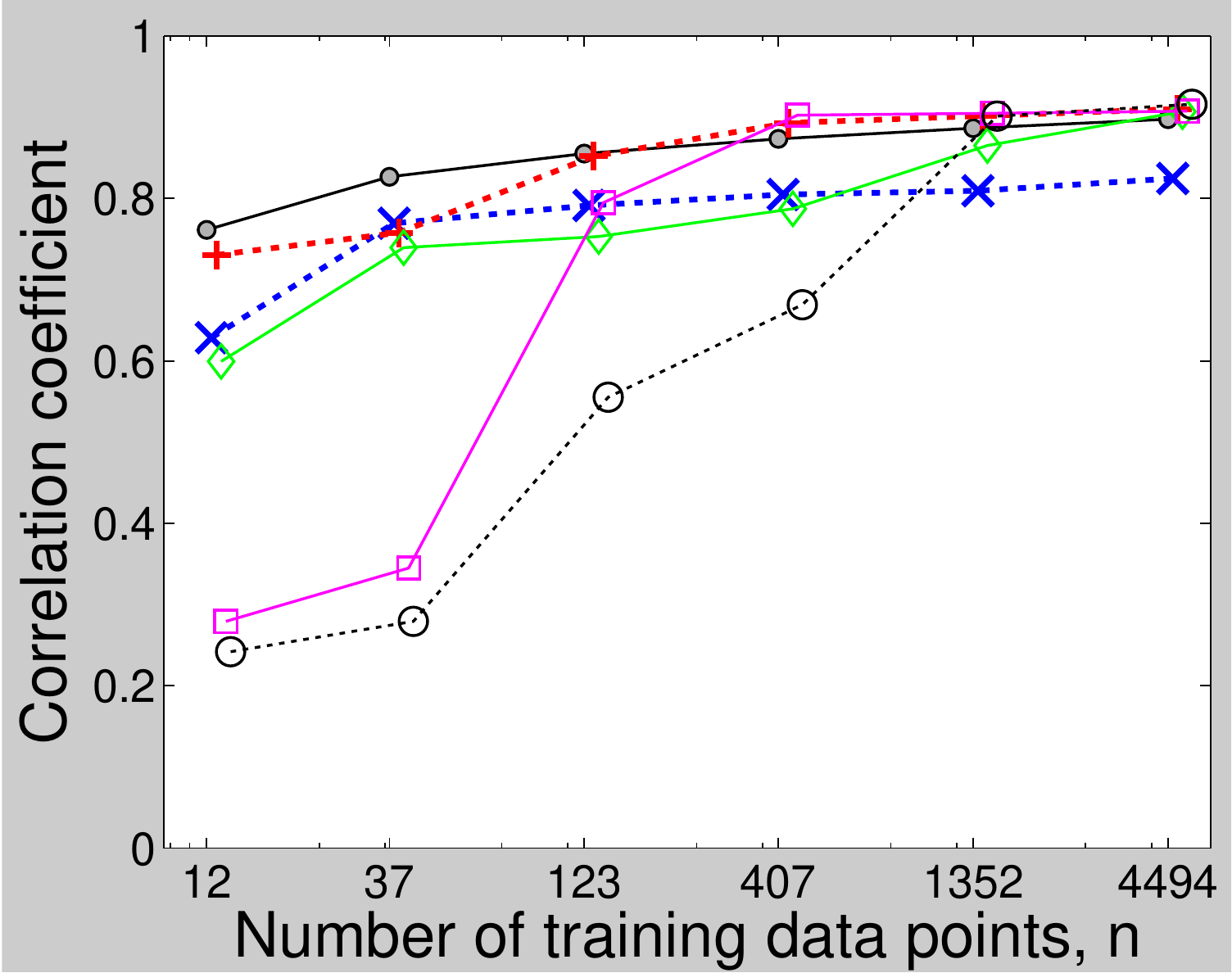}}
            ~~
\subfigure[\lkh{}-\TSPLIB{}\label{fig:scalingN-just-instances-lkh}]{\includegraphics[scale=0.25]{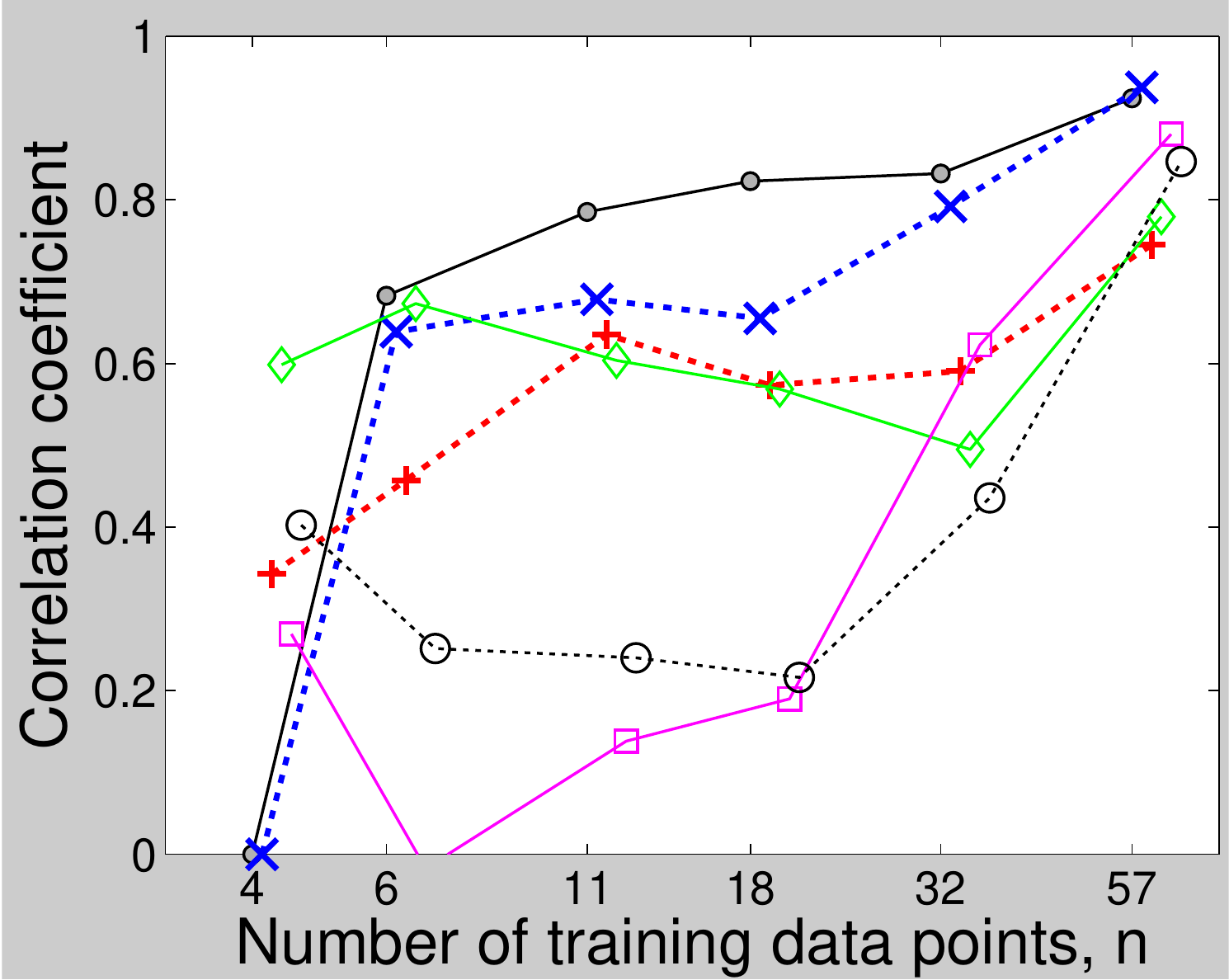}}
				}
    \caption{Prediction quality for varying numbers of training instances.
    For each model and number of training instances, we plot the mean (taken across 10 cross-validation folds) correlation coefficient (CC) between true and predicted runtimes for new test instances; larger CC is better, 1 is perfect.
Figures D.11--D.13 (in the online appendix) show equivalent plots for the other benchmarks.
    \label{fig:feature_space_scaling_def_10fold_cv}}
\end{figure}

We now study how the performance of EPM techniques changes based on the quantity of training data available.
Figure \ref{fig:feature_space_scaling_def_10fold_cv} visualizes this relationship for six representative benchmarks; data for all benchmarks appears in the online appendix.
Here and in the following, we use CC rather than RMSE for such scaling plots, for two reasons. First, RMSE plots are often cluttered due to outlier instances for which prediction accuracy is poor (particularly for the ridge regression methods). Second, plotting CC facilitates performance comparisons across benchmarks, since CC $\in [-1,1]$.

\hide{
The most striking result is that \revision{in many cases} ridge regression performed very poorly with small training sets.
The exception were highly homogeneous data sets, such as \CORLAT{}, for which few data points sufficed for any of the
methods to make quite accurate predictions.
Even for the two TSP benchmarks, where ridge regression performed best based on a large training set, it
performed poorly with a smaller training set (see Figures \ref{fig:scalingN-just-instances-concorde} and \ref{fig:scalingN-just-instances-lkh}).
}

Overall, random forests performed best across training set sizes.
Both versions of ridge regression (SP and RR) performed poorly for small training sets.
\revision{This observation is significant, since most past work employed ridge regression with large amounts of data (\eg{}, in \SATzilla{}~\cite{SATzilla-Full}), only measuring its performance in what turns out to be a favourable condition for it.}

\section{Performance Predictions for New Parameter Configurations}\label{sec:rms_better}

We now move from predicting a single algorithm's runtime across a distribution of \emph{instances} to predicting runtime across a family of algorithms (achieved by changing a given solver's \emph{parameter settings} or \emph{configurations}).
\revision{
For parameterized algorithms, there are four ways in which we can assess the prediction quality achieved by a model:
\begin{enumerate}
    \item \textbf{Predictions for training configurations on training instances.}
    Predictions for this most basic case are useful for succinctly modeling known algorithm performance data.
    Interestingly, several methods already perform poorly here. 
    \item \textbf{Predictions for training configurations on test instances.}
    Such predictions can be used to make a per-instance decision about which of a set of given parameter configurations will perform best on a previously unseen test instance,
		for example in algorithm selection~\cite{SmithMilesAlgoSelectionSurvey,SATzilla-Full,Hydra,isac}.
    \item \textbf{Predictions for test configurations on training instances.}
    This case is important in algorithm configuration, where the goal is to find high-quality
    parameter configurations for the given training instances~\cite{HutHooLey11-SMAC,HutHooLey12-ParallelAC}.
    \item \textbf{Predictions for test configurations on test instances.}
    This most general case 
    is the most natural ``pure prediction'' problem (see \cite{RidKud07,ChiGoe10}). It is also important for per-instance algorithm configuration, where one could use a model to search for the configuration that is most promising for a previously-unseen test instance~\cite{HutHamHooLey06b}.
\end{enumerate}
We can understand the evaluation in the previous section as a special case of 2, where we only consider an algorithm's default configuration, but vary instances.
We now consider the converse case 3, where instances do not vary, but configurations do. We consider case 4, in which we aim to generalize across both parameter configurations and instances, in Section \ref{sec:combination_ehm_rsm}.}

\subsection{Parameter Configuration Spaces}\label{sec:parameter_spaces}

\begin{table}[t]
    {\scriptsize
\centering
    \begin{tabular}{ccccc}
            \toprule
      \textbf{Algorithm} & \textbf{Parameter type} & \textbf{\# parameters of this type} & \textbf{\# values considered} & \textbf{Total \# configurations}\\
            \midrule
                & Categorical & 10 & 2--20& \\
                        \spear{} & Integer & 4 & 5--8 & $8.34 \times 10^{17}$\\
                        & Continuous & 12 & 3--6 & \\
      \addlinespace[.8em]
            & Boolean & 6 & 2& \\
            \cplex{} & Categorical & 45 & 3--7& $1.90 \times 10^{47}$\\
         & Integer & 18 & 5--7& $$\\
            & Continuous & 7 & 5--8& \\
        \bottomrule
        \end{tabular}
        \small
    \caption{Algorithms and characteristics of their parameter configuration spaces. \label{tab:solvers}
    }
  }
\end{table}

Here and in Section \ref{sec:combination_ehm_rsm}, we study two highly parameterized algorithms for two different \revision{problems}: \spear{} for SAT and \cplex{} for MIP.

\revision{For the industrial SAT solver \spear{}~\cite{babic08}, we used the same parameter configuration space as in previous work \cite{HutBabHooHu07}.}
This includes 26 parameters, out of which ten are categorical, four are integral, and twelve are continuous.
The categorical parameters mainly control heuristics for variable and value selection, clause sorting, and resolution ordering, and also enable or disable optimizations, such as the pure literal rule. The continuous and integer parameters mainly deal with activity, decay, and elimination of variables and clauses, as well as with the randomized restart interval and percentage of random choices; we discretized each of them to between three and eight values.
In total, and based on our discretization of continuous parameters, \spear{} has $8.34 \times 10^{17}$ different configurations.

\revision{For the commercial MIP solver IBM ILOG \cplex{}, we used the same configuration space with 76 parameters as in previous work \cite{HutHooLey10-mipconfig}.}
These parameters exclude all \cplex{} settings that change the problem formulation (\eg{}, the optimality gap below which a solution is considered optimal). They include 12 preprocessing parameters (mostly categorical); 17 MIP strategy parameters (mostly categorical); 11 categorical parameters deciding how aggressively to use which types of cuts; 9 real-valued MIP ``limit'' parameters; 10 simplex parameters (half of them categorical); 6 barrier optimization parameters (mostly categorical); and 11 further parameters.
In total, and based on our discretization of continuous parameters, these parameters gave rise to $1.90 \times 10^{47}$ unique configurations.

\subsection{Experimental Setup}\label{sec:parameter_spaces_exp_setup}

\revision{For the experiments in this and the next section, we gathered runtime data for \spear{} and \cplex{} by executing each of them with 1\,000 randomly sampled parameter configurations. We ran each solver on instances from distributions for which we expected it to yield state-of-the-art performance:
\spear{} on  \SWV{} and \IBM{}; \cplex{} on all MIP instance distributions discussed in the previous section.
The runtime data for this and the next section was gathered on the 840-node Westgrid cluster Glacier (each of whose nodes is equipped with two 3.06 GHz Intel Xeon 32-bit processors and \mbox{2--4 GB} RAM). Due to the large number of algorithm runs required
for the experiments described in Section~\ref{sec:combination_ehm_rsm}, we restricted the cutoff time of each single algorithm run to 300 seconds (compared to the 3\,000 seconds for the runs with the default parameter setting used in Section \ref{sec:ehms_better}).
In the following, we consider the performance of EPMs as parameters vary, but instance
features do not; thus, here we used only one instance from each distribution and have no
use for instance features. For each dataset, we selected the easiest benchmark instance amongst the ones for which the default parameter configuration required more than ten seconds on our reference machines.
As before, we used 10-fold cross validation to assess the accuracy of our model predictions for previously unseen parameter configurations.}

\subsection{Predictive Quality}\label{sec:rsm-empirical}

\hide{
orig submission, no tests:
\begin{table}[t]
    {\scriptsize
     \setlength{\tabcolsep}{2pt}
\centering
    \begin{tabular}{l@{\hskip 1em}cccccc@{\hskip 2.5em}cccccc}
    \toprule
 & \multicolumn{6}{c}{\textbf{RMSE}} & \multicolumn{6}{c}{\textbf{Time to learn model (s)}}\\
\cmidrule(r{2.25em}){2-7}\cmidrule{8-13}
\textbf{Domain} &RR &SP &NN &PP &RT &RF &RR &SP &NN &PP &RT &RF\\
\midrule
\cplex{}-\BIGMIX{} & 0.26 & 0.34 & 0.4 & \textbf{0.24} & 0.33 & 0.25 &  5.4 & \textbf{0.84} & 3.9 & 36 & 4.3 & 3\\
\cplex{}-\CORLAT{} & 0.56 & 0.67 & 0.76 & \textbf{0.53} & 0.75 & 0.55 &  5.6 & \textbf{2.6} & 3.8 & 36 & 4.2 & 3\\
\cplex{}-\REG{} & 0.43 & 0.5 & 0.6 & 0.42 & 0.49 & \textbf{0.38} &  5.4 & \textbf{2.2} & 3.7 & 30 & 3.9 & 2.9\\
\cplex{}-\RCW{} & \textbf{0.2} & 0.25 & 0.3 & 0.21 & 0.28 & 0.21 &  5.5 & \textbf{0.53} & 3.7 & 38 & 2.2 & 2\\
\addlinespace[\interrowspace]
\spear{}-\IBM{} & \textbf{0.25} & 0.75 & 0.74 & \textbf{0.25} & 0.31 & 0.28 &  3.1 & \textbf{0.25} & 2.7 & 13 & 1.6 & 1.5\\
\spear{}-\SWV{} & 0.36 & 0.52 & 0.59 & \textbf{0.35} & 0.41 & 0.36 &  2.8 & \textbf{0.22} & 2.6 & 13 & 1.6 & 1.5\\
\bottomrule
        \end{tabular}
        \small
    \caption{Quantitative comparison of models for runtime predictions on previously unseen parameter configurations.
    We report 10-fold cross-validation performance. Lower RMSE is better (0 is optimal).
    Table B.5 (in the online appendix) provides additional results (correlation coefficients and log likelihoods).
    \label{tab:configuration_space_def}}
  }
\end{table}
}

\begin{table}[t]
    {\scriptsize
     \setlength{\tabcolsep}{2pt}
\centering
    \begin{tabular}{l@{\hskip 1em}cccccc@{\hskip 2.5em}cccccc}
    \toprule
 & \multicolumn{6}{c}{\textbf{RMSE}} & \multicolumn{6}{c}{\textbf{Time to learn model (s)}}\\
\cmidrule(r{2.25em}){2-7}\cmidrule{8-13}
\textbf{Domain} &RR &SP &NN &PP &RT &RF &RR &SP &NN &PP &RT &RF\\
\midrule
\cplex{}-\BIGMIX{} &0.26 &0.34 &0.38 &$\bm{0.24}$ &0.33 &$\bm{0.25}$ & 5.33 &$\bm{0.75}$ &3.66 &34.26 &4.24 &2.98\\
\cplex{}-\CORLAT{} &$\bm{0.56}$ &0.67 &0.78 &$\bm{0.53}$ &0.75 &$\bm{0.55}$ & 5.36 &$\bm{2.48}$ &3.88 &32.53 &4.19 &$\bm{3}$\\
\cplex{}-\REG{} &0.43 &0.5 &0.63 &0.42 &0.49 &$\bm{0.38}$ & 5.35 &$\bm{2.09}$ &3.62 &29.28 &4 &2.86\\
\cplex{}-\RCW{} &$\bm{0.2}$ &0.25 &0.29 &$\bm{0.21}$ &0.28 &$\bm{0.21}$ & 5.32 &$\bm{0.43}$ &3.65 &33.6 &2.25 &1.93\\
\addlinespace[\interrowspace]
\spear{}-\IBM{} &$\bm{0.25}$ &0.75 &0.74 &$\bm{0.25}$ &0.31 &0.28 & 2.94 &$\bm{0.17}$ &2.61 &11.3 &1.62 &1.51\\
\spear{}-\SWV{} &$\bm{0.36}$ &0.52 &0.57 &$\bm{0.35}$ &0.41 &$\bm{0.36}$ & 2.79 &$\bm{0.14}$ &2.6 &12.49 &1.68 &1.52\\
\bottomrule
        \end{tabular}
        \small
    \caption{Quantitative comparison of models for runtime predictions on previously unseen parameter configurations.
    We report 10-fold cross-validation performance. Lower RMSE is better (0 is optimal). \revision{Boldface indicates performance not statistically significantly different from the best method in each row.}
    Table D.6 (in the online appendix) provides additional results (correlation coefficients and log likelihoods).
    \label{tab:configuration_space_def}}
  }
\end{table}

\begin{figure}[tbp]
    {\scriptsize
 \setlength{\tabcolsep}{2pt}
\centering
    \begin{tabular}{ccccc}
      ~ & SPORE-FoBa & Neural network & Projected process & Random Forest \\
     \begin{sideways}\spear{}-\IBM{}\end{sideways} &
\includegraphics[scale=0.2]{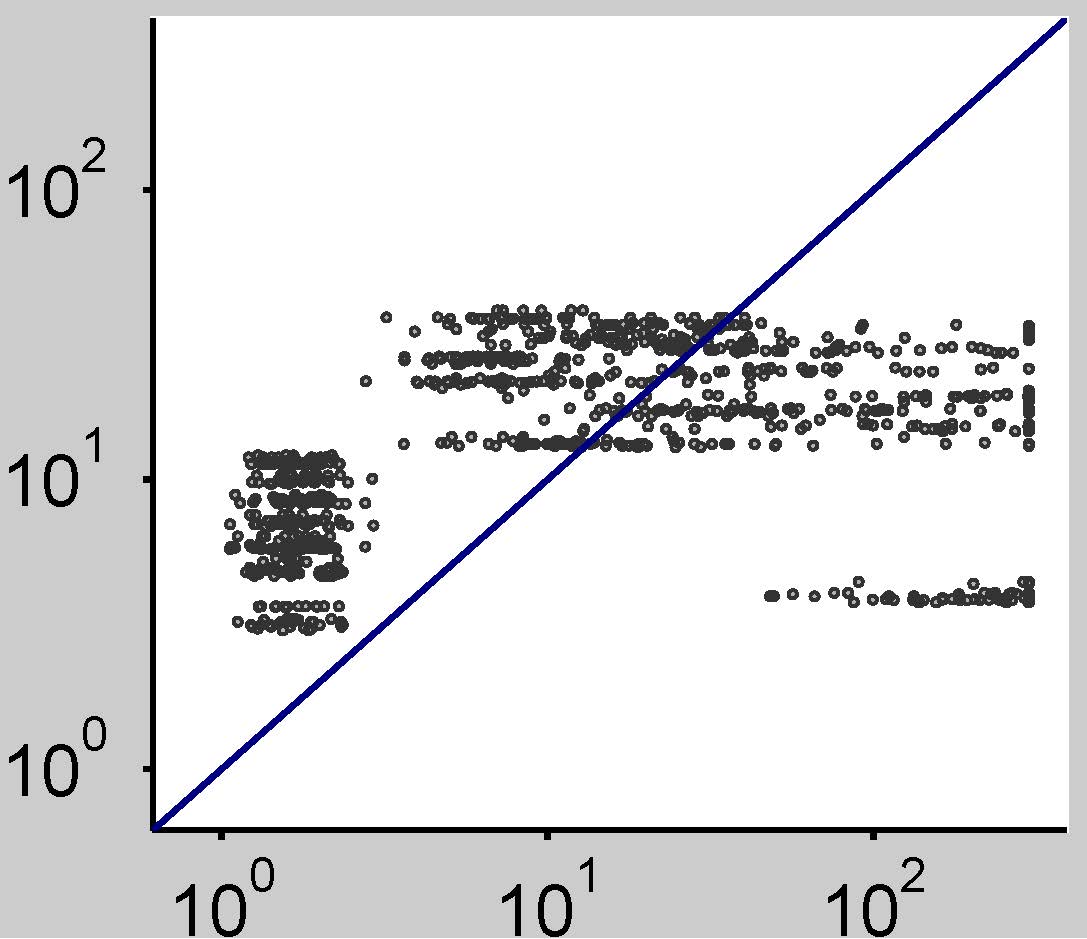} &
\includegraphics[scale=0.2]{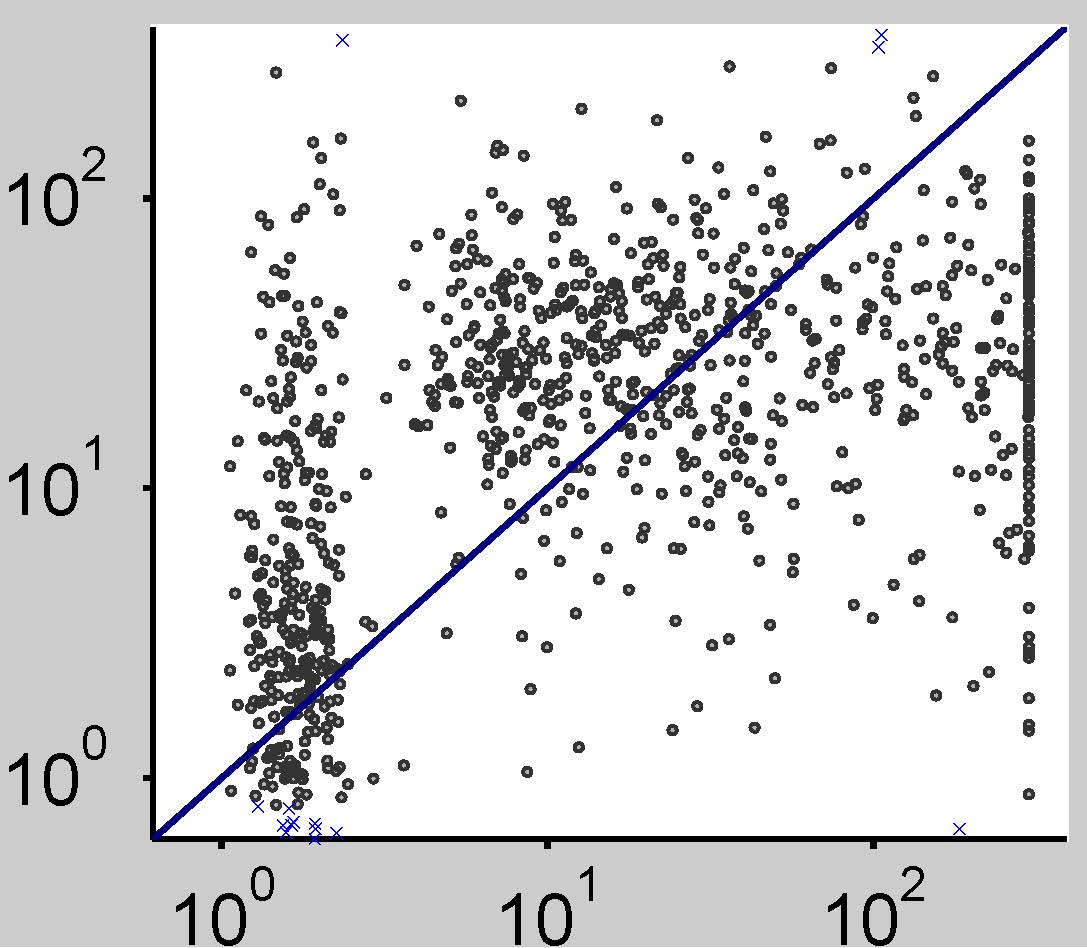} &
\includegraphics[scale=0.2]{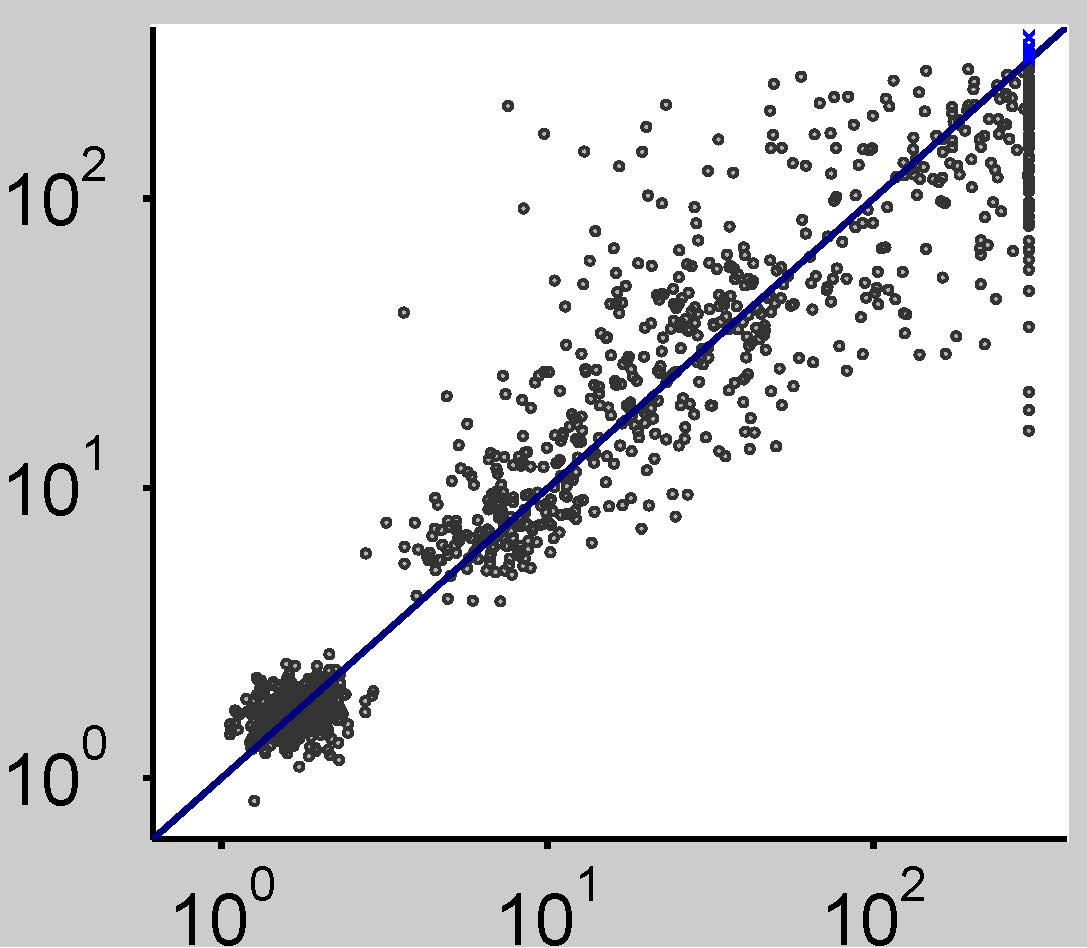} &
\includegraphics[scale=0.2]{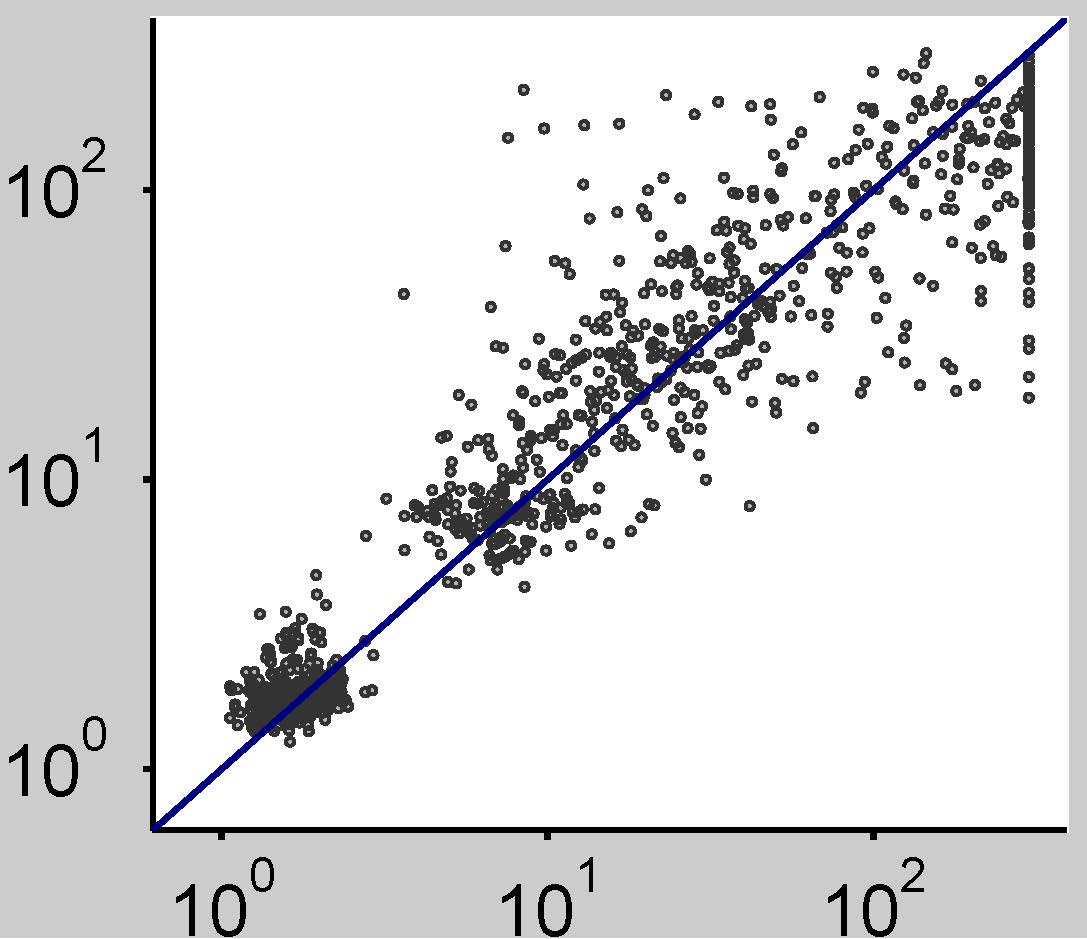}\\
     \begin{sideways}\cplex{}-\CORLAT{}\end{sideways} &
\includegraphics[scale=0.2]{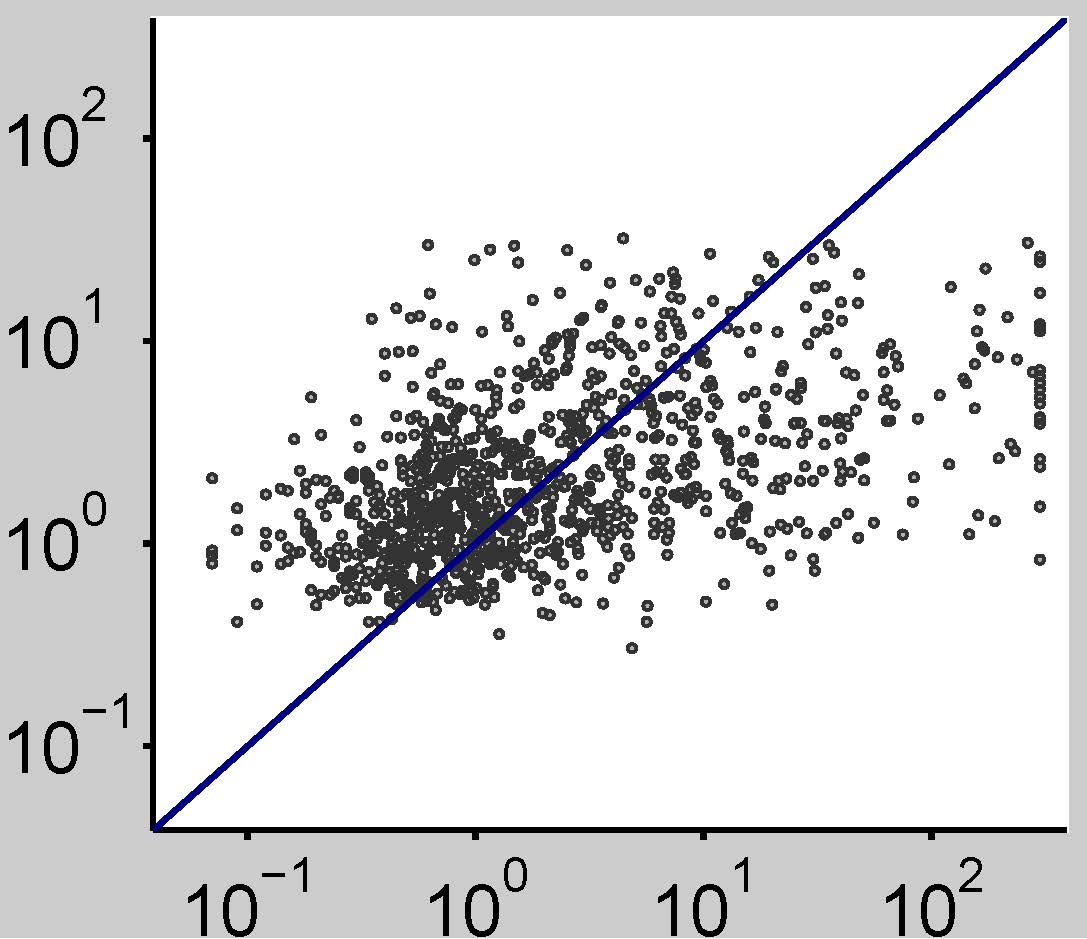} &
\includegraphics[scale=0.2]{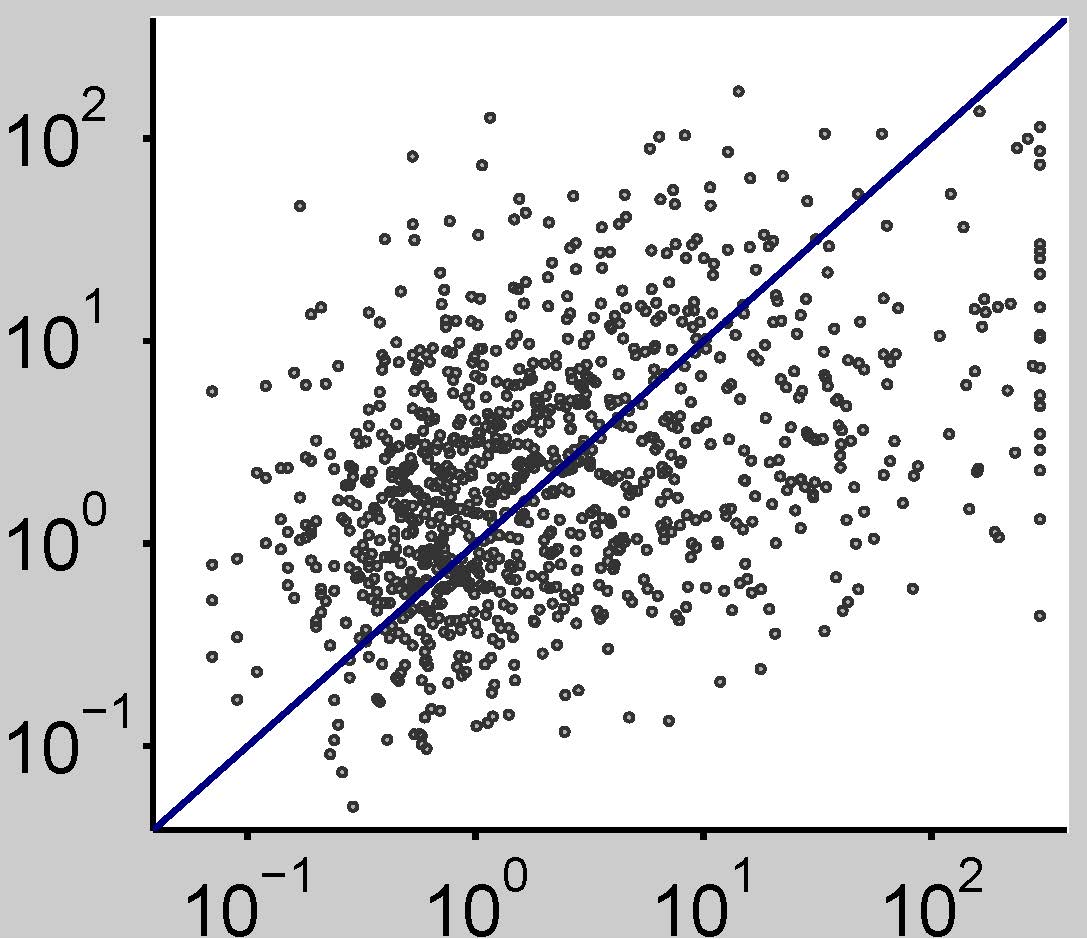} &
\includegraphics[scale=0.2]{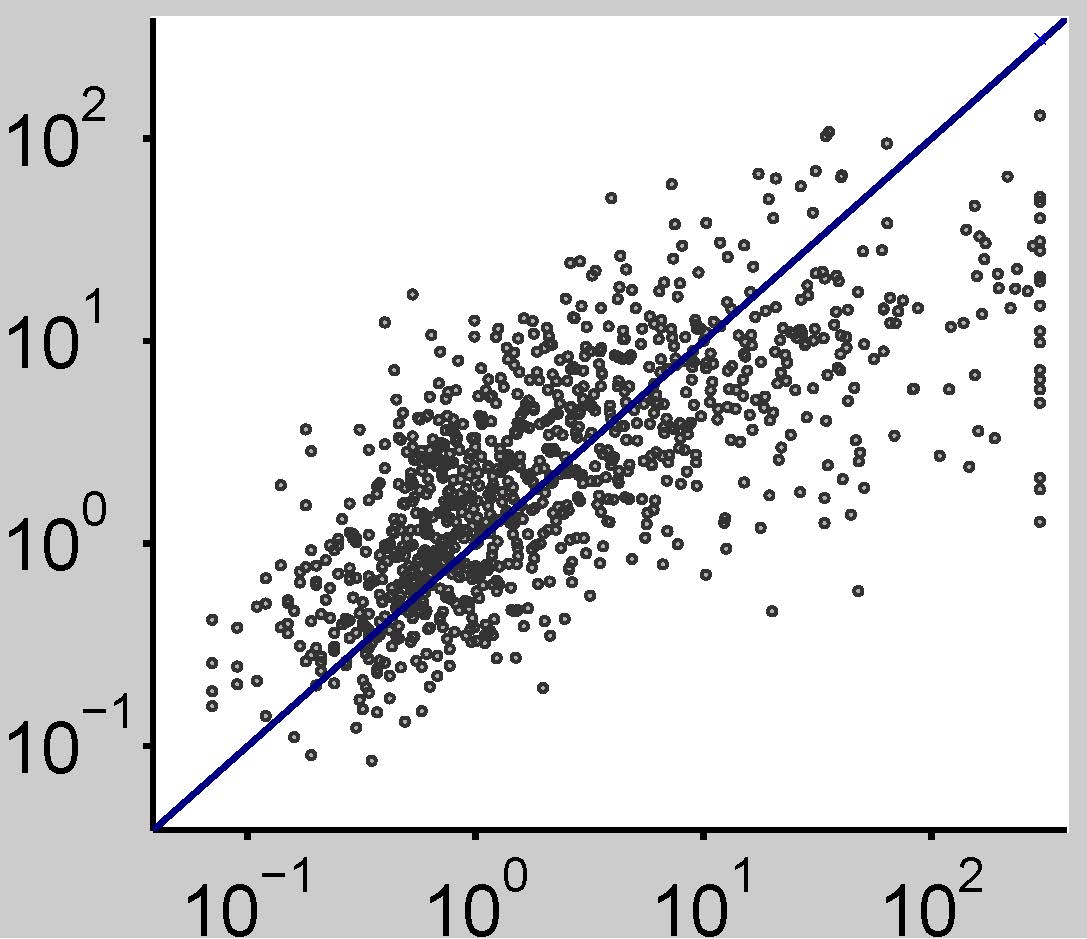} &
\includegraphics[scale=0.2]{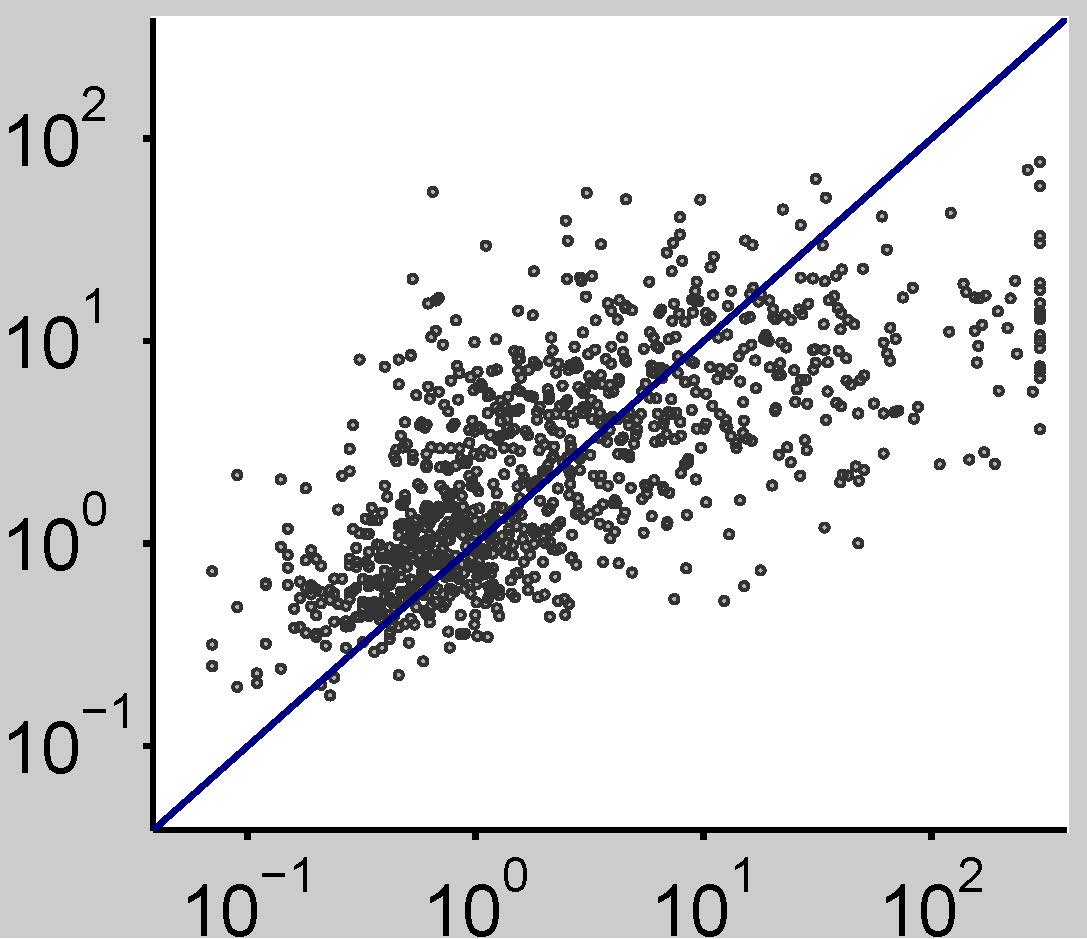}\\

    \end{tabular}
        \small
    \caption{Visual comparison of models for runtime predictions on previously unseen parameter configurations. In each scatter plot, the $x$-axis denotes true runtime and the $y$-axis cross-validated runtime as predicted by the respective model. Each dot represents one parameter configuration. Figures D.14 and D.15 (in the online appendix) provide results for all domains and also show the performance of regression trees and ridge regression variant RR (whose predictions were similar to random forests and projected processes, with somewhat larger spread for regression trees). \label{fig:configuration_space_def}}
  }
\end{figure}

Table \ref{tab:configuration_space_def} quantifies the performance of all models on all benchmark
\revision{problems}, and Figure \ref{fig:configuration_space_def} visualizes predictions.
Again, we see that qualitatively, solver runtime as a function of parameter settings could be predicted quite well by most methods, even as runtimes varied by  factors of over 1\,000 (see Figure \ref{fig:configuration_space_def}).
We observe that projected processes, random forests, and ridge regression variant RR consistently outperformed regression trees; this is significant, as regression trees are the only model that has previously been used for predictions in configuration spaces with categorical parameters~\citep{BarMar04}.
On the other hand, the poor performance of neural networks and of SPORE-FoBa (which mainly differs from variant RR in its feature expansion and selection) underlines that selecting the right (combinations of) features is not straightforward.
Overall, the best performance was achieved by projected processes (applying our kernel function for categorical parameters from Section \ref{sec:GPR-kernel}). As in the previous section, however, random forests were also either best or very close to the best for every data set.

\subsection{Predictive Quality with Sparse Training Data} \label{sec:scaling_N_just_params}

\begin{figure}[tpb]
\centering
        \mbox{
\subfigure[\cplex{}-\BIGMIX{}]{\includegraphics[scale=0.25]{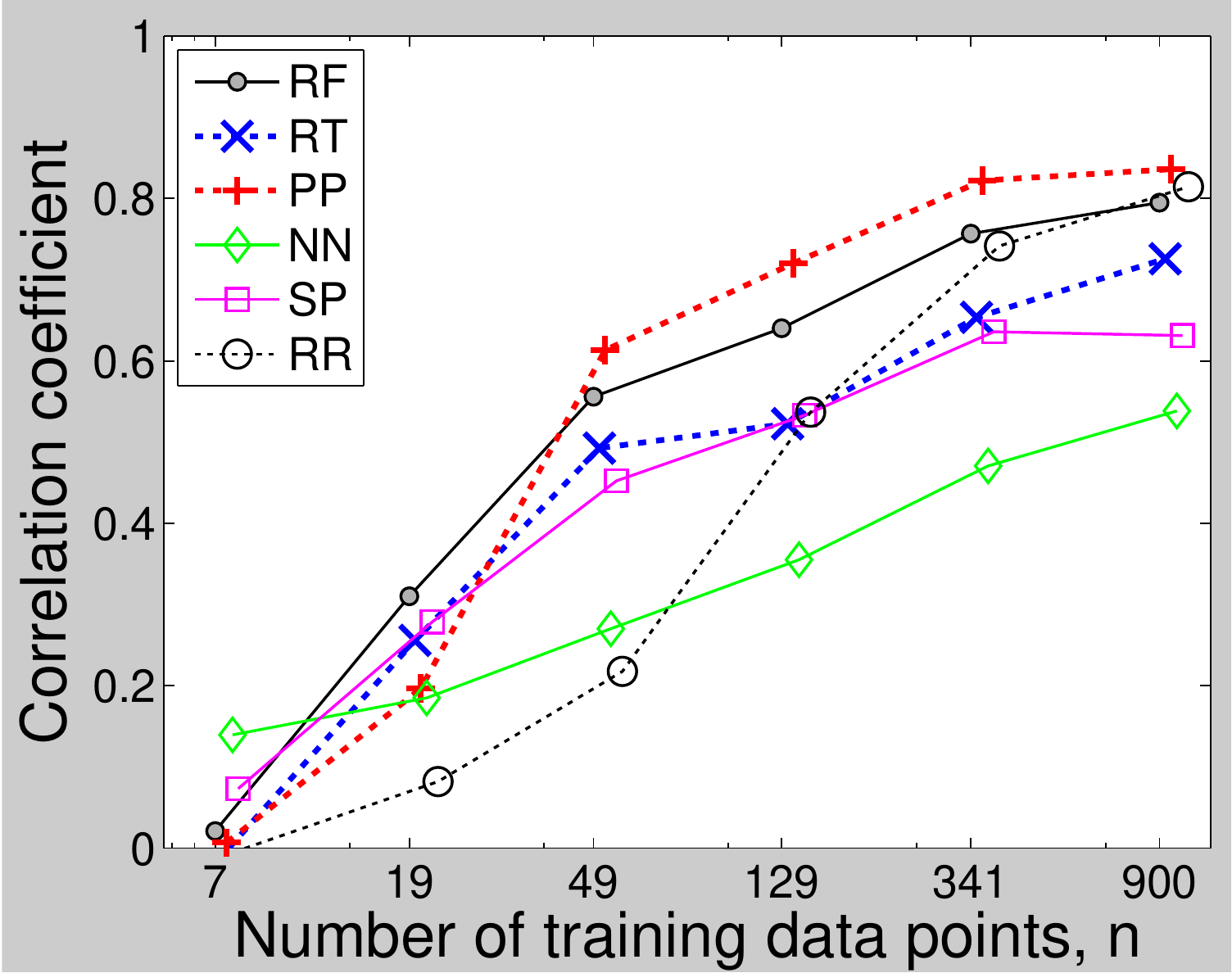}}
            ~~
\subfigure[\cplex{}-\CORLAT{}]{\includegraphics[scale=0.25]{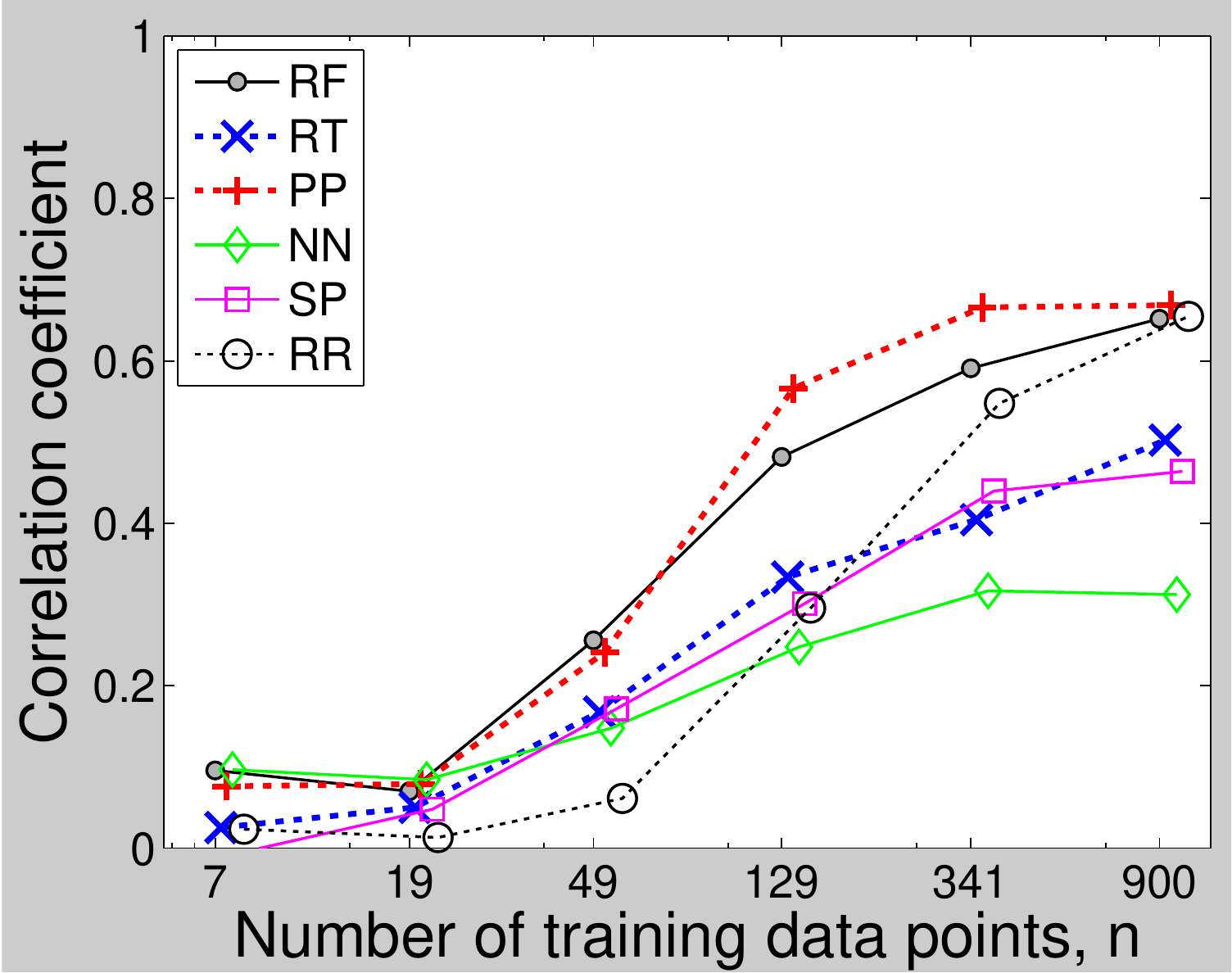}}
\subfigure[\spear{}-\IBM{}]{\includegraphics[scale=0.25]{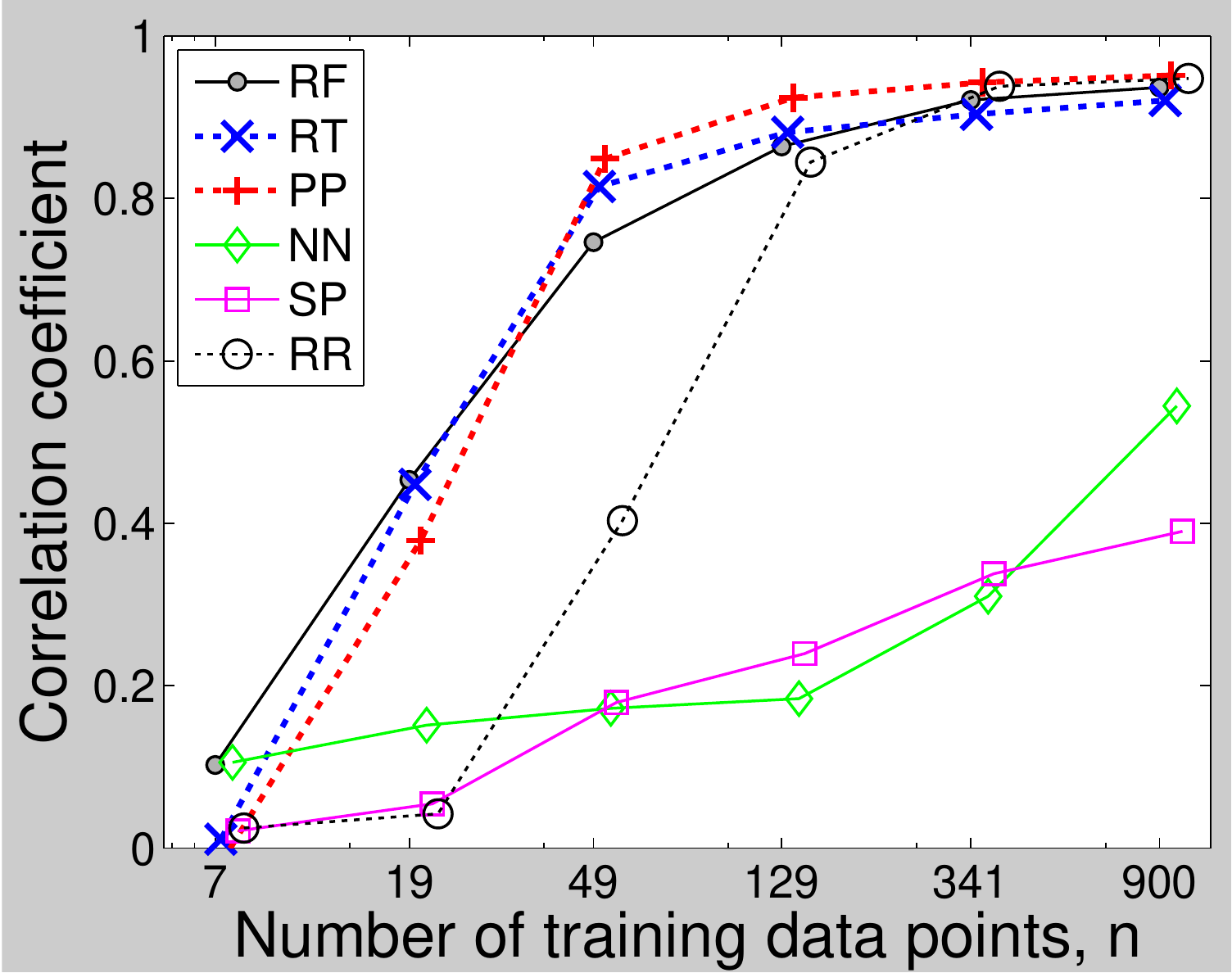}}
        }
    \caption{Quality of predictions in the configuration space, as dependent on the number of training configurations.
    For each model and number of training instances, we plot mean $\pm$ standard deviation of the
correlation coefficient (CC) between true and predicted runtimes for new test configurations.\label{fig:configuration_space_scaling_def}
Figure D.16 (in the online appendix) shows equivalent results for all benchmarks.}
\end{figure}

Results remained similar when varying the number of training configurations.
As Figure \ref{fig:configuration_space_scaling_def} shows, projected processes performed best overall, closely followed by random forests.
Ridge regression variant RR often produced poor predictions when trained using a relatively small number of training data points, but performed well when given sufficient data. Finally, both SPORE-FoBa and neural networks performed relatively poorly regardless of the amount of data given.

\section{Performance Predictions in the Joint Space of Instance Features and Parameter Configurations}\label{sec:combination_ehm_rsm}

We now consider more challenging prediction problems for parameterized algorithms. In the first experiments discussed here (Sections~\ref{sec:matrix_generalizing_full} and \ref{sec:joint-sparse-data}) we tested predictions on the most challenging case, where both configurations and instances are previously unseen. Later in the section (Section~\ref{sec:all-generalizations}) we evaluate predictions made on all four combinations of training/test instances and training/test configurations.

\subsection{Experimental Setup} \label{sec:exp_setup_matrix}

\revision{For the experiments in this section, we used \spear{} and \cplex{} with the same configuration spaces as in Section~\ref{sec:rms_better}
and the same $M=1\,000$ randomly sampled configurations. We ran each of these configurations on all of the $P$ problem instances in
each of our instance sets (with $P$ ranging from 604 to 2\,000), generating runtime data that can be thought of as a $M \times P$ matrix.
We split both the $M$ configurations and the $P$ instances into training and test sets of equal size (using uniform random permutations). We then trained our EPMs on a fixed number of $n$ randomly selected combinations of the $P/2$ training instances and $M/2$ training configurations.}

\revision{We note that while a sound empirical evaluation of our methods 
required gathering a very large amount of data, such extensive experimentation is not required to use them in practice.
The execution of the runs used in this section (between 604\,000 and 2\,000\,000 per instance distribution) took over 60 CPU years, with time requirements for individual data sets ranging between 1.3 CPU years (\spear{} on \SWV{}, where many runs took less than a second) and 18 CPU years (\cplex{} on \RCW{}, where most runs timed out). However, as we will demonstrate in Section \ref{sec:joint-sparse-data}, our methods often yield surprisingly accurate predictions based on data that can be gathered overnight on a single machine.}

\hide{
We used the same runtime matrix data for \spear{} and \cplex{} described in Section~\ref{sec:parameter_spaces_exp_setup}: for each benchmark, this includes the runtimes for $M$ configurations on each of $P$ instances.
We split both the $M$ configurations and the $P$ instances into training and test sets of equal size (using uniform random permutations). We then trained our EPMs on a fixed number of $n$ randomly-selected combinations of the $P/2$ training instances and $M/2$ training configurations.
}

\subsection{Predictive Quality} \label{sec:matrix_generalizing_full}

\begin{table}[t]
\setlength{\tabcolsep}{2.5pt}
{\scriptsize
\centering
\begin{tabular}{l@{\hskip 1.5em}cccccc@{\hskip 2.5em}cccccc}
\toprule
 & \multicolumn{6}{c}{\textbf{RMSE}} & \multicolumn{6}{c}{\textbf{Time to learn model (s)}}\\
 \cmidrule(r{2.25em}){2-7}\cmidrule{9-13}
\textbf{Domain} &RR &SP &NN &PP &RT &RF &RR &SP &NN &PP &RT &RF\\
\midrule
\cplex{}-\BIGMIX{} & $>10^{100}$ & 4.5 & 0.68 & 0.78 & 0.74 & \textbf{0.55} &  \textbf{25} & 34 & 49 & 84 & 52 & 47\\
\cplex{}-\CORLAT{} & 0.53 & 0.57 & 0.56 & 0.53 & 0.67 & \textbf{0.49} &  \textbf{26} & 27 & 52 & 76 & 46 & 40\\
\cplex{}-\REG{} & \textbf{0.17} & 0.19 & 0.19 & 0.19 & 0.24 & \textbf{0.17} &  23 & \textbf{14} & 50 & 77 & 32 & 31\\
\cplex{}-\RCW{} & 0.1 & 0.12 & 0.12 & 0.12 & 0.12 & \textbf{0.09} &  24 & \textbf{13} & 45 & 78 & 25 & 24\\
\cplex{}-\CORLATREG{} & 0.41 & 0.43 & 0.42 & 0.42 & 0.52 & \textbf{0.38} &  \textbf{26} & 37 & 54 & 88 & 47 & 43\\
\cplex{}-\CORLATREGRCW{} & 0.35 & 0.37 & 0.37 & 0.39 & 0.43 & \textbf{0.32} &  \textbf{29} & 35 & 48 & 81 & 38 & 37\\
\addlinespace[\interrowspace]
\spear{}-\IBM{} & 0.58 & 11 & 0.54 & 0.52 & 0.57 & \textbf{0.44} &  \textbf{15} & 31 & 41 & 70 & 36 & 30\\
\spear{}-\SWV{} & 0.58 & 0.61 & 0.63 & 0.54 & 0.55 & \textbf{0.44} &  \textbf{15} & 42 & 41 & 69 & 42 & 28\\
\spear{}-\SWVIBM{} & 0.65 & 0.69 & 0.65 & 0.65 & 0.59 & \textbf{0.45} &  \textbf{17} & 35 & 39 & 70 & 41 & 32\\
\bottomrule
\end{tabular}
        \small
\caption{Root mean squared error (RMSE) obtained by various models for runtime predictions on unseen instances and configurations. \revision{Boldface indicates the best average performance in each row.}
For \cplex{}-\BIGMIX{}, RR had a few extremely poorly predicted outliers, with the maximal prediction of $\log{}_{10}$ runtime exceeding $10^{100}$ (\ie{}, a runtime prediction above $10^{10^{100}}$); thus, we can only bound its RMSE from below.  
    Models were based on 10\,000 data points. Table D.7 (in the online appendix) provides additional results (correlation coefficients and log likelihoods).
    \label{tab:combined_space_def}}
}
\end{table}

\hide{
Domain & \multicolumn{6}{|c|}{Pearson correlation coefficient} & \multicolumn{6}{|c}{Log-likelihood}\\
 &RR &SP &NN &PP &RT &RF &RR &SP &NN &PP &RT &RF\\
\cplex{}-\BIGMIX{} & -0 & 0.12 & 0.77 & 0.7 & 0.75 & \textbf{0.85} &  -Inf & -1e+11 & -2.3e+09 & -0.98 & -1.2 & \textbf{-0.59}\\
\hline
\cplex{}-\CORLAT{} & 0.94 & 0.93 & 0.94 & 0.94 & 0.91 & \textbf{0.95} &  -1.1 & -1.6e+09 & -1.6e+09 & -0.73 & -1.1 & \textbf{-0.51}\\
\hline
\cplex{}-\REG{} & 0.53 & 0.36 & 0.43 & 0.49 & 0.38 & \textbf{0.54} &  -0.93 & -1.8e+08 & -1.8e+08 & \textbf{0.33} & -0.95 & 0.09\\
\hline
\cplex{}-\RCW{} & 0.57 & -0 & 0.34 & 0.33 & 0.48 & \textbf{0.66} &  -0.92 & -7.3e+07 & -6.8e+07 & \textbf{0.75} & -0.93 & 0.19\\
\hline
\cplex{}-\CORLATREG{} & \textbf{0.95} & 0.94 & 0.94 & 0.94 & 0.92 & \textbf{0.95} &  -1 & -9.3e+08 & -8.9e+08 & -0.36 & -1.1 & \textbf{-0.24}\\
\hline
\cplex{}-\CORLATREGRCW{} & 0.95 & 0.94 & 0.94 & 0.93 & 0.92 & \textbf{0.96} &  -0.98 & -7e+08 & -6.7e+08 & \textbf{-0.1} & -1 & -0.11\\
\hline
\spear{}-\IBM{} & 0.93 & 0.22 & 0.93 & 0.94 & 0.93 & \textbf{0.96} &  -1.1 & -5.9e+11 & -1.5e+09 & -0.73 & -1.1 & \textbf{-0.35}\\
\hline
\spear{}-\SWV{} & 0.92 & 0.91 & 0.91 & 0.93 & 0.93 & \textbf{0.96} &  -1.1 & -1.9e+09 & -2e+09 & -0.72 & -1.1 & \textbf{-0.35}\\
\hline
\spear{}-\SWVIBM{} & 0.91 & 0.9 & 0.91 & 0.91 & 0.93 & \textbf{0.96} &  -1.1 & -2.4e+09 & -2.1e+09 & -0.88 & -1.1 & \textbf{-0.42}\\
\hline
}

\begin{figure}[tbp]
    {\scriptsize
 \setlength{\tabcolsep}{2pt}
\centering
    \begin{tabular}{ccccc}
      ~ & Ridge regression (RR) & Neural network & Projected process & Random Forest \\
     \begin{sideways}\cplex{}-\BIGMIX{}\end{sideways} &
\includegraphics[scale=0.2]{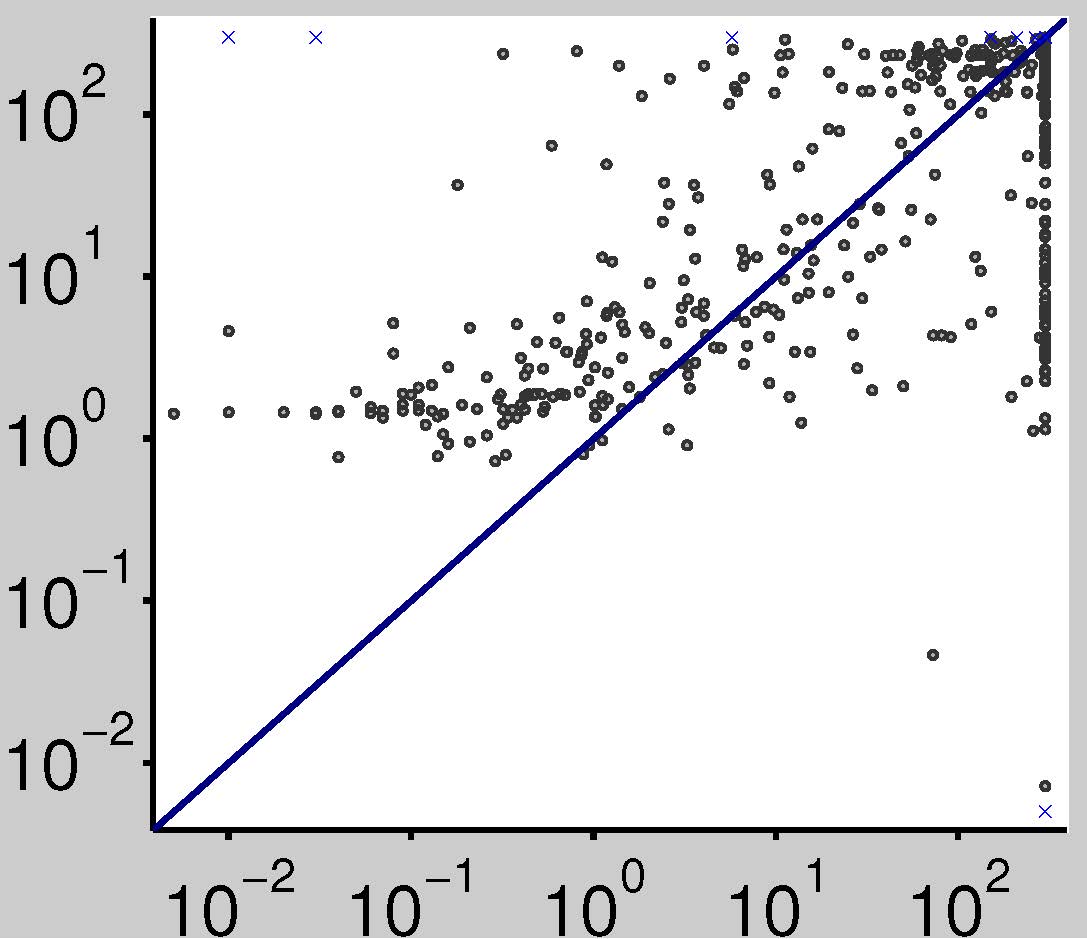} &
\includegraphics[scale=0.2]{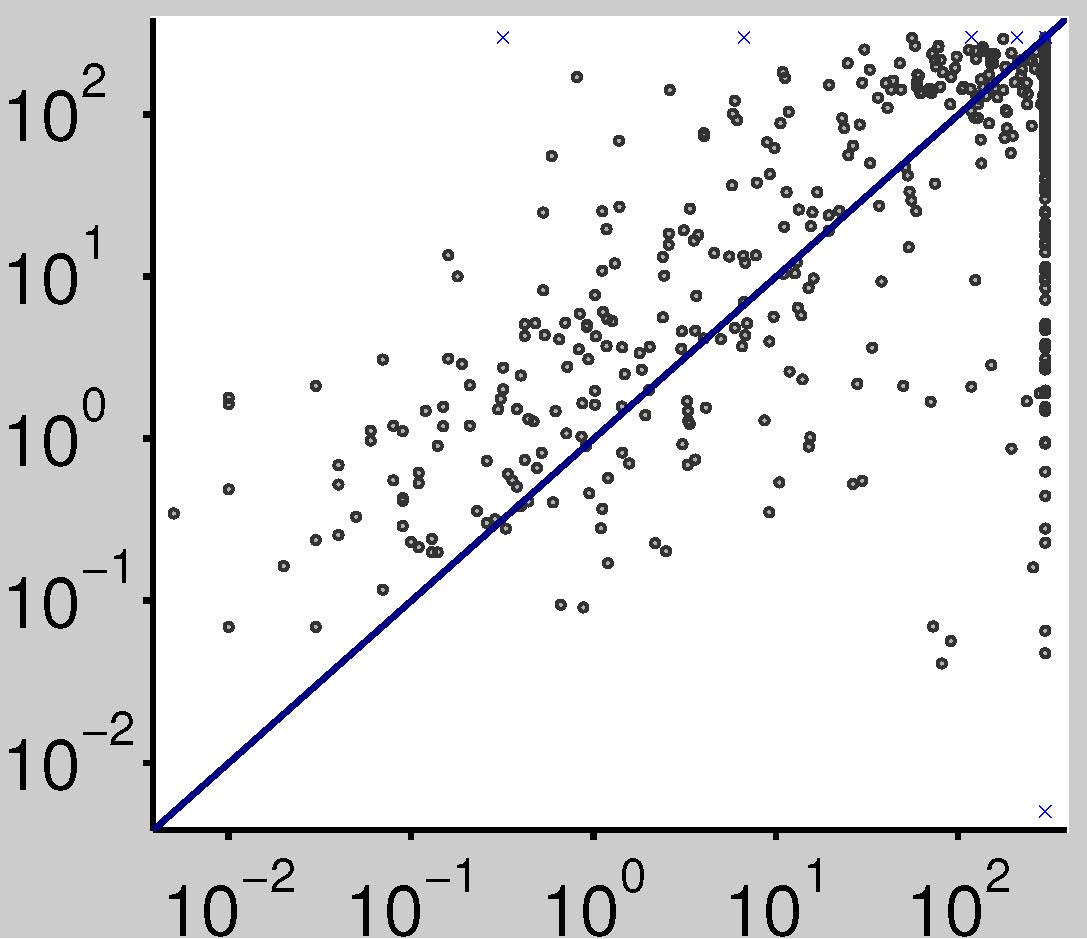} &
\includegraphics[scale=0.2]{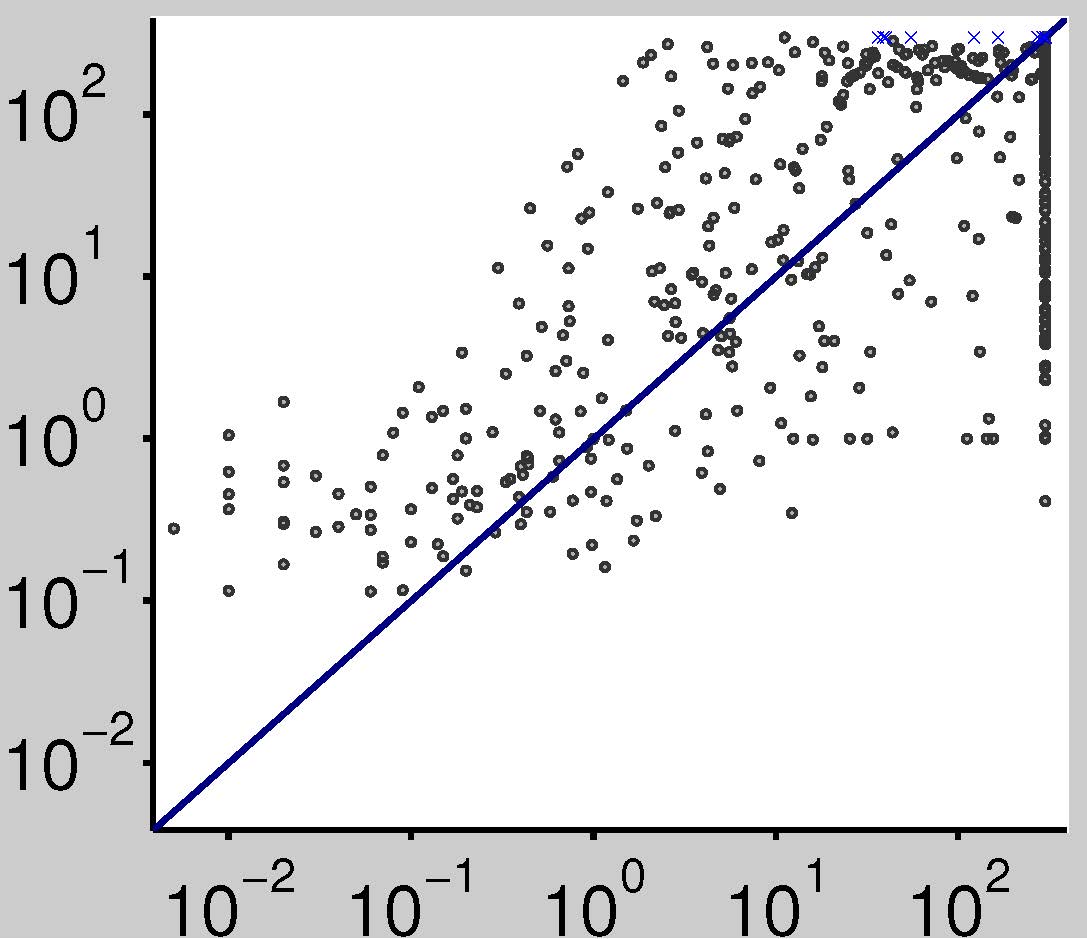} &
\includegraphics[scale=0.2]{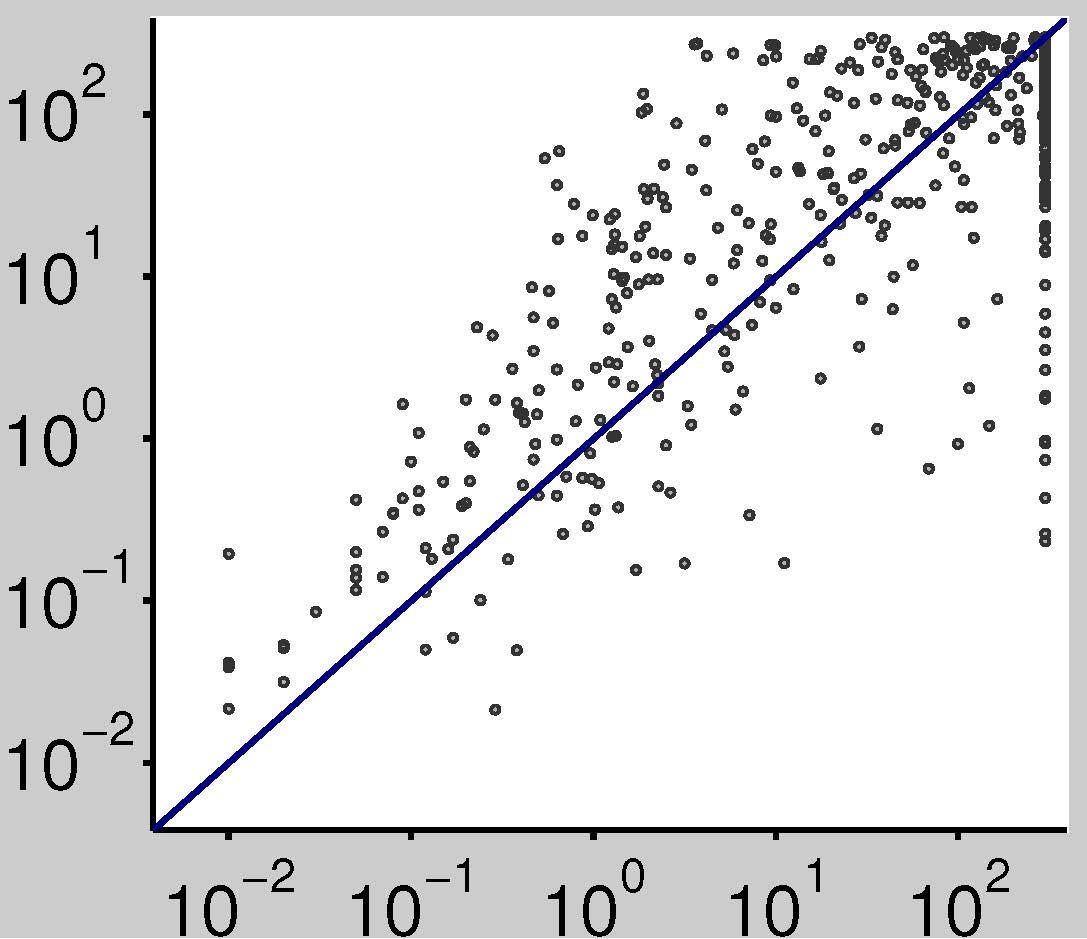} \\
            \hlinespace{}
     \begin{sideways}\spear{}-\SWVIBM{}\end{sideways} &
\includegraphics[scale=0.2]{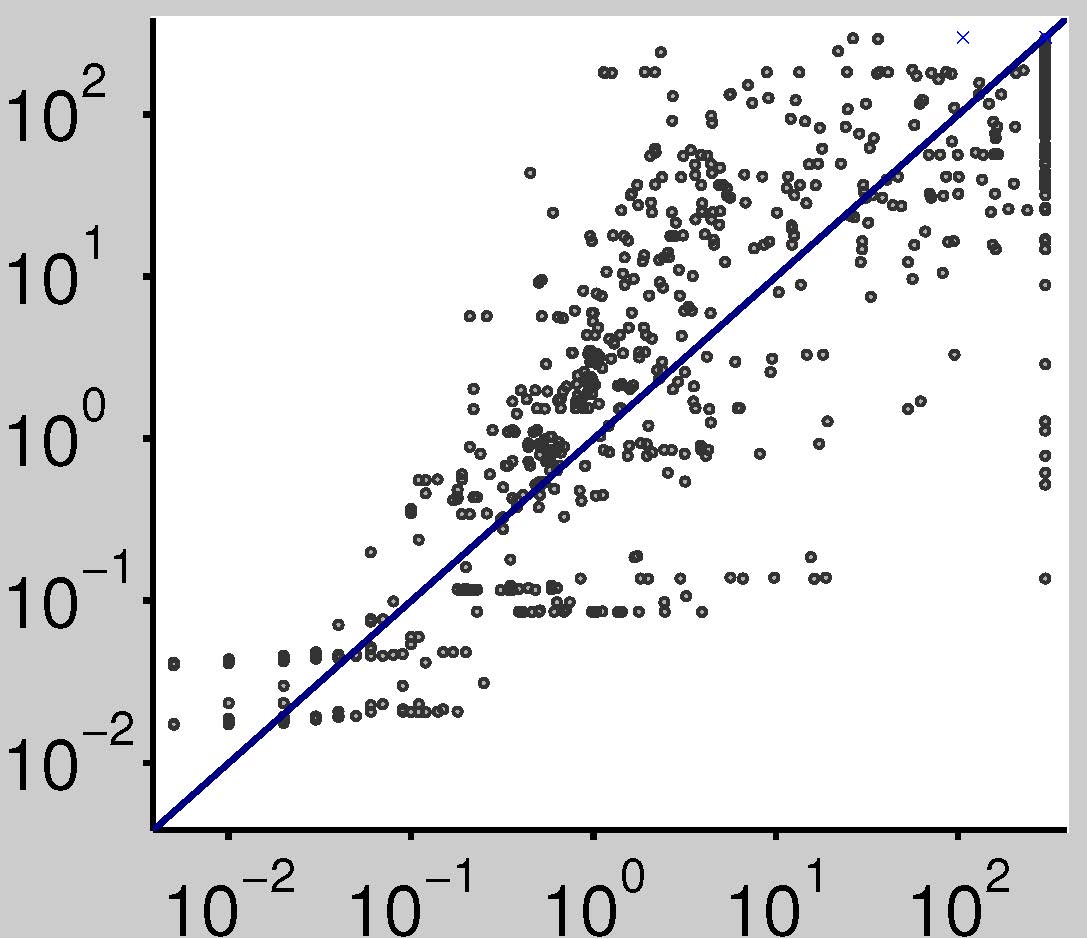} &
\includegraphics[scale=0.2]{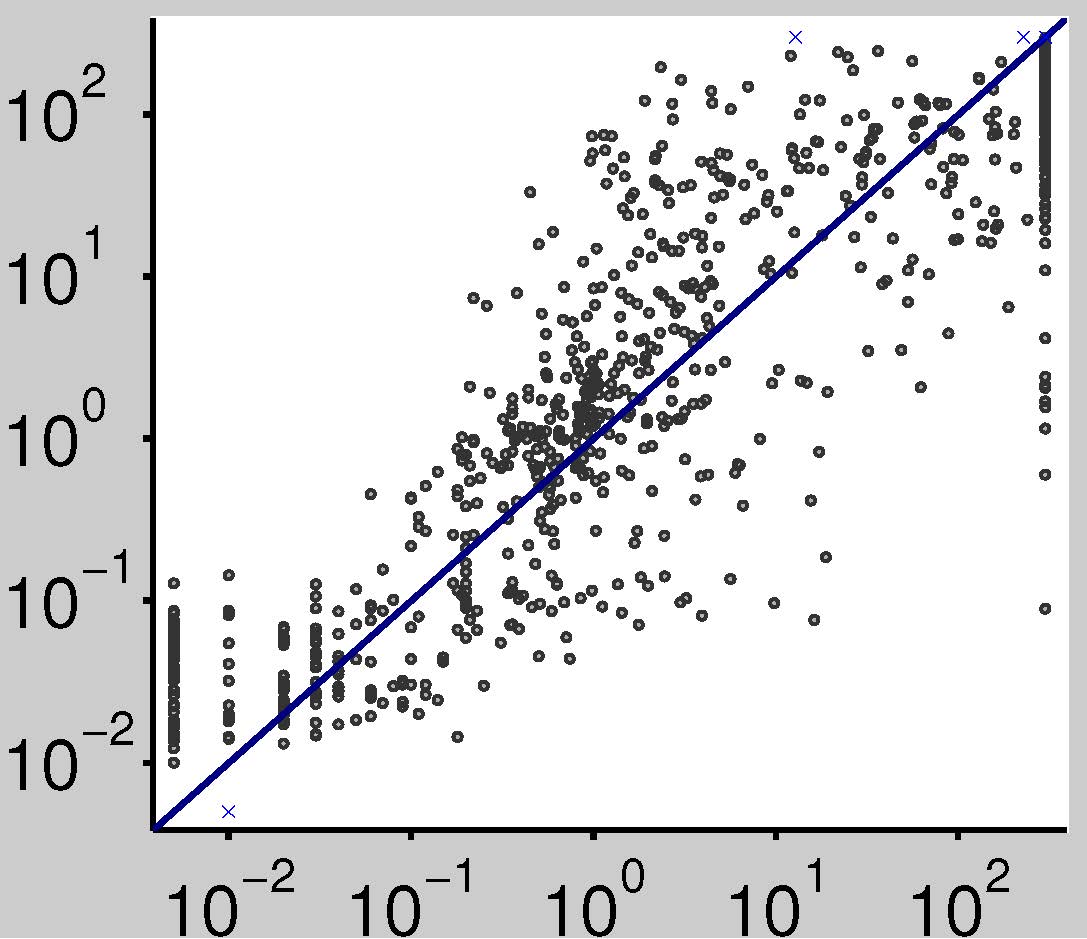} &
\includegraphics[scale=0.2]{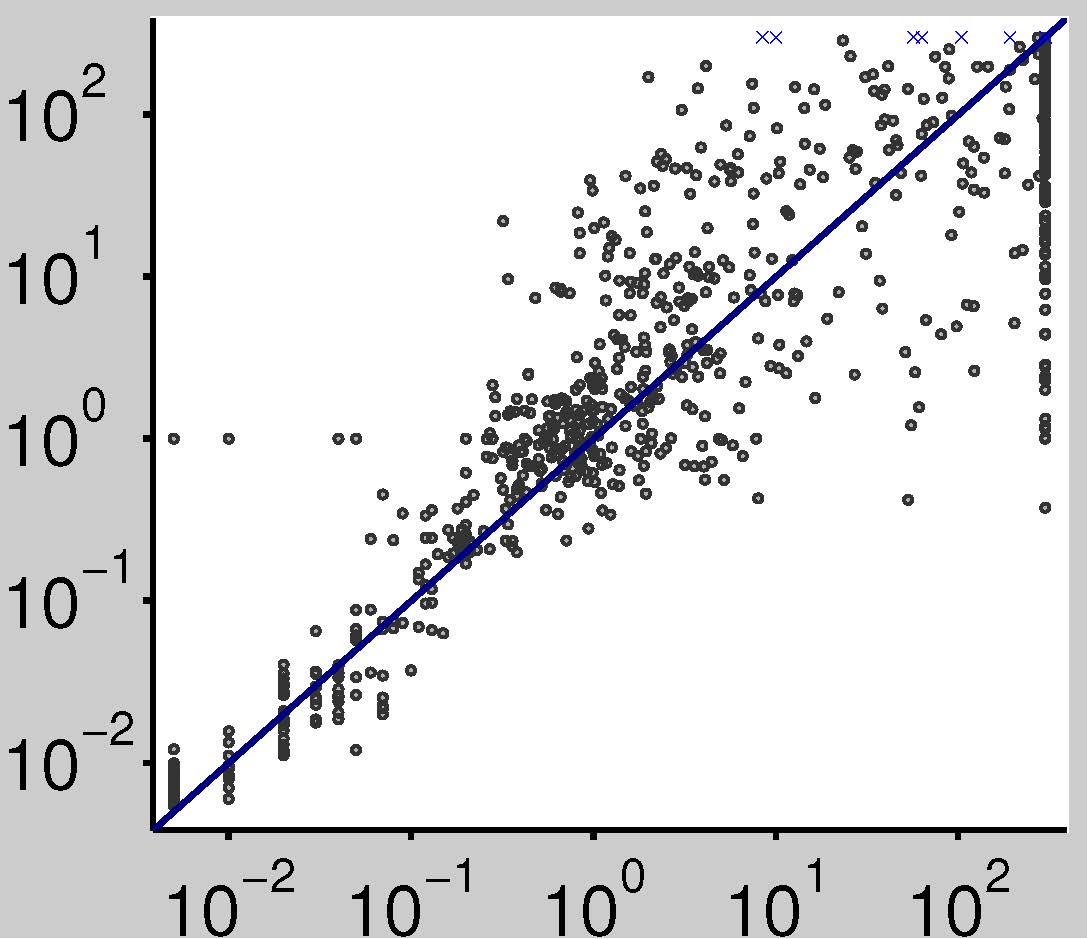} &
\includegraphics[scale=0.2]{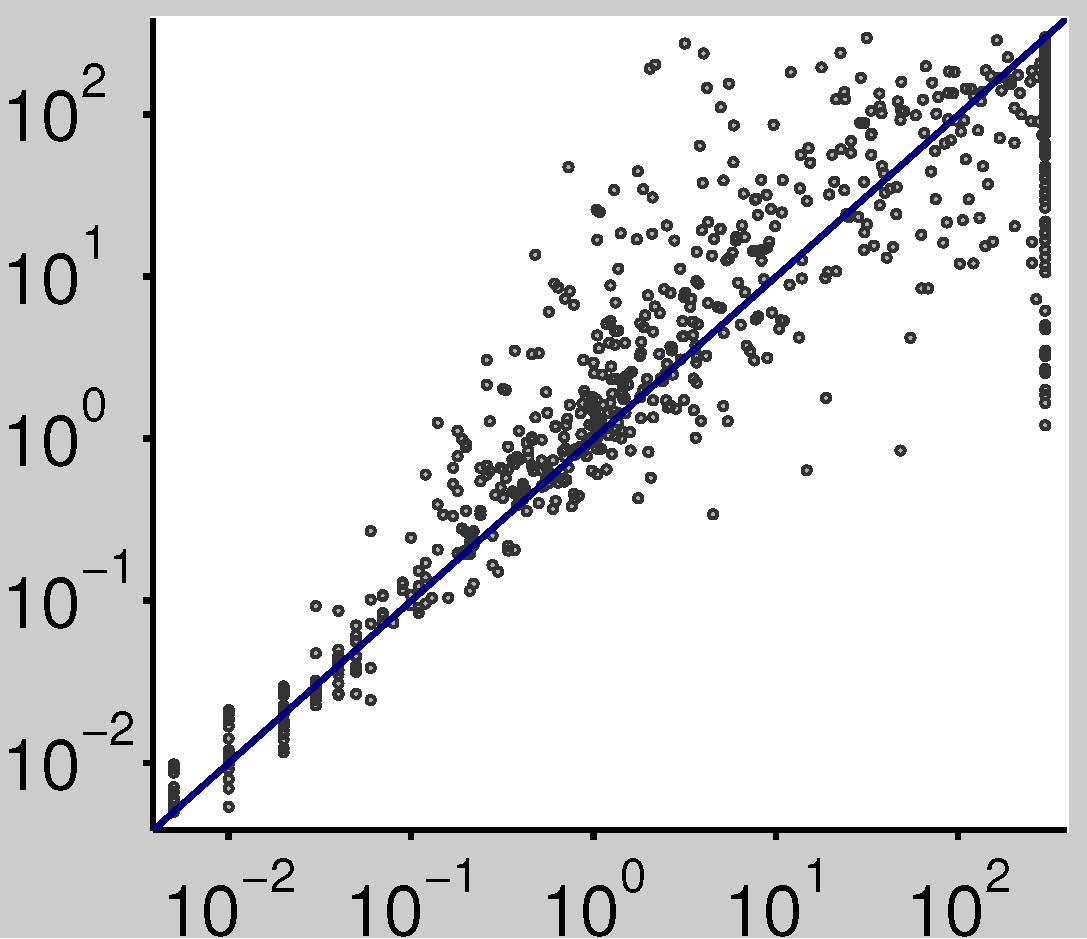} \\
    \end{tabular}
        \small
    \caption{Visual comparison of models for runtime predictions on pairs of previously unseen test configurations and instances. In each scatter plot, the $x$-axis shows true runtime and the $y$-axis cross-validated runtime as predicted by the respective model. Each dot represents one combination of an unseen instance and parameter configuration. Figures D.17--D.19 (in the online appendix) include all domains and also show the performance of SPORE-FoBa (very similar to RR) and regression trees (similar to RF, somewhat larger spread).
    \label{fig:combined_space_def}}
  }
\end{figure}

We now examine the most interesting case, where test instances and configurations were both previously unseen.
Table \ref{tab:combined_space_def} provides quantitative results of model performance based on $n=10\,000$ training data points,
and Figure \ref{fig:combined_space_def} visualizes performance.
Overall, we note that the best models generalized to new configurations \emph{and} to new instances almost as well as to either alone (compare to Sections \ref{sec:ehms_better} and \ref{sec:rms_better}, respectively).
On the most heterogeneous data set, \cplex{}-\BIGMIX{}, we once again witnessed extremely poorly predicted outliers for the ridge regression variants, but in all other cases, the models captured the large spread in runtimes (above 5 orders of magnitude) quite well.
As in the experiments in Section \ref{sec:basic_pred_perf}, the tree-based approaches,
which are able to model different regions of the input space independently,
performed best on the most heterogeneous data sets.
Figure \ref{fig:combined_space_def} also shows some qualitative differences in predictions: for example,
ridge regression, neural networks, and projected processes sometimes overpredicted the runtime of the shortest runs, while the tree-based methods did not have this problem.
Random forests performed best in all cases, which is consistent with their robust predictions in both the instance and the configuration space observed earlier.

\subsection{Predictive Quality with Sparse Training Data} \label{sec:joint-sparse-data}

\begin{figure}[tpb]
\centering
        \mbox{
\subfigure[\cplex{}-\BIGMIX{}]{\includegraphics[scale=0.25]{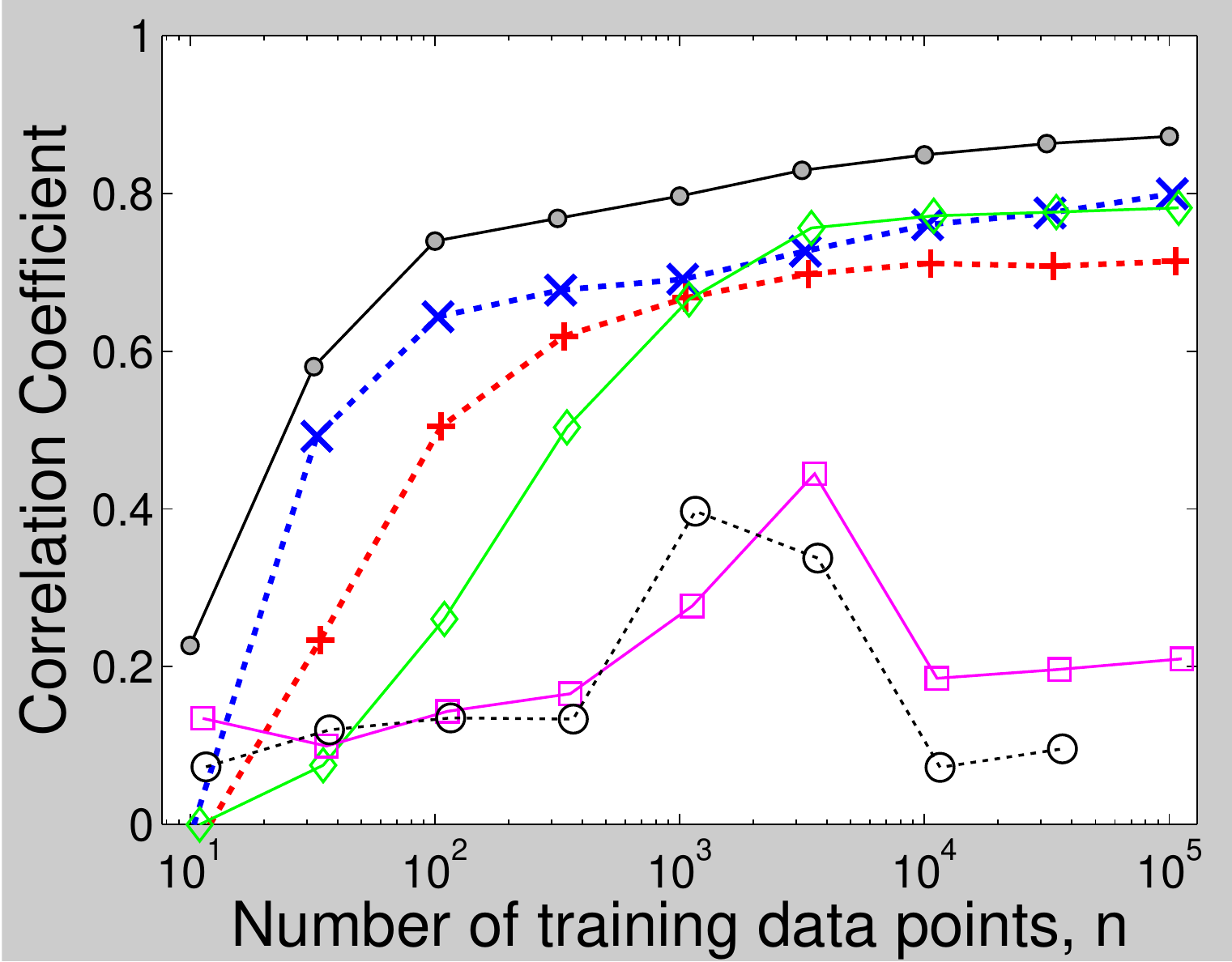}}
            ~~
\subfigure[\cplex{}-\CORLAT{}]{\includegraphics[scale=0.25]{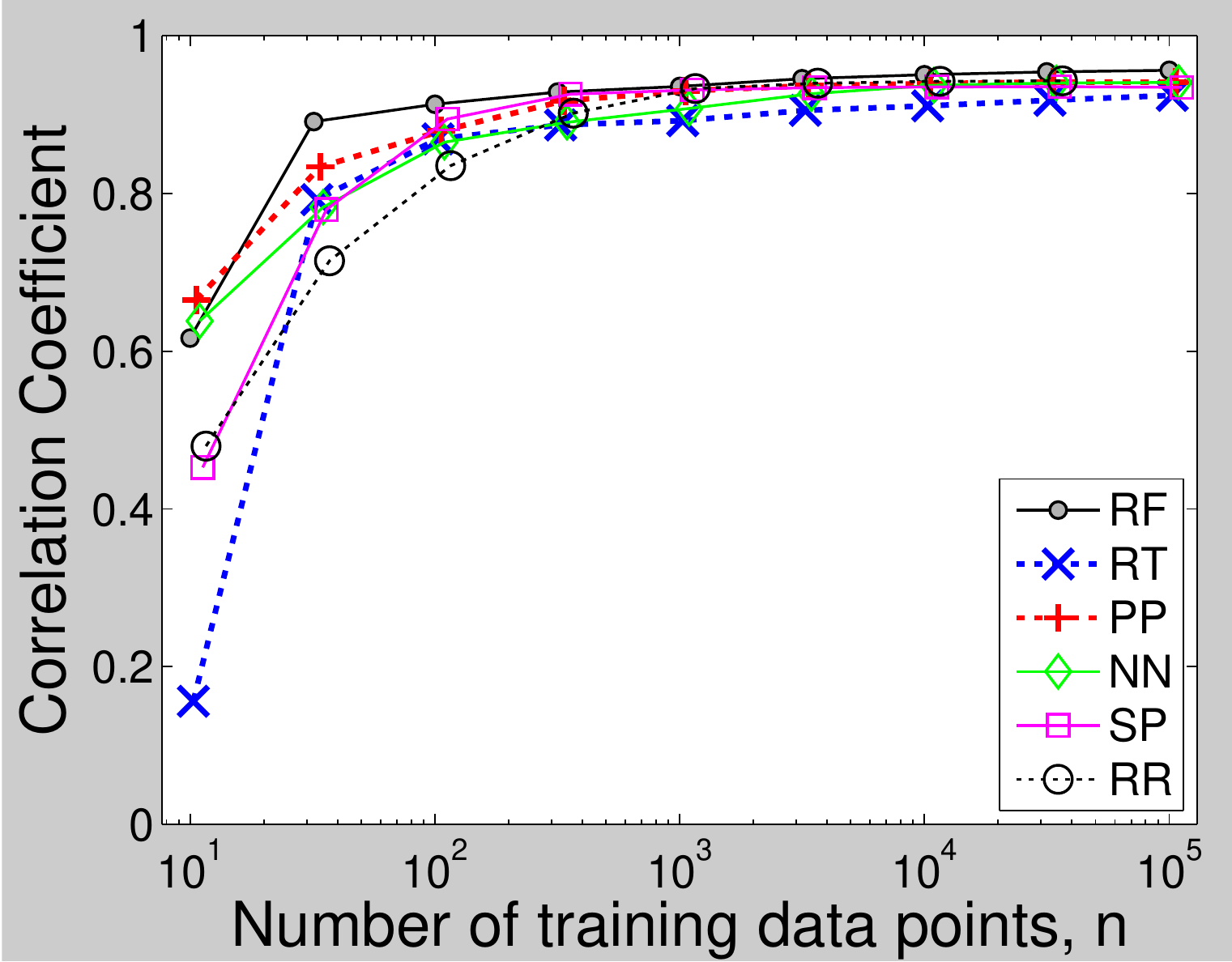}}
            ~~
\subfigure[\spear{}-\SWVIBM{}]{\includegraphics[scale=0.25]{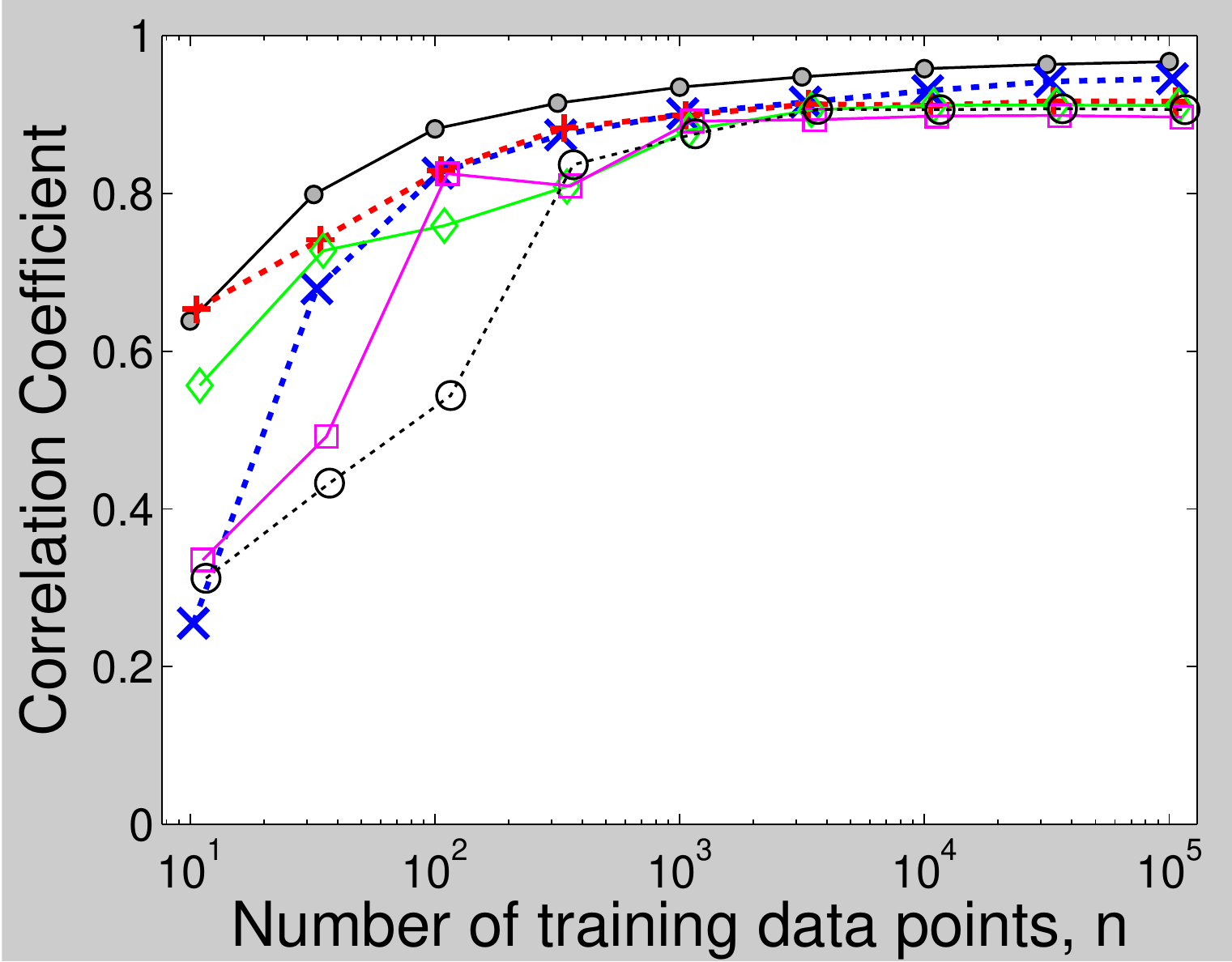}}
        }
    \caption{Quality of predictions in the joint instance/configuration space as a function of the number of training data points.
    For each model and number of training data points, we plot mean correlation coefficients between true and predicted runtimes for new test instances and configurations. We omit standard deviations to avoid clutter, but they are very high for the two ridge regression variants. Figure D.20 (in the online appendix) shows corresponding, and qualitatively similar, results for all benchmarks.\label{fig:combined_space_scaling_def}}
\end{figure}

Next, we studied the amount of data that was actually needed to obtain good predictions, varying the number $n$ of randomly selected combinations of training instances and configurations.
Figure \ref{fig:combined_space_scaling_def} shows the correlation coefficients achieved by the various methods as a function
of the amount of training data available. Overall, we note that most models already performed remarkably well (yielding correlation coefficients of 0.9 and higher) based on a few hundred training data points.
This confirmed the practicality of our methods: on a single machine, it takes at most 12.5 hours to execute 150 algorithm runs with a cutoff time of 300 seconds. Thus, even users without access to a cluster can expect to be able to execute sufficiently many algorithm runs overnight to build a decent empirical performance model for their algorithm and instance distribution of interest.
Examining our results in some more detail, the ridge regression variants again had trouble on the most heterogeneous benchmark \cplex{}-\BIGMIX{}, but otherwise performed quite well.
Overall, random forests performed best across different training set sizes. Naturally, all methods required more data to make good predictions for heterogeneous benchmarks (\eg{}, \cplex{}-\BIGMIX{}) than for relatively homogeneous ones (\eg{}, \cplex{}-\CORLAT{}, for which the remarkably low number of 30 data points already yielded correlation coefficients exceeding 0.9).

\subsection{Evaluating Generalization Performance in Instance and Configuration Space}\label{sec:all-generalizations}

\begin{table}[t]
\centering
 \setlength{\tabcolsep}{2.15pt}
   \scriptsize
   \begin{tabular}{l@{\hskip .8em}l@{\hskip 1em}cccccc@{\hskip 2.5em}cccccc}
   \toprule
 &  & \multicolumn{6}{c}{\textbf{Training configurations}} & \multicolumn{6}{c}{\textbf{Test configurations}}\\
\cmidrule(r{2.25em}){3-8}\cmidrule{9-14}
\textbf{Domain} & \textbf{Instances}&RR &SP &NN &PP &RT &RF &RR &SP &NN &PP &RT &RF\\
\midrule
\multirow{2}{*}{\cplex{}-\BIGMIX{}} & Training & 0.6 & 0.6 & 0.55 & 0.65 & 0.59 & \textbf{0.43} &  0.6 & 0.6 & 0.56 & 0.65 & 0.62 & \textbf{0.45}\\
~ & Test & $>10^{100}$ & 4.5 & 0.67 & 0.78 & 0.71 & \textbf{0.54} &  $>10^{100}$ & 4.5 & 0.68 & 0.78 & 0.74 & \textbf{0.55}\\
\addlinespace[\interrowspace]
\multirow{2}{*}{\cplex{}-\CORLAT{}} & Training & 0.5 & 0.55 & 0.47 & 0.49 & 0.54 & \textbf{0.39} &  0.52 & 0.56 & 0.54 & 0.51 & 0.64 & \textbf{0.46}\\
~ & Test & 0.51 & 0.55 & 0.5 & 0.51 & 0.58 & \textbf{0.42} &  0.53 & 0.57 & 0.56 & 0.53 & 0.67 & \textbf{0.49}\\
\addlinespace[\interrowspace]
\multirow{2}{*}{\cplex{}-\REG{}} & Training & 0.15 & 0.18 & 0.15 & 0.16 & 0.17 & \textbf{0.12} &  \textbf{0.16} & 0.18 & 0.18 & 0.17 & 0.22 & \textbf{0.16}\\
~ & Test & 0.17 & 0.19 & 0.17 & 0.18 & 0.19 & \textbf{0.14} &  \textbf{0.17} & 0.19 & 0.19 & 0.19 & 0.24 & \textbf{0.17}\\
\addlinespace[\interrowspace]
\multirow{2}{*}{\cplex{}-\RCW{}} & Training & 0.09 & 0.11 & 0.09 & 0.1 & 0.08 & \textbf{0.06} &  0.1 & 0.12 & 0.11 & 0.11 & 0.12 & \textbf{0.09}\\
~ & Test & 0.09 & 0.12 & 0.09 & 0.11 & 0.08 & \textbf{0.06} &  0.1 & 0.12 & 0.12 & 0.12 & 0.12 & \textbf{0.09}\\
\addlinespace[\interrowspace]
\multirow{2}{*}{\cplex{}-\CORLATREG{}} & Training & 0.39 & 0.41 & 0.37 & 0.4 & 0.45 & \textbf{0.32} &  0.4 & 0.42 & 0.41 & 0.41 & 0.49 & \textbf{0.36}\\
~ & Test & 0.4 & 0.42 & 0.38 & 0.41 & 0.47 & \textbf{0.34} &  0.41 & 0.43 & 0.42 & 0.42 & 0.52 & \textbf{0.38}\\
\addlinespace[\interrowspace]
\multirow{2}{*}{\cplex{}-\CORLATREGRCW{}} & Training & 0.33 & 0.35 & 0.33 & 0.36 & 0.38 & \textbf{0.28} &  0.34 & 0.36 & 0.36 & 0.37 & 0.41 & \textbf{0.31}\\
~ & Test & 0.34 & 0.37 & 0.34 & 0.38 & 0.4 & \textbf{0.29} &  0.35 & 0.37 & 0.37 & 0.39 & 0.43 & \textbf{0.32}\\
\addlinespace[\interrowspace]
\multirow{2}{*}{\spear{}-\IBM{}} & Training & 0.57 & 0.64 & 0.5 & 0.48 & 0.43 & \textbf{0.34} &  0.57 & 0.64 & 0.51 & 0.48 & 0.45 & \textbf{0.36}\\
~ & Test & 0.57 & 11 & 0.53 & 0.52 & 0.57 & \textbf{0.42} &  0.58 & 11 & 0.54 & 0.52 & 0.57 & \textbf{0.44}\\
\addlinespace[\interrowspace]
\multirow{2}{*}{\spear{}-\SWV{}} & Training & 0.52 & 0.56 & 0.56 & 0.46 & 0.37 & \textbf{0.3} &  0.52 & 0.56 & 0.57 & 0.47 & 0.43 & \textbf{0.34}\\
~ & Test & 0.57 & 0.61 & 0.62 & 0.53 & 0.51 & \textbf{0.4} &  0.58 & 0.61 & 0.63 & 0.54 & 0.55 & \textbf{0.44}\\
\addlinespace[\interrowspace]
\multirow{2}{*}{\spear{}-\SWVIBM{}} & Training & 0.63 & 0.66 & 0.61 & 0.62 & 0.48 & \textbf{0.36} &  0.63 & 0.66 & 0.61 & 0.62 & 0.5 & \textbf{0.38}\\
~ & Test & 0.64 & 0.69 & 0.64 & 0.65 & 0.58 & \textbf{0.43} &  0.65 & 0.69 & 0.65 & 0.65 & 0.59 & \textbf{0.45}\\
\bottomrule
   \end{tabular}
   \caption{Root mean squared error (RMSE) obtained by various empirical performance models for predicting the runtime 
   based on combinations of paramater configurations and intance features. We trained on $10\,000$ randomly-sampled combinations of training configurations and instances, and report  performance for the four combinations of training/test instances and training/test configurations. \revision{Boldface indicates the model with the best performance.}
    \label{tab:combined_space_4quadrants_def}}
\end{table}

\begin{figure}[tbp]
    {\scriptsize
 \setlength{\tabcolsep}{2pt}
\centering
    \begin{tabular}{ccccc}
      ~ & $\Pi_{train}, \vTheta_{train}$ & $\Pi_{train}, \vTheta_{test}$ & $\Pi_{test}, \vTheta_{train}$ & $\Pi_{test}, \vTheta_{test}$\\
      \begin{sideways}~~~True runtime\end{sideways} &
\includegraphics[scale=0.18]{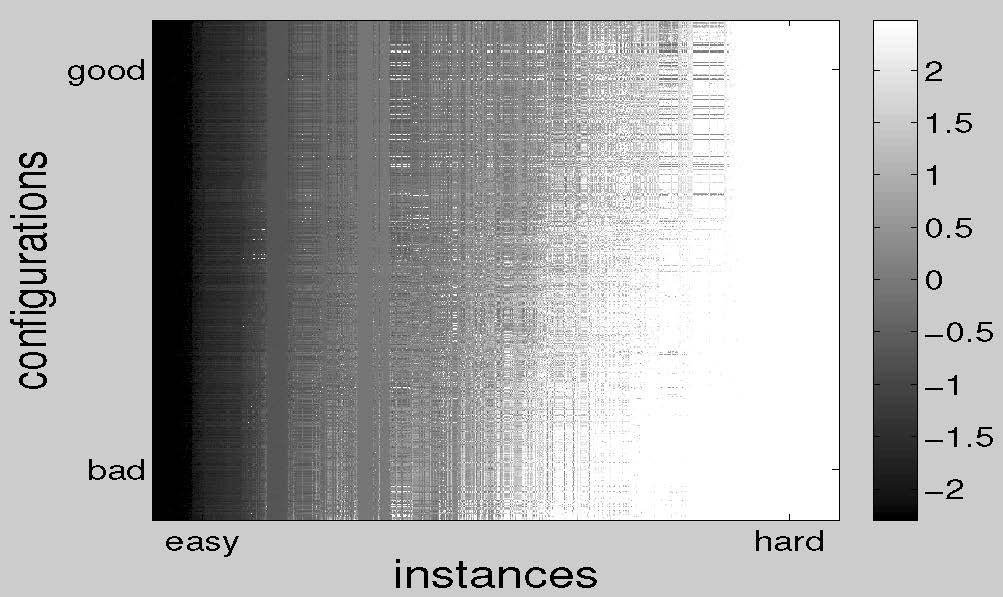} &
\includegraphics[scale=0.18]{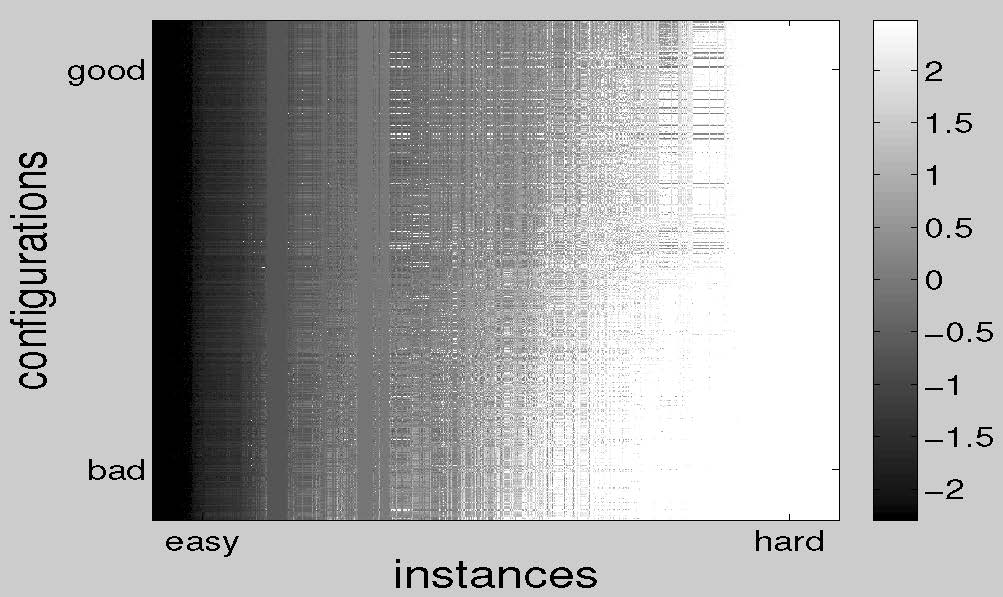} &
\includegraphics[scale=0.18]{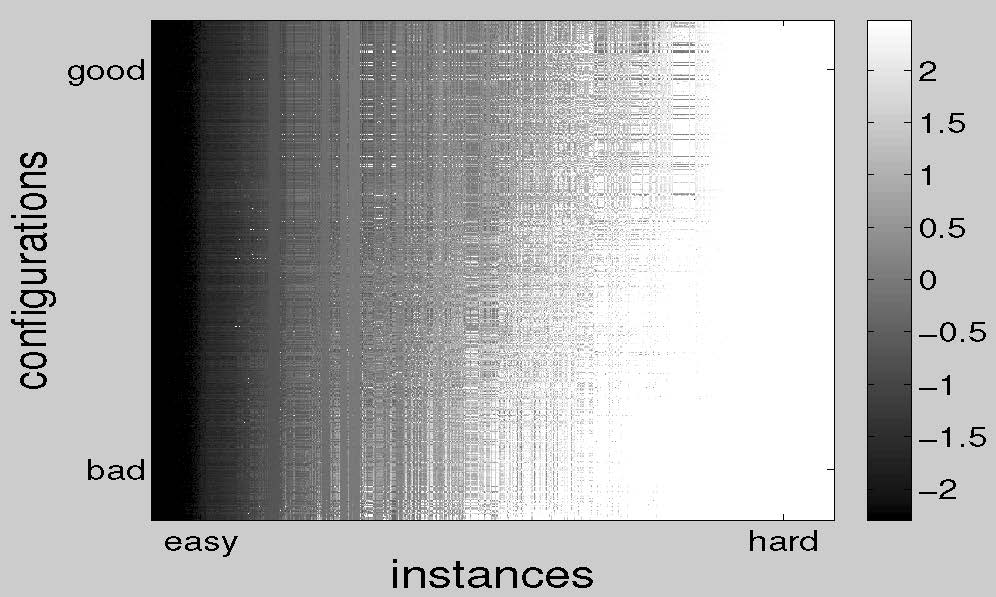} &
\includegraphics[scale=0.18]{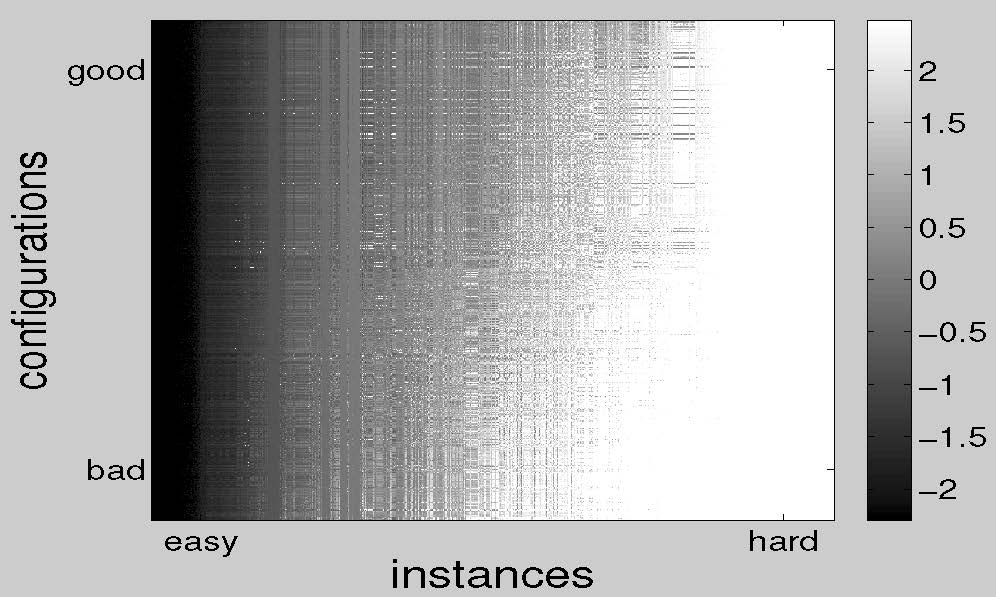}
\hlinespace{}
      \begin{sideways}~~~Random forest\end{sideways} &
\includegraphics[scale=0.18]{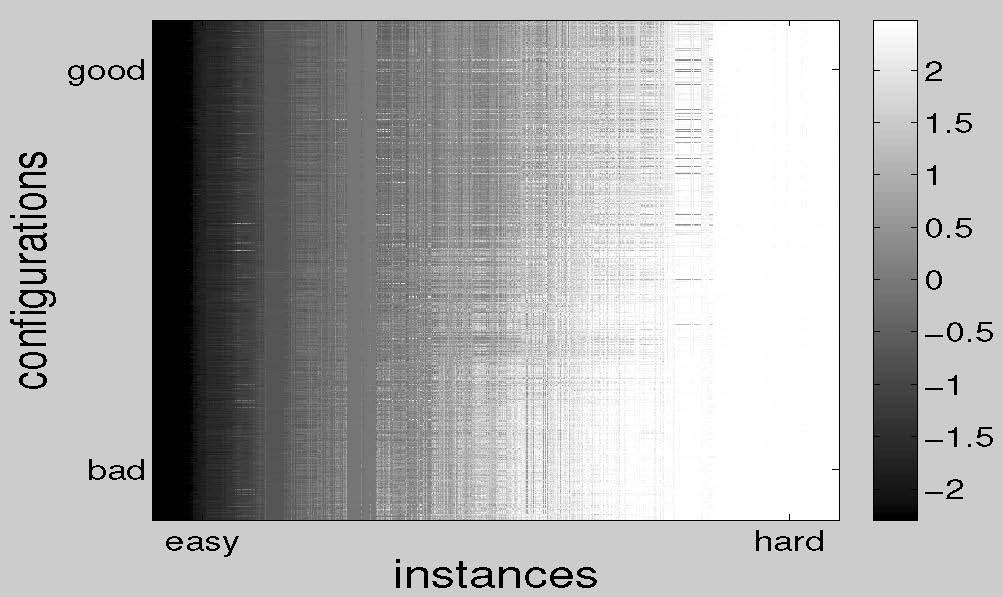} &
\includegraphics[scale=0.18]{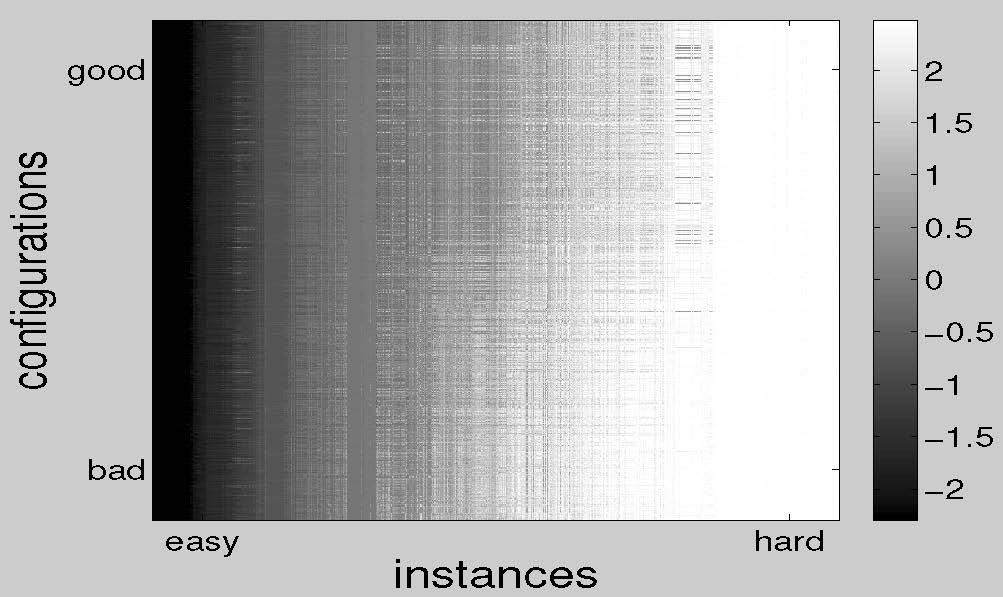} &
\includegraphics[scale=0.18]{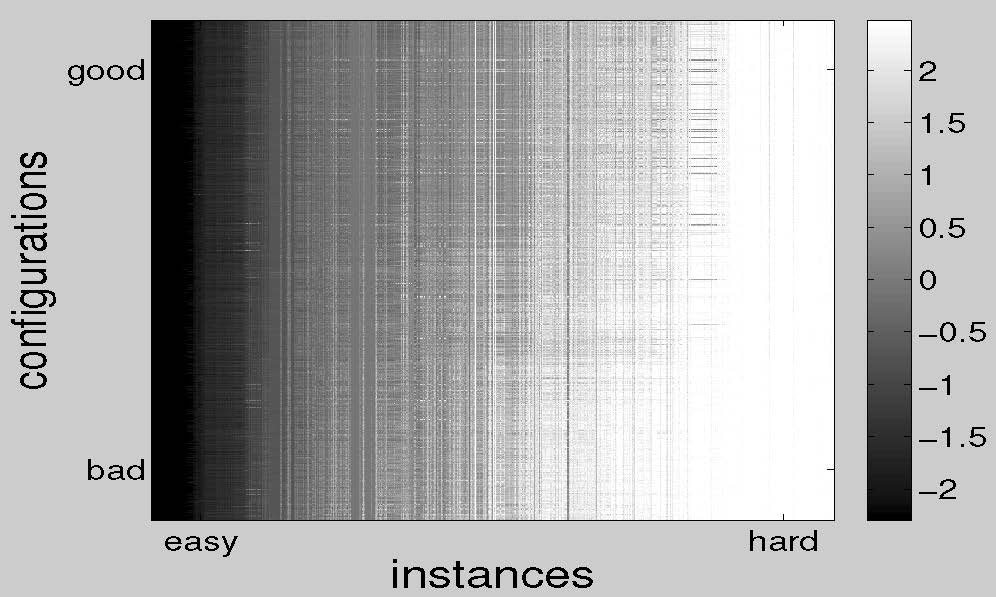} &
\includegraphics[scale=0.18]{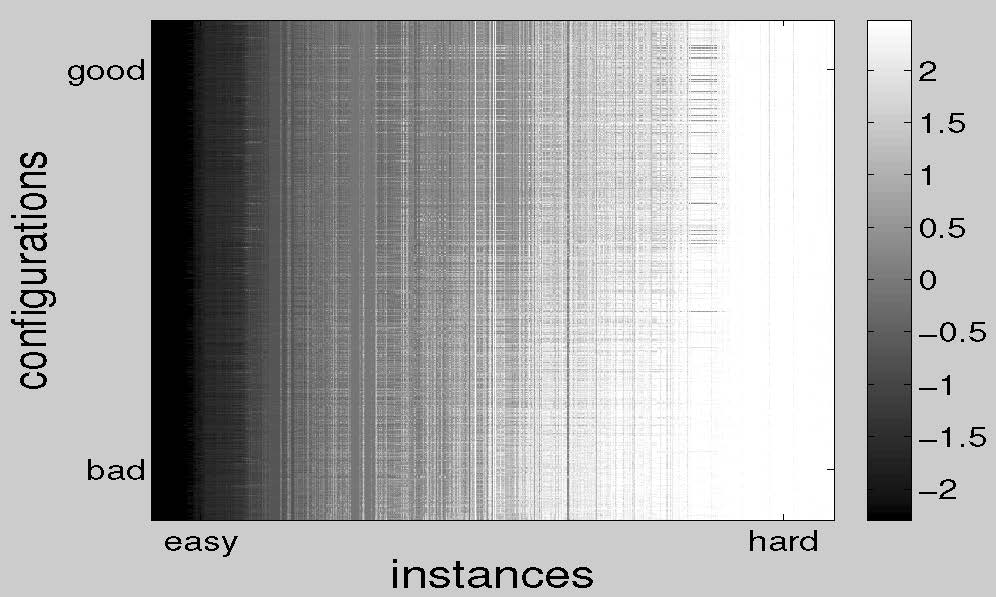}
\hlinespace{}
      \begin{sideways}Ridge regression (RR)\end{sideways} &
\includegraphics[scale=0.18]{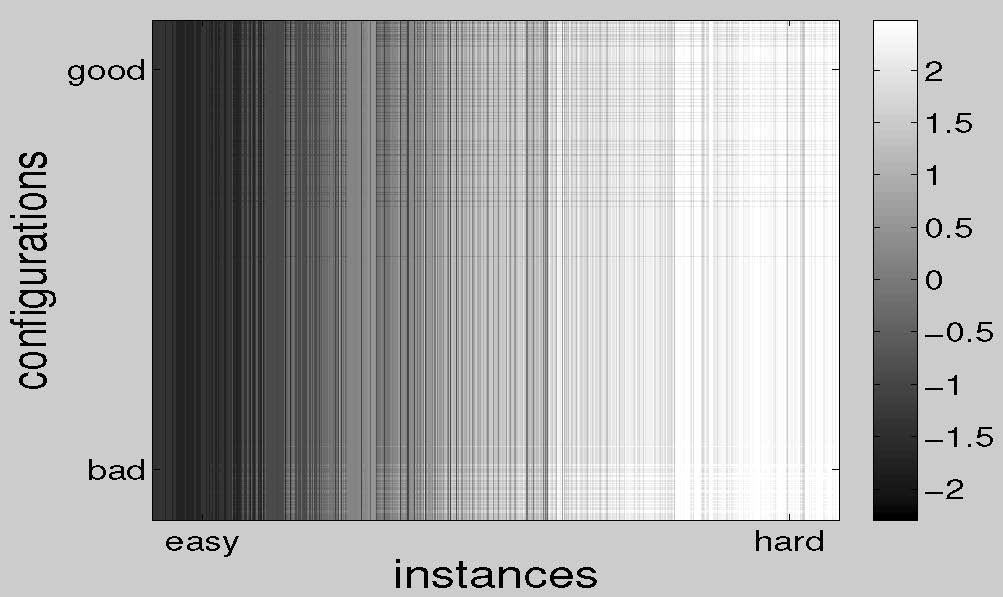} &
\includegraphics[scale=0.18]{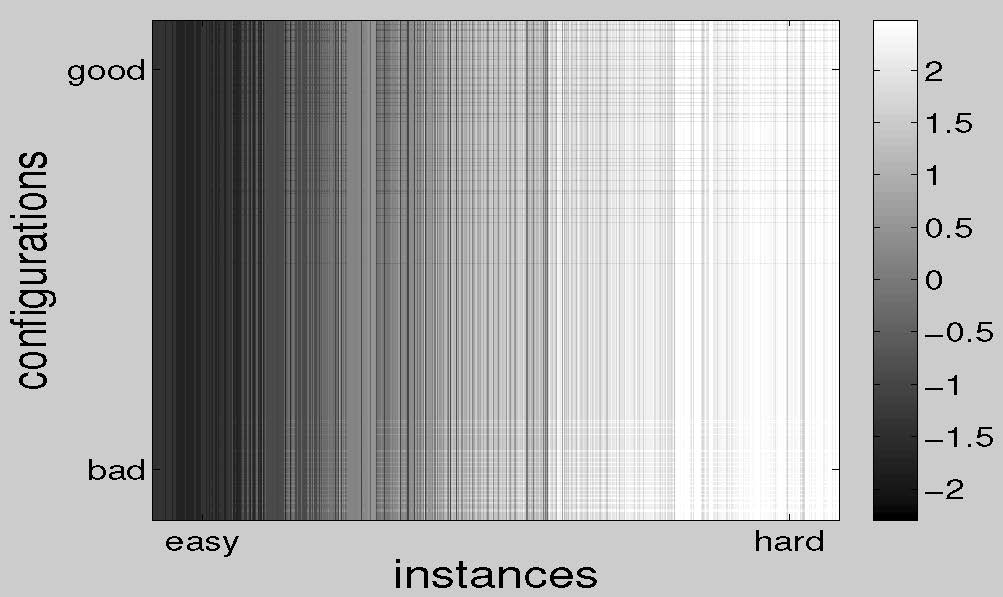} &
\includegraphics[scale=0.18]{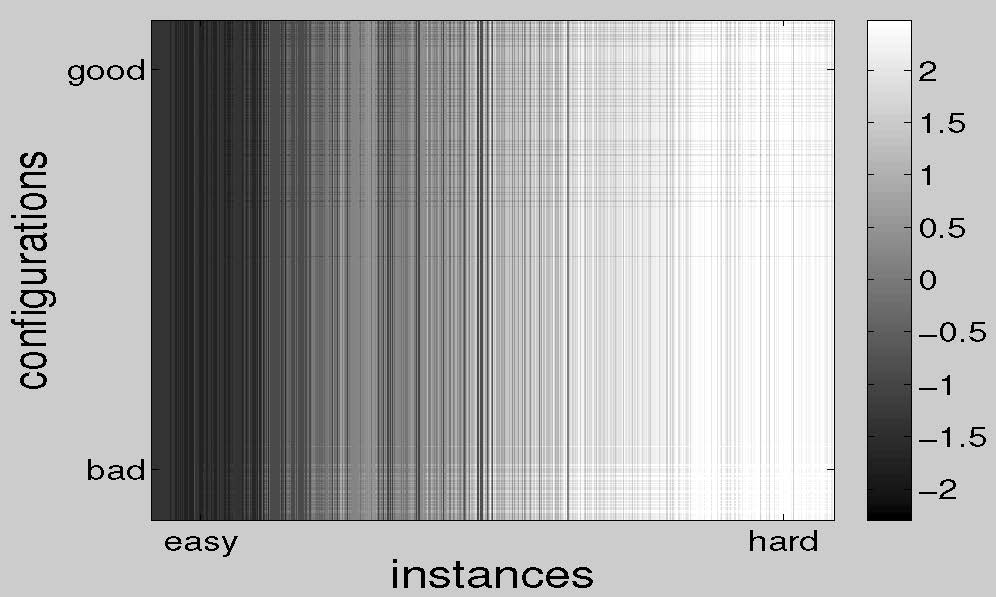} &
\includegraphics[scale=0.18]{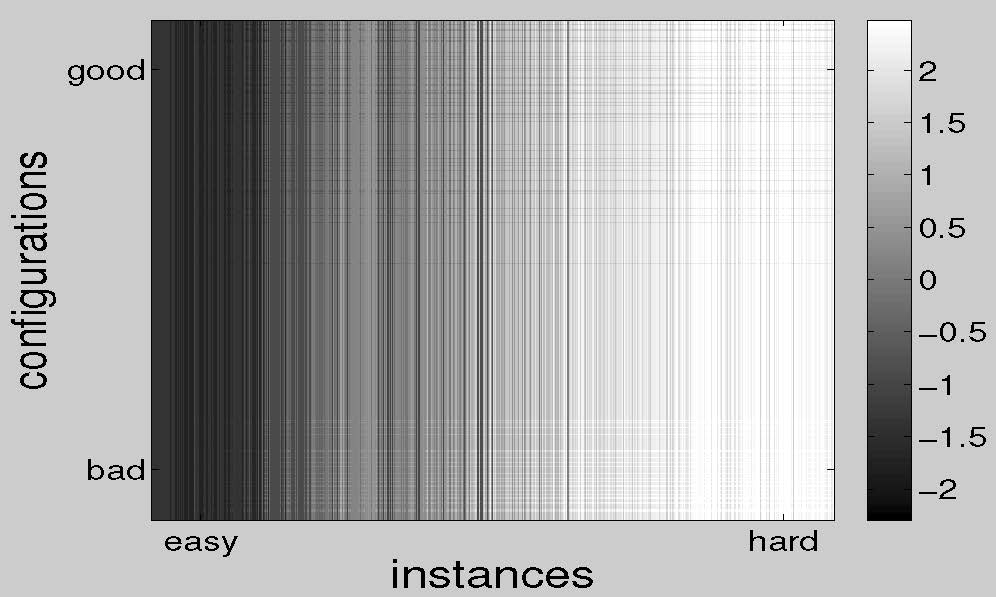}\\
   \end{tabular}
        \small
    \caption{True and predicted runtime matrices for dataset \spear{}-\SWVIBM{}, for all combinations of training/test instances ($\Pi_{train}$ and $\Pi_{test}$, respectively) and training test configurations ($\vTheta_{train}$ and $\vTheta_{test}$, respectively). 
		\revision{For example, the top left heatmap shows the true runtimes for the cross product of 500 training configurations of \spear{} and the 684 training instances of the \SWVIBM{} benchmark set.
Darker greyscale values represent faster runs, \ie{}, instances on the right side of each heatmap are hard (they take longer to solve), and configurations at the top of each heapmap are good (they solve instances faster).} (Plots for all models and benchmarks are given in Figures D.21--D.29, in the online appendix.) The predicted matrix of regression trees (not shown) is visually indistinguishable from that of random forests, and those of all other methods closely resemble that of ridge regression.\label{fig:matrix-pred-spear-SWVIBM_def}}
    }
\end{figure}

\begin{figure}[tbp]
    {\scriptsize
 \setlength{\tabcolsep}{2pt}
\centering
    \begin{tabular}{ccccc}
      ~ & $\Pi_{train}, \vTheta_{train}$ & $\Pi_{train}, \vTheta_{test}$ & $\Pi_{test}, \vTheta_{train}$ & $\Pi_{test}, \vTheta_{test}$\\
      \begin{sideways}~~~True runtime\end{sideways} &
\includegraphics[scale=0.18]{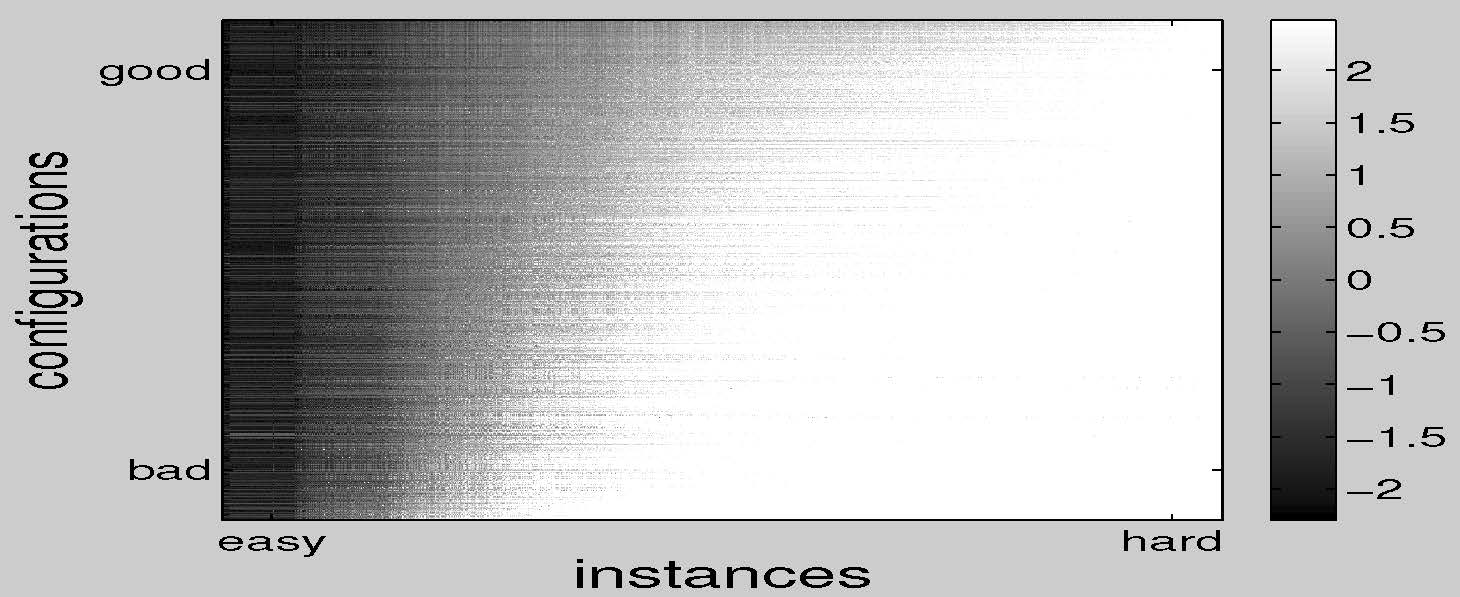} &
\includegraphics[scale=0.18]{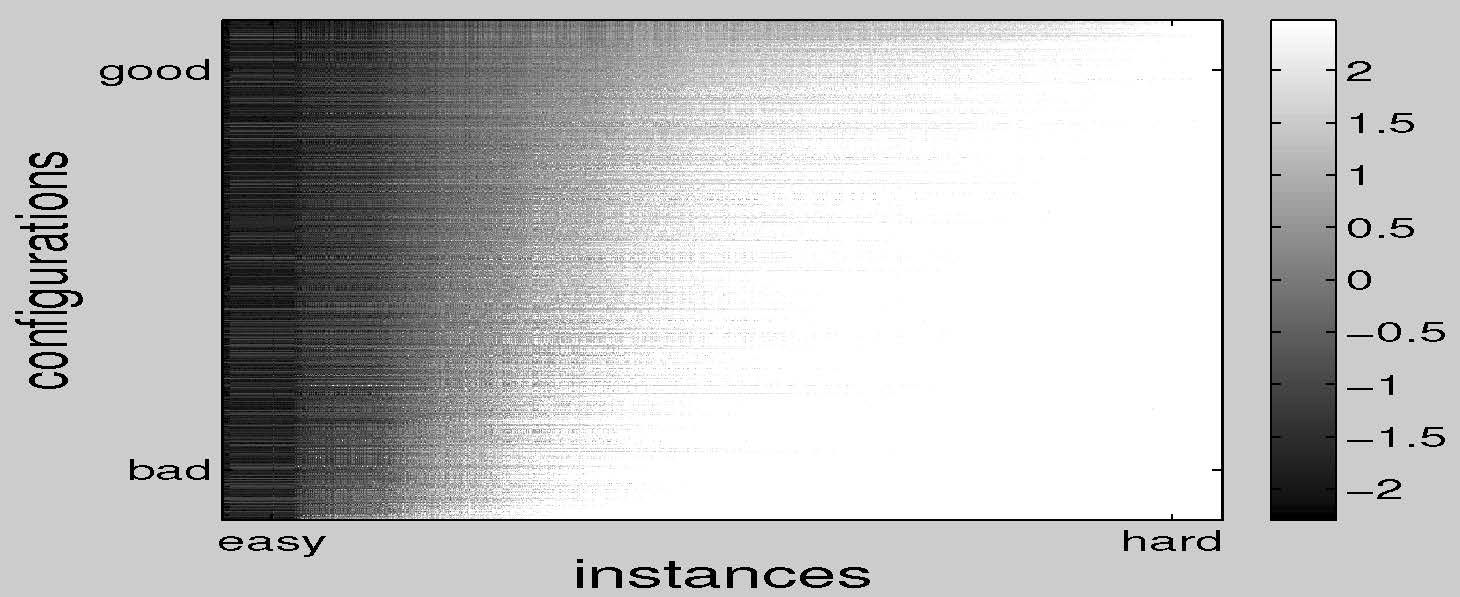} &
\includegraphics[scale=0.18]{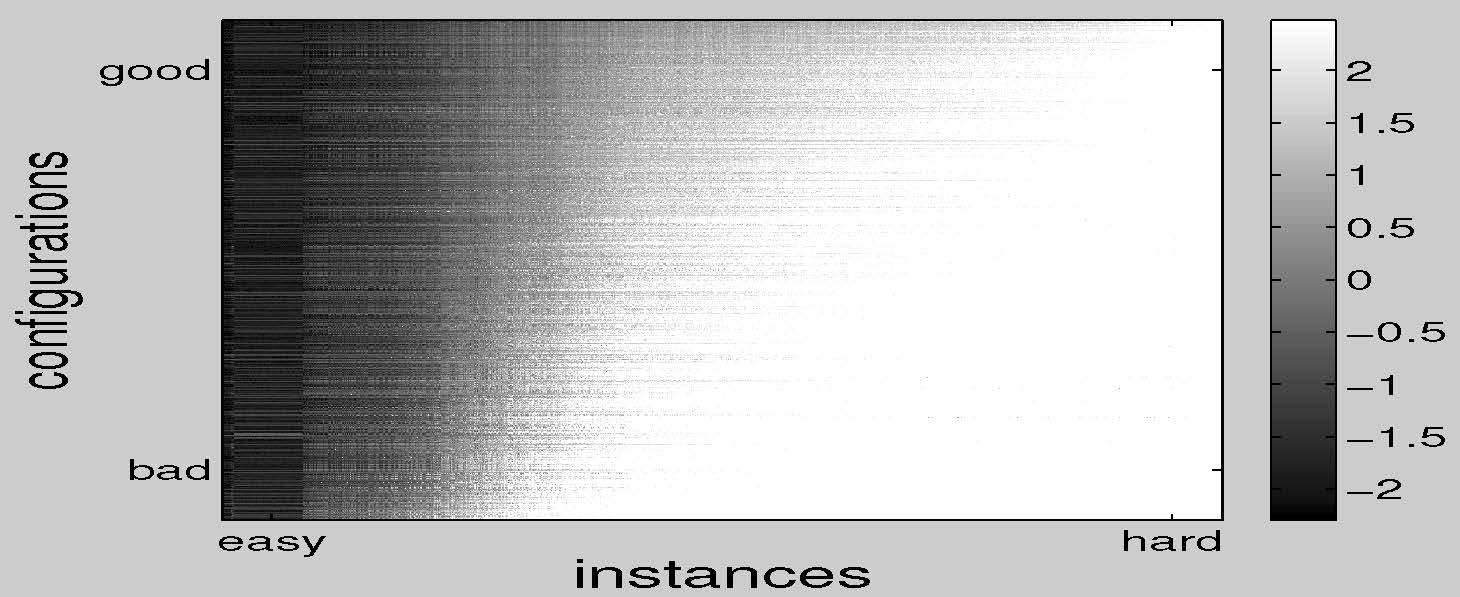} &
\includegraphics[scale=0.18]{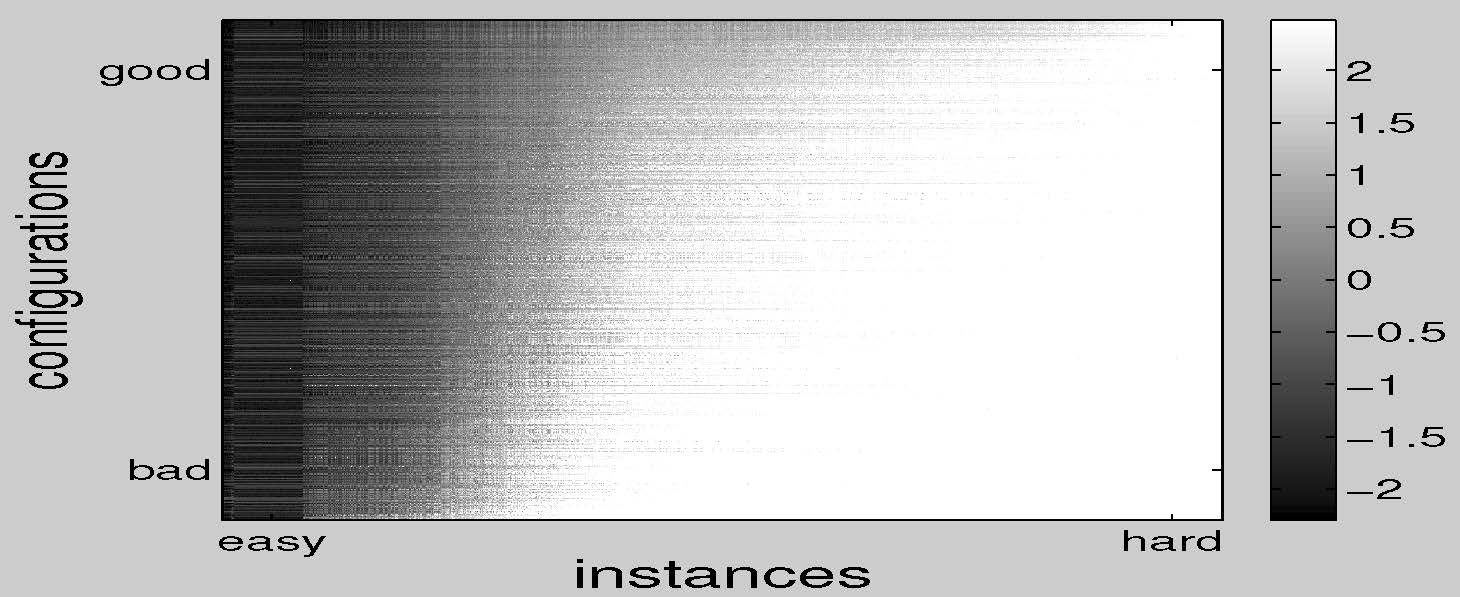}
\hlinespace{}
      \begin{sideways}~~~Random forest\end{sideways} &
\includegraphics[scale=0.18]{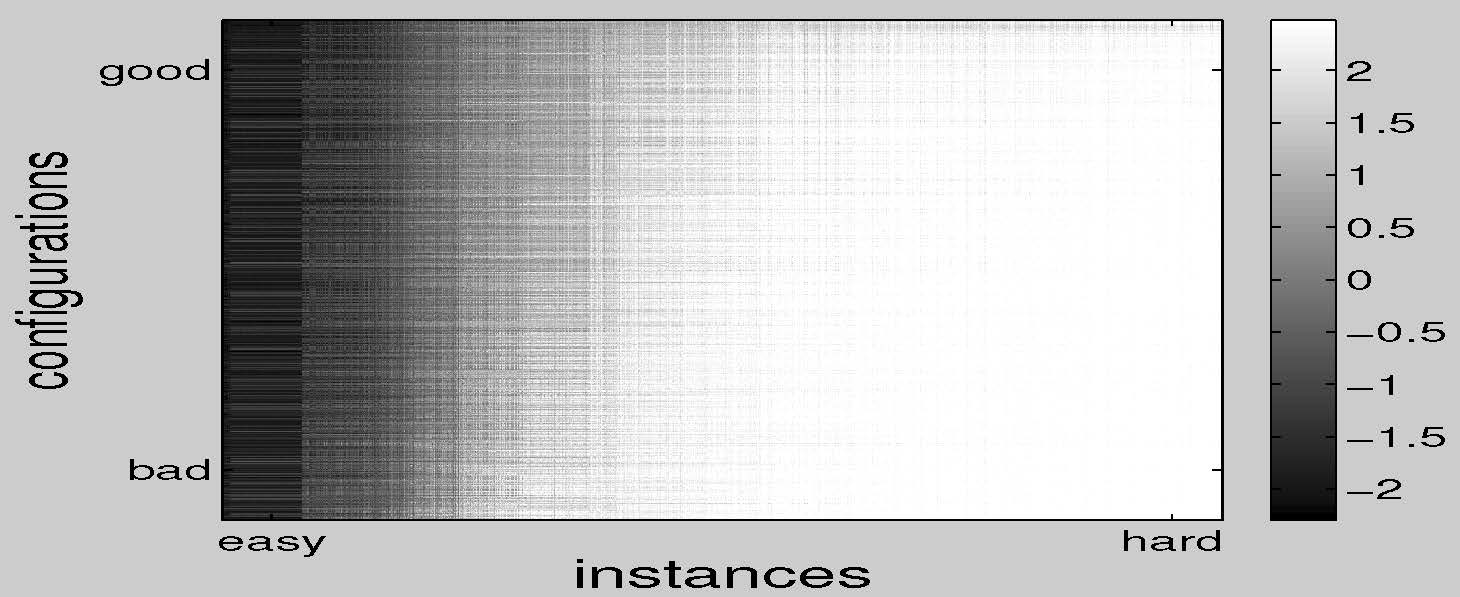} &
\includegraphics[scale=0.18]{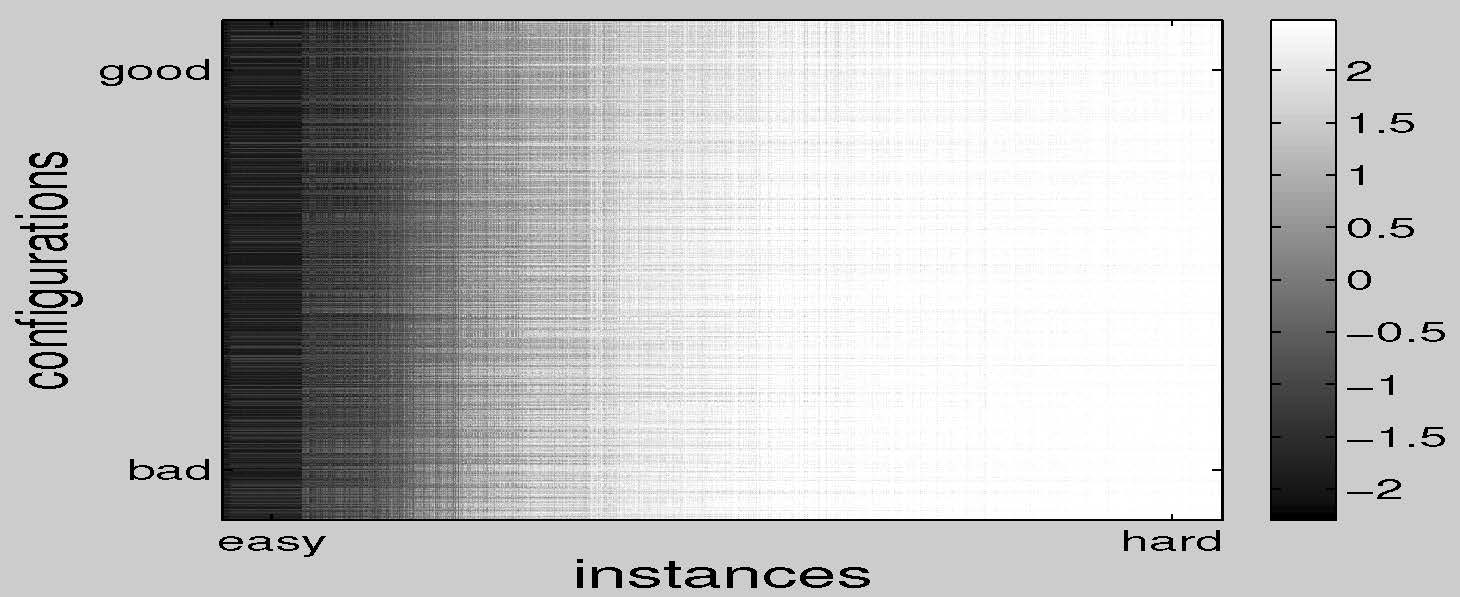} &
\includegraphics[scale=0.18]{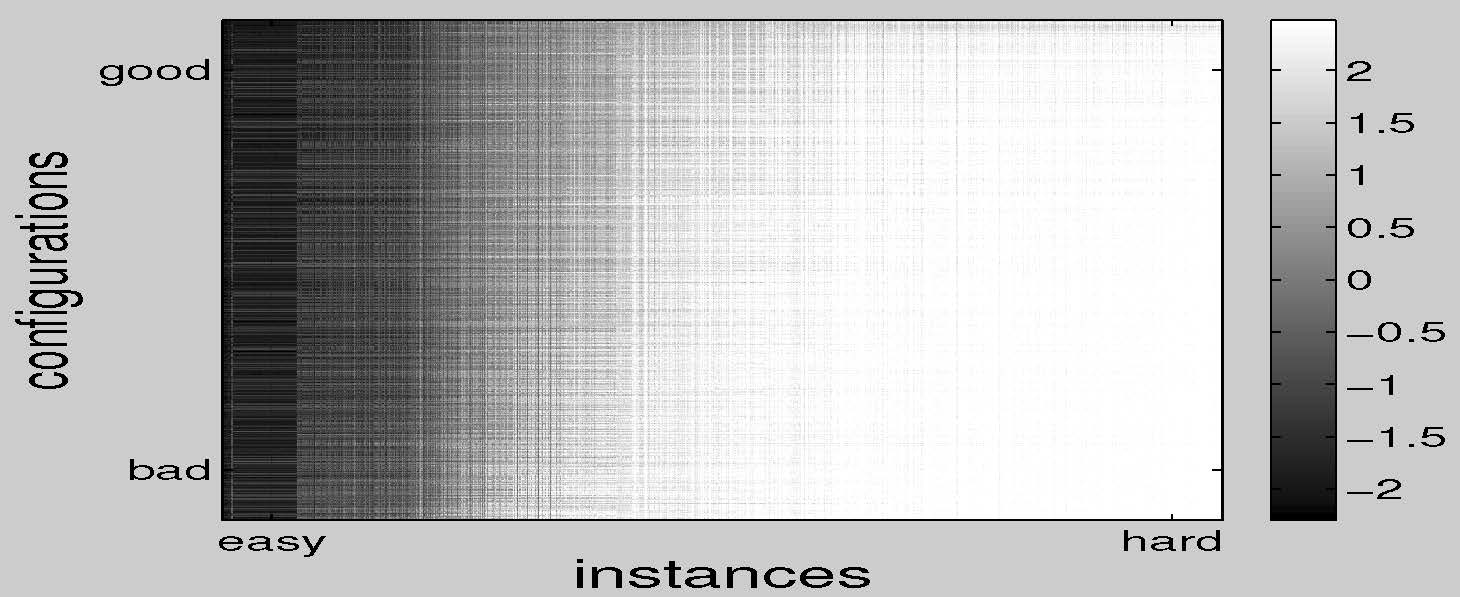} &
\includegraphics[scale=0.18]{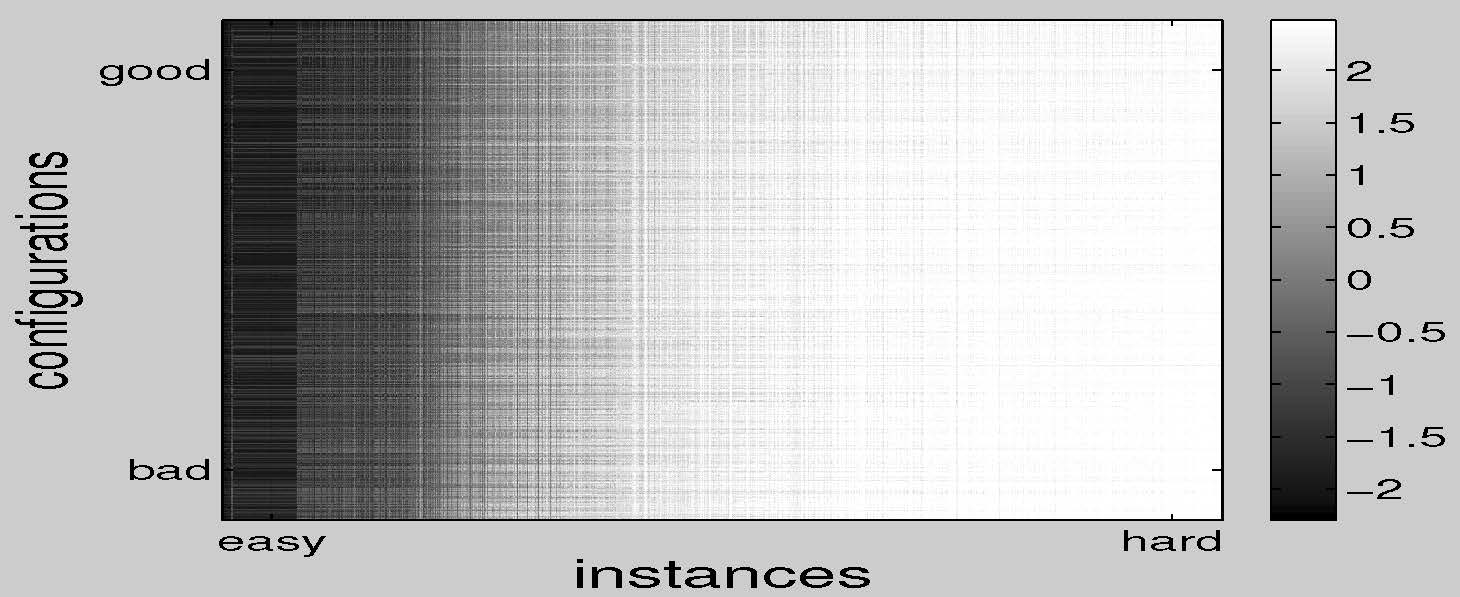}
\hlinespace{}
      \begin{sideways}Ridge regression (RR)\end{sideways} &
\includegraphics[scale=0.18]{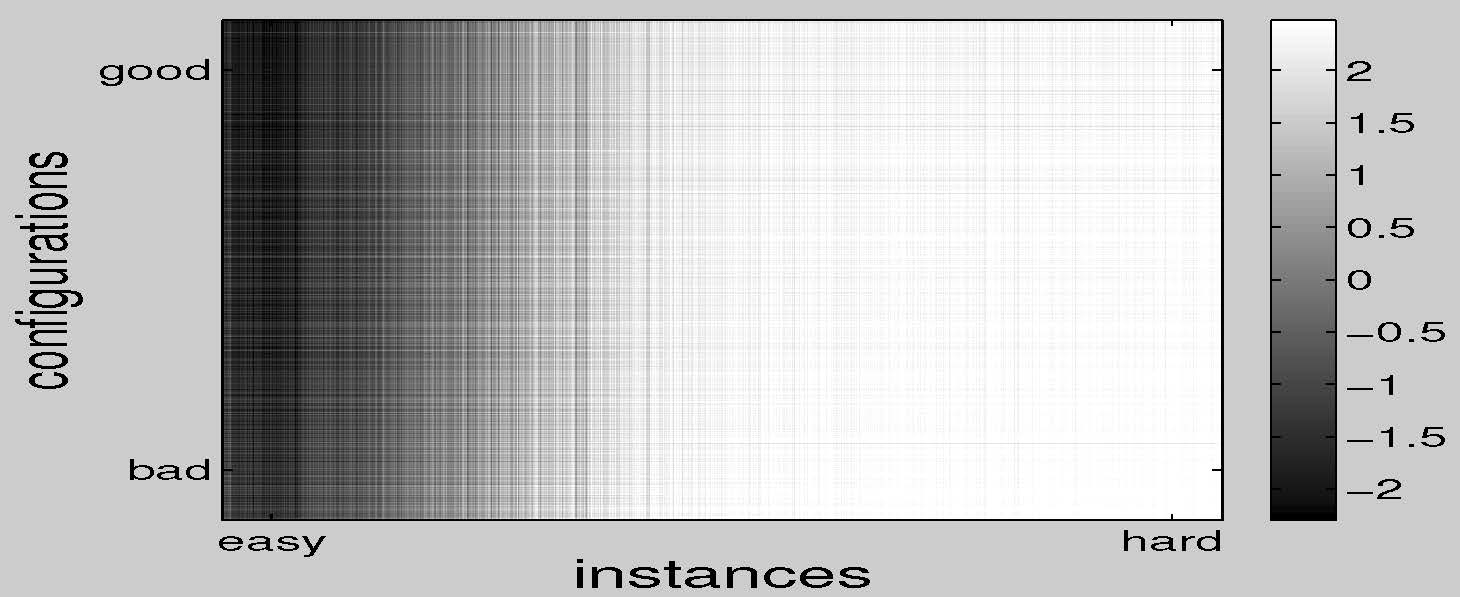} &
\includegraphics[scale=0.18]{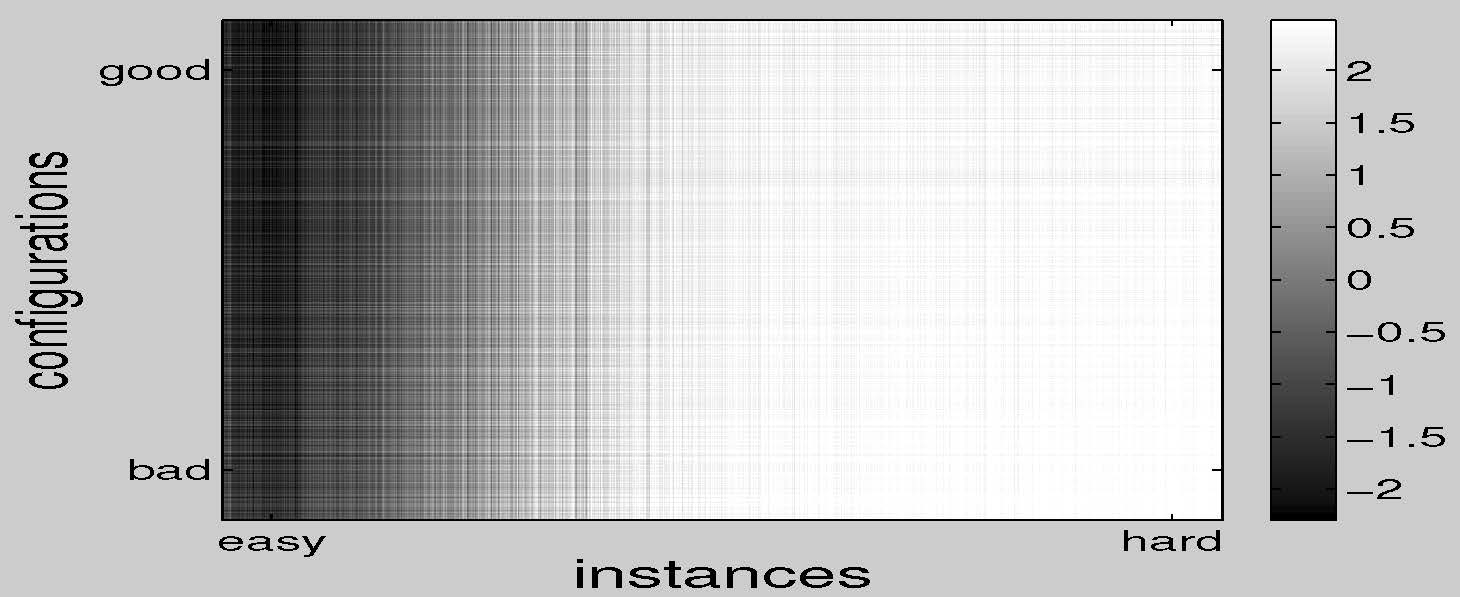} &
\includegraphics[scale=0.18]{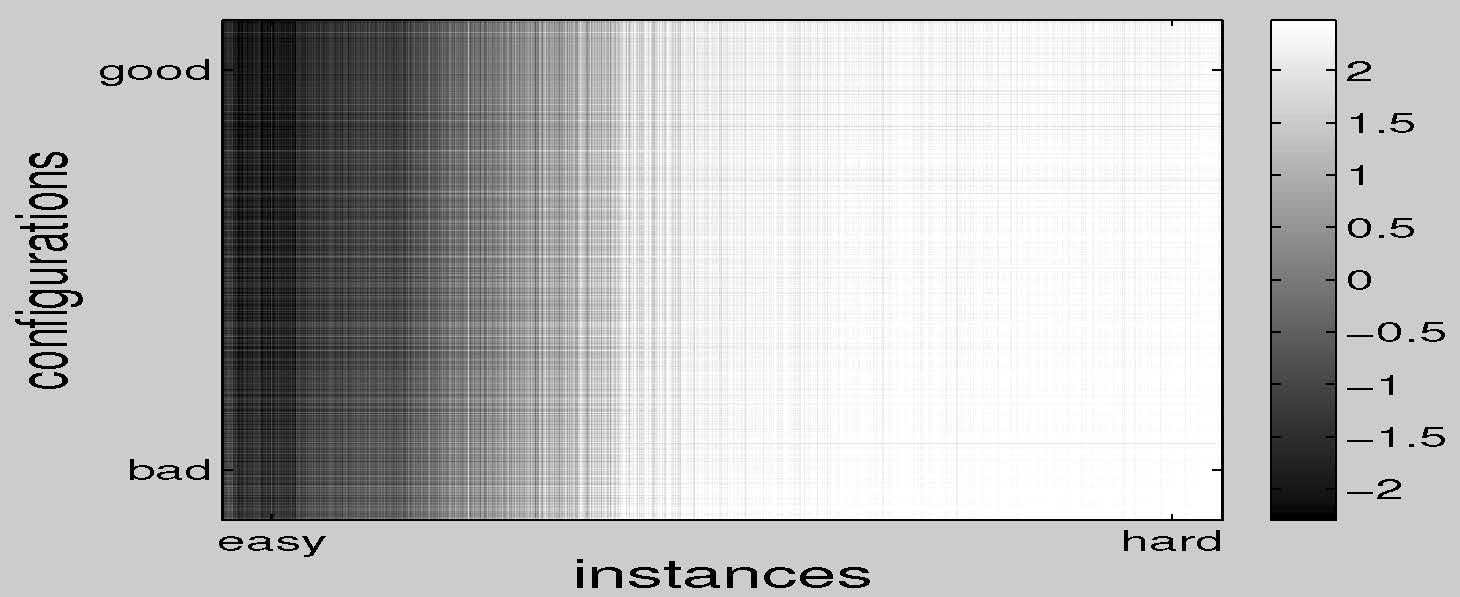} &
\includegraphics[scale=0.18]{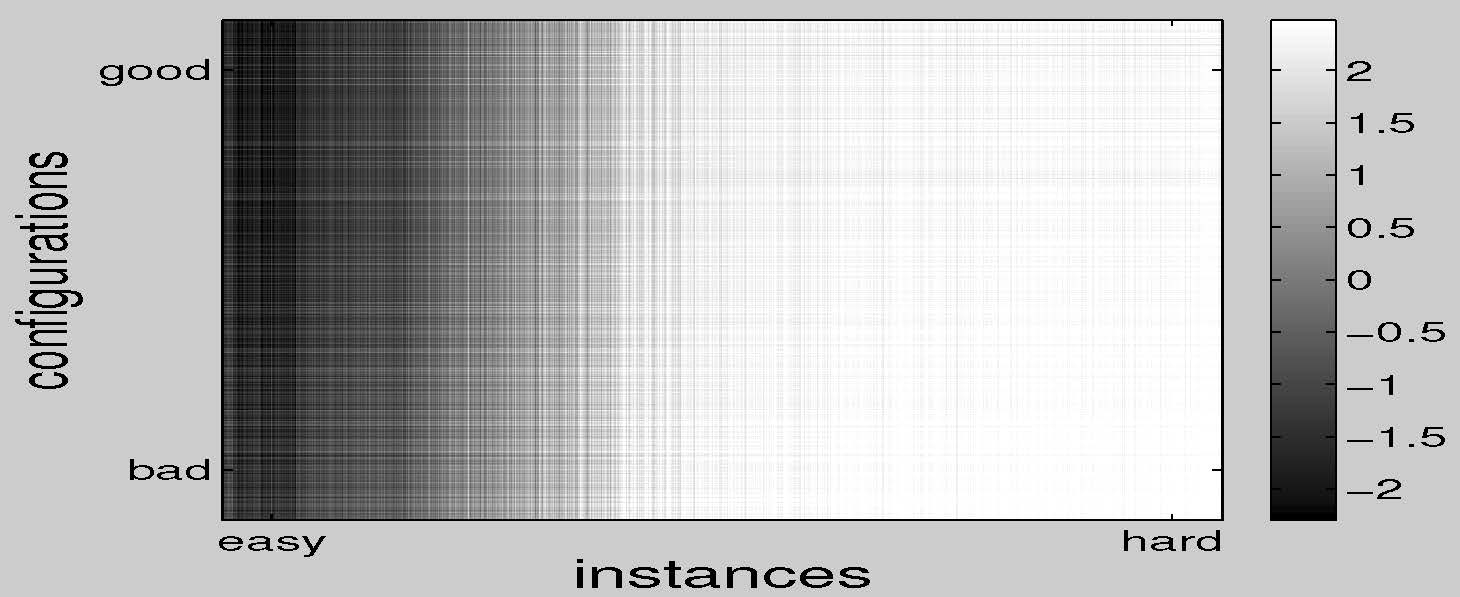}\\
   \end{tabular}
        \small
    \caption{Same type of data as in Figure \ref{fig:matrix-pred-spear-SWVIBM_def} but for dataset \cplex{}-\CORLAT{}. \label{fig:matrix-pred-cplex-CORLAT_def}}
}
\end{figure}

\begin{figure}[tbp]
    {\scriptsize
 \setlength{\tabcolsep}{2pt}
\centering
    \begin{tabular}{ccccc}
      ~ & RF with 100 points & RF with 1\,000 points & RF with all points & True runtimes\\
      \begin{sideways}\spear{}-\SWVIBM{}\end{sideways} &
\includegraphics[scale=0.18]{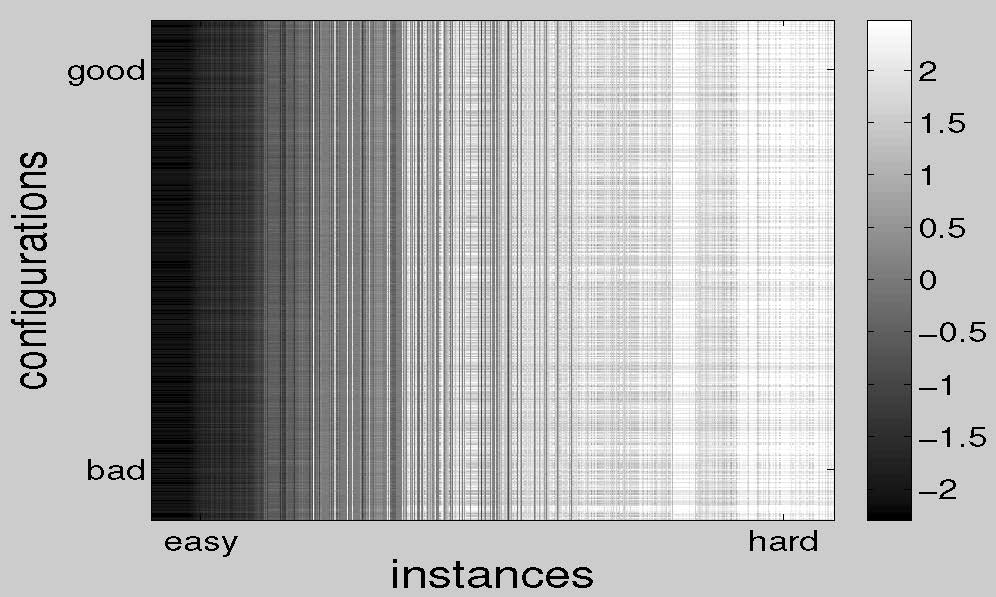} &
\includegraphics[scale=0.18]{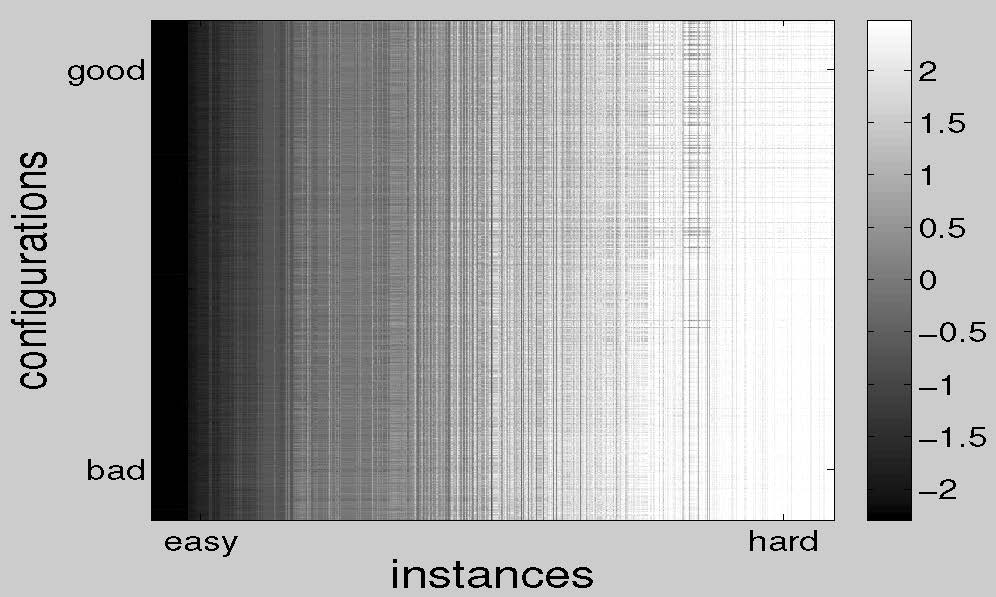} &
\includegraphics[scale=0.18]{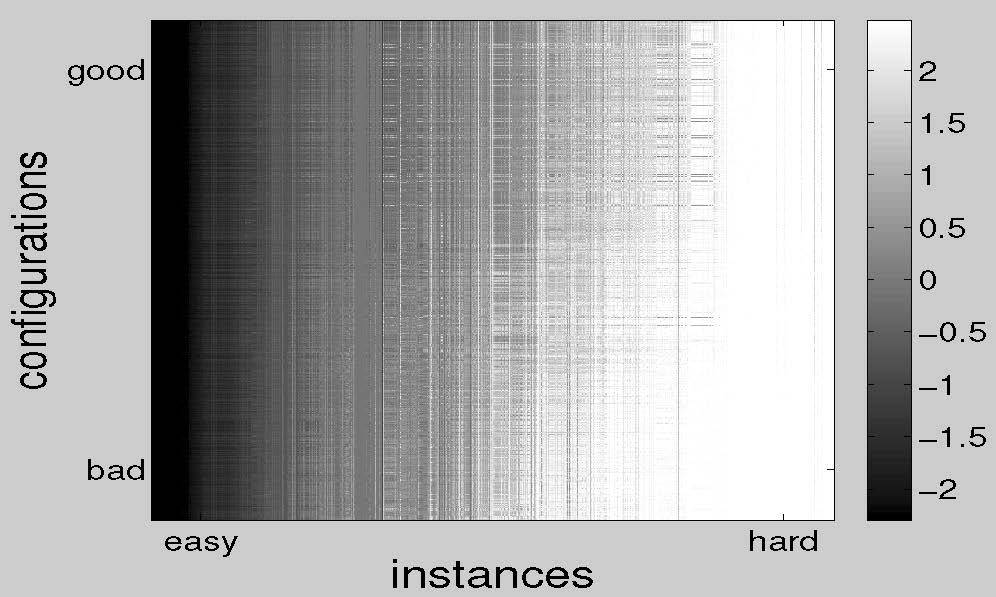} &
\includegraphics[scale=0.18]{figures__default_matrix__2d__redone__SPEAR-ibm-swv-al-truematrix-testC_testI}
\hlinespace{}
      \begin{sideways}\cplex{}-\CORLAT{}\end{sideways} &
\includegraphics[scale=0.18]{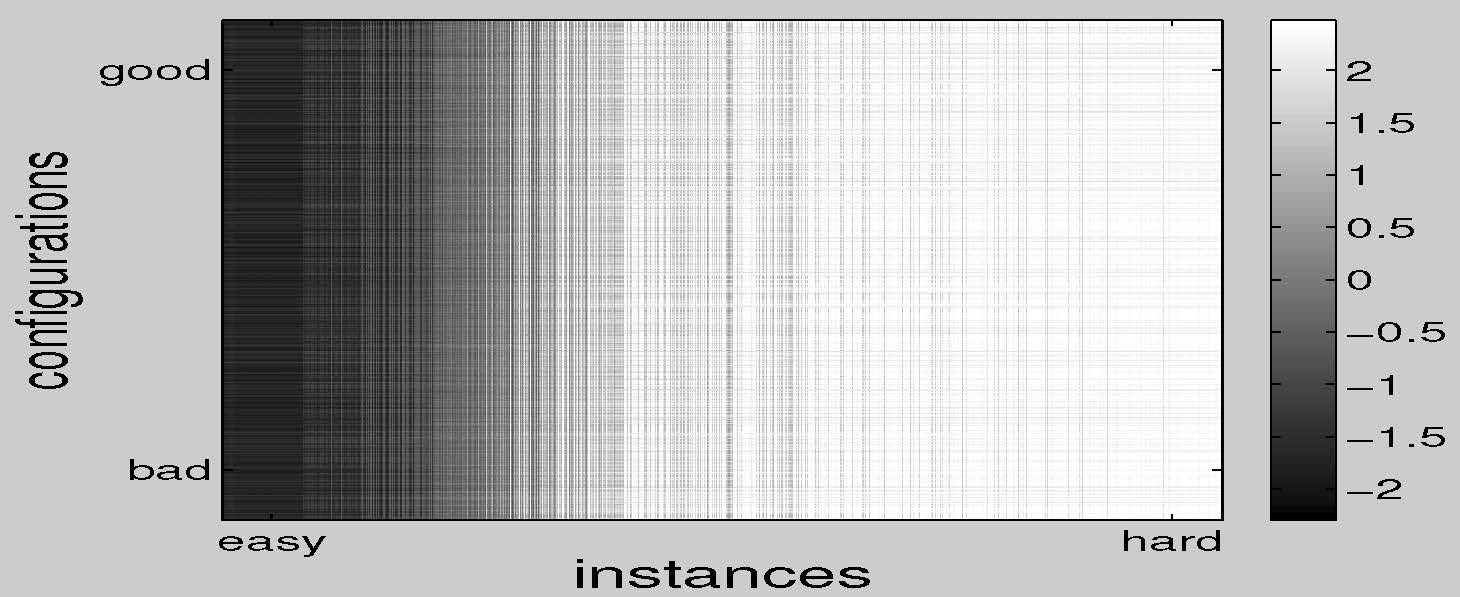} &
\includegraphics[scale=0.18]{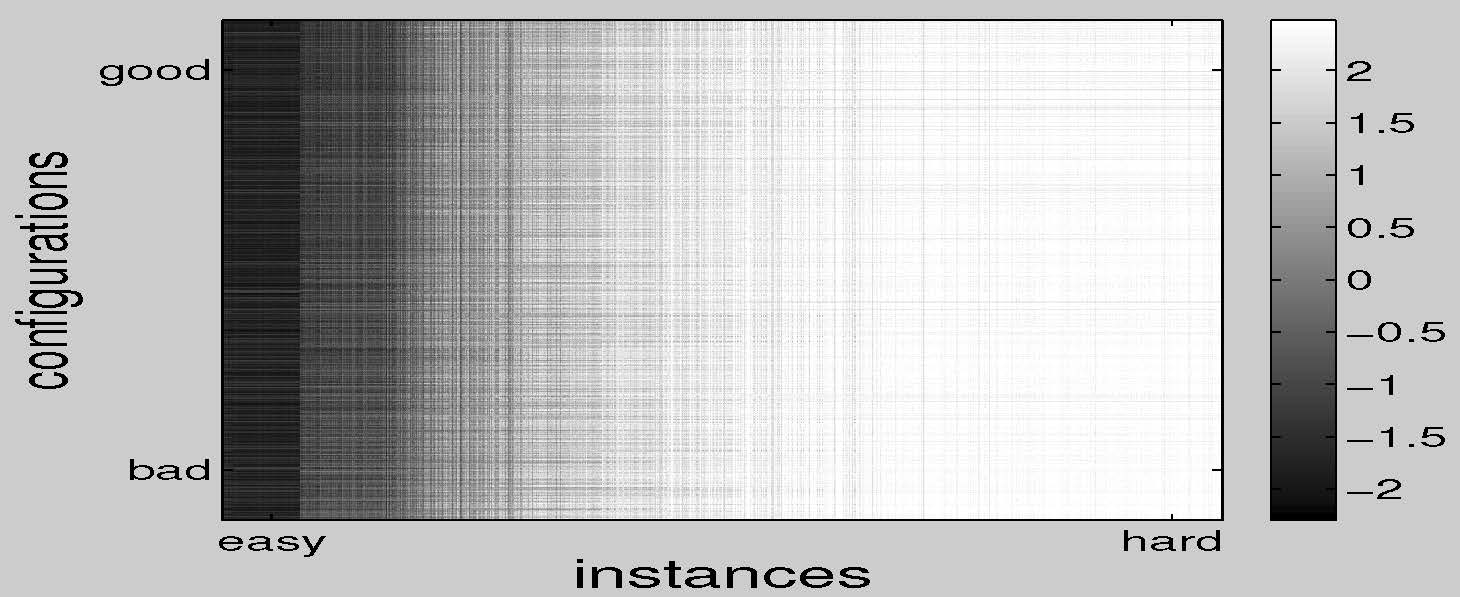} &
\includegraphics[scale=0.18]{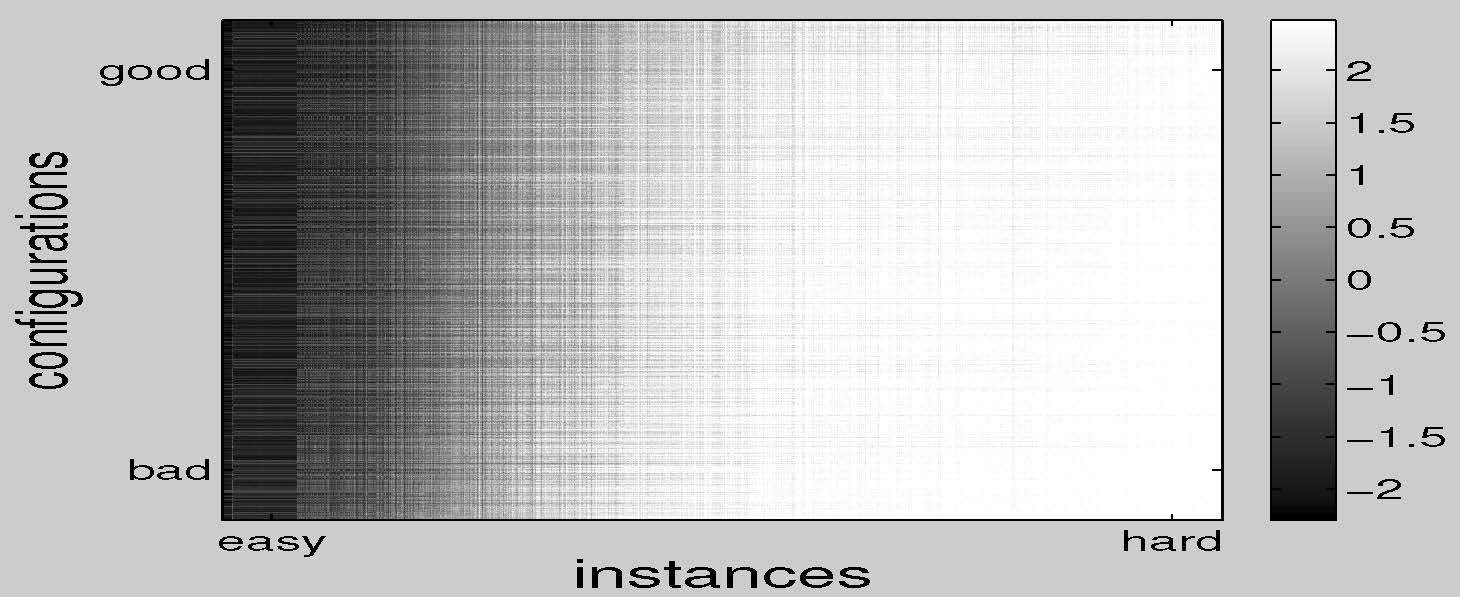} &
\includegraphics[scale=0.18]{figures__default_matrix__2d__redone__CPLEX12-cat-CORLAT-truematrix-testC_testI}
   \end{tabular}
        \small
    \caption{Predicted runtime matrices with different number of training data points, compared to true runtime matrix. ``All points'' means the entire crossproduct of training instances and training configurations (342\,500 data points for \spear{}-\SWVIBM{} and 500\,000 for \cplex{}-\CORLAT{}). (Plots for all benchmarks are given in Figure D.30 in the online appendix.) \label{fig:combined_space_matrix_varyingN_def}}
  }
\end{figure}

Now, we study all four combinations of predictions on training/test instances
and training/test configurations. \revision{(See the beginning of Section \ref{sec:rms_better} for a description of each scenario.)}
Our results are summarized in Table \ref{tab:combined_space_4quadrants_def} and Figures \ref{fig:matrix-pred-spear-SWVIBM_def} and \ref{fig:matrix-pred-cplex-CORLAT_def}. For the figures, we sorted instances by average hardness (across configurations), and parameter configurations by average performance (across instances), generating a heatmap with instances on the $x$-axis, configurations on the $y$-axis, and greyscale values representing algorithm runtime for given configuration/instance combinations.
We compare heatmaps representing true runtimes against those based on the predictions obtained from each of our models.
Here, we only show results for the two \revision{scenarios} where the performance advantage of random forests (the overall best method based on our results
reported so far) over the other methods was highest (the heterogeneous data set \spear{}-\SWVIBM{}) and lowest (the homogeneous data set \cplex{}-\CORLAT{}); heatmaps for all data sets and model types are given in Figures D.21--D.29 in the online appendix.

Figure \ref{fig:matrix-pred-spear-SWVIBM_def} shows the results for benchmark \spear{}-\SWVIBM{}.
It features one column for each of the four combinations of training/test instances and training/test configurations, allowing us to visually assess how well the respective generalization works for each of the models.
We note that in this case, the true heatmaps are almost indistinguishable from those predicted by random forests (and regression trees). Even for the most challenging case of unseen problem instances and parameter configurations, the tree-based methods captured the non-trivial interaction pattern between instances and parameter configurations.
On the other hand, the non-tree-based methods (ridge regression variants, neural networks, and projected processes) only captured instance hardness, failing to distinguish good from bad configurations even in the simplest case of predictions for training instances and training configurations.

Figure \ref{fig:matrix-pred-cplex-CORLAT_def} shows the results for benchmark \cplex{}-\CORLAT{}.
For the simplest case of predictions on training instances and configurations, the tree-based methods yielded predictions close to the true runtimes, capturing both instance hardness and performance of parameter configurations. In contrast, even in this simple case, the other methods only captured instance hardness, predicting all configurations to be roughly equal in performance.
Random forests generalized better to test instances than to test configurations (compare the 3rd and 2nd columns of Figure \ref{fig:matrix-pred-cplex-CORLAT_def}); this trend is also evident quantitatively in Table \ref{tab:combined_space_4quadrants_def} for all \cplex{} benchmarks.
Regression tree predictions were visually indistinguishable from those of random forests; this strong qualitative performance is remarkable, considering that quantitatively they performed worse than other methods in terms of measures such as RMSE (see the results for \cplex{}-\CORLAT{} in Table \ref{tab:combined_space_4quadrants_def}).

Finally, we investigated once more how predictive quality depends on the quantity of training data, focusing on random forests (Figure \ref{fig:combined_space_matrix_varyingN_def}).
For \spear{}-\SWVIBM{}, 100 training data points sufficed to obtain random forest models that captured the most salient features (\eg{}, they correctly determined the simplicity of the
roughly 20\% easiest instances);
more training data points gradually improved qualitative predictions, especially in distinguishing good from bad configurations.
Likewise, for \cplex{}-\CORLAT{}, salient features (\eg{}, the simplicity of the roughly 25\% easiest instances) could be detected based on 100 training data points, and more training data improved qualitative predictions to capture some of the differences between good and bad configurations.
Overall, increases in the training set size yielded diminishing returns, and even predictions based on the entire cross-product of training instances and parameter configurations \revision{(i.e.{}, between 151\,000 and 500\,000 runs)} were not much different from those based on a subset of 10\,000 samples \revision{(representing 2\% to 6.6\% of the entire training data)}.

\section{Improved Handling of Censored Runtimes in Random Forests} \label{sec:censoring}

Most past work on predicting algorithm runtime has treated algorithm runs that were terminated prematurely at a so-called \emph{captime} $\kappa$ as if they finished at time $\kappa$. Thus, we adopted the same practice in the model comparisons we have described so far (using captimes of 3\,000 seconds for the runs in Section \ref{sec:ehms_better} and 300 seconds for the runs in Sections \ref{sec:rms_better} and \ref{sec:combination_ehm_rsm}). Now, we revisit this issue for 
random forests.

Formally, terminating an algorithm run after a captime (or \emph{censoring threshold}) $\kappa$ yields a \emph{right-censored} data point: we learn that $\kappa$ is a lower bound on the actual time the algorithm run required.
Let $y_i$ denote the \emph{actual} (unknown) runtime of algorithm run $i$. Under partial right censoring, our training data is $(\mathbf{x}_i, z_i, c_i)_{i=1}^n$, where $\mathbf{x}_i$ is our usual input vector (a vector of instance features, parameter values, or both combined), $z_i \in \mathds{R}$ is a (possibly censored) runtime observation, and $c_i \in \{0,1\}$ is a censoring indicator such that $z_i = y_i$ if $c_i=0$ and $z_i < y_i$ if $c_i=1$.

Observe that the typical, simplistic strategy for dealing with censored data produces biased models; intuitively, treating slow runs as though they were faster than they really were biases our training data downwards, and hence likewise biases predictions.
Statisticians, mostly in the literature on so-called ``survival analysis'' from actuarial science, have developed strategies for building unbiased regression models based on censored data~\cite{nelson2003applied}. (Actuaries need to predict when people will die, given mortality data and the ages of people still living.)
Gagliolo et al. \cite{GagSch06AIMATH,Gagliolo2010Survival} were the first to use techniques from this literature for runtime prediction. Specifically, they used a method for handling censored data in parameterized probabilistic models and employed the resulting models to construct dynamic algorithm portfolios.
In the survival analysis literature, \emcite{Josef79} described an iterative procedure for handling censored data points in linear regression models. We employed this technique to improve the runtime predictions made by our portfolio-based algorithm selection method \satzilla{}~\cite{cp-satzilla07}.
While to the best of our knowledge, no other methods from this literature have been applied to algorithm runtime prediction, there exist several candidates for consideration in future work.
In Gaussian processes, one could use approximations to handle the non-Gaussian observation likelihoods resulting from censorship; for example, \emcite{Ert07} described a Laplace approximation for handling right-censored data.
Random forests~(RFs) have previously been adapted to handle censored data~\cite{Seg88,HotEtAl04}, but the classical methods yield non-parametric Kaplan--Meier estimators that are undefined beyond the largest uncensored data point.
Here, we describe a simple improvement of the method by \emcite{Josef79} for use with random forests that we developed
in the context of handling censored data in model-based algorithm configuration\red{~\cite{HutHooLey11-censoredBO,HutHooLey13:censoredBOarXiv}}.

We denote the probability density function (PDF) and cumulative distribution function (CDF) of a Normal distribution by $\varphi$ and $\Phi$, respectively.
Let $\mathbf{x}_i$ be an input with censored runtime $\kappa_i$.
Given a Gaussian predictive distribution $\gauss(\mu_i, \sigma^{2}_i)$,
the truncated Gaussian distribution $\gauss(\mu_i, \sigma^{2}_i)_{\ge \kappa_i}$
is defined by the PDF
\begin{eqnarray}
\nonumber
p(y) =
\left\{
\begin{array}{ll}
0 & y<\kappa_i\\
\frac{1}{\sigma_i} \varphi(\frac{x-\mu_i}{\sigma_i}) / (1-\Phi(\frac{\mu_i-\kappa_i}{\sigma_i}))
& y \ge \kappa_i.\\
\end{array}
\right.
\end{eqnarray}

The method of Schmee and Hahn~\cite{Josef79} is an Expectation Maximization (EM) algorithm.
Applied to an RF model as its base model, that algorithm would first fit an initial RF using only uncensored data and then iterate between the following steps:
\begin{enumerate}
        \item[(E)] For each tree $T$ in the RF and each $i$ s.t.\ $c_i=1$: $\hat{y}_i^{(T)} \leftarrow \text{mean of } \gauss(\mu_i, \sigma^{2}_i)_{\ge z_i}$;
        \item[(M)] Refit the RF using $\left(\vtheta_i, \hat{y_i}^{(T)}\right)_{i=1}^n$ as the basis for tree $T$.
\end{enumerate}

Here, $\gauss(\mu_i, \sigma^{2}_i)_{\ge z_i}$ denotes the predictive distribution of the current RF for data point $i$,
\blue{\emph{truncated at $z_i$, that is, conditioned on the fact that it is at least as large as $z_i$}}.
While the mean of
$\gauss(\mu_i, \sigma^{2}_i)_{\ge z_i}$
is the best \emph{single} value to impute for the $i$th data point,
in the context of RF models this approach yields overly confident predictions: all trees would perfectly agree on the predictions for censored data points.
To preserve our uncertainty about the true runtime of censored runs, \red{we can change the E step to:}
\begin{enumerate}
        \item[(E$\prime$)] For each tree $T$ in the RF and each $i$ s.t.\ $c_i\!=\!1$: $\hat{y}_i^{(T)} \! \leftarrow \!\! \text{ sample from } \gauss(\mu_i, \sigma^{2}_i)_{\ge z_i}$.
\vspace*{-0.2cm}
\end{enumerate}
\red{In order to guarantee convergence, we also keep the assignment of bootstrap data points to each of the forest's trees fixed across iterations and draw the samples for each censored data point in a stratified manner; for brevity, we refer the reader to \cite{HutHooLey13:censoredBOarXiv} for the precise details.}
Our resulting modified variant of Schmee \& Hahn's algorithm takes our prior uncertainty into account when computing the posterior predictive distribution, thereby avoiding overly confident predictions. 
As an implementation detail, to avoid potentially large outlying predictions above the known maximal runtime of $\kappa_{max} = 300$ seconds, we ensure that the mean imputed value does not exceed $\kappa_{max}$.\footnote{In Schmee \& Hahn's algorithm, this simply means imputing $\min \{\kappa_{max}, \text{mean}(\gauss(\mu_i, \sigma^{2}_i)_{\ge z_i})\}$.
In our sampling version, it amounts to keeping track of the mean $m_i$ of the imputed samples for each censored data point $i$ and subtracting $m_i - \kappa_{max}$ from each sample for data point $i$ if $m_i > \kappa_{max}$.} (In the absence of censored runs --- the case addressed in the major part of our work  --- this mechanism is not needed, since all predictions are linear combinations of observed runtimes and are thus upper-bounded by their maximum.)

\subsection{Experimental Setup} \label{sec:censoring-exp-setup}

We now experimentally compare Schmee \& Hahn's procedure and our modified version to two baselines: ignoring censored data points altogether and treating data points that were censored at the captime $\kappa$ as uncensored data points with runtime $\kappa$.
We only report results for the most interesting case of predictions for previously unseen parameter configurations and instances.
We used the 9 benchmark distributions from Section \ref{sec:combination_ehm_rsm}, artificially censoring the training data at different thresholds below the actual threshold.
We experimented with two different types of capped data: (1)~data with a fixed censoring threshold across all data points,
and (2)~data in which the thresholds were instance-specific (specifically, we set the threshold for all runs on an instance to the runtime of the best of the 1\,000 configurations on that instance). The fixed threshold represents the sort of data generated by experimental studies like those from the previous sections of this paper, while the instance-specific threshold models practical applications of EPMs in model-based algorithm configuration procedures~\cite{HutHooLey11-censoredBO}.
For both types of capped data and for all prediction strategies, we measured both predictive error (using RMSE as in the rest of the paper) and the quality of uncertainty estimates (using log likelihood, LL, as defined in Section \ref{sec:exp_setup_eh}) on the uncensored part of the test data.

\subsection{Experimental Results} \label{sec:censoring-exp}

\begin{figure}[tbp]
    {\scriptsize
 \setlength{\tabcolsep}{2pt}
\centering
    \begin{tabular}{ccccc}
      ~ & Drop censored data & Pretend all uncensored & Schmee \& Hahn & Sampling S\&H \\
      \begin{sideways}Fixed threshold\end{sideways} &
\includegraphics[scale=0.20]{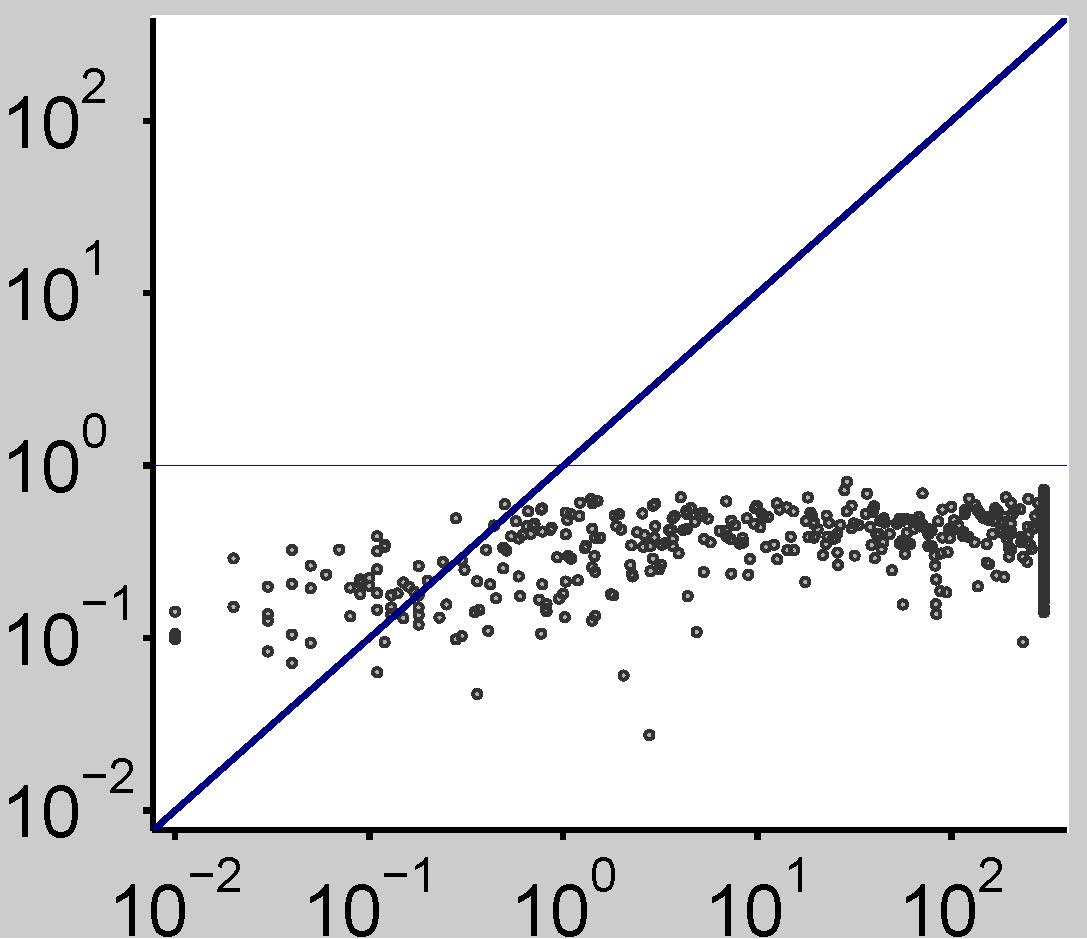} &
\includegraphics[scale=0.20]{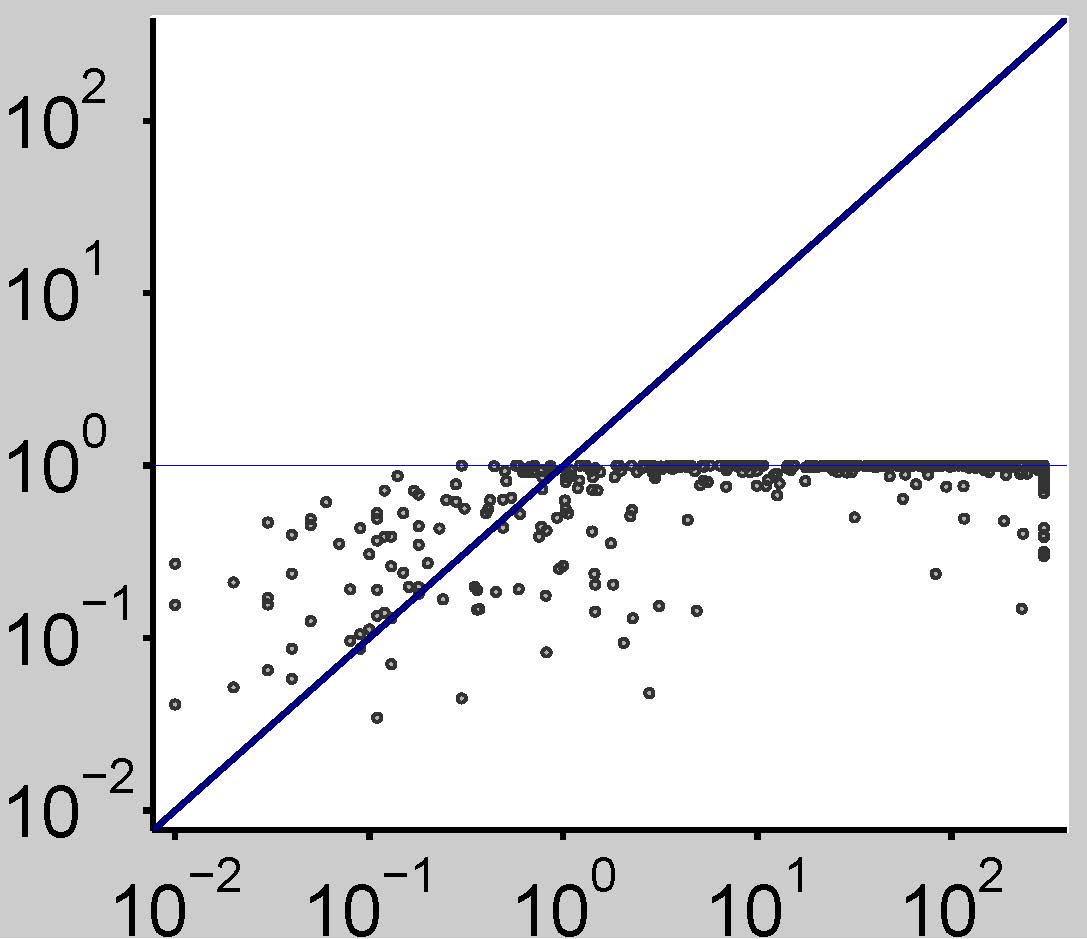} &
\includegraphics[scale=0.20]{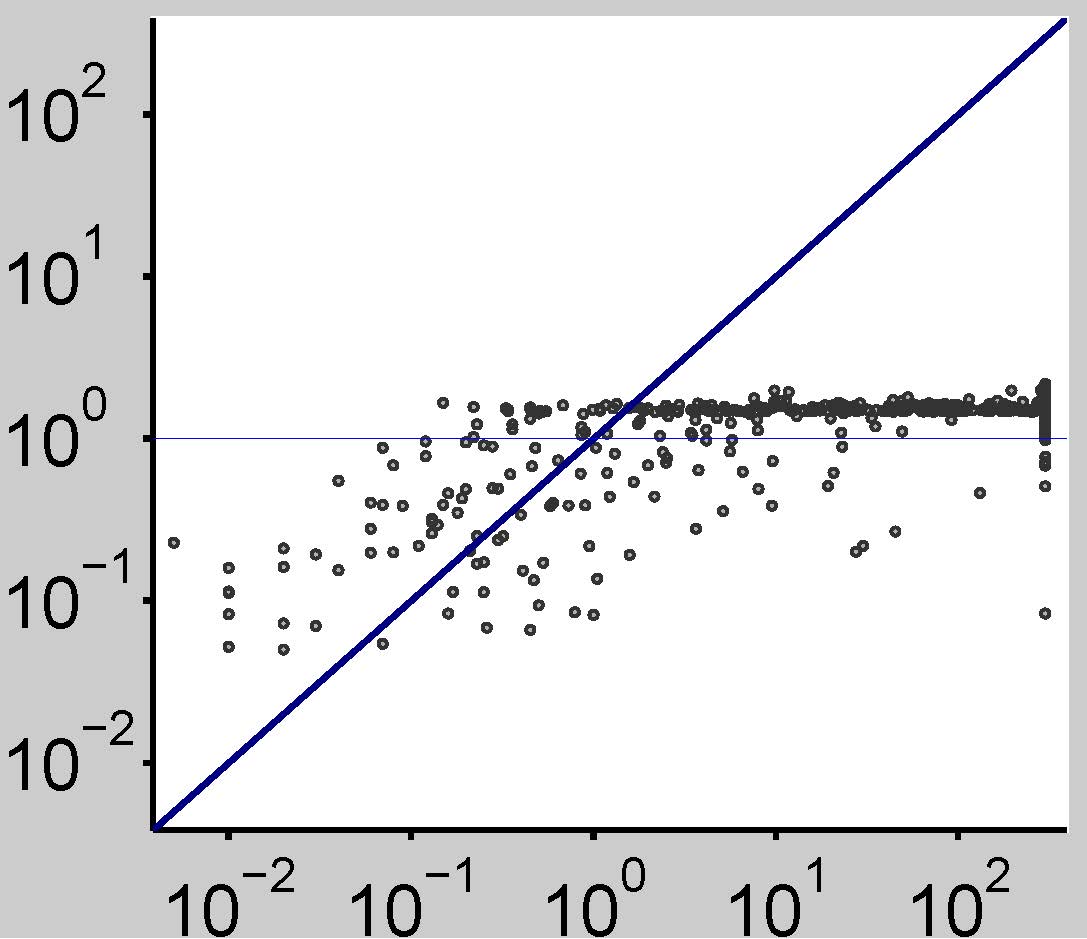} &
\includegraphics[scale=0.20]{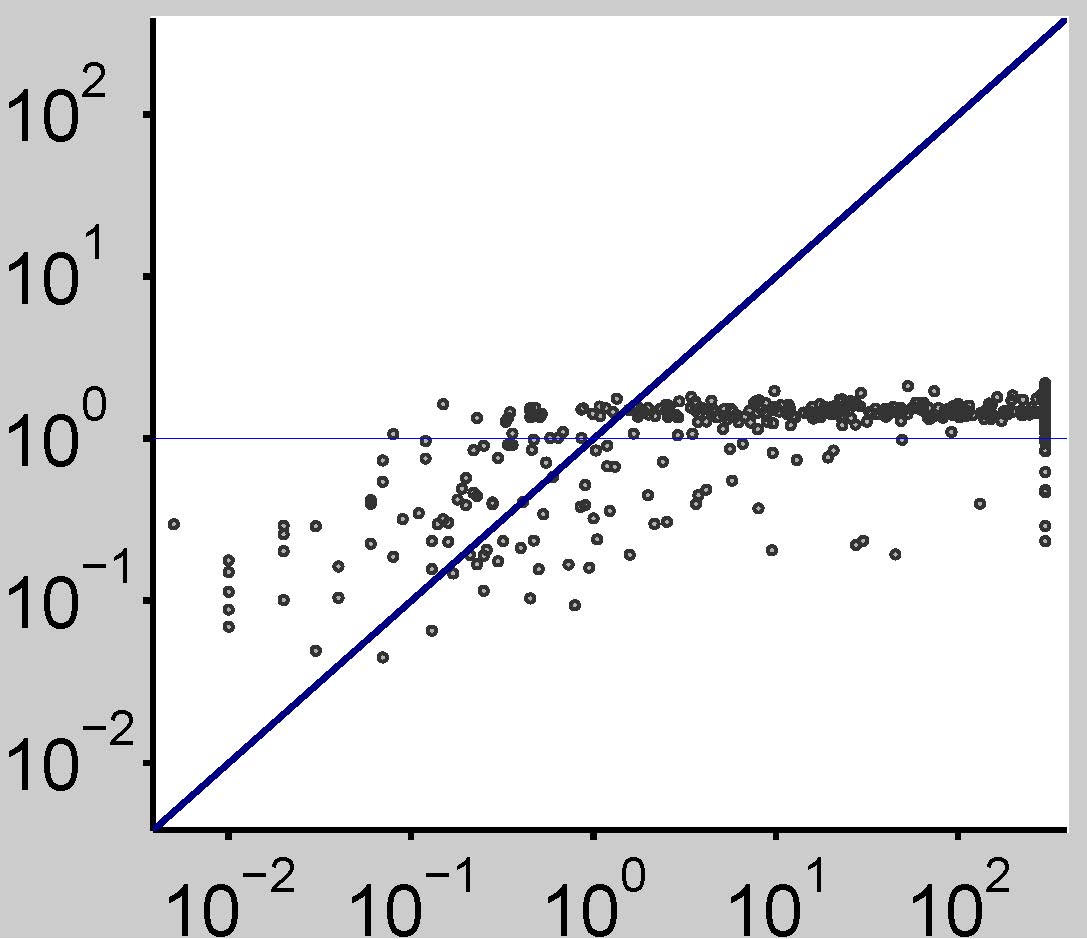}
\hlinespace{}
      \begin{sideways}Varying threshold\end{sideways} &
\includegraphics[scale=0.20]{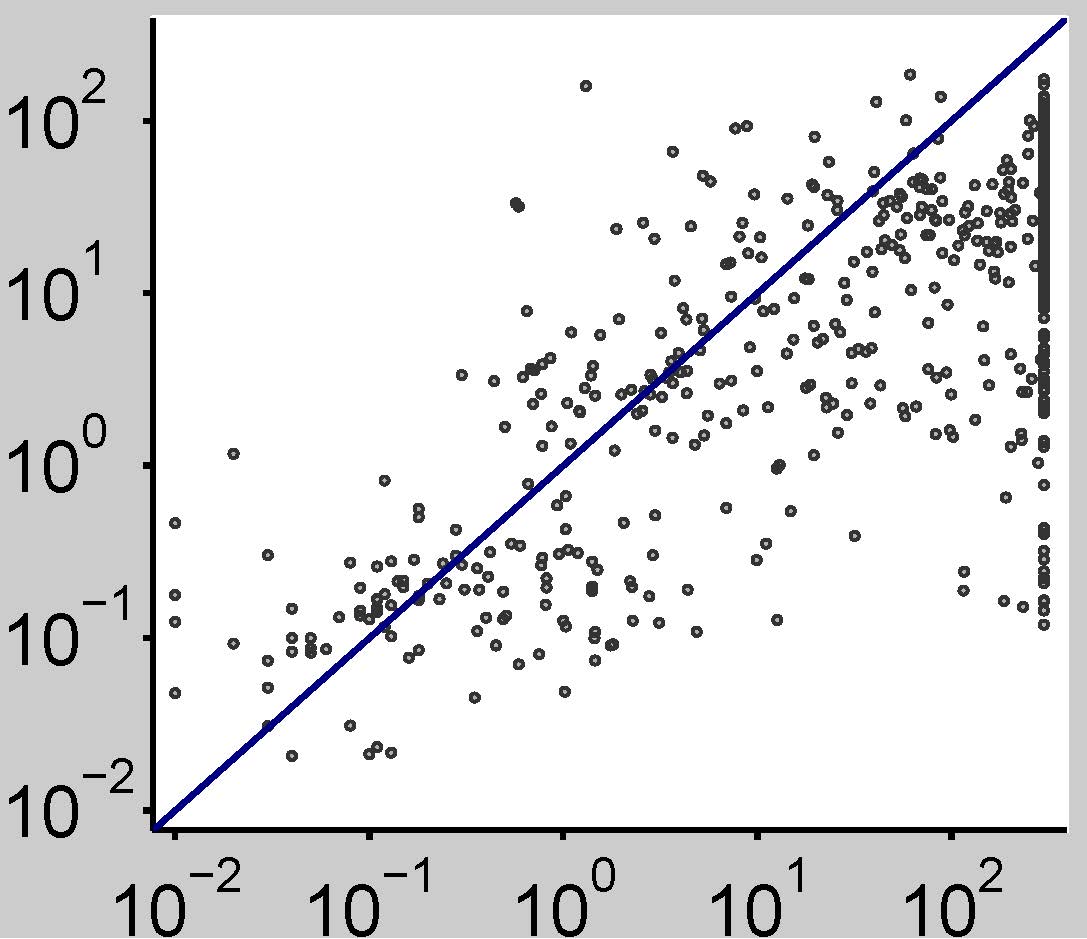} &
\includegraphics[scale=0.20]{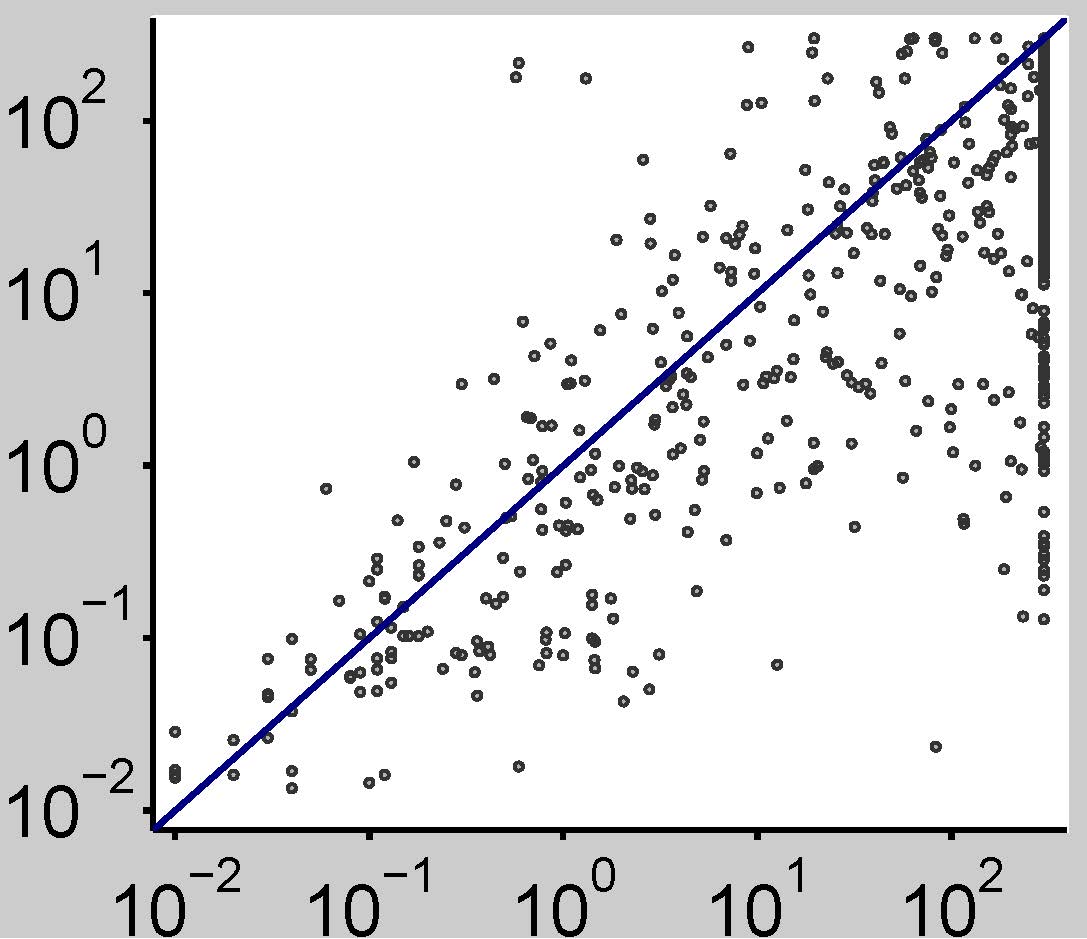} &
\includegraphics[scale=0.20]{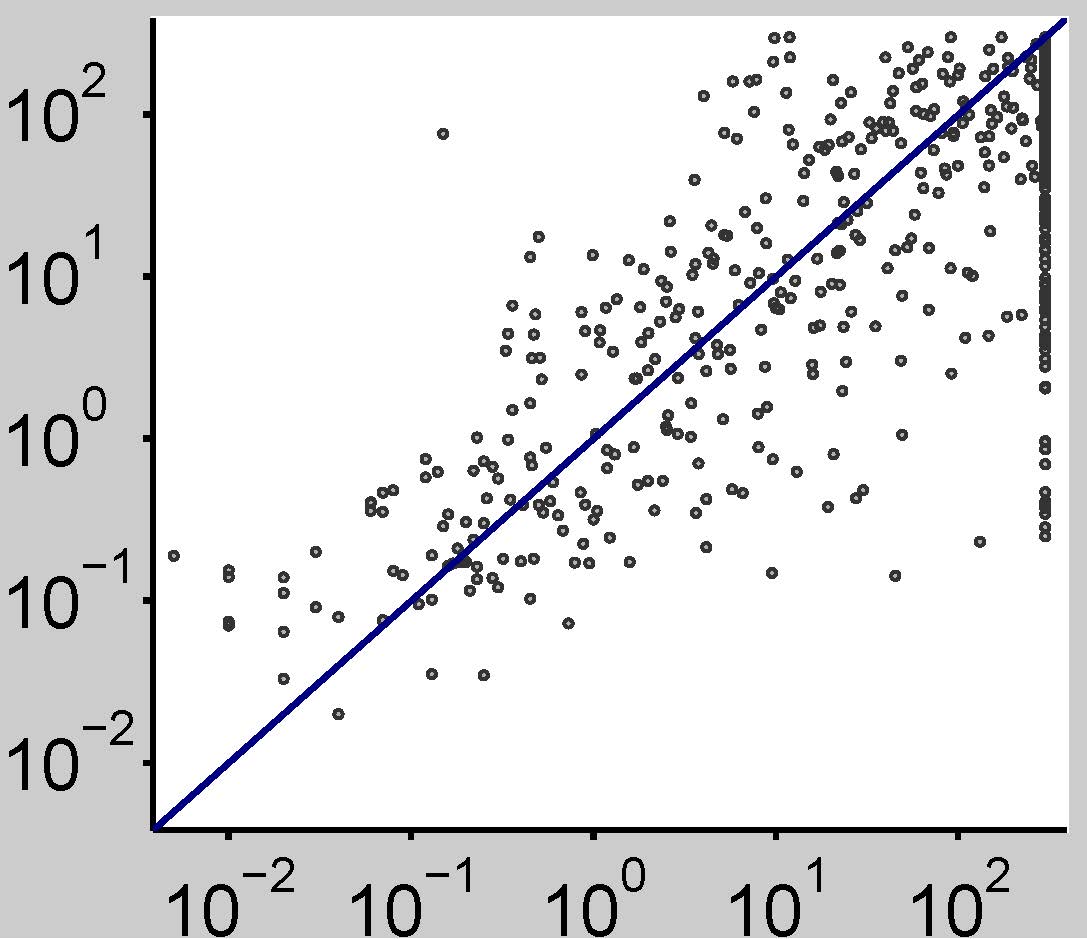} &
\includegraphics[scale=0.20]{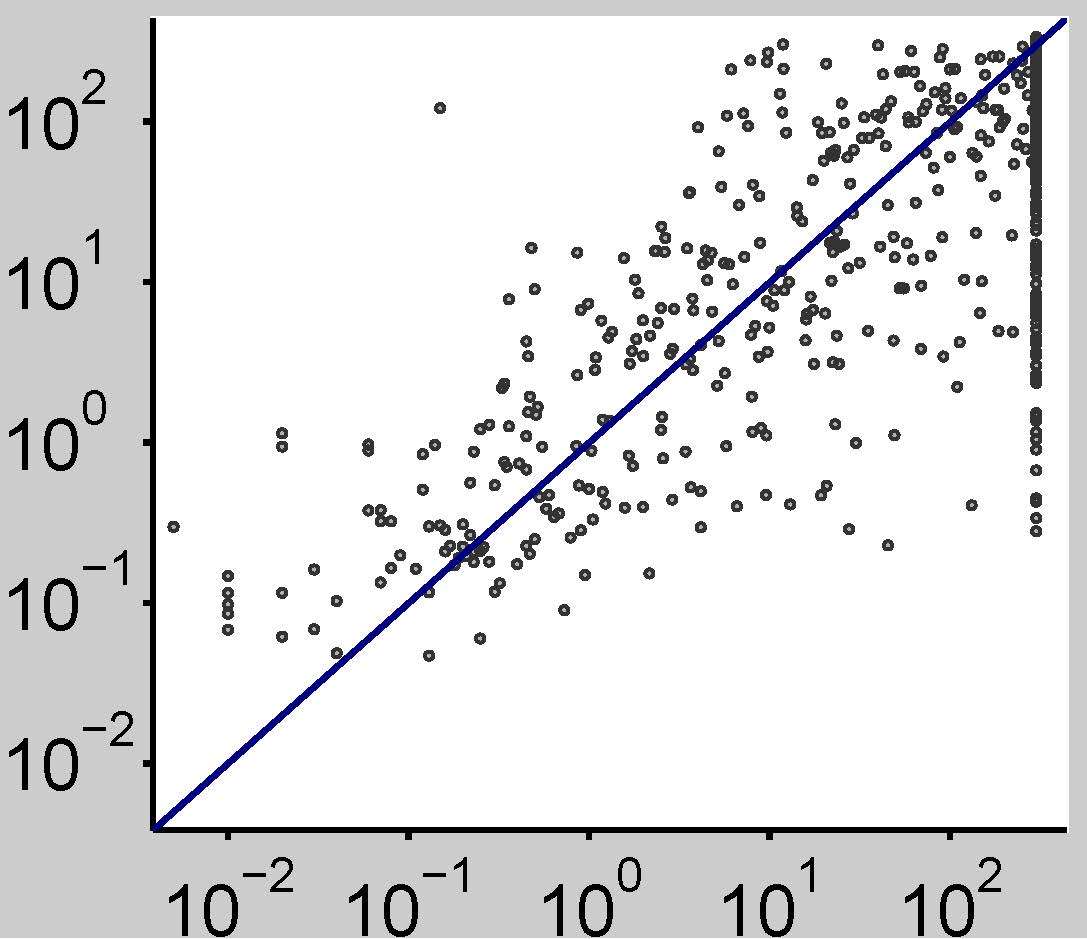}
    \end{tabular}
        \small
    \caption{True and predicted runtime of various ways of handling censored data in random forests, for scenario \cplex{}-\BIGMIX{}
    with fixed censoring threshold of one second during training (top) and varying threshold (bottom). In each scatterplot, the $x$-axis indicates true runtime and the $y$-axis cross-validated runtime as predicted by the respective model. Each dot represents one instance. Analogous figures for all benchmarks are given in Figures D.31--D.39 (in the online appendix). \label{fig:censoring}}
  }
\end{figure}

Figure \ref{fig:censoring} illustrates the raw predictions for one benchmark, demonstrating the qualitative differences between the four methods for treating capped data.
In the case of a fixed censoring threshold $\kappa$, simply dropping censored data yielded consistent underestimates (see the top-left plot of Figure \ref{fig:censoring}), while treating censored data as uncensored at $\kappa$ yielded good predictions up to $\kappa$ (but not above); this strategy is thus reasonable when no predictions beyond $\kappa$ are required (which is often the case; \eg{}, throughout the main part of this article). The Schmee \& Hahn variants performed similarly up to $\kappa$, but yielded unbiased predictions up to about two times $\kappa$.
\revision{We note that a factor of two is not very much compared to the orders of magnitude variation we observe in our data.
Much better predictions of larger runtimes can be achieved by using the instance-specific captimes discussed above (see the lower half of Figure \ref{fig:censoring}), and we thus advocate the use of
such varying captimes in order to enable better scaling to larger captimes.}

\revision{A quantitative analysis (described in Section D.4 of the online appendix) showed that in the fixed-threshold case, dropping censored data led to the worst prediction errors; treating censored data as uncensored improved results; and using the Schmee \& Hahn variants further reduced prediction errors. However, with fixed thresholds, the Schmee \& Hahn variants often yielded poor uncertainty estimates because they imputed similar values (close to the fixed threshold) for all censored data points, yielding too little variation across trees, and thus also yielded overconfident predictions.
In contrast, for data with varying captimes, treating censored data as uncensored often performed worse than simply dropping it, and the Schmee \& Hahn variants (in particular our new one) yielded both competitive uncertainty estimates and the lowest prediction error. Finally, we found these qualitative findings to be robust with respect to how aggressively (i.e., how low) the captimes were chosen. Overall, random forests handled censored data reasonably well. We note that other models might be better suited to \emph{extrapolating} from training data with short captimes to obtain accurate runtime predictions for long algorithm runs.}

\section{Conclusions} \label{sec:conclusion}

In this article, we assessed and advanced the state of the art in predicting the performance of algorithms for hard combinatorial problems. We proposed new techniques for building predictive models, with a particular focus on improving prediction accuracy for parameterized algorithms, and also introduced a wealth of new features for three of the most widely studied NP-hard problems (SAT, MIP and TSP)
that benefit all models. We conducted the largest experimental study of which we are aware---predicting the performance of 11 algorithms on 35 instance distributions from SAT, MIP and TSP---comparing our new modeling approaches with
a comprehensive set of methods from the literature.
We showed that our new approaches---chiefly those based on random forests, but also approximate Gaussian processes---offer the best performance, whether we consider predictions for previously unseen problem instances for parameterless algorithms, new parameter settings for a parameterized algorithm running on a single problem instance, or parameterized algorithms being run both with new parameter values and on previously unseen problem instances.
We also demonstrated in each of these settings that very accurate predictions (correlation coefficients between predicted and true runtime exceeding 0.9) are possible based on very small amounts of training data (only hundreds of runtime observations).
Finally, we demonstrated how our best-performing model, random forests, could be improved further by better handling data from prematurely terminated runs.
Overall, we showed that our methods are fast, general, and achieve good, robust performance. We hope they will be useful to a wide variety of researchers who seek to model algorithm performance for algorithm analysis, scheduling, algorithm portfolio construction, automated algorithm configuration, and other applications.
The Matlab source code for our models, the data and source code to reproduce our experiments, and an online appendix containing additional experimental results, are available online at {\footnotesize\url{http://www.cs.ubc.ca/labs/beta/Projects/EPMs}}.

\small
\section*{Acknowledgments}

We thank Kevin Murphy for many valuable discussions regarding Gaussian processes and random forests, and Jonathan Shen for proofreading an early version of this paper.
We also thank the anonymous reviewers for comments that helped us to significantly improve the paper.

\appendix

\section{Details on Benchmark Instance Sets}\label{app:benchmarks}

This appendix gives more information about our instance benchmarks. For the SAT benchmarks, the number of variables and clauses are given for the original instance (before preprocessing). (In contrast, \cite{Hut09:phd} reported these numbers after preprocessing, explaining the differences in reported values for \IBM{} and \SWV{}.)

\subsection{SAT benchmarks}\label{app:benchmarks-sat}

\paragraph{\INDU}
This benchmark data set comprises 1\,676 instances from the industrial categories
of the 2002--2009 SAT competitions as well as from the 2006, 2008 and 2010 SAT Races. These instances contain an average of $111\,000$ variables and $689\,187$ clauses, with respective standard deviations of $318\,955$ and $1\,510\,764$, and respective maxima of $9\,685\,434$ variables and $14\,586\,886$ clauses.

\vspace*{-0.15cm}
\paragraph{\HAND}
This benchmark data set comprises 1\,955 instances from the handmade categories of the 2002-2009 SAT Competitions. These instances contain an average of $4\,968$ variables and $82\,594$ clauses, with respective standard deviations of $21\,312$ and $337\,760$, and respective maxima of $270\,000$ variables and $4\,333\,038$ clauses.

\vspace*{-0.15cm}
\paragraph{\RAND}
This benchmark data set comprises 3\,381 instances from the random categories of the 2002-2009 SAT Competitions. These instances contain an average of $1\,048$ variables and $6\,626$ clauses, with respective standard deviations of $2\,593$ and $11\,221$, and respective maxima of $19\,000$ variables and $79,800$ clauses.

\vspace*{-0.15cm}
\paragraph{\INDUHANDRAND} This set is the union of INDU, HAND, and RAND.

\vspace*{-0.15cm}
\paragraph{\IBM}
This set of SAT-encoded bounded model checking instances comprises 765
instances generated by \emcite{zarpas05benchmarking}; these instances were selected as the instances in 40 randomly-selected folders from the IBM Formal Verification Benchmarks Library.
These instances contained an average of $96\,454$ variables and $413\,143$ clauses, with respective standard deviations of $169\,859$ and $717\,379$, and respective maxima of $1\,621\,756$ variables and $6\,359\,302$ clauses.

\vspace*{-0.15cm}
\paragraph{\SWV}
This set of SAT-encoded software verification instances comprises
604 instances generated with the CALYSTO static checker~\cite{babic07structural-abs},
used for the verification of five programs: the spam filter Dspam, the SAT solver HyperSAT, the Wine Windows OS emulator, the gzip archiver, and a component of xinetd (a secure version of inetd).
These instances contain an average of $68\,935$ variables and $206\,147$ clauses, with respective standard deviations of $56\,966$ and $181\,714$, and respective maxima of $280\,972$ variables and $926\,872$ clauses.

\vspace*{-0.15cm}
\paragraph{\RANDSAT}
This set contains 2\,076 satisfiable instances (proved by at least one winning solver from the previous SAT competitions) from data set \RAND{}. These instances contain an average of $1\,380$ variables and $8\,042$ clauses, with respective standard deviations of $3\,164$ and $13\,434$, and respective maxima of $19\,000$ variables and $79,800$ clauses.

\subsection{MIP benchmarks}\label{app:benchmarks-mip}
\paragraph{\BIGMIX}
This highly heterogenous mix of publicly available Mixed Integer Linear Programming (MILP) benchmarks comprises 1\,510 MILP instances. The instances in this set have an average of 8\,610 variables and 4\,250 constraints, with respective standard deviations of $34\,832$ and $21\,009$, and respective maxima of $550\,539$ variables and $550\,339$ constraints.

\vspace*{-0.15cm}
\paragraph{\CORLAT}
This set comprises 2\,000 MILP instances based on real data used for the construction of a wildlife corridor for grizzly bears in the Northern Rockies region (the instances were described by \citet{ghs08:connection} and made available to us by Bistra Dilkina).
All instances had $466$ variables; on average they had $486$ constraints (with standard deviation $25.2$ and a maximum of $551$).

\vspace*{-0.15cm}
\paragraph{\RCW}
This set comprises 1\,980 MILP instances from a computational sustainability project. These instances model the spread of the endangered red-cockaded woodpecker, conditional on decisions about certain parcels of land to be protected. We generated 1\,980 instances (20 random instances for each combination of 9 maps and 11 budgets), using the generator from~\cite{rcw} with the same parameter setting as used in that paper, except a smaller sample size of 5. All instances have $82\,346$ variables; on average, they have $328\,816$ constraints (with a standard deviation of only $3$ and a maximum of $328\,820$).

\vspace*{-0.15cm}
\paragraph{\REG}
This set comprises 2\,000 MILP-encoded instances of the winner determination problem in combinatorial auctions. We generated 2\,000 instances using the {\smaller\texttt{regions}} generator from the Combinatorial Auction Test Suite~\cite{LeyPeaSho00}, with the number of bids selected uniformly at random from between 750 and 1250, and a fixed bids/goods ratio of 3.91 (following~\cite{LeyNudSho09:jacm}). They have an average of $1\,129$ variables and $498$ constraints, with respective standard deviations of $73$ and $32$ and respective maxima of $1\,255$ variables and $557$ constraints.

\subsection{TSP benchmarks}\label{app:benchmarks-tsp}
\paragraph{\PORTGEN}
This set comprises 4\,993 uniform random Euclidean 2-dimensional TSP instances generated by the random TSP generator, {\smaller\texttt{portgen}}~\cite{url:tspgen}. The number of nodes was randomly selected from 100 to 1\,600, and the generated TSP instances contain an average of $849$ nodes with a standard deviation of $429$ and a maximum of $1\,599$ nodes.

\vspace*{-0.15cm}
\paragraph{\PORTCGEN}
This set comprises 5\,001 random clustered Euclidean 2-dimensional TSP instances generated by the random TSP generator, portcgen~\cite{url:tspgen}. The number of nodes was randomly selected from 100 to 1\,600, and the number of clusters was set to 1\% of the number of nodes. The generated TSP instances contain an average of $852$ nodes with a standard deviation of $432$ and a maximum of $1\,599$ nodes.

\vspace*{-0.15cm}
\paragraph{\TSPLIB} This set contains a subset of the prominent TSPLIB ({\smaller\url{http://comopt.ifi.uni-heidelberg.de/software/TSPLIB95/}}) repository. We only included the 63 instances for which both our own feature computation code and the code by \emcite{MilvHem11} completed successfully (ours succeeded on 23 additional instances). These 63 instances have $931 \pm 1376$ nodes, with a range from $100$ to $5\,934$.

\bibliographystyle{newapa}

\renewcommand{\baselinestretch}{0.97}
\setlength\bibsep{0.16cm}

\footnotesize{\bibliography{EPMs}}

\end{document}